\documentclass[letterpaper,journal]{IEEEtran}
\usepackage{amsmath,amsfonts}
\usepackage{algorithmic}
\usepackage{algorithm}
\usepackage{array}
\usepackage[caption=false,font=normalsize,labelfont=sf,textfont=sf]{subfig}
\usepackage{textcomp}
\usepackage{stfloats}
\usepackage{url}
\usepackage{verbatim}
\usepackage{graphicx}
\usepackage{cite}
\usepackage{enumitem}

\usepackage{subfig}
\usepackage{bm}
\usepackage{enumerate}
\usepackage{amssymb}%
\usepackage{multirow}%
\usepackage{amsthm}%
\usepackage{adjustbox}
\usepackage{makecell}
\usepackage{booktabs}    
\usepackage{placeins}
\usepackage{makecell}
\usepackage{float} 
\usepackage{booktabs}
\usepackage{tikz}
\usetikzlibrary{arrows.meta, positioning, calc, shapes.geometric, backgrounds, decorations.pathmorphing}

\usepackage{xr}
\usepackage[caption=false, font=normalsize, labelfont=sf, textfont=sf]{subfig}

\begin{document}

\title{Multivariate Spatio-Temporal Neural Hawkes Processes}

\author{{Christopher Chukwuemeka, Hojun You, and Mikyoung Jun*}

\thanks{*Corresponding author: Mikyoung Jun
(Email: mjun@central.uh.edu)}}


\maketitle

\begin{abstract}
We propose a Multivariate Spatio-Temporal Neural Hawkes Process for modeling complex multivariate event data with spatio-temporal dynamics. The proposed model extends continuous-time neural Hawkes processes by integrating spatial information into latent state evolution through learned temporal and spatial decay dynamics, enabling flexible modeling of excitation and inhibition without predefined triggering kernels. By analyzing fitted intensity functions of deep learning–based temporal Hawkes process models, we identify a modeling gap in how fitted intensity behavior is captured beyond likelihood-based performance, which motivates the proposed spatio-temporal approach. Simulation studies show that the proposed method successfully recovers sensible temporal and spatial intensity structure in multivariate spatio-temporal point patterns, while existing temporal neural Hawkes process approach fails to do so. An application to terrorism data from Pakistan further demonstrates the proposed model’s ability to capture complex spatio-temporal interaction across multiple event types.
\end{abstract}


\section{Introduction}
\IEEEPARstart{I}{n} recent years, the analysis of spatio-temporal point patterns has gained significant attentions in various scientific applications, including the study of earthquakes \cite{ogata1998space, kwon2023flexible}, epidemiology \cite{dong2023non}, crimes \cite{mohler2011self, dong2022spatio}, terrorist attacks \cite{jun2024flexible}, and social network analysis \cite{reinhart2018review, yuan2021fast}. Understanding spatio-temporal characteristics of point pattern of events is crucial for comprehending their dynamics and making correct predictions. Spatio-temporal point processes provide a versatile statistical framework for understanding event occurrence patterns and explaining the relationships between events of different types\cite{daley2003introduction, diggle2006spatio, wang2020deep}. 

Two of the most popular point process frameworks are Log-Gaussian Cox Processes (LGCP) \cite{moller1998log} and Hawkes processes \cite{hawkes1971spectra}. 
LGCP is a type of Cox process where the (log-transformed) intensity of the Poisson process is modeled as a Gaussian process. LGCP framework is frequently used to explain local clustering behavior of point patterns as the covariance structure of the Gaussian process naturally induce clustering of points in space and/or time. Hawkes processes, often referred to as self-exciting (or self-modulating) process, describe point patterns where past events excite (or modulate) subsequent events. Hawkes processes have been traditionally applied in seismology such as \cite{ogata1988statistical}, but recently they have gained popularity for applications in finance, criminology, and epidemology \cite{park2021investigating, kwon2023flexible, schoenberg2023estimating, jun2024flexible, bernabeu2025spatio, das2025likelihood}. In this work, we focus on the Hawkes process framework, targeting multivariate spatio-temporal point patterns from terrorism with complex self excitation or inhibition structure, both marginally and jointly. 

There have been developments of univariate and multivariate spatio-temporal Hawkes process models in parametric settings \cite{reinhart2018review,yuan2019multivariate}. However, most of these models are restrictive in the sense that their triggering structure is given as a fixed parametric function of spatial and/or temporal lag, and thus not flexible. Furthermore, their spatio-temporal separablility is often assumed. Therefore, these approaches do not allow dynamic change of triggering structure in space or time domains. Many of these models require specific stability conditions that make computation of likelihood complex and computationally challenging.  Recently, there have been some efforts to enhance statistical estimation methods \cite{yuan2019multivariate, yuan2021fast, siviero2024flexible} or model flexibility \cite{jun2024flexible}. Yet, much of these work assume specific ``triggering" structure for excitation, both marginally and jointly, which is limited for many real applications. For examples, \cite{yuan2019multivariate, yuan2021fast, siviero2024flexible} imposed non-negative excitation effect on the triggering effect. Later, \cite{jun2024flexible} relaxed the non‑negativity constraint and propose non-separable and nonstationary triggering kernels. However, their triggering functions are still specified through predefined parametric families, and increasing flexibility introduces more parameters, leading to substantial computational burden.

With the advent of deep learning, adapting neural networks in Hawkes process models have been studied in the literature \cite{mei2017neural, zhang2020self, chen2021neuralstpp, zhu2021imitation, dong2022spatio, yang2022attentive, zhou2022neural, yuan2023spatio, li2024beyond, cheng2025deep}.
Neural temporal models capture temporal triggering effects using recurrent or attention-based architectures, such as the continuous-time LSTM-based Neural Hawkes process \cite{mei2017neural} and self-attentive Hawkes models \cite{zhang2020self, yang2022attentive}. We kindly refer readers to \cite{xue2024easytpp} for detailed benchmarks of temporal neural Hawkes processes. In the spatio-temporal domain, \cite{chen2021neuralstpp} combined continuous normalizing flows with attention mechanism to explain continuous evolution of the process, but suffers from inefficient ODE solver. \cite{zhu2021imitation} modeled the conditional intensity with Gaussian diffusion kernels and learned the conditional intensity by imitation learning. \cite{dong2022spatio} proposed a deep non-stationary parameterization of the spatio-temporal triggering function using low-rank neural representations. \cite{yuan2023spatio} introduced a diffusion-based generative spatio-temporal point process model that learns complex space–time event distributions beyond standard parametric triggering functions. \cite{li2024beyond} addressed a score-matching-based neural spatio-temporal Hawkes framework that avoids explicit likelihood normalization and enables both point prediction and uncertainty quantification for event times and locations. We refer readers to \cite{cheng2025deep} for a comprehensive survey of deep spatio-temporal point processes. To the best of our knowledge, however, these methods do not effectively explain cross-triggering structures between different event types despite the fact that they attempt to capture spatio-temporal interactions in Hawkes processes.

\subsection{Motivating example - Terrorism data}
\label{sec:motivation}

\noindent Our work is motivated by the real world application in modeling spatio-temporal patterns of terrorist attacks. 
Terrorism, ``the premeditated use or threat to use violence by individuals or subnational groups to obtain a political or social objective'' \cite{enders2006distribution}[pg. 4] is an important topic of research due to its serious impact with thousands of casualties annually, countless injuries \cite{stein1999medical}, and lasting psychological \cite{rubin2007enduring} and economic \cite{sandler2008economic} damages. 
Terrorism data are naturally spatio-temporal point patterns with complex spatial and temporal characteristics, which require flexible models. 

We are interested in modeling spatio-temporal point pattern of occurrences of terrorist attacks in Pakistan from the Global Terrorism Database \cite{start}. Particularly, we focus on four major terrorist groups in the country, namely Tehrik-i-Taliban Pakistan (TTP), Balochistan Republican Army (BRA), Balochistan Liberation Army (BLA), and Balochistan Liberation Front (BLF). TTP, also known as Pakistan Taliban, was formed in 2007 and was designated as a Foreign Terrorist Organization under U.S. law in 2010. It has been identified as one of the largest and deadliest terrorist groups in the country and maintains ties with other jihadist organizations including Al Qaeda and the Afghan Taliban \cite{kronstadt_23}. The other three, BRA, BLA, and BLF, are Baloch separatist groups who mainly operate in the resource-rich southwestern province of Balochistan, Pakistan. Although they share the broad goal of greater autonomy for Balochistan, they remain distinct in terms of leadership, preferred targets, and tactical methods \cite{verma_25_baloch}. These organizational differences motivate the use of a multivariate spatio-temporal framework that can capture both their individual attack patterns and potential interactions among them.

\begin{figure}[!htbt]
\centering
  \subfloat[]{\includegraphics[width=.5\columnwidth]{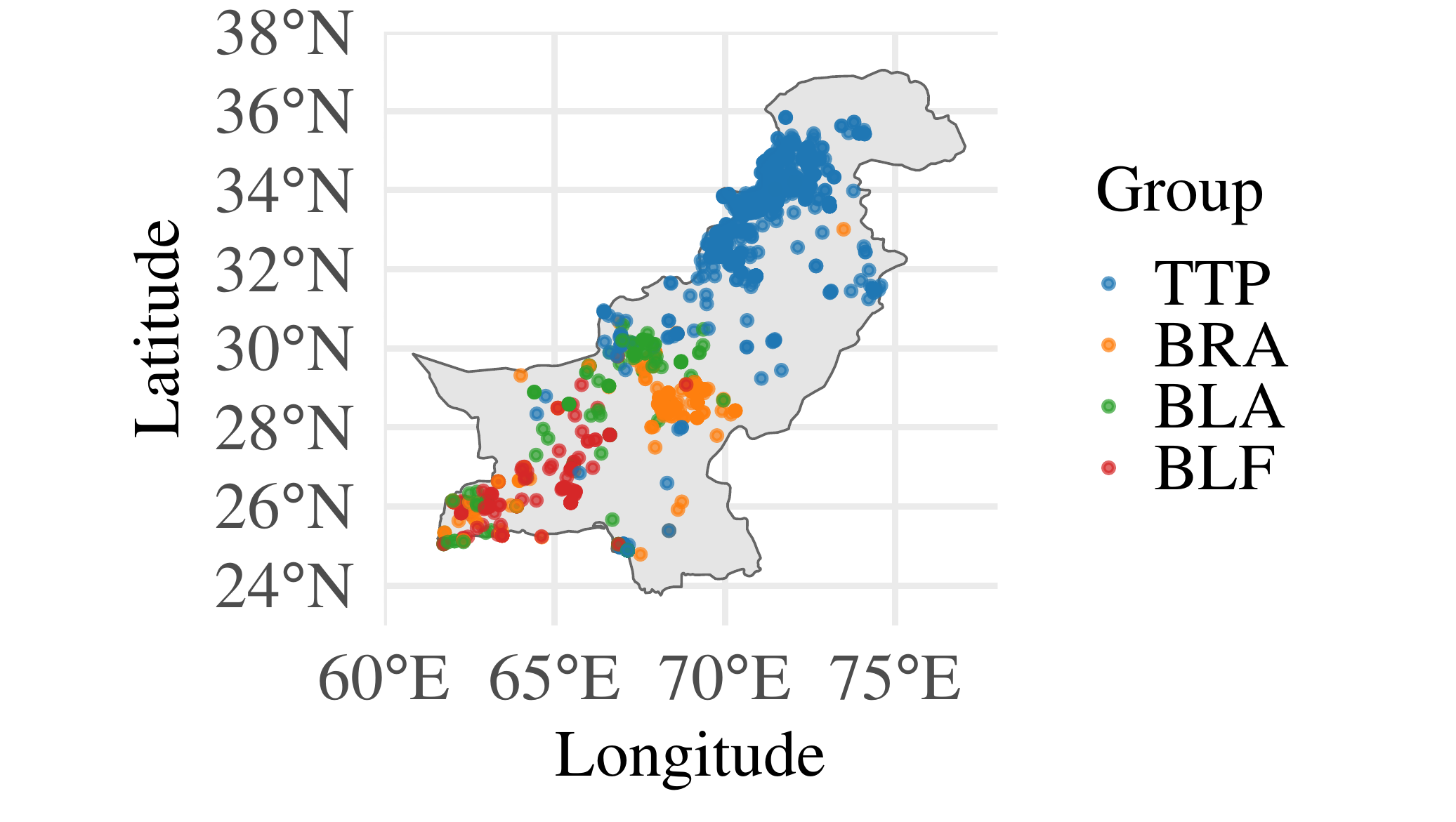}}
  \hfill
  \subfloat[]{\includegraphics[width=.5\columnwidth]{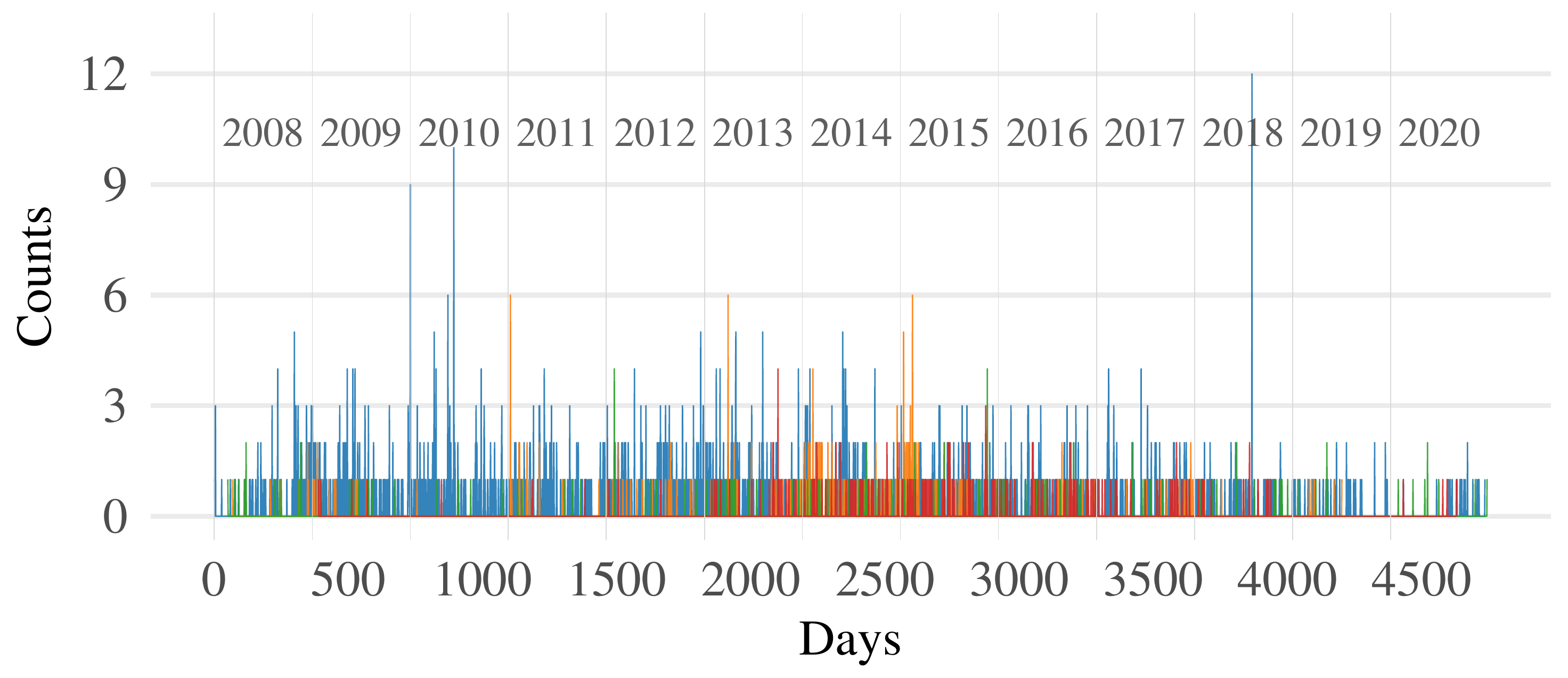}}

  \caption{(a): Location of terrorist attacks by four major groups in Pakistan during 2008-2020. (b): Daily counts of attacks for the same data.}
  
  \label{fig:conflict_zones}
\end{figure}

Figure~\ref{fig:conflict_zones} (a) shows a spatial point pattern of terrorist attacks carried out by the four groups during 2008-2020. Clearly, spatial patterns of these attacks are complex and there are interesting interaction between groups. Overall, attack locations are clustered marginally and the attacks by TTP are overall separated from the attack locations of the other three groups. Even within Baloch groups, there are some spatial separation. 
Figure~\ref{fig:conflict_zones} (b) shows temporal pattern of daily counts of terrorist attacks by the four groups during the same period. Although there is no clear time trend of counts for any of the groups, temporal patterns of each group are distinct and complex. 
Both figures clearly demonstrate the need for flexible point process models that can accommodate complex triggering structure (both marginally and jointly) in a spatio-temporal domain.

There is limited work in the development of flexible spatial and spatio-temporal point process models in terrorism study. Rather, it is common that point pattern data are aggregated at count level and typically models such as Poisson and negative binomial regressions are applied, ignoring relationship between points across space and/or time. See \cite{zhu2021promiseperilspointprocess} for discussions regarding this point. Recently, \cite{jun2024flexible} developed a bivariate, nonstationary parametric Hawkes process framework for spatial point patterns of terrorist attacks. While their models incorporate complex cross-triggering structure as well as nonstationary nature of spatial triggering, they are also limited due to the parametric structure of the model. For instance, the triggering structure (whether there is a triggering or inhibition) is fixed over time, which may be limited for complex nature of terrorist attack patterns. They consider a bivariate data set but more than two processes would significantly increase computational challenges. For instance, the stability condition for some of the parameters in the parametric triggering functions to satisfy are complex when there are more than two processes. Number of parameters to be estimated using maximum likelihood estimation method would grow rapidly with the increase of number of processes and many of these parameters may be hard to identify with the likelihood surface being practically flat.  

\section{Neural Temporal Point Process Models}

\noindent In this section, we introduce the definition of Hawkes processes and review currently available neural network-based models applicable to point pattern data under the Hawkes process framework.

\subsection{Preliminaries}\label{subsection:preliminary}

\noindent Consider a sequence of (temporal) events $\{(k_i, t_i)\}_{i=1}^n$, where $k_i \in \{1,\ldots,K\}$ denotes the event type and $t_i \in \mathbb{R}^+$ denotes the event time. Let $\mathcal H_t = \{(k_i, t_i): t_i < t\}$ denote the history of all events prior to time $t$. The event distribution in a point process is often characterized by its conditional intensity, $\lambda_k(t \mid \mathcal H_t)$, which is defined as
$$\lambda_k(t \mid \mathcal H_t) = \lim_{\Delta t \to 0} \frac{\mathbb{P}\big(\text{type }k\text{ event in }[t,t+\Delta t)\mid \mathcal H_t\big)}{\Delta t}.$$
Thus, the conditional intensity represents the instantaneous rate at which an event of type $k$ is expected to occur at time $t$, given the history $\mathcal H_t$.

{For a multivariate temporal Hawkes process,  the conditional intensity for event type $k$ is specified as
$$\lambda_k(t \mid \mathcal H_t)
= \mu_k (t) + \sum_{i: t_i < t}
\phi_{k,k_i}(t - t_i).$$
Here, $\mu_k (t) \geq  0$ is the baseline intensity for event type $k$ and it is often modeled as a constant. The function $\phi_{k,k_i}(\cdot)$ is a triggering kernel that quantifies the influence of past type-$k_i$ events on the future occurrence rate of type $k$ events. The collection $\{\phi_{k,k_i}\}$ encodes both self-triggering ($k=k_i$) and cross-triggering ($k\neq k_i$) effects, and may allow either excitation or inhibition.

\subsection{Review}\label{subsection:review}

\noindent The ``EasyTPP" \cite{xue2024easytpp} provides a significant effort to standardize the temporal point process (TPP) modeling and they provide a unified, open benchmarking platform with consolidated data, models, evaluation tools, and documentation. We review six foundational neural TPP methods implemented in the EasyTPP benchmark, highlighting their core innovations and limitations. 
Below, we provide a concise summary of those methods discussed in EasyTPP and used for our comparison. 
\begin{itemize}[leftmargin=*]
\item \emph{Recurrent Marked Temporal Point Process} (RMTPP,  \cite{du2016recurrent}): RMTPP integrates recurrent neural networks (RNNs) with temporal point processes, encoding event history into hidden states and feeds this state through a soft-plus transform to obtain the intensity. This is one of the earliest deep TPP models. It struggles with long-term dependencies due to RNN limitations and imposes restrictive parametric forms on the intensity function.

\item \emph{Neural Hawkes Process} (NHP, \cite{mei2017neural}): 
Building on RMTPP,  NHP replaces the discrete-time RNN with a continuous-time LSTM whose memory decays exponentially between events, thereby permitting both excitation and inhibition in a principled way. The model must evaluate intractable integrals via thinning-based Monte-Carlo sampling, which might become computationally expensive for dense event streams.

\item \emph{Self-Attentive Hawkes Process} (SAHP, \cite{zhang2020self}): This approach replaces recurrent architectures with self-attention mechanisms (transformer-style) to capture long-range dependencies more effectively. This discrete-time formulation sacrifices continuous-time granularity and the full attention matrix grows quadratically with sequence length, making both memory and time cost prohibitive for long sequences.

\item \emph{Transformer Hawkes Process} (THP, \cite{zuo2020transformer}): This method adapts a full Transformer encoder-decoder architecture to directly model event sequences by incorporating continuous-time positional encodings,  improving temporal resolution over SAHP. While this approach achieves state-of-the-art predictive accuracy, it requires high memory requirements inherent to self-attention. Furthermore, the model is constrained by predefined temporal kernels and requires careful tuning of its embeddings to avoid periodic artifacts.

\item \emph{Neural ODE-based TPP} (ODETPP, \cite{chen2021neuralstpp}):  It leverages neural ordinary differential equations to model hidden state trajectories between events, offering smoothly varying intensities that better respect continuous time. The iterative nature of ODE solvers introduces significant inference latency and numerical instability with stiff dynamics as every likelihood evaluation now calls a (potentially stiff) ODE solver.

\item \emph{Attentive Neural Hawkes Process}(AttNHP, \cite{yang2022attentive}): This is a hybrid model that combines continuous-time state evolution, such as a continuous-time LSTM, with an attention gate. This attention mechanism selectively re-weights past events, enabling the model to capture dependencies across multiple time scales and highlight salient triggers. However, this approach introduces several challenges. The extra gating layers increase architectural complexity and enlarge the parameter space, which can slow convergence and lead to training instability. Furthermore, the model is sensitive to hyperparameter tuning, and the learned attention weights are not always straightforward to interpret.

\end{itemize}

\subsection{Comparison of Methods}
\label{sec:comparison}

\noindent We perform an experiment to compare these six methods discussed in Section~\ref{subsection:review} with a simulated bivariate temporal Hawkes process. We consider various settings with long and short temporal triggering distance and the joint triggering between the two processes. We simulated the data using an R package {\it hawkes} \cite{hawkes} and the true intensity functions are given with Ogata-type triggering function: for $k=1,2$, 
\begin{align}
    \lambda_k(t)  = \mu_k +\sum_{l=1}^2 \sum_{i: t_i^{(l)} <t} \alpha_{kl}\cdot \exp\left\{-\beta_{k} (t-t_i^{(l)})\right\}.\label{compare}\end{align} Here, $t_i^{(l)}$ denotes the time of $i$'th event for $l$-th process, $l=1,2$.
We simulated 50 sequences over the interval independently and used 36 for training, 10 for validation, and 4 for testing, with the true intensity given in \eqref{compare} with 
\begin{align}
&\mu_1=\mu_2=0.3,~~(\beta_1,\beta_2)=(1.3,0.4), \nonumber\\
&(\alpha_{11}, \alpha_{12}, \alpha_{21}, \alpha_{22}) = (0.15,0.02,0.01,0.15). \label{compare-values}
\end{align}
Figure~\ref{fig:true_intensity_sim} shows the true conditional intensity of the bivariate process.

\begin{figure}[htbp!]
\centering
  \includegraphics[width=.3\textwidth]{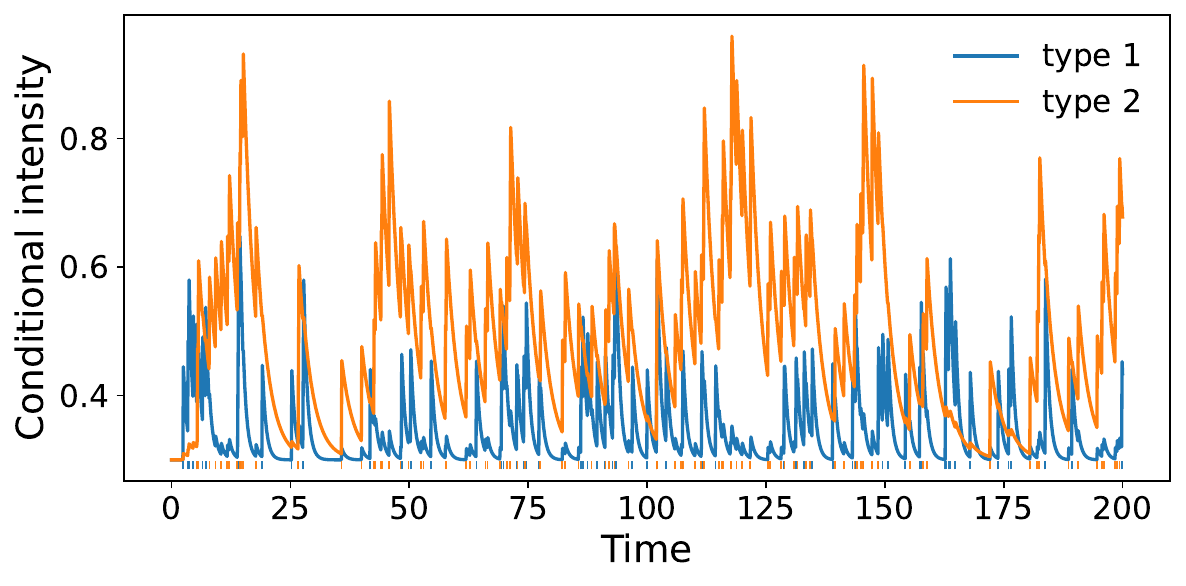}
  \caption{True intensity for a simulated data example used to compare six different neural network approaches with \eqref{compare} and \eqref{compare-values}.}
  \label{fig:true_intensity_sim} 
\end{figure}


The training and evaluation of the six models were conducted using the EasyTPP open‑source benchmarking framework \cite{xue2024easytpp}, which provides standardized implementation tools.
The source code for each model was obtained directly from the official EasyTPP GitHub repository\footnote{\url{https://github.com/ant-research/EasyTemporalPointProcess/tree/main}}, and the entire training pipeline, including data preprocessing, model configuration, optimization, and evaluation, was carried out according to the detailed guidelines provided in the project documentation\footnote{\url{https://ant-research.github.io/EasyTemporalPointProcess/index.html}}. 
To ensure a fair and reproducible comparison, all models were trained under identical experimental conditions, the same optimizer, early stopping based on validation log‑likelihood, and consistent data splits. 
The hyperparameters for each neural TPP model were adopted directly from the final configuration tables reported in \cite{xue2024easytpp} (see \ref{sec:supp_easytpp} in the supplementary material for more details).  The convergence plots of training for each method are given in 
 \ref{sec:supp_easytpp}  in the supplementary material. 

The six methods of consideration exhibit fundamental trade-offs: continuous-time models (NHP, ODETPP) achieve high temporal resolution at higher computational cost, while attention-based approaches (SAHP, THP, AttNHP) capture long-range dependencies efficiently at the expense of continuous-time granularity. This is empirically validated in Table \ref{tab:model_complexity} in section \ref{sec:supp_easytpp} of the supplementary materials: ODETPP is the slowest by a wide margin, while NHP and the hybrid AttNHP are substantially slower than pure attention models (SAHP, THP) under identical training budgets. 

Figure \ref{fig:temp_biv_simulation1} shows fitted conditional intensities for each method with the simulated temporal processes over a time interval [0,200]. 
In all the methods other than NHP, the fitted temporal conditional intensities do not adequately resemble the ground truth intensity. They not only struggle with capturing marginal temporal patterns of the intensity curves, but also fail to capture the joint relationship between the two temporal processes.  
\cite{xue2024easytpp} mainly considers quantitative measures such as negative log-likelihood and RMSE of event timing to compare the models. These metrics are important in assessing overall performance of models. However, they do not evaluate models' ability in capturing the underlying conditional intensity structure in time, which is the core quantity that defines excitation, inhibition, and temporal dynamics in Hawkes processes. As a result, models may appear competitive under standard prediction-based metrics despite producing fitted intensities that deviate significantly from the true generative structure.

One notable observation is that SAHP exhibits degenerate likelihood-driven solutions due to the absence of explicit continuous-time parameterization. Although SAHP imposes an exponential-like functional form on the conditional intensity, it does not constrain the magnitude or variability of the associated coefficients, which are reparameterized at every event and can therefore vary across time intervals. This flexibility allows the model to enforce extremely rapid intensity decay after each event, effectively suppressing the intensity over most of the inter-event interval and substantially reducing the non-event integral. As illustrated in Figure~\ref{fig:convplots_2k_epochs}in section \ref{subsec:supp_sahp_runs} of the supplementary material, SAHP learns the overall temporal trend early in training but later converges to a collapsed solution that maximizes likelihood primarily by minimizing the non-event contribution. The resulting intensity functions are highly unrealistic and fail to reflect plausible temporal excitation dynamics. In contrast, models such as NHP and THP impose stronger constraints on temporal decay through globally shared parameters, which prevents likelihood-driven collapse.

As is shown in 
Figure \ref{fig:temp_biv_simulation1}, the performance of NHP is particulary strong among the six methods. It is the only method that provides fitted intensity curves comparable to the ground truth in Figure~\ref{fig:true_intensity_sim}. This can be attributed to a key architectural property of NHP that distinguishes it from other neural TPP models: NHP maintains a latent state that evolves continuously between events. This state directly determines the conditional intensity, rather than recomputing intensity parameters only at event occurrences. 

Based on this finding, our work will extend \cite{mei2017neural} continuous-time architecture to spatio-temporal domains.

\begin{figure}[!ht]
\centering
  \subfloat[NHP]{\includegraphics[width=0.5\columnwidth]{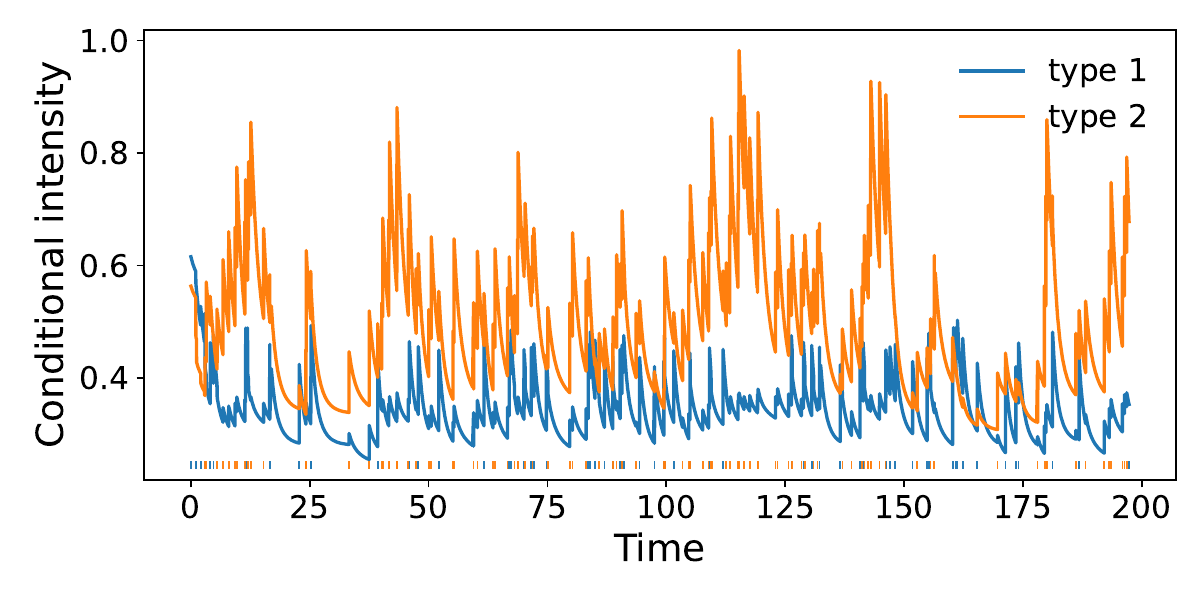}}
  \hfill
  \subfloat[AttNHP]{\includegraphics[width=0.5\columnwidth]{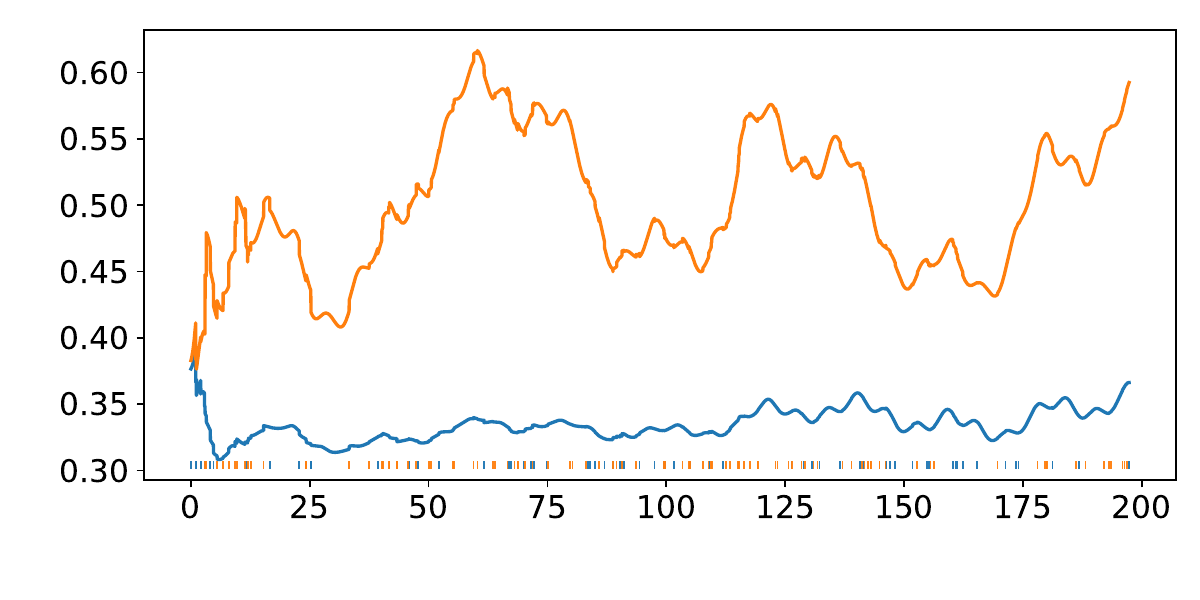}}
  
  \vspace{-8pt}

  \subfloat[SAHP]{\includegraphics[width=0.5\columnwidth]{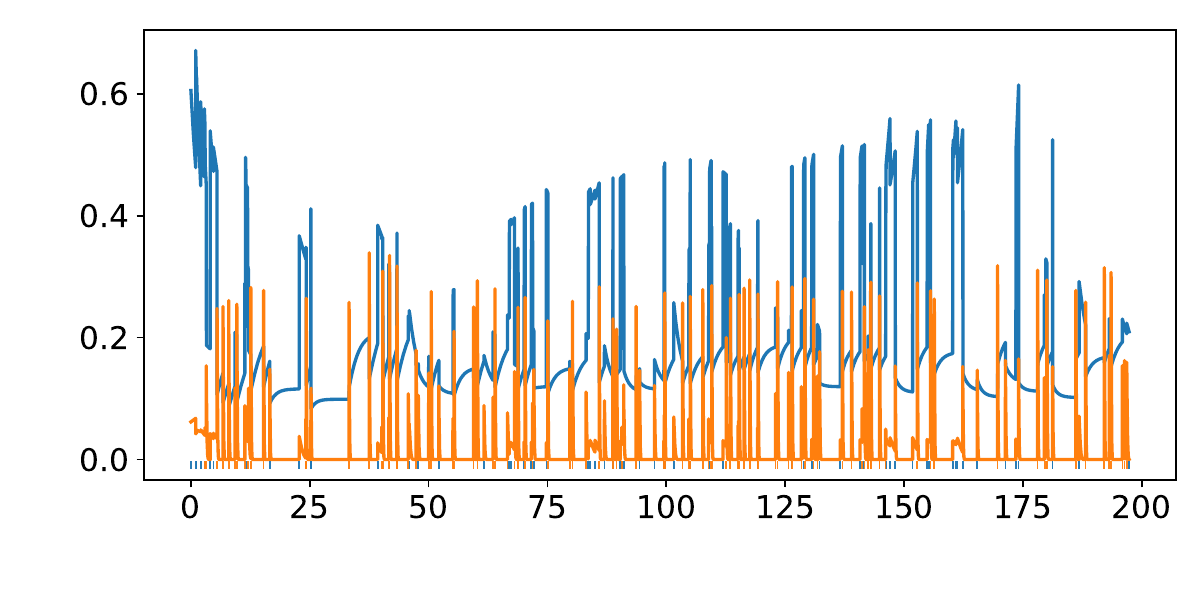}}
  \hfill
  \subfloat[ODETPP]{\includegraphics[width=0.5\columnwidth]{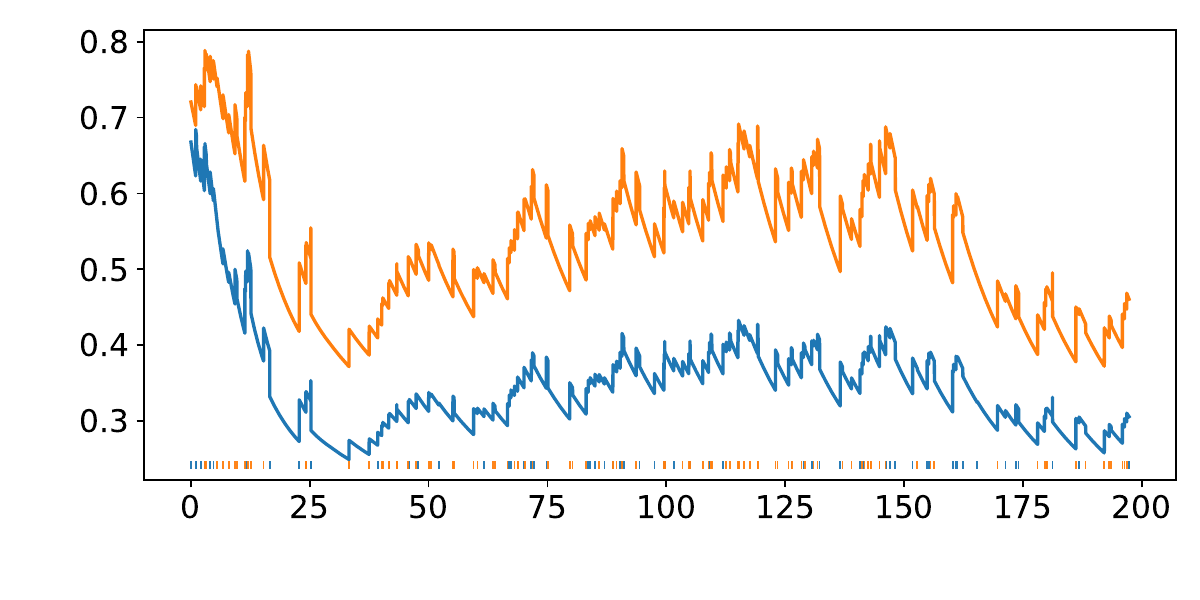}}

  \vspace{-8pt}

  \subfloat[THP]{\includegraphics[width=0.5\columnwidth]{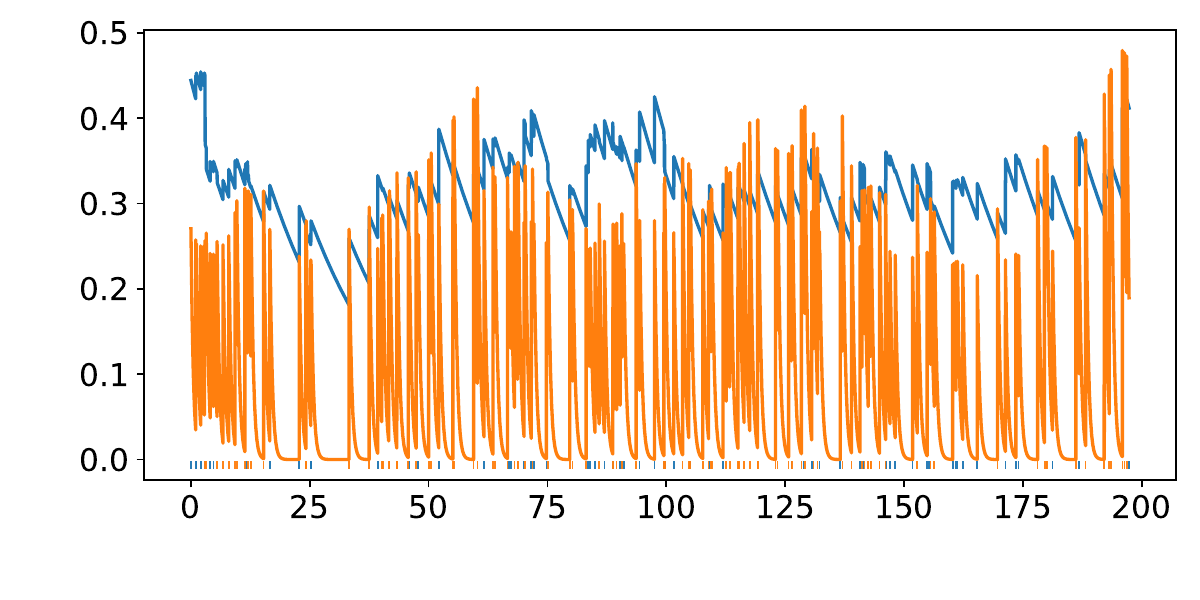}}
  \hfill
  \subfloat[RMTPP]{\includegraphics[width=0.5\columnwidth]{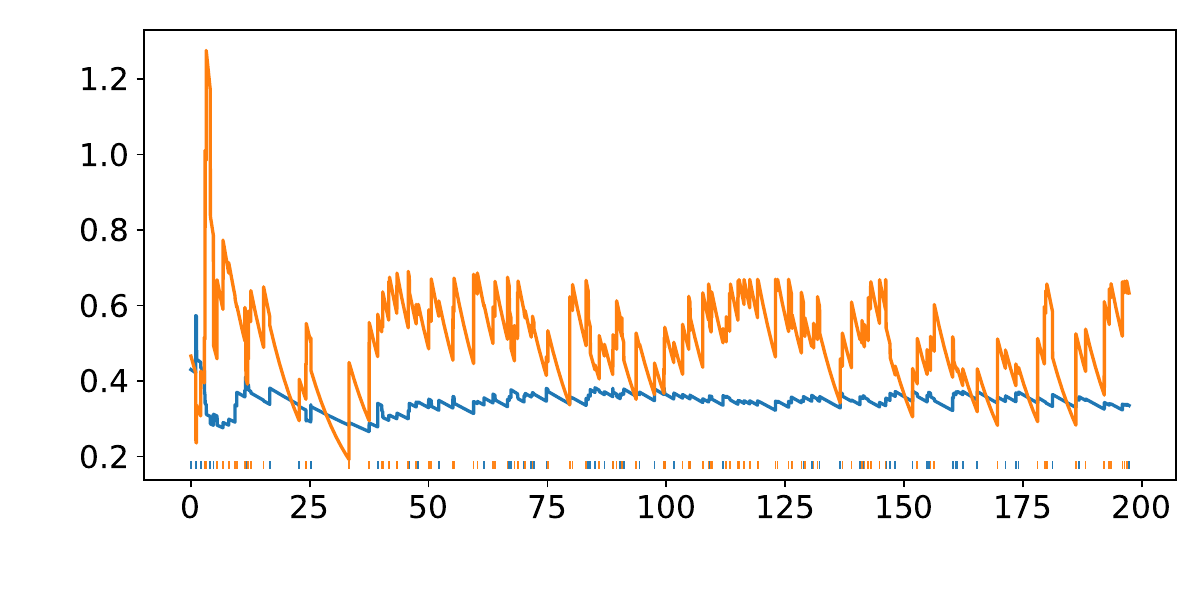}}

  \caption{Fitted conditional intensity for the simulation data with six methods in EasyTPP. Each model is trained until convergence based on validation log-likelihood.}
  \label{fig:temp_biv_simulation1}
\end{figure}

\subsection{Temporal NHP method}
\label{sec:temporalnhp}

Let us now discuss essence of temporal NHP method in order to prepare for the spati-temporal extension in the next Section. 
From \cite{mei2017neural} in the temporal setting, the conditional intensity for event type $k$ at time $t$ is given by
$$\lambda_k(t) = f_k(\bm w_k^T\mathbf{h}(t)), $$
where $\bm w_k \in \mathbb{R}^D$ is a trainable weight vector for event type $k$ and $\mathbf{h}(t) \in \mathbb{R}^D$ is the continuous hidden state, serving as a learned sufficient statistic of the event history $\mathcal{H}_t$. The function $f_k(\cdot)$ is a positive link function such as softplus. The hidden state $\mathbf{h}(t)$ is derived from the following continuous-time LSTM structure.
\begin{equation}\label{eq:temporal_hidden}
  \mathbf{h}(t) = \mathbf{o}_{i} \;\odot\;\bigl[\,2\,\sigma\bigl(2\,\mathbf{c}(t)\bigr) - 1\bigr],
  \quad
  t \in (t_{i-1},\,t_i],
\end{equation}
where $\sigma(\cdot)$ is the sigmoid function, $\odot$ is elementwise product. $\mathbf{o}_{i-1}$ is the LSTM’s output gate immediately after event $i-1$, which determines how much past information is exposed to the intensity function. $\mathbf{c}(t)$ is memory cell from continuous-time LSTM where it encodes how past events influence the likelihood of future events at time $t$.

At event times, the memory cell, $\mathbf c(t)$, is updated discretely using standard LSTM gating equations. Between events, each dimension of the memory cell, $c_d(t)$ ($d=1, \ldots, D$), evolves deterministically according to an exponential decay toward a learned steady state. Specifically, for $t \in (t_{i-1},\,t_i]$, the $d$-th component of the memory cell satisfies 
\begin{equation}\label{eq:temporal_decay}
  c_d(t) = \bar c_{d,i-1} + (c_{d,i-1} - \bar c_{d,i-1}) e^{-\delta_{d,i-1}(t - t_{i-1})},
\end{equation}
where $d = 1,\dots,D$.
Here, $c_{d,i-1}$ denotes the value of the $d$-th memory cell immediately after the (i-1)-th event, while $\bar c_{d,i-1}$ is the corresponding steady-state target toward which the cell decays in the absence of further events. The decay rate $\delta_{d,i-1} > 0$ controls how quickly past information is forgotten along dimension d; larger values correspond to faster decay.

Training is carried out by maximizing the log-likelihood of observed sequences, requiring evaluation of both event log-intensities and integrated intensities over inter-event times.
This architecture enables NHP to generalize classical self-exciting processes, while capturing richer temporal dependencies, long-term memory, and both excitatory and inhibitory interactions among events.

\section{Proposed Methodology}

\label{sec:TNHP}

\noindent We now propose our method by extending the NHP method to multivariate spatio-temporal Hawkes processes. We associate each event with a spatial location and consider a sequence $\{(k_i, t_i, \bm s_i)\}_{i=1}^n$, where $\bm s_i \in \mathbb{R}^d$ denotes the location of event $i$ (with $d=2$ for planar spatial data). Let
$\mathcal H_t = \{(k_i,t_i,\bm s_i): t_i < t\}$ denote the spatio-temporal history prior to time $t$. The conditional intensity is defined as $\lambda_k(\bm s,t | \mathcal H_t)$, representing the instantaneous rate at which an event of type $k$ occurs at location $s$ and time $t$, given the past history. In a multivariate spatio-temporal Hawkes process, the conditional intensity is specified as
$$\lambda_k(\bm s,t | \mathcal H_t) = \mu_k(\bm s, t) + \sum_{i: t_i < t} \phi_{k, k_i}(\bm s - \bm s_i,\, t - t_i),$$
where $\mu_k(\bm s, t) \ge 0$ is a baseline intensity for event type $k$, and $\phi_{k, k_i}(\Delta \bm s, \Delta t)$ is a spatio-temporal triggering kernel that characterizes how a past type $k_i$ event at displacement $\Delta \bm s =\bm s - \bm s_i$ and time lag $\Delta t = t - t_i$ modulates the future occurrence rate of type $k$ events. 

For notational simplicity, we omit the explicit conditioning on the event history $\mathcal{H}_t$ in conditional intensity functions throughout this section. As demonstrated in Section~\ref{sec:comparison}, NHP is flexible to capture various bivariate triggering structure in the temporal case, compared to five other methods from EasyTPP. With flexible continuous-time RNN structures, it is able to account for complex changes in triggering structures over time, unlike fixed parametric intensity functions in multivariate Hawkes process models \cite{yuan2019multivariate, jun2024flexible, siviero2024flexible}. The hidden state evolves according to a continuous-time LSTM between events, which allows the model to capture both excitation and inhibition effects as well as long-term temporal dependencies. 

\subsection{Multivariate Spatio-Temporal Neural Hawkes Process} 

\noindent We now extend the NHP method discussed in Section~\ref{sec:temporalnhp} to accommodate spatio-temporal point patterns. We refer to the existing method presented in Section~\ref{sec:temporalnhp} as Multivariate Temporal NHP (MTNHP) and the proposed method here as Multivariate Spatio-Temporal NHP (MSTNHP). The extension is straightforward in that we now consider spatial locations of events in addition to event types and event times, and the memory cell and hidden state functions are defined over both space and time. In \cite{mei2017neural}, the history ($\mathcal{H}_t$) is transformed into $\bm h(t)$, which serves as a sufficient statistics of the history. We incorporate spatial information into $\bm h(\bm s, t)$ to take a similar role as $\bm h(t)$ in \eqref{eq:temporal_hidden} as follows.
\[\begin{aligned}
    \lambda_k(\bm s, t) &= f_k(\bm w_k^T\bm h(\bm s,t))  \\
    \bm h(\bm s, t) &= \bm o_i \odot (2\sigma(2\bm c(\bm s, t))-1),
\end{aligned}\]
where k $ \in \{1,\dots,K\} $ indexes event types, $\bm w_k \in \mathbb{R}^D$ is a trainable weight vector for type $k$, with D denoting the dimension of $\bm h(\bm s,t)$, $\bm o_i$ is a counterpart of the output gate in LSTM, $\sigma(\cdot)$ is the sigmoid function and $\bm c(\bm s, t)$ represents the continuous counterpart of the memory cell in LSTM. Finally, the activation $f_k(\cdot)$ is specified as a softplus function, \[   f_k(x)    \;=\;    \tau_k\,\log\bigl(1 + \exp(x/\tau_k)\bigr),   \qquad    \tau_k > 0, \] that ensures $\lambda_k(\mathbf{s},t) > 0$. In our experiments, we set $\tau_k=1$ for all event types.

One of the key changes here from the temporal case is the cell-decay in \eqref{eq:temporal_decay} and now $d$-th component of the spatio-temporal memory cell, $c_d(\mathbf{s},t)$, is defined as 

\begin{align}\label{eq:spatio_temporal_decay}
  c_d(\mathbf{s},t) =& \bar c_{d,i-1}+\\ & (c_{d,i-1} - \bar c_{d,i-1}) e^{-\delta^{(t)}_{d,i-1}(t - t_{i-1}) - \delta^{(s)}_{d,i-1}\|\mathbf{s} - \mathbf{s}_{i-1}\|},\nonumber 
\end{align}
where $\mathbf{s}_{i-1}\in\mathbb{R}^2$ is the location of event $i-1$, $\delta^{(t)}_{d,i-1} > 0$ is the learned time‐decay rate for cell $d$ after event $i-1$, $\delta^{(s)}_{d,i-1} > 0$ is the learned spatial‐decay rate for cell $d$ after event $i-1$, and $\|\mathbf{s} - \mathbf{s}_{i-1}\|$ is the Euclidean distance between location $\mathbf{s}$ and the previous event’s location $\mathbf{s}_{i-1}$.
See Figure~\ref{fig:mstnhp_architecture} for the illustration of how the method works for a bivariate spatio-temporal point process for the demonstration of its flexible triggering structure. 

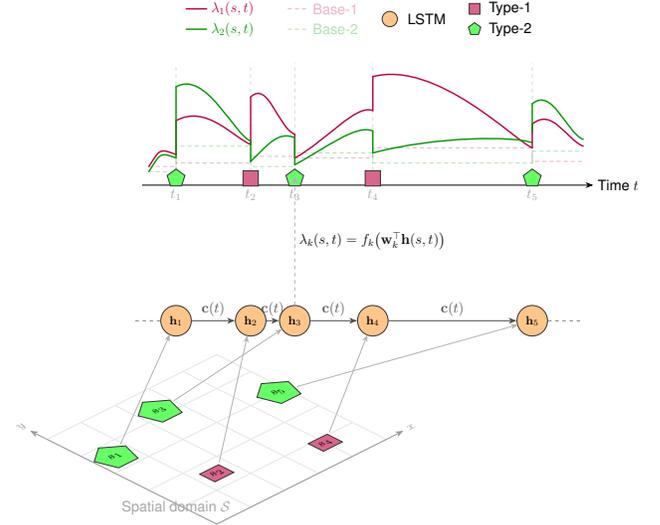
\begin{figure}[hbpt!]
    \centering
    \resizebox{0.96\columnwidth}{!}{%
        \begin{tikzpicture}[
    node distance=2.2cm,
    >=Stealth,
    lstm/.style={circle, draw=black!80, fill=orange!45, minimum size=9mm, thick},
    type1/.style={rectangle, fill=purple!60, minimum size=4.5mm, draw=black!80},
    type2/.style={regular polygon, regular polygon sides=5, fill=green!60, minimum size=5.5mm, draw=black!80},
    intensity1/.style={purple, very thick, line cap=round, line join=round},
    intensity2/.style={green!60!black, very thick, line cap=round, line join=round},
    base1/.style={purple, dashed, thick, opacity=0.3, line cap=round},
    base2/.style={green!60!black, dashed, thick, opacity=0.3, line cap=round},
    grid_style/.style={gray!20, very thin},
    smalltxt/.style={font=\sffamily\large},
    labeltxt/.style={font=\sffamily\large\bfseries}
]

\begin{scope}[yshift=-6.0cm, xshift=1.2cm, every node/.append style={yslant=0.5,xslant=-1}, yslant=0.5,xslant=-1]
    \draw[grid_style] (0,0) grid (5,5);
    \draw[thick, ->, gray!80] (0,0) -- (5.5,0) node[right] {$x$};
    \draw[thick, ->, gray!80] (0,0) -- (0,5.5) node[above] {$y$};
    \node[type2] (ev1_s) at (0.5, 3.5) {\(s_1\)}; 
    \node[type1] (ev2_s) at (1.5, 1.5) {\(s_2\)};
    \node[type2] (ev3_s) at (2.5, 4.2) {\(s_3\)};
    \node[type1] (ev4_s) at (4.0, 0.8) {\(s_4\)};
    \node[type2] (ev5_s) at (4.8, 3.0) {\(s_5\)};
\end{scope}
\node[smalltxt, gray!70] at (0,-5.5) {Spatial domain $\mathcal{S}$};

\foreach \i/\pos in {1/0, 2/2.2, 3/3.5, 4/5.8, 5/10.5} {
    \node[lstm] (h\i) at (\pos,0) {$\mathbf{h}_\i$};
}
\foreach \i/\j in {1/2, 2/3, 3/4, 4/5} {
    \draw[->, thick, black!70] (h\i) -- node[above, smalltxt] {$\mathbf{c}(t)$} (h\j);
}
\draw[dashed, gray] (-1.2,0) -- (h1); \draw[dashed, gray] (h5) -- (12.0,0);
\foreach \i in {1,2,3,4,5} { \draw[->, gray!60, shorten <=2pt] (ev\i_s) -- (h\i); }

\begin{scope}[yshift=4.0cm]
    
    \draw[->, thick] (-1,0) -- (12.3,0) node[right, smalltxt] {Time $t$};

    \coordinate (E1) at (0,0); \coordinate (E2) at (2.2,0); \coordinate (E3) at (3.5,0); \coordinate (E4) at (5.8,0); \coordinate (E5) at (10.5,0);

    \foreach \P/\lab in {E1/$t_1$,E2/$t_2$,E3/$t_3$,E4/$t_4$,E5/$t_5$}{

        \draw[gray!30, dashed, thick] (\P) -- ++(0,3.6);

        \node[smalltxt, gray!70, anchor=north] at (\P) {\lab};
    }

    \node[type2] at ($(E1)+(0,0.20)$) {}; \node[type1] at ($(E2)+(0,0.20)$) {}; \node[type2] at ($(E3)+(0,0.20)$) {}; \node[type1] at ($(E4)+(0,0.20)$) {}; \node[type2] at ($(E5)+(0,0.20)$) {};

    \draw[base1] (-0.8,0.55) -- (0,0.55) |- (2.2,0.65) |- (3.5,0.85) |- (5.8,0.80) |- (10.5,1.10) |- (12,0.70);

    \draw[base2] (-0.8,0.40) -- (0,0.40) |- (2.2,1.15) |- (3.5,0.55) |- (5.8,0.95) |- (10.5,0.65) |- (12,1.00);

    \newcommand{\greatshape}[6]{

        \draw[#1] (#2,#4) .. controls (#2+0.4*#3-#2*0.4, #4+0.5) and (#3-0.5, #5+0.2) .. (#3,#5);
    }

    \greatshape{intensity1}{-0.8}{0.0}{0.55}{0.90}{0.5}

    \draw[intensity1] (0.0,0.90) -- (0.0,1.90); 

    \greatshape{intensity1}{0.0}{2.2}{1.90}{1.20}{0.8}

    \draw[intensity1] (2.2,1.20) -- (2.2,2.60); 

    \greatshape{intensity1}{2.2}{3.5}{2.60}{1.50}{0.4}

    \draw[intensity1] (3.5,1.50) -- (3.5,0.80); 

    \greatshape{intensity1}{3.5}{5.8}{0.80}{2.20}{0.9}

    \draw[intensity1] (5.8,2.20) -- (5.8,3.20); 

    \greatshape{intensity1}{5.8}{10.5}{3.20}{1.10}{0.7}

    \draw[intensity1] (10.5,1.10) -- (10.5,1.80); 

    \greatshape{intensity1}{10.5}{12.0}{1.80}{1.15}{0.5}

    \greatshape{intensity2}{-0.8}{0.0}{0.40}{0.80}{0.5}

    \draw[intensity2] (0.0,0.80) -- (0.0,2.90); 

    \greatshape{intensity2}{0.0}{2.2}{2.90}{1.30}{0.8}

    \draw[intensity2] (2.2,1.30) -- (2.2,0.70); 

    \greatshape{intensity2}{2.2}{3.5}{0.70}{1.40}{0.4}

    \draw[intensity2] (3.5,1.40) -- (3.5,0.60); 

    \greatshape{intensity2}{3.5}{5.8}{0.60}{1.60}{0.9}

    \draw[intensity2] (5.8,1.60) -- (5.8,0.95); 

    \greatshape{intensity2}{5.8}{10.5}{0.95}{1.25}{0.7}

    \draw[intensity2] (10.5,1.55) -- (10.5,2.40); 

    \greatshape{intensity2}{10.5}{12.0}{2.40}{1.35}{0.5}

    \begin{scope}[shift={(0.3, 5.2)}]
        \draw[intensity1] (0,0) -- (0.6,0) node[right, smalltxt] {$\lambda_1(s,t)$};
        \draw[intensity2] (0,-0.6) -- (0.6,-0.6) node[right, smalltxt] {$\lambda_2(s,t)$};
        \draw[base1] (3.0,0) -- (3.6,0) node[right, smalltxt] {Base-1};
        \draw[base2] (3.0,-0.6) -- (3.6,-0.6) node[right, smalltxt] {Base-2};
        
        \node[lstm, scale=0.5] at (6.0, -0.3) {}; \node[right=0.4cm, smalltxt] at (6.0, -0.3) {LSTM};
        \node[type1, scale=0.7] at (8.5, 0) {}; \node[right=0.3cm, smalltxt] at (8.5, 0) {Type-1};
        \node[type2, scale=0.7] at (8.5, -0.6) {}; \node[right=0.3cm, smalltxt] at (8.5, -0.6) {Type-2};
    \end{scope}
\end{scope}

\draw[->, gray!60, dashed, thick] (h3.north) -- ++(0,3.8) 
    node[midway, right, smalltxt, black] {$\lambda_k(s,t)=f_k\bigl(\mathbf{w}_k^\top\mathbf{h}(s,t)\bigr)$};

\end{tikzpicture}%
    }
   \caption{Architecture of the Proposed MSTNHP. The model processes spatio-temporal events $(t_i, \mathbf{s}_i)$ through a continuous-time LSTM, where intensities $\lambda_k$ and baselines jump at event times and decay smoothly between them.}
    \label{fig:mstnhp_architecture}
\end{figure}

Given a spatial domain $\mathcal{S}$ (e.g.\ a bounding box or polygon containing all events), the total intensity over all types and all locations is
\[
  \Lambda(t)
  \;=\;
  \int_{\mathcal{S}} \sum_{k=1}^K \lambda_k(\mathbf{s},t)\,d\mathbf{s},
\]
and the log‐likelihood of an observed sequence $\{(k_i,t_i,\mathbf{s}_i)\}_{i : t_i \le T}$ is
\begin{equation}\label{eq:log_likelihood}
  \ell = \sum_{i : t_i \le T} \log\bigl(\lambda_{k_i}(\mathbf{s}_i,\,t_i)\bigr)-\int_{t=0}^T \int_{\mathcal{S}} \sum_{k=1}^K \lambda_k(\mathbf{s},t)\,d\mathbf{s}\,dt.
\end{equation}
As in \cite{mei2017neural}, the double integral in \eqref{eq:log_likelihood} is approximated by Monte Carlo sampling over $[0,T]\times\mathcal{S}$.
We also follow the training procedure in \cite{mei2017neural}, including Monte Carlo sampling of the integral, optimizer settings and early stopping.


\subsection{Model Complexity and Sensitivity to Number of Parameters}
\label{sec:parameters}

\noindent In neural network architectures, model complexity is primarily governed by the number of hidden units $D$ in a layer. This  determines the representational capacity of the neural architecture: a larger $D$ increases the number of trainable parameters, allowing the model to capture more complex patterns, but at the cost of higher computational demand and a greater risk of overfitting.  Selecting an appropriate $D$ is therefore a critical hyperparameter tuning step.
It is important to emphasize that the primary goal of this work is not the development of a new neural architecture or an exhaustive architectural search. Instead, our focus lies in the effective mathematical and methodological integration of spatial and temporal components. Consequently, we utilize established neural structures to demonstrate the efficacy of our spatio-temporal formulation.

For our proposed spatio-temporal models, we followed the tuning guidelines outlined in \cite{xue2024easytpp} to determine a suitable hidden size. Constrained by available hardware resources, we explored a range of values to account for the increased complexity of modeling both geographical coordinates and event timings simultaneously. The final architecture was selected by choosing the hidden unit size that yielded the {highest\ validation\ set\ likelihood}, ensuring the model is powerful enough to capture the underlying patterns in the data while maintaining stable optimization and generalization.

As shown in the original work, \cite{mei2017neural} (e.g., the Retweets experiment), increasing the number of parameters generally leads to better performance on held-out data, indicating that the models benefit from greater representational capacity. However, this is not a strict monotonic relationship, as demonstrated in our own simulations where a smaller number of hidden units sometimes produced better results for certain synthetic settings. Ultimately, the number of hidden units ($D$) remains a crucial hyperparameter requiring careful tuning for each specific dataset and task to optimize the expressiveness of the model against the risk of overfitting.

\section{Simulation study}
\label{sec:simulation_setting}

\noindent We now show how MSTNHP performs with simulated data with various triggering structure. 
We consider a few commonly used spatio-temporal triggering structure in the literature including the framework of SMASH \cite{li2024beyond} and the one in \cite{zhang2020self}, to generate bivariate synthetic datasets. All simulations were done on a single NVIDIA Tesla V100 with 10~CPU cores per job.

\subsection{Simulation setting}

\noindent We consider four simulation settings (namely, Biv 1-4). For each of them, the spatial domain is a unit square and time interval is {[0,100]}. For all cases (Biv 1-4), we use constant baseline intensities with \(\mu_1 = \mu_2 = 0.1\).

The triggering structures of first three simulation settings (Biv 1-3) are given by \cite{li2024beyond}; the spatio-temporal triggering function is given by $$\phi_{k,l}(\Delta \mathbf{s}, \Delta t) 
= \alpha_{k l} \, h_{k,l}(\Delta \mathbf{s})\, g_{k,l}(\Delta t)$$ with  
\begin{align}\label{SMASH}
h_{k,l}(\Delta \mathbf{s}) = \frac{1}{2\pi\sigma_{kl}^2 }
\exp\!\left(-\frac{\|\Delta \mathbf{s}\|^2}{2\sigma_{kl}^2}\right) \mbox{ and } \nonumber\\ g_{k,l}(\Delta t) = \beta_{kl} \, e^{-\beta_{kl} \Delta t} \, \mathbf{1}_{\{\Delta t > 0\}}.
\end{align}
Table~\ref{tab:mle_estimates_synthetic} summarizes the parameter values for the triggering functions. 
They are chosen in a way that we can make various pairwise comparison between two of the settings. For instance, Biv 1 and Biv 2 have the same spatio-temporal triggering structure but have different cross-triggering structure, Biv 1 simply triggers marginally as well as jointly but Biv 2 has joint inhibition while marginally triggering. Biv 3 has longer triggering ranges compared to Biv 1.

\begin{table}[!hbpt]
\renewcommand{\arraystretch}{1.3}
\caption{Parameter values for the separable triggering structure in \eqref{SMASH}}
\label{tab:mle_estimates_synthetic}
\centering
\footnotesize
\begin{tabular}{lccc}
\toprule
Parameters & Biv 1 & Biv 2 & Biv 3 \\
\midrule
$\alpha_{11}, \alpha_{12},$ & 0.25, 0.1, & 0.25, -0.1, & 0.25, 0.1, \\
$\alpha_{21}, \alpha_{22}$ & 0.1, 0.25  & -0.1, 0.25  & 0.1, 0.25  \\
\midrule
$\beta_{11}, \beta_{12},$  & 0.3, 0.3,  & 0.3, 0.3,   & 0.1, 0.1,  \\
$\beta_{21}, \beta_{22}$  & 0.3, 0.3   & 0.3, 0.3    & 0.1, 0.1   \\
\midrule
$\sigma_{11}^2, \sigma_{12}^2,$ & 0.5, 0.5, & 0.5, 0.5, & 0.5, 0.25, \\
$\sigma_{21}^2, \sigma_{22}^2$ & 0.5, 0.5  & 0.5, 0.5  & 0.25, 0.5  \\
\bottomrule
\end{tabular}
\end{table}

Another bivariate spatio-temporal triggering function considered (Biv 4), similar to the one used in  \cite{zhang2020self}, is given by: 
\begin{align}
    g_{1,1}(\Delta \mathbf{s}, \Delta t) &= 0.15(0.5+\Delta t)^{-1.3}e^{-2\|\Delta \mathbf{s}\|}, \nonumber \\
    g_{1,2}(\Delta \mathbf{s}, \Delta t) &= 0.03e^{-0.3\Delta t -2 \|\Delta \mathbf{s}\|}, \nonumber\\
    g_{2,1}(\Delta \mathbf{s}, \Delta t) &= \{0.05e^{-0.2\Delta t} + 0.16e^{-0.8\Delta t}\}e^{-2 \|\Delta \mathbf{s}\|}, \nonumber\\
    g_{2,2}(\Delta \mathbf{s}, \Delta t) &= \max(0, \tfrac{\sin\Delta t}{8})e^{-2 \|\Delta \mathbf{s}\|}, \, \Delta t \in [0,4]. \label{sim2}
\end{align}
This cross-triggering structure is more complex than those of Biv 1-3, especially regarding the temporal triggering structure.

\begin{table}[!hbpt]
\renewcommand{\arraystretch}{1.3}
\caption{Statistics of simulated datasets for biv 1--4}
\label{tab:dataset_stats}
\centering
\footnotesize
\setlength{\tabcolsep}{4pt}
\begin{tabular}{lrrrrrrr} 
\toprule
Dataset & \# Param. & \multicolumn{3}{c}{\# Event Tokens} & \multicolumn{3}{c}{Sequence Length} \\
\cmidrule(lr){3-5} \cmidrule(lr){6-8}
& & Train & Valid & Test & Min & Mean & Max \\
\midrule
Biv 1 & 20,960 & 86,520 & 4,692 & 4,900 & 60 & 97 & 134 \\
Biv 2 & 20,960 & 77,106 & 4,181 & 4,304 & 59 & 87 & 119 \\
Biv 3 & 20,960 & 87,313 & 4,726 & 4,944 & 60 & 97 & 135 \\
Biv 4 & 82,880 & 133,615 & 7,131 & 7,693 & 88 & 149 & 242 \\
\bottomrule
\end{tabular}
\end{table}

Table~\ref{tab:dataset_stats} provides the statistics of all simulated dataset as well as the number of parameters used to fit each dataset. Due to more complex spatio-temporal triggering structure of the model in Biv 4, more events are generated, and more parameters are used. 
We established a consistent training protocol across the four synthetic settings, Biv 1 through Biv 4. Each setting involved the simulation of 1,000 independent sequences. The full dataset was segmented using a 900:50:50 split into training, validation, and testing sets.
The final model checkpoint was selected based on the epoch that achieved the highest validation set likelihood (see section \ref{sec:supp_mstnhp}
 of the supplementary material for convergence plots). This ensures that any reported performance metrics on the independent test set are derived from the model that demonstrated optimal generalization capabilities during training, mitigating potential overfitting to the training data. 

\subsection{Simulation Results}

\noindent Figure~\ref{fig:biv1-4_images} shows, for all four bivariate simulation settings (Biv~1–4), the true temporal intensity functions together with the corresponding fitted intensities obtained from MSTNHP. In each case, the fitted curves closely track the overall shape and magnitude of the ground truth intensities, capturing both the baseline activity level and the bursts induced by self and cross excitation. For the purely excitatory configurations (Biv~1, Biv~3, and Biv~4, where all interaction kernels are nonnegative), the model reproduces the sharp increases in intensity following event clusters. For the mixed excitatory/inhibitory design (Biv~2, where the cross-type triggering coefficients $\alpha_{12}$ and $\alpha_{21}$ are negative), it also recovers the dampening of intensity in periods following inhibitory interactions. Small discrepancies are mainly confined to regions with very sparse events or at the boundaries of the observation window, where the effective amount of information is limited.

To assess how well the model recovers the spatial structure, we further visualize cumulative time-averaged spatial intensity maps at the final horizon $\tau =$ 100 for all four bivariate settings as in Figure~\ref{fig:biv1-4_spatial}. For each event type, we compute the cumulative mean intensity $\bar{\lambda}_k^{(\tau)}(s)$ by averaging $\lambda_k(t,s)$ over all times $t \le \tau$ at each spatial location $s$. The maps in the main text therefore represent the stabilized long-run spatial footprint over [0,100], while the full evolution over  horizons $\tau \in \{$20,40,60,80,100$\}$ for each setting and event type is provided in section \ref{sec:supp_mstnhp} of the supplementary material. The fitted maps closely reproduce the dominant spatial patterns of the ground truth process, indicating that the model not only matches the temporal dynamics but also captures the spatial intensity structure reasonably well.

\begin{figure}[!hbpt]
\centering
  \subfloat[Biv1 - true]{\includegraphics[width=0.5\columnwidth]{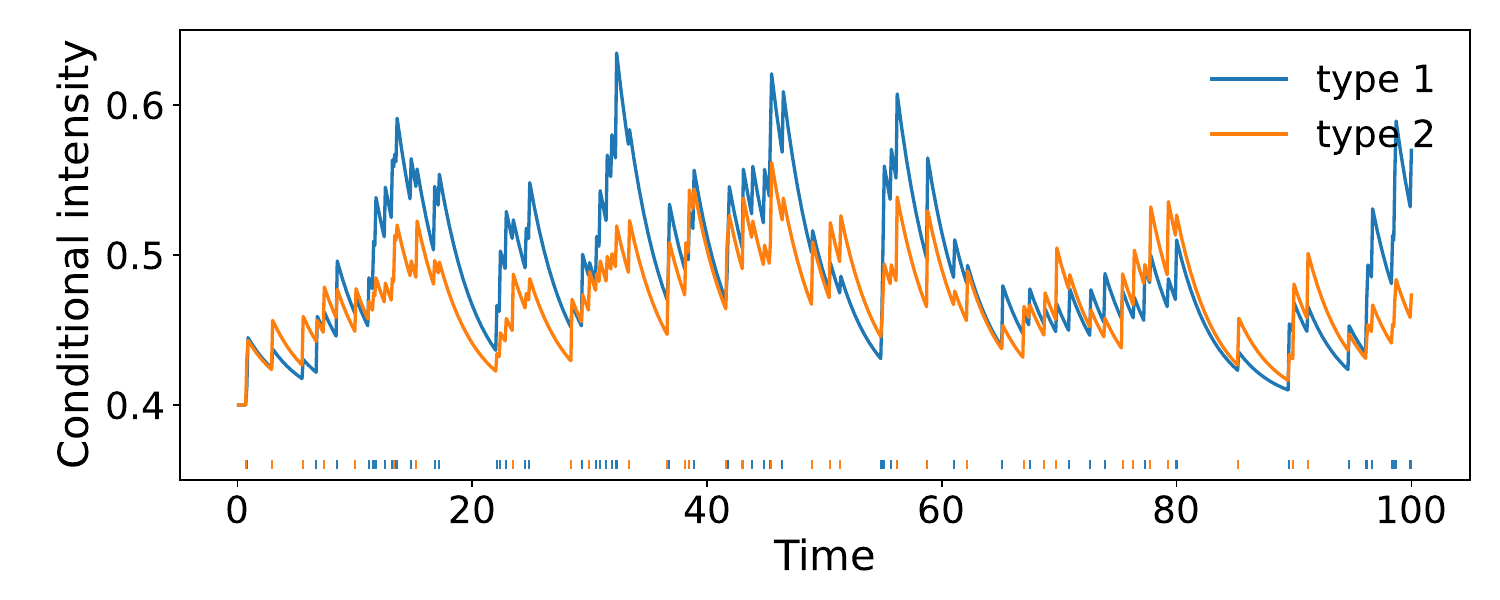}}
  \hfill
  \subfloat[Biv1 - fitted]{\includegraphics[width=0.5\columnwidth]{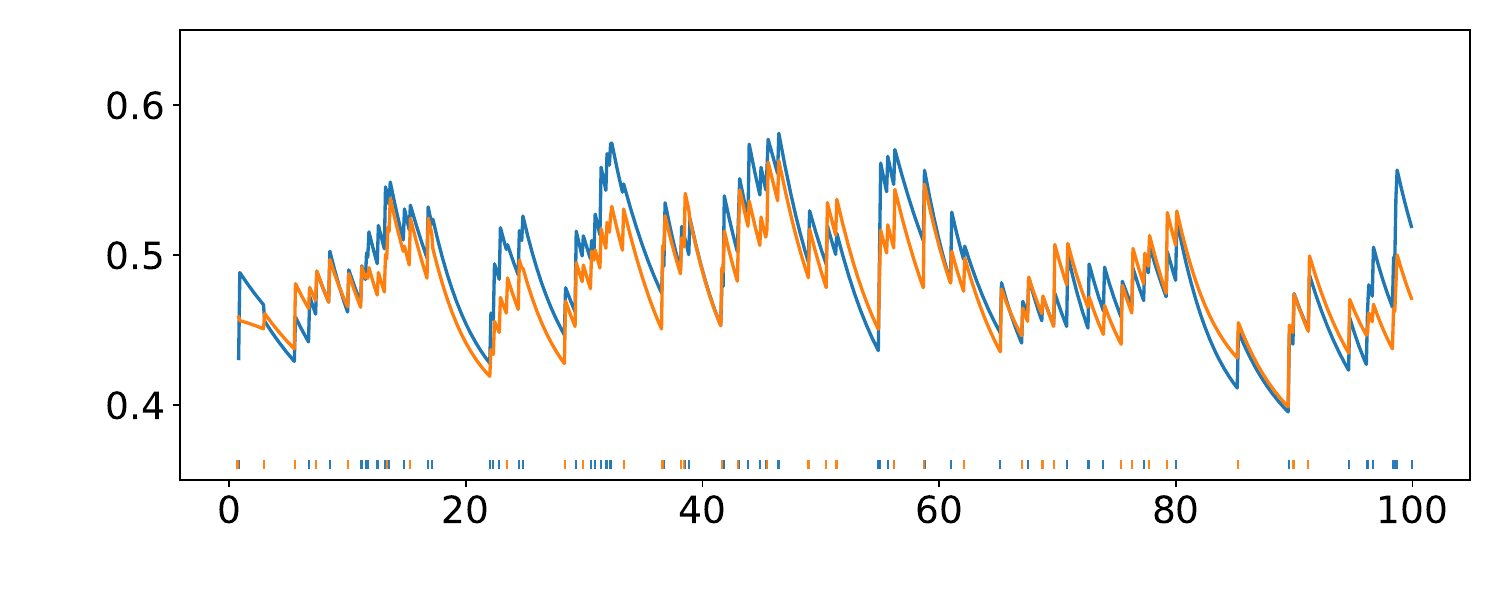}}
  
  \vspace{-8pt}

  \subfloat[Biv2 - true]{\includegraphics[width=0.5\columnwidth]{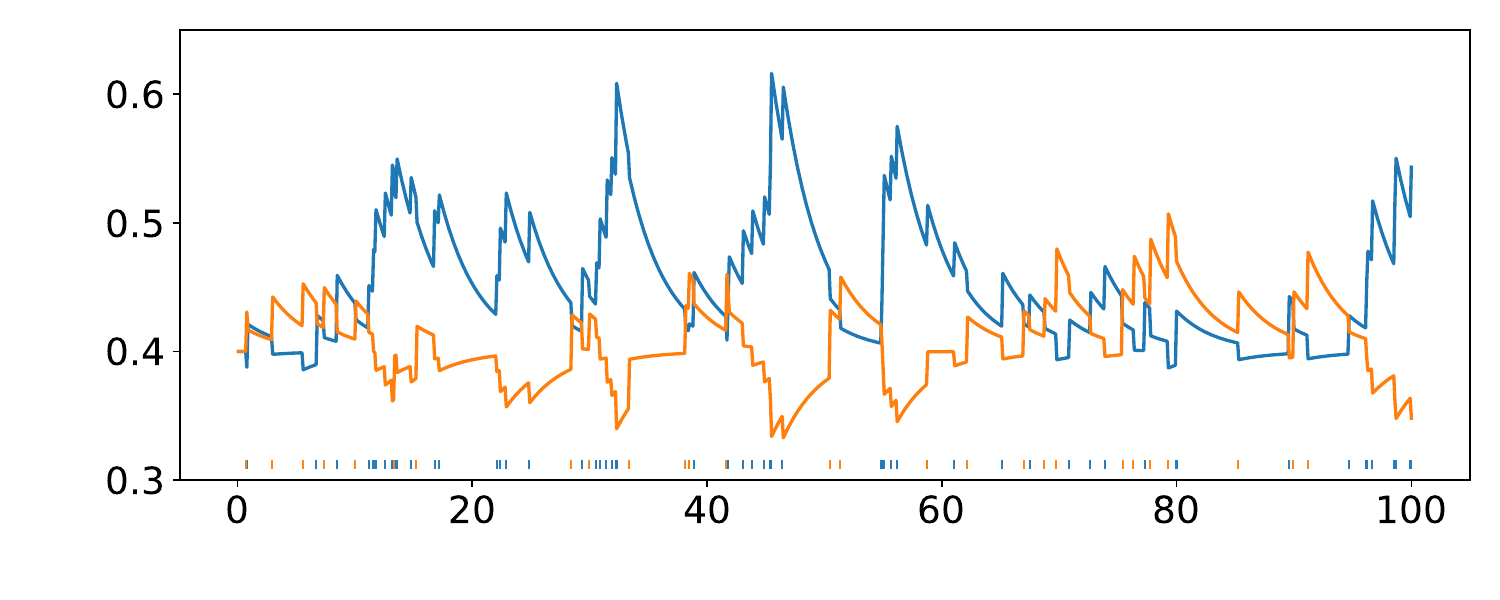}}
  \hfill
  \subfloat[Biv2 - fitted]{\includegraphics[width=0.5\columnwidth]{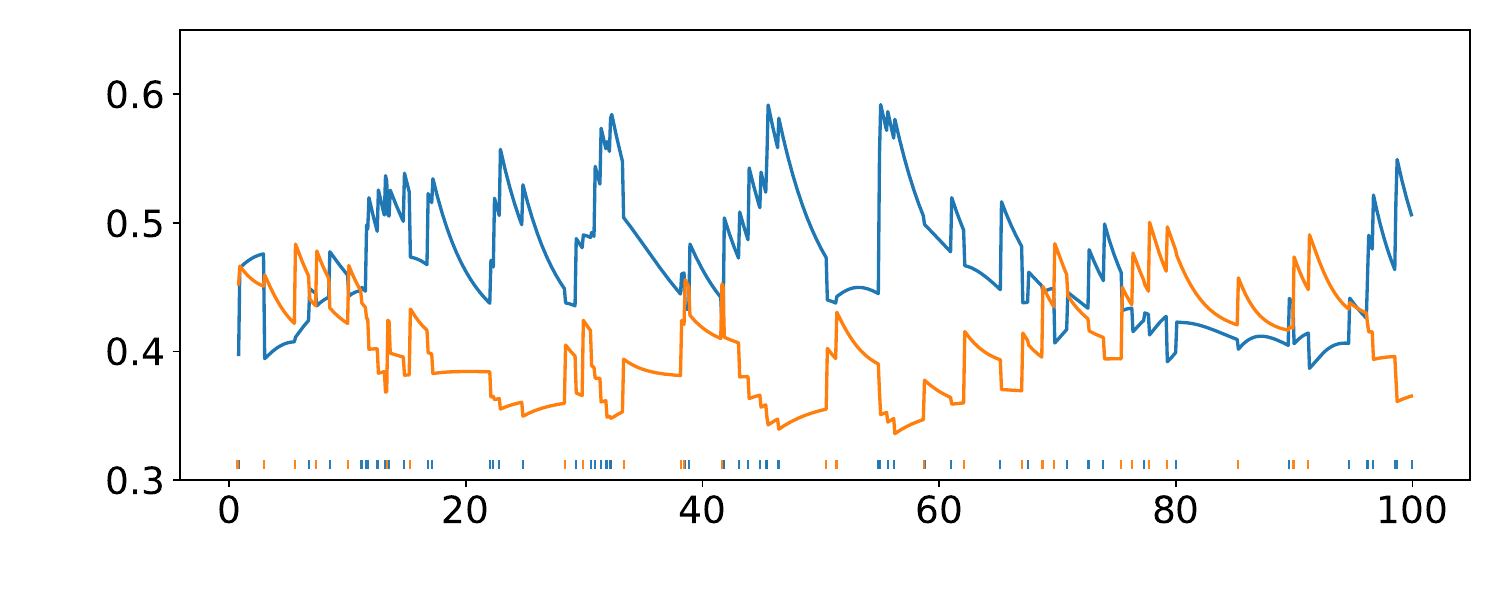}}

  \vspace{-8pt}

  \subfloat[Biv3 - true]{\includegraphics[width=0.5\columnwidth]{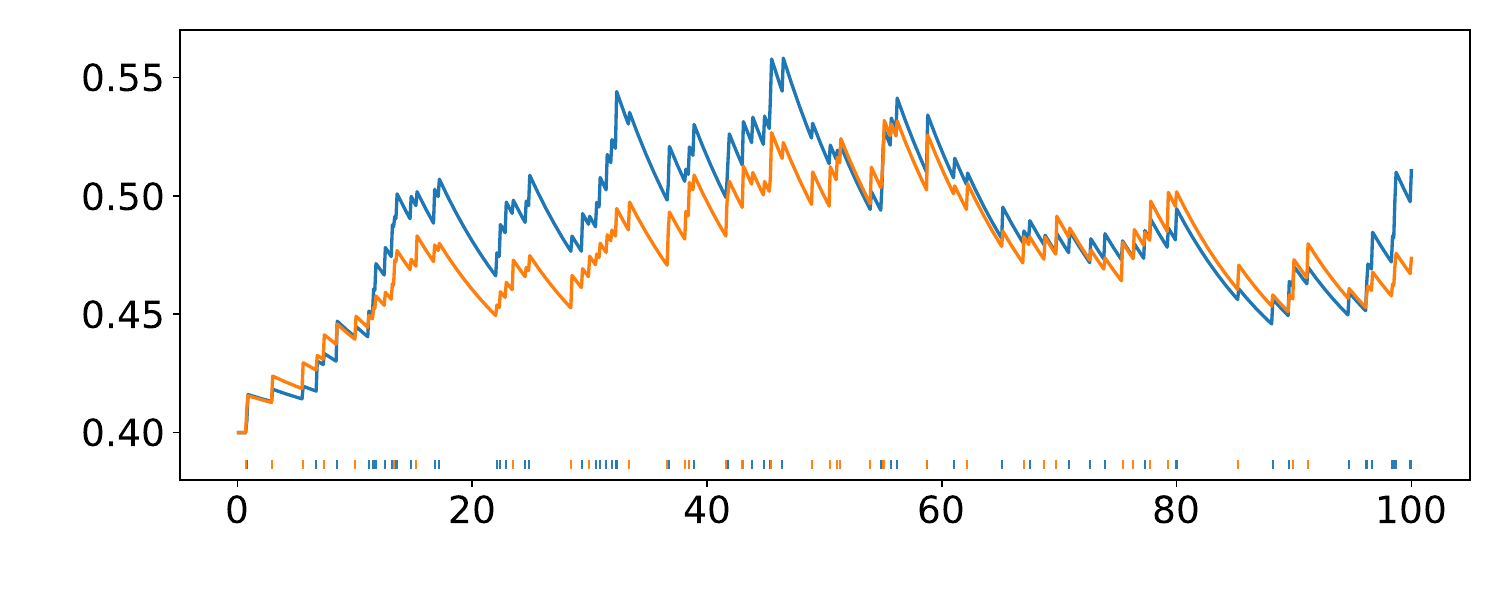}}
  \hfill
  \subfloat[Biv3 - fitted]{\includegraphics[width=0.5\columnwidth]{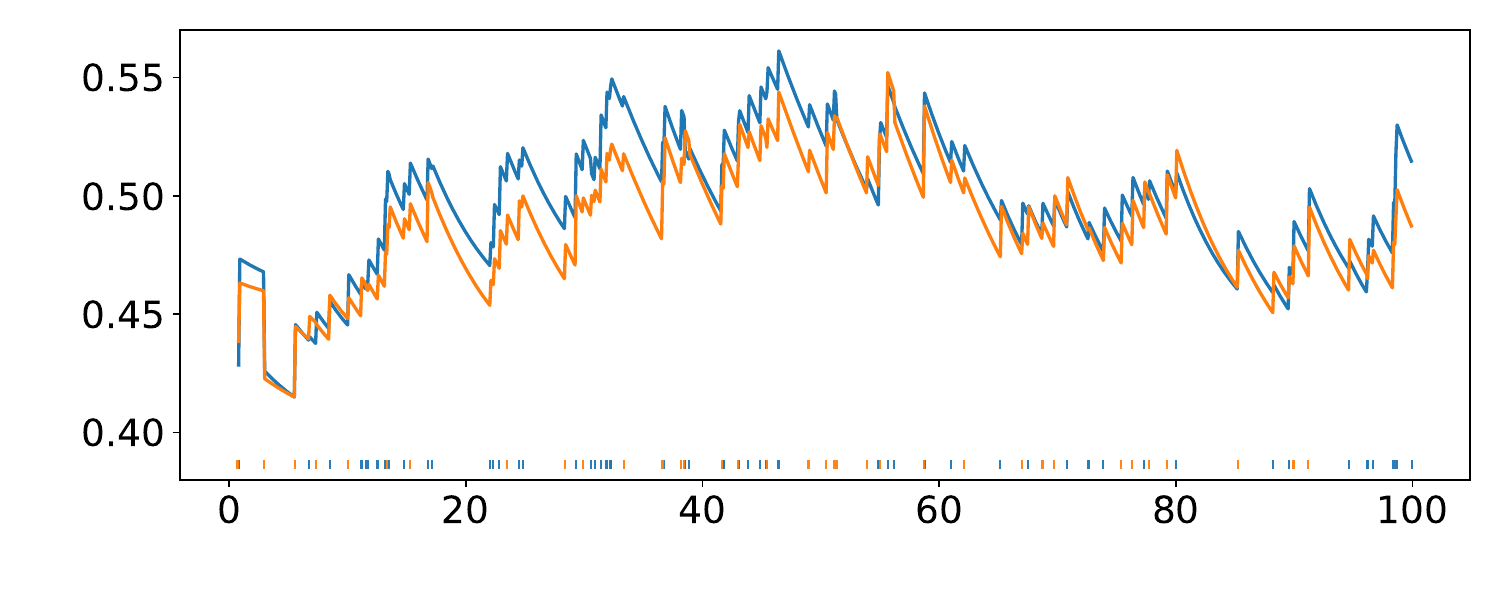}}

   \vspace{-8pt}

  \subfloat[Biv4 - true]{\includegraphics[width=0.5\columnwidth]{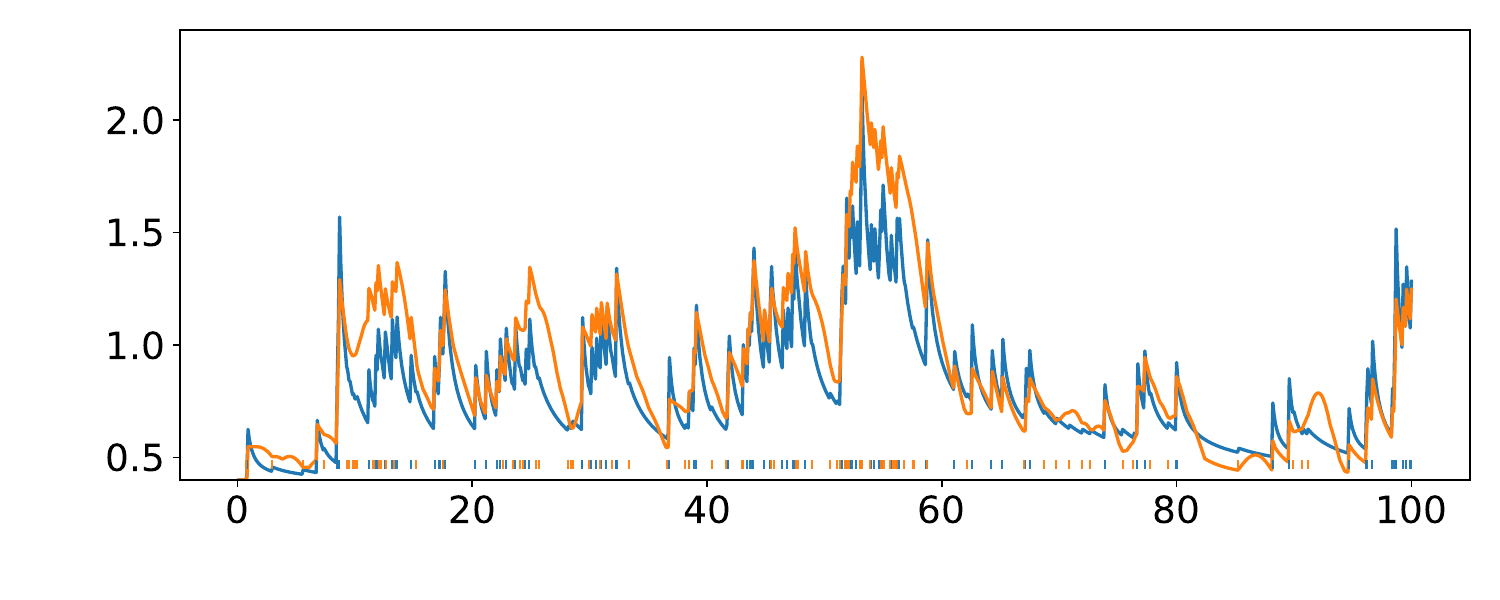}}
  \hfill
  \subfloat[Biv4 - fitted]{\includegraphics[width=0.5\columnwidth]{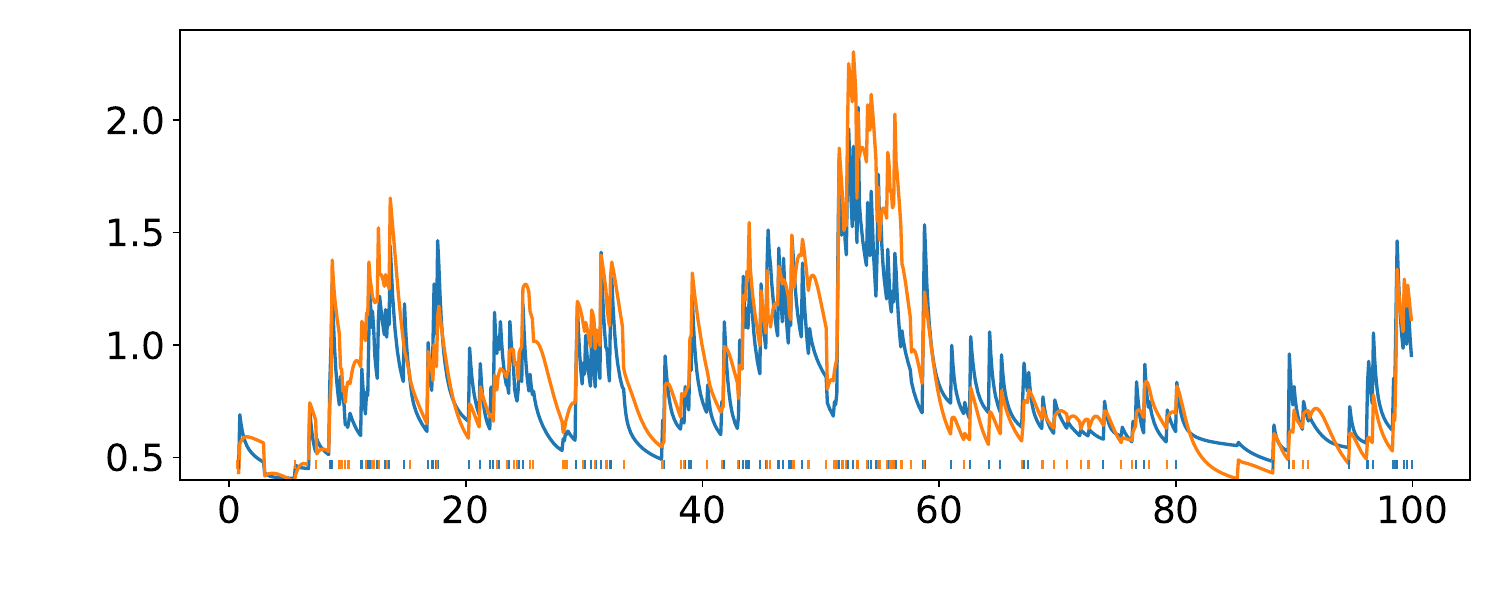}}

  \caption{True vs. Fitted temporal intensities for simulated datasets using parameter configurations from Table \ref{tab:mle_estimates_synthetic}.}
  \label{fig:biv1-4_images}
\end{figure}

\begin{figure}[!hbpt]
\centering
\subfloat{\includegraphics[width=0.24\columnwidth]{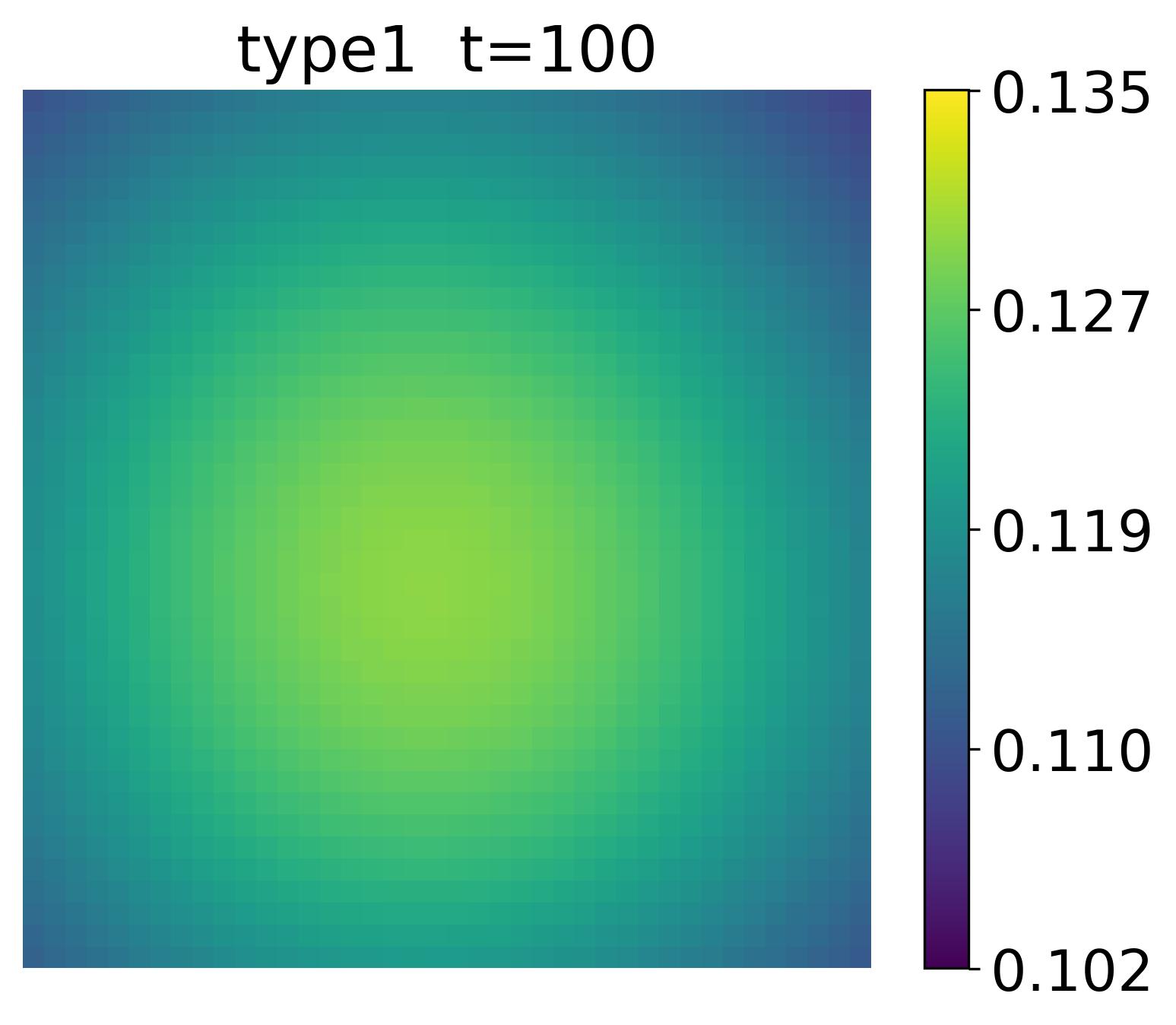}} \hspace{-2pt}
\subfloat{\includegraphics[width=0.24\columnwidth]{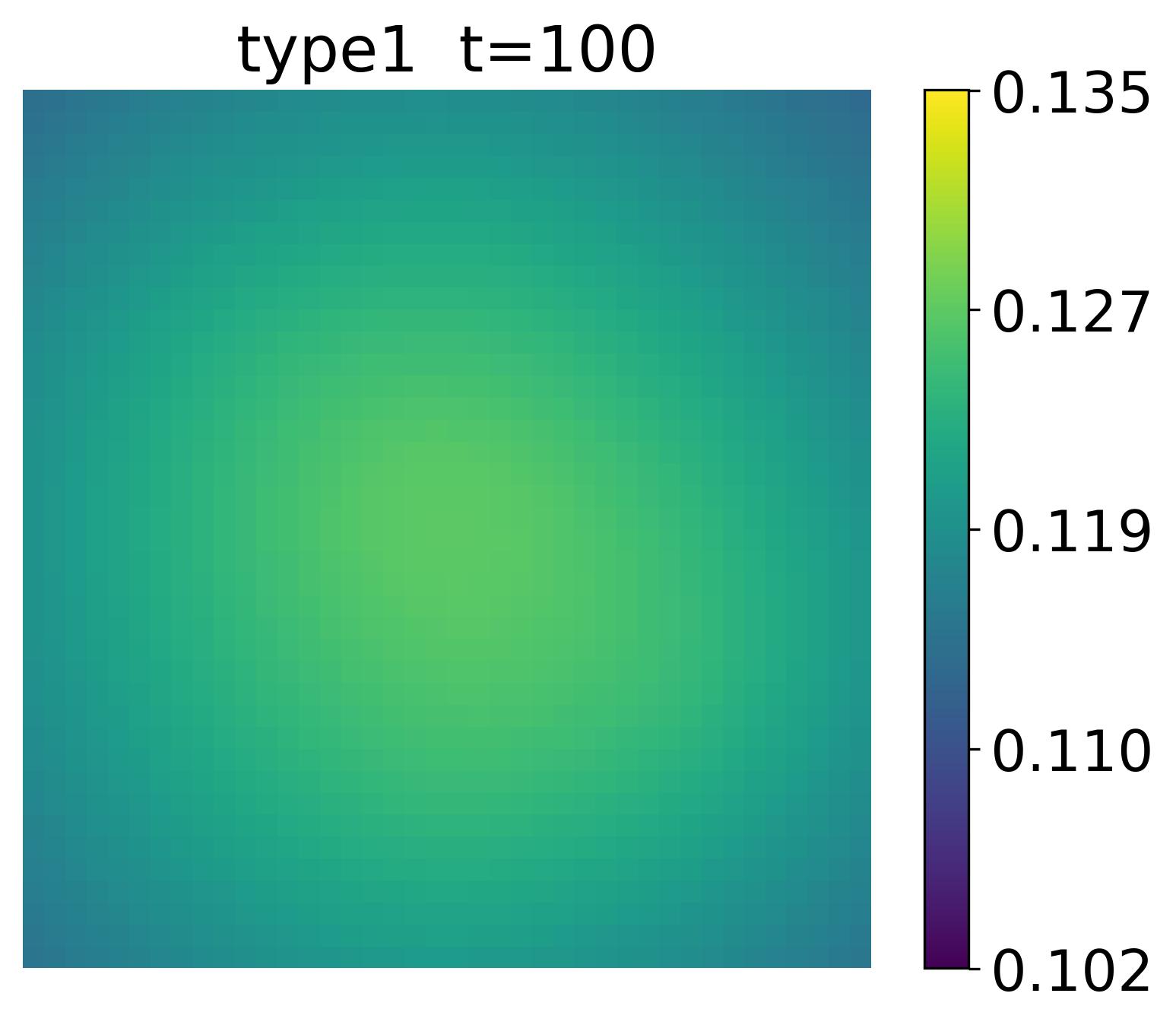}} \hspace{-2pt}
\subfloat{\includegraphics[width=0.24\columnwidth]{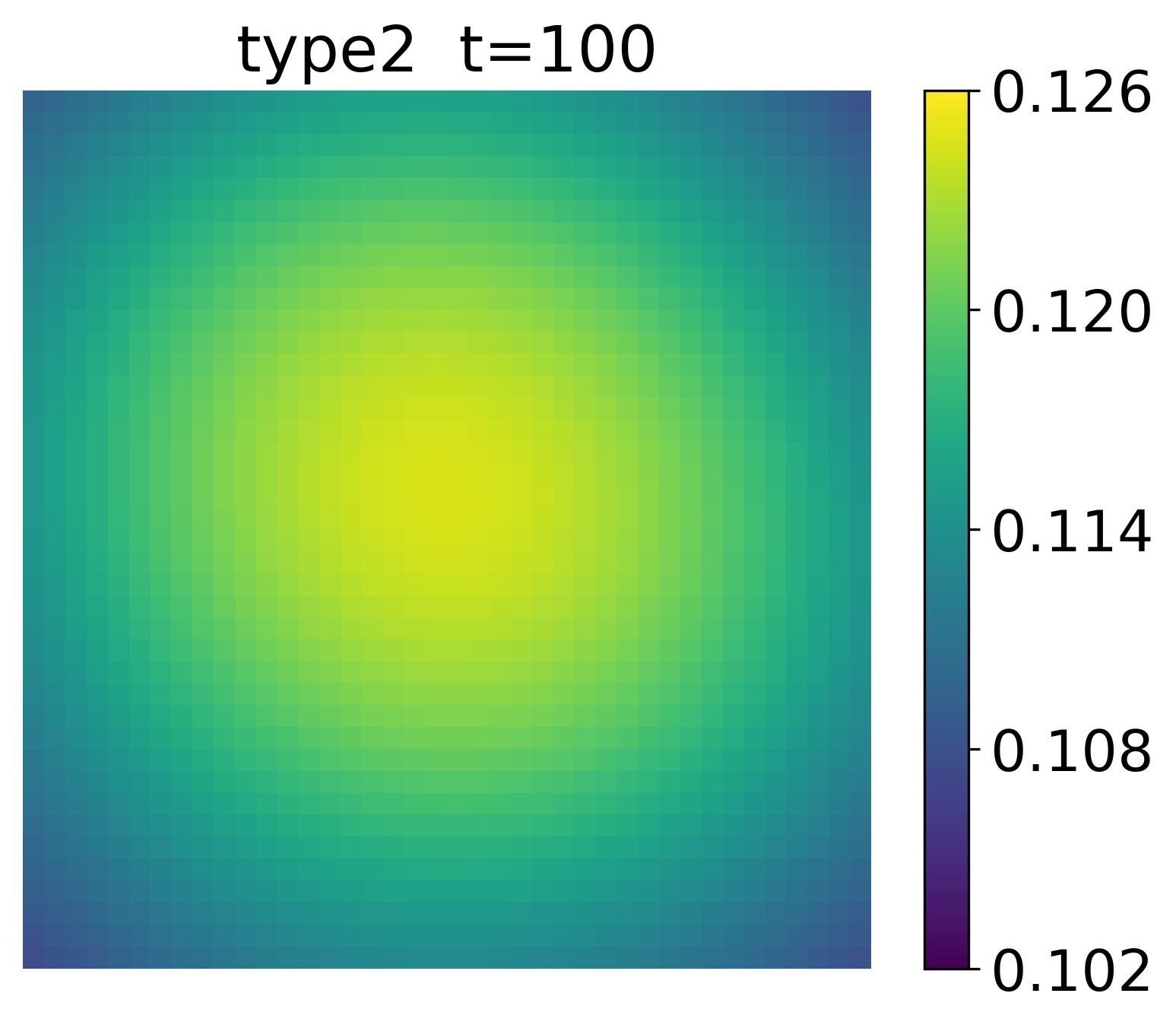}} \hspace{-2pt}
\subfloat{\includegraphics[width=0.24\columnwidth]{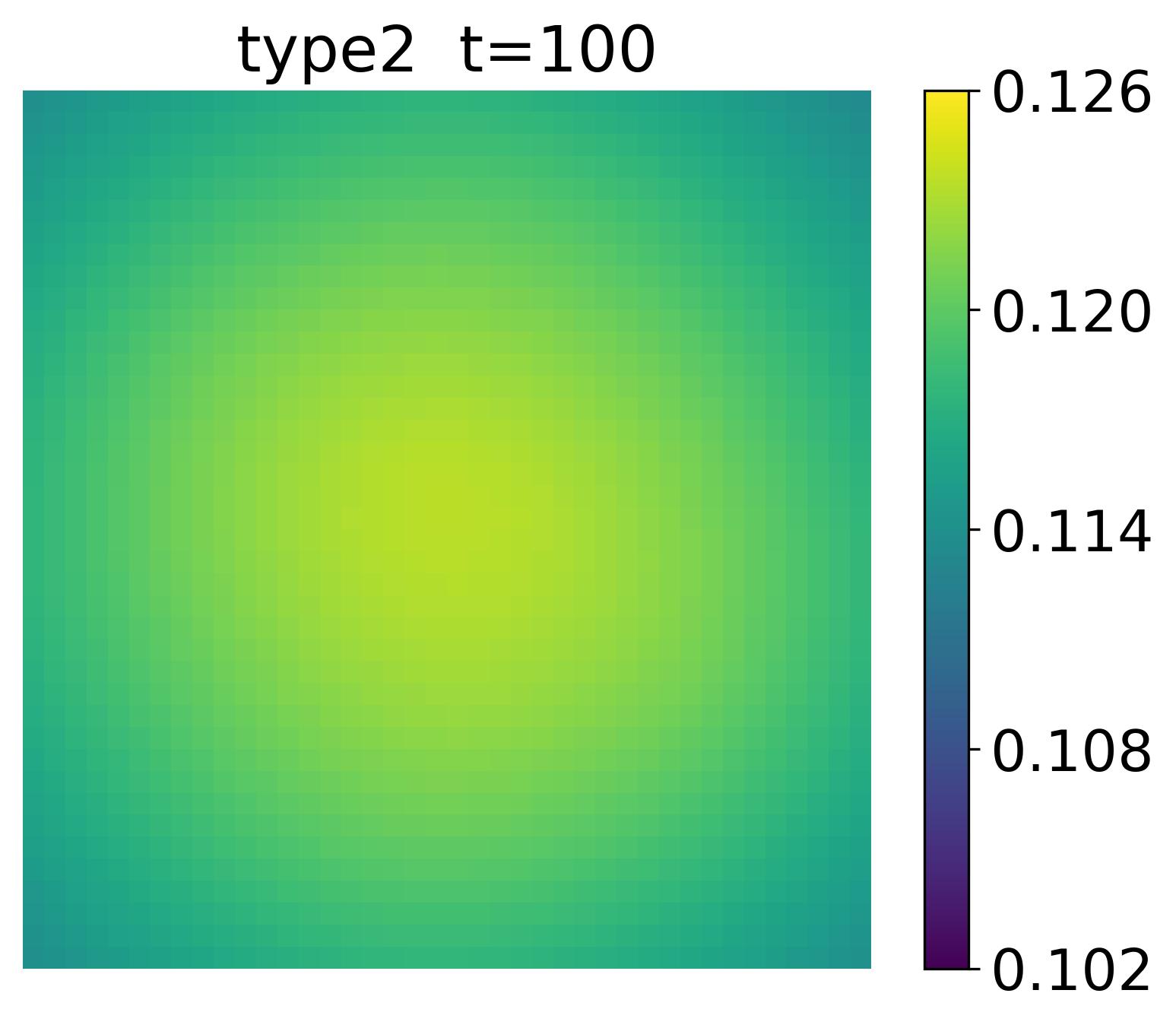}}

\vspace{-8pt} 

\subfloat{\includegraphics[width=0.24\columnwidth]{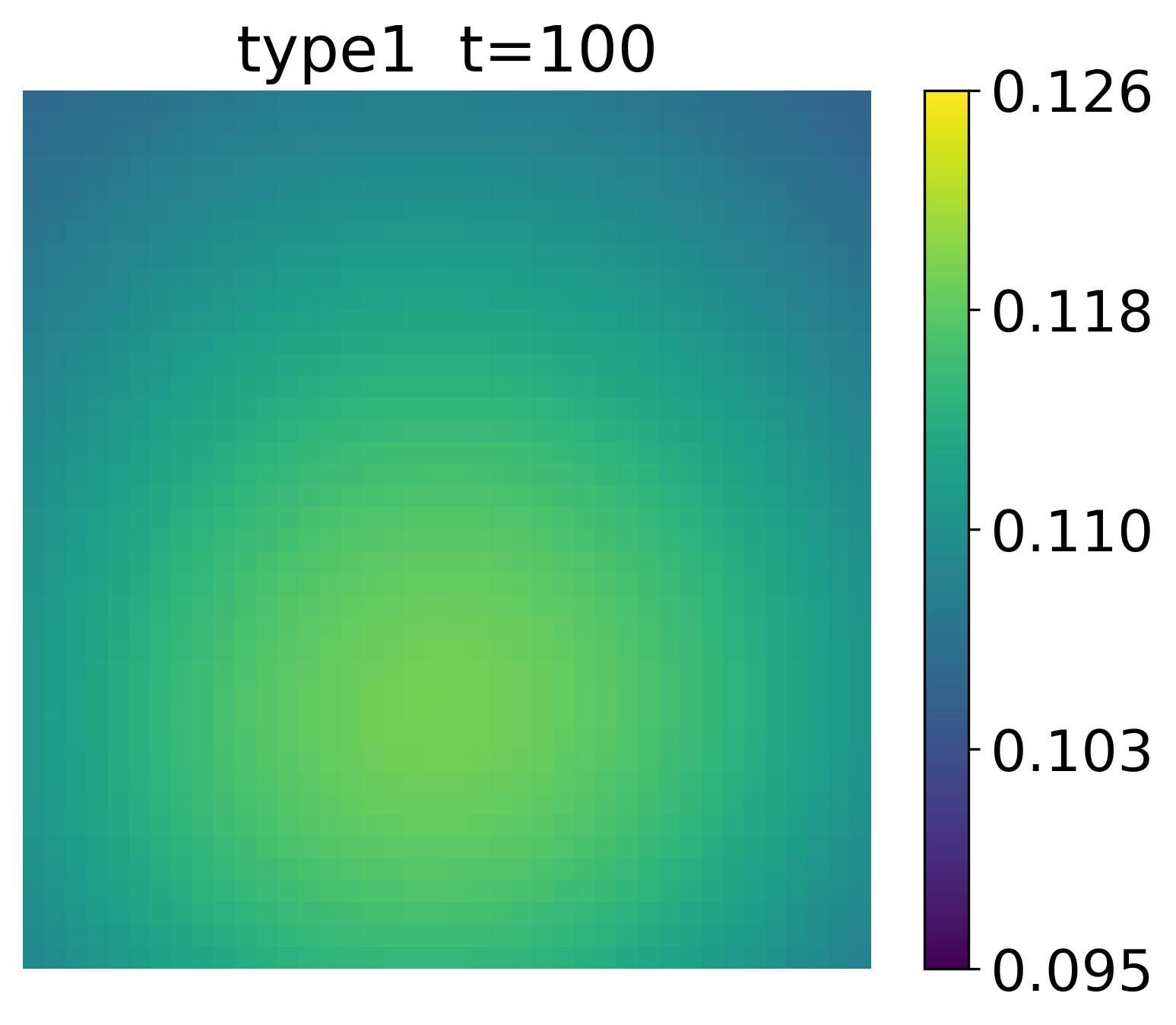}} \hspace{-2pt}
\subfloat{\includegraphics[width=0.24\columnwidth]{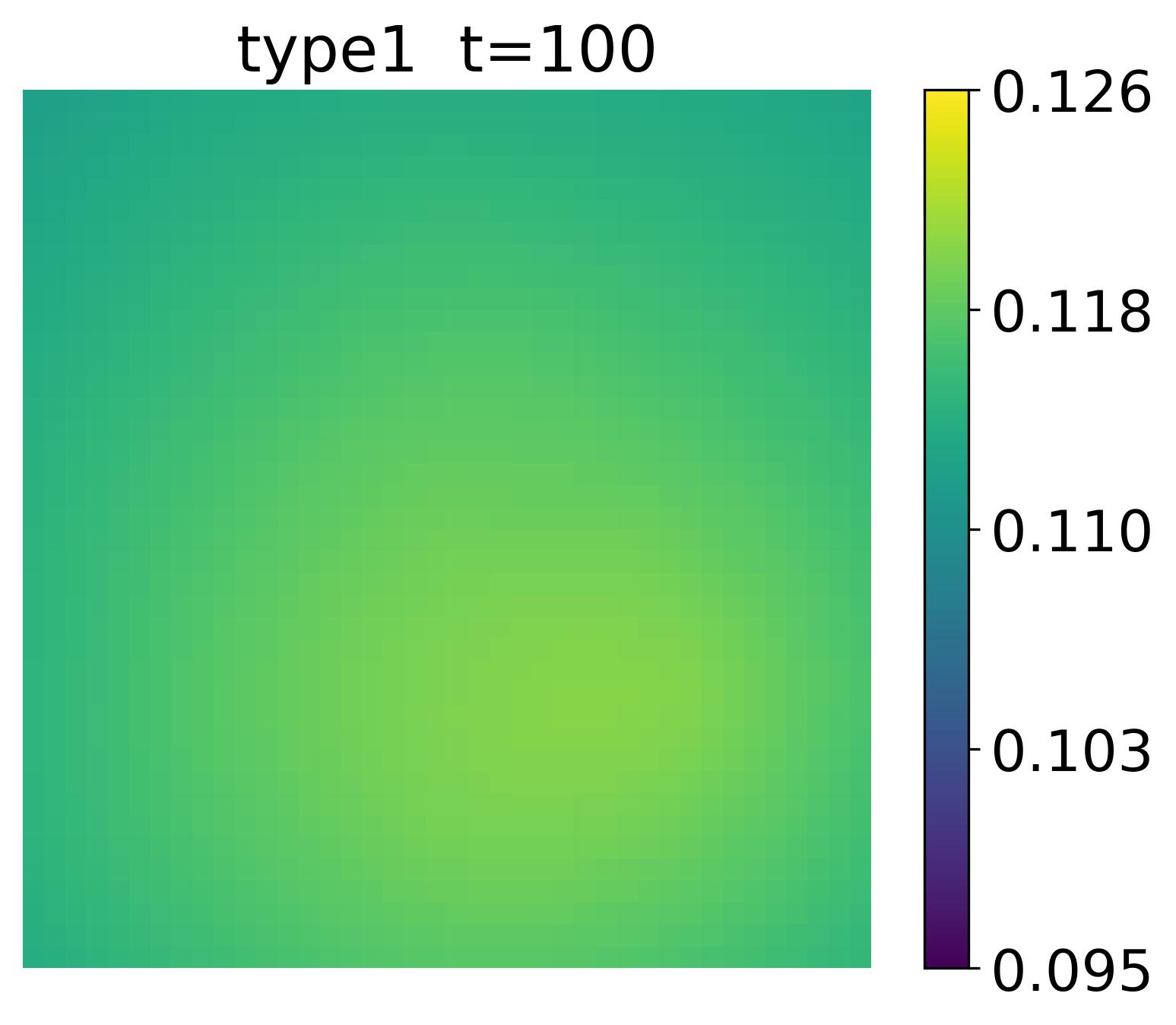}} \hspace{-2pt}
\subfloat{\includegraphics[width=0.24\columnwidth]{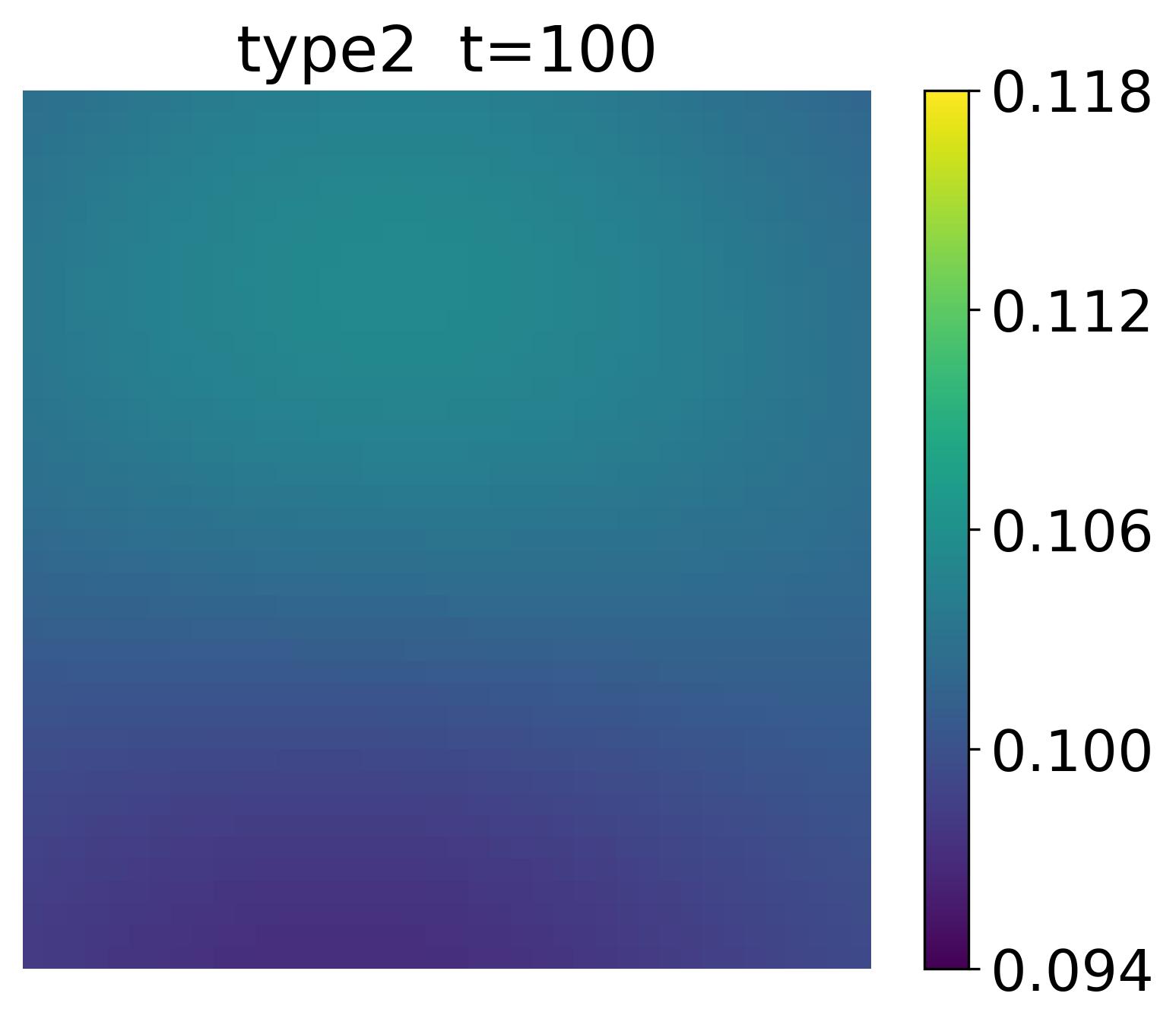}} \hspace{-2pt}
\subfloat{\includegraphics[width=0.24\columnwidth]{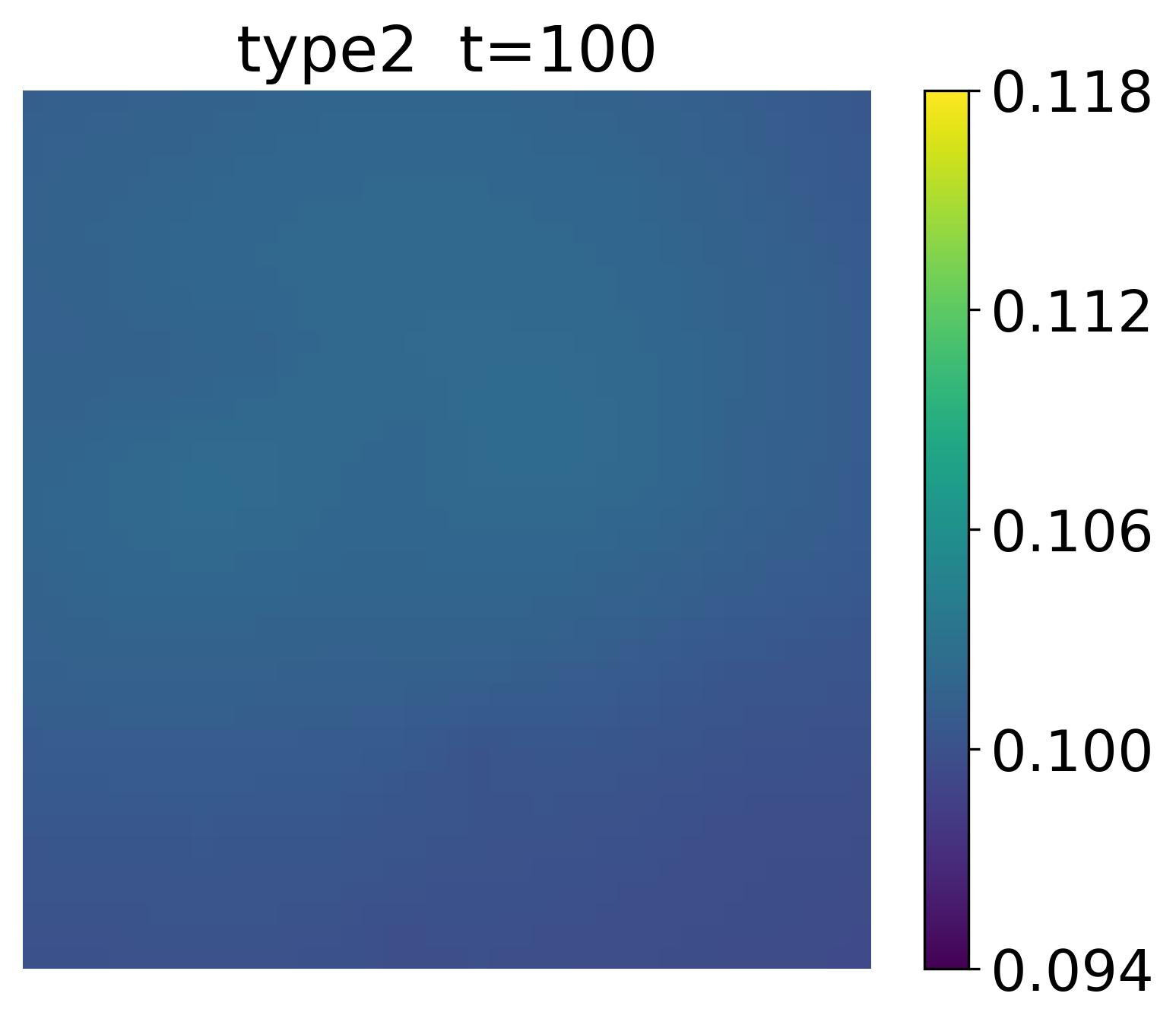}}

\vspace{-8pt}

\subfloat{\includegraphics[width=0.24\columnwidth]{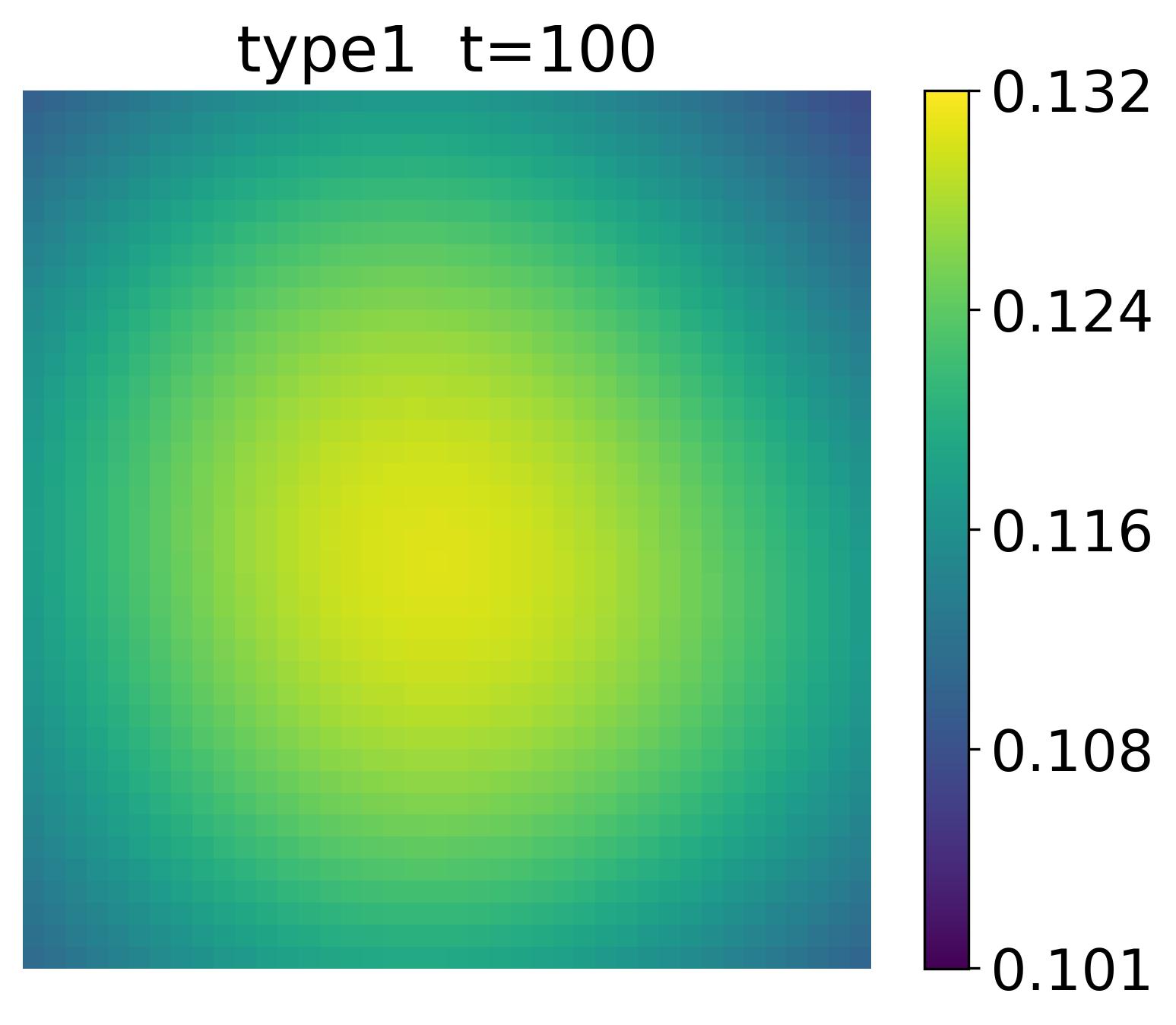}} \hspace{-2pt}
\subfloat{\includegraphics[width=0.24\columnwidth]{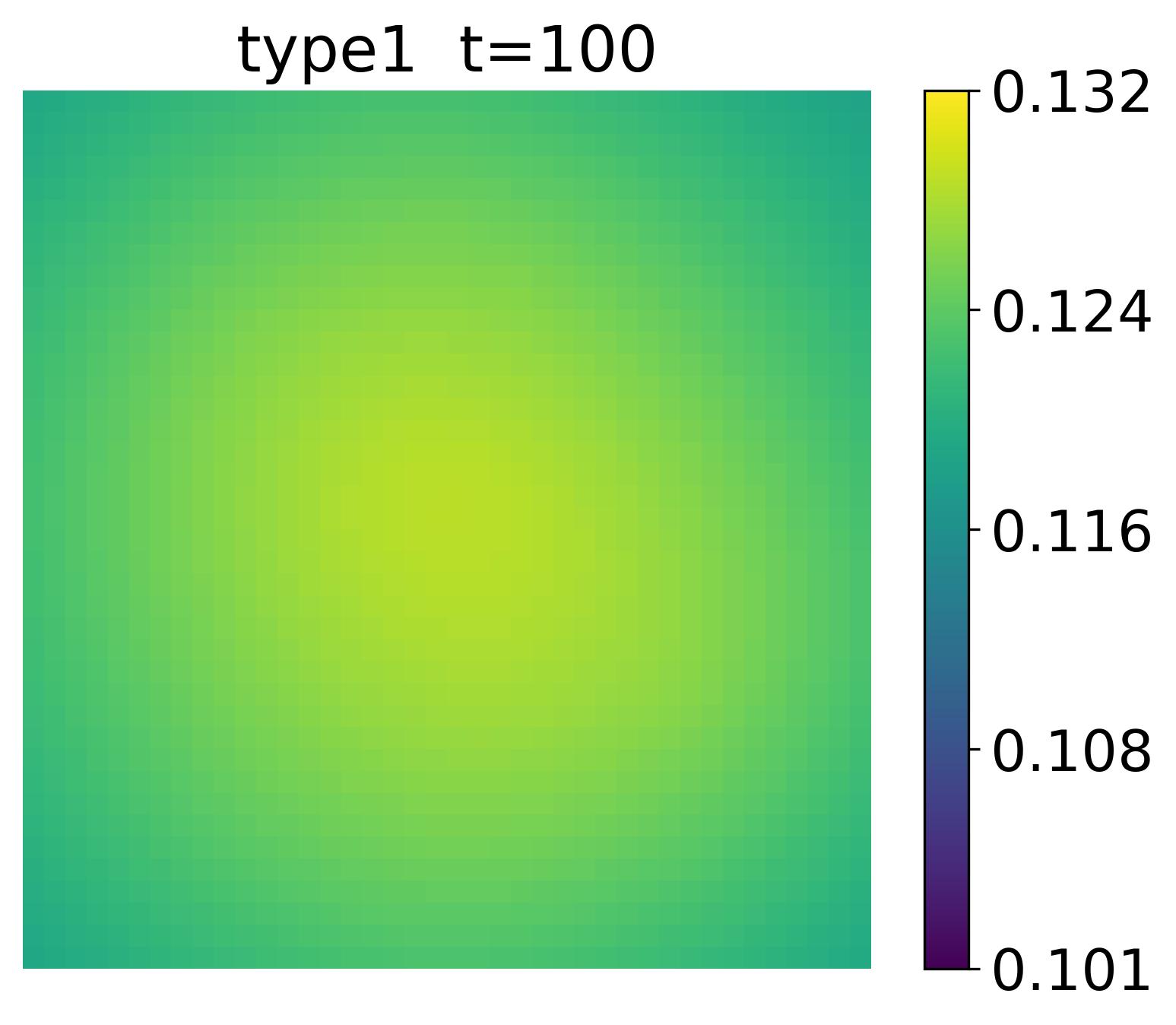}} \hspace{-2pt}
\subfloat{\includegraphics[width=0.24\columnwidth]{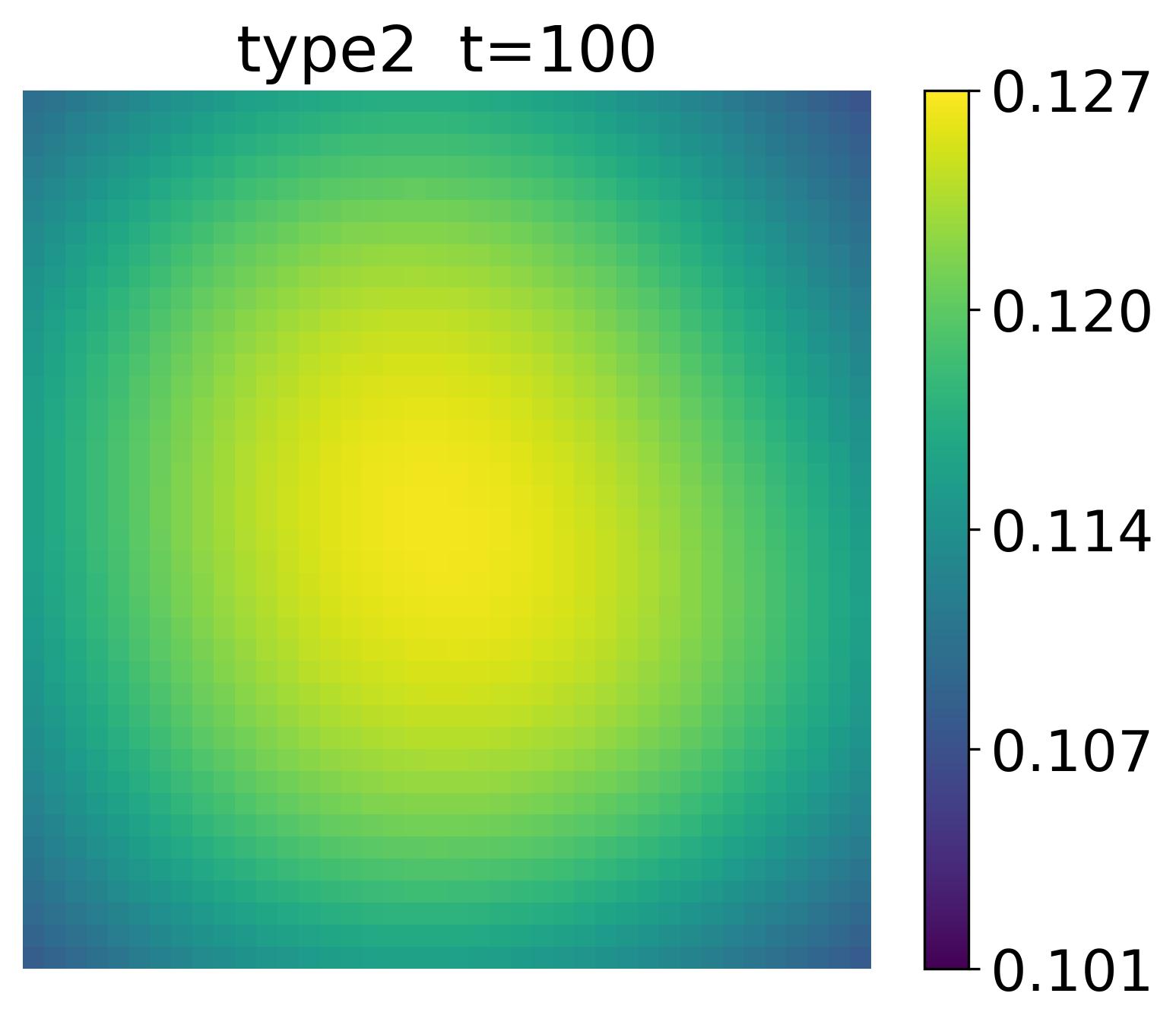}} \hspace{-2pt}
\subfloat{\includegraphics[width=0.24\columnwidth]{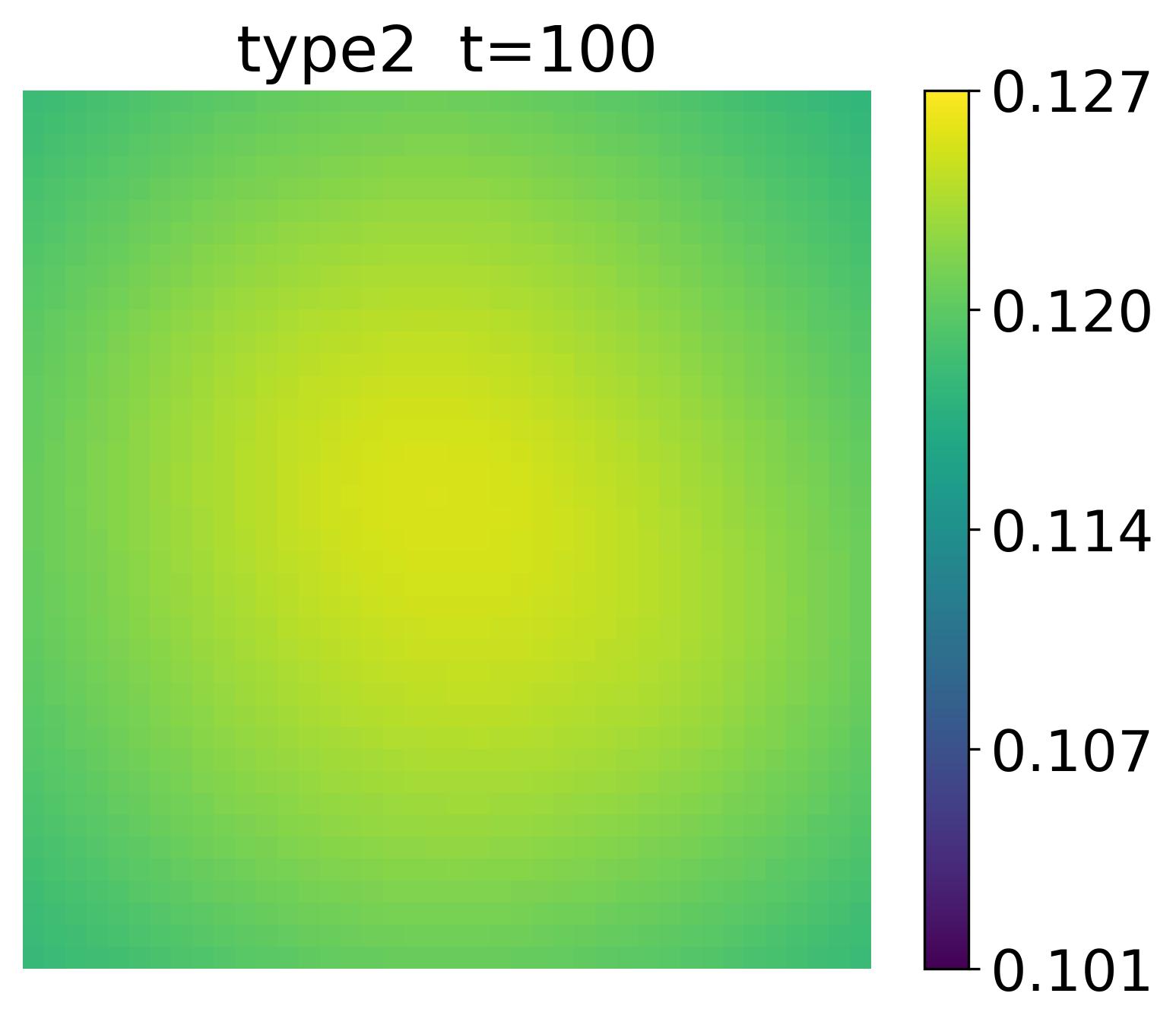}}

\vspace{-8pt}

\subfloat{\includegraphics[width=0.24\columnwidth]{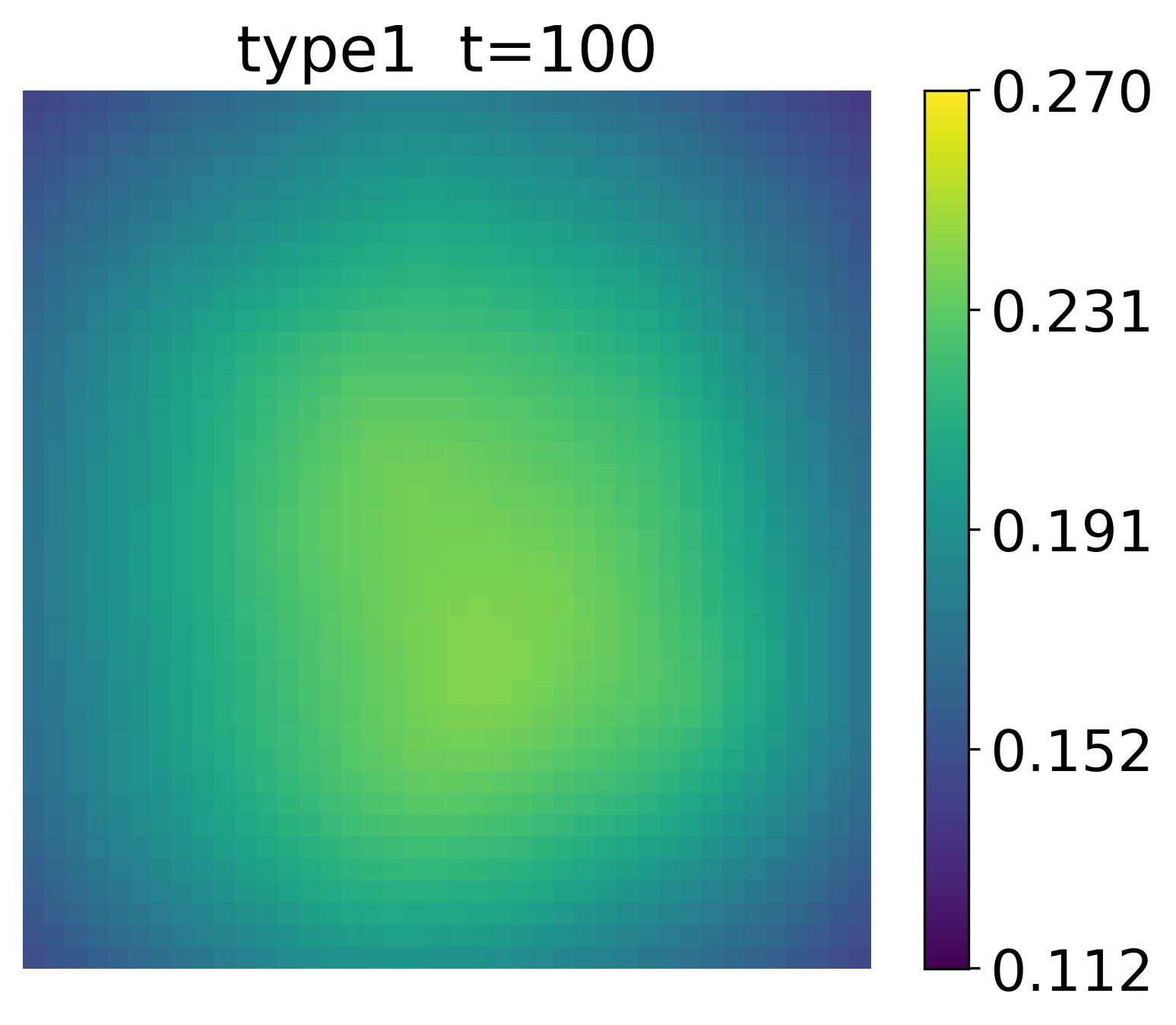}} \hspace{-2pt}
\subfloat{\includegraphics[width=0.24\columnwidth]{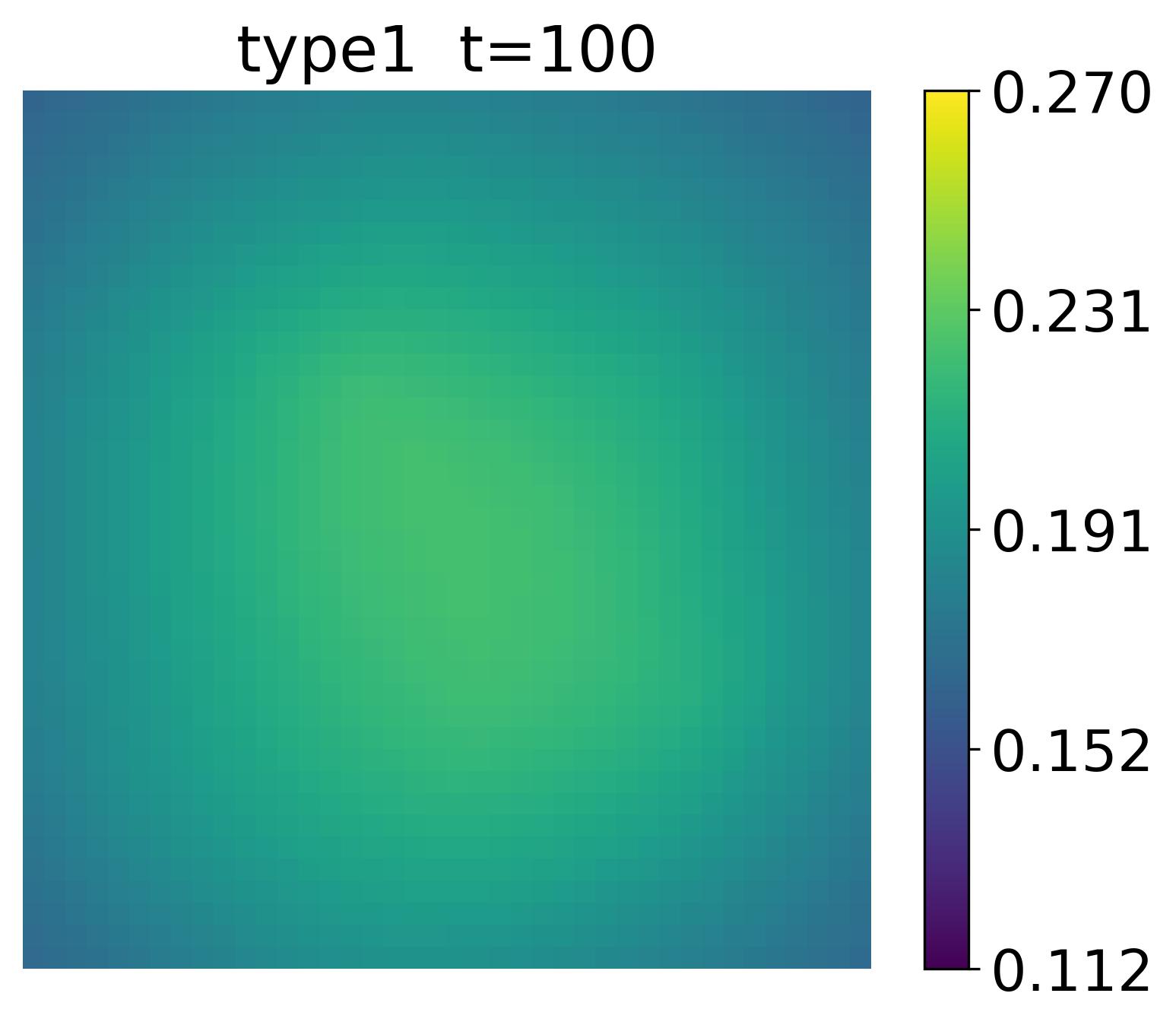}} \hspace{-2pt}
\subfloat{\includegraphics[width=0.24\columnwidth]{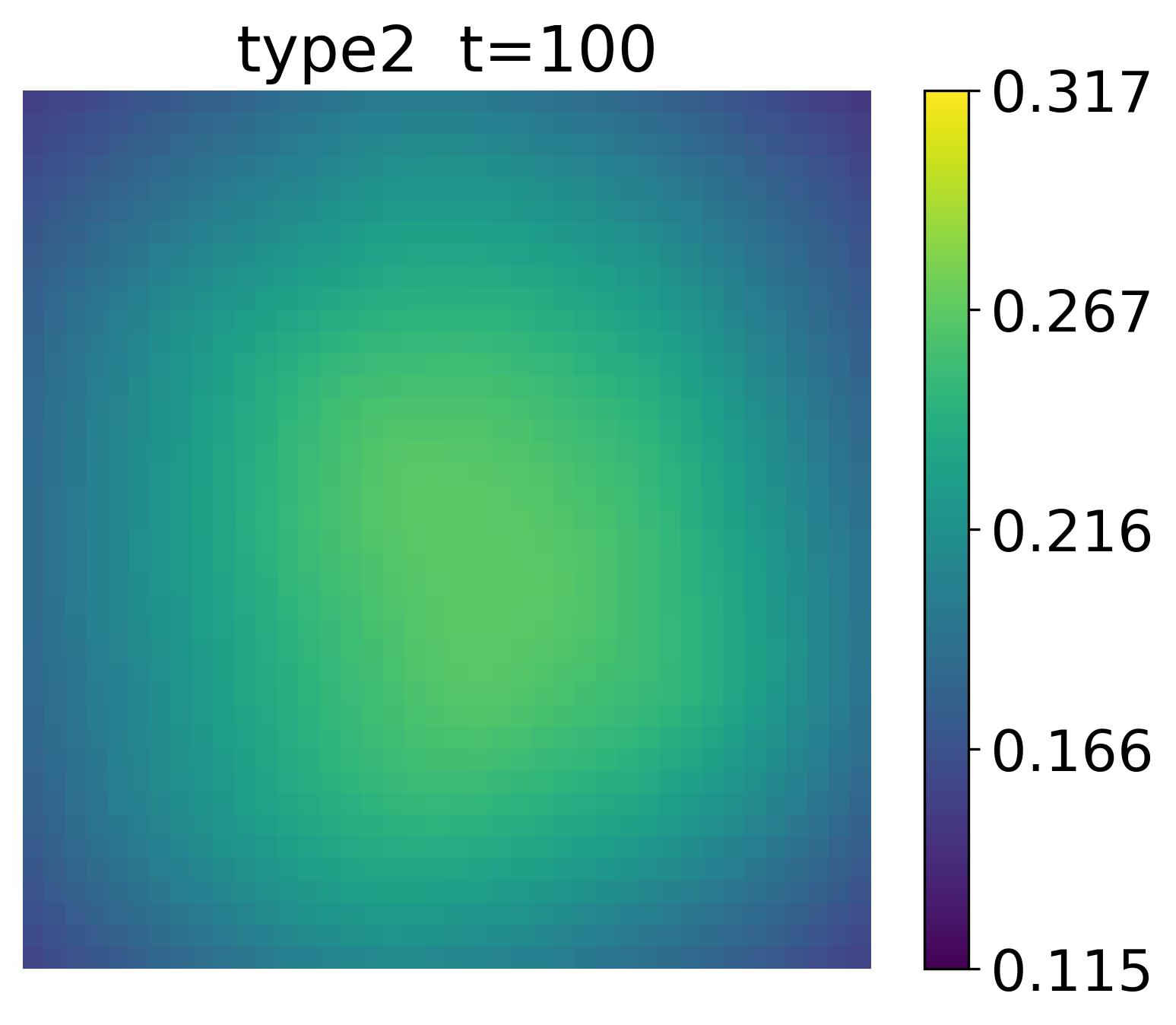}} \hspace{-2pt}
\subfloat{\includegraphics[width=0.24\columnwidth]{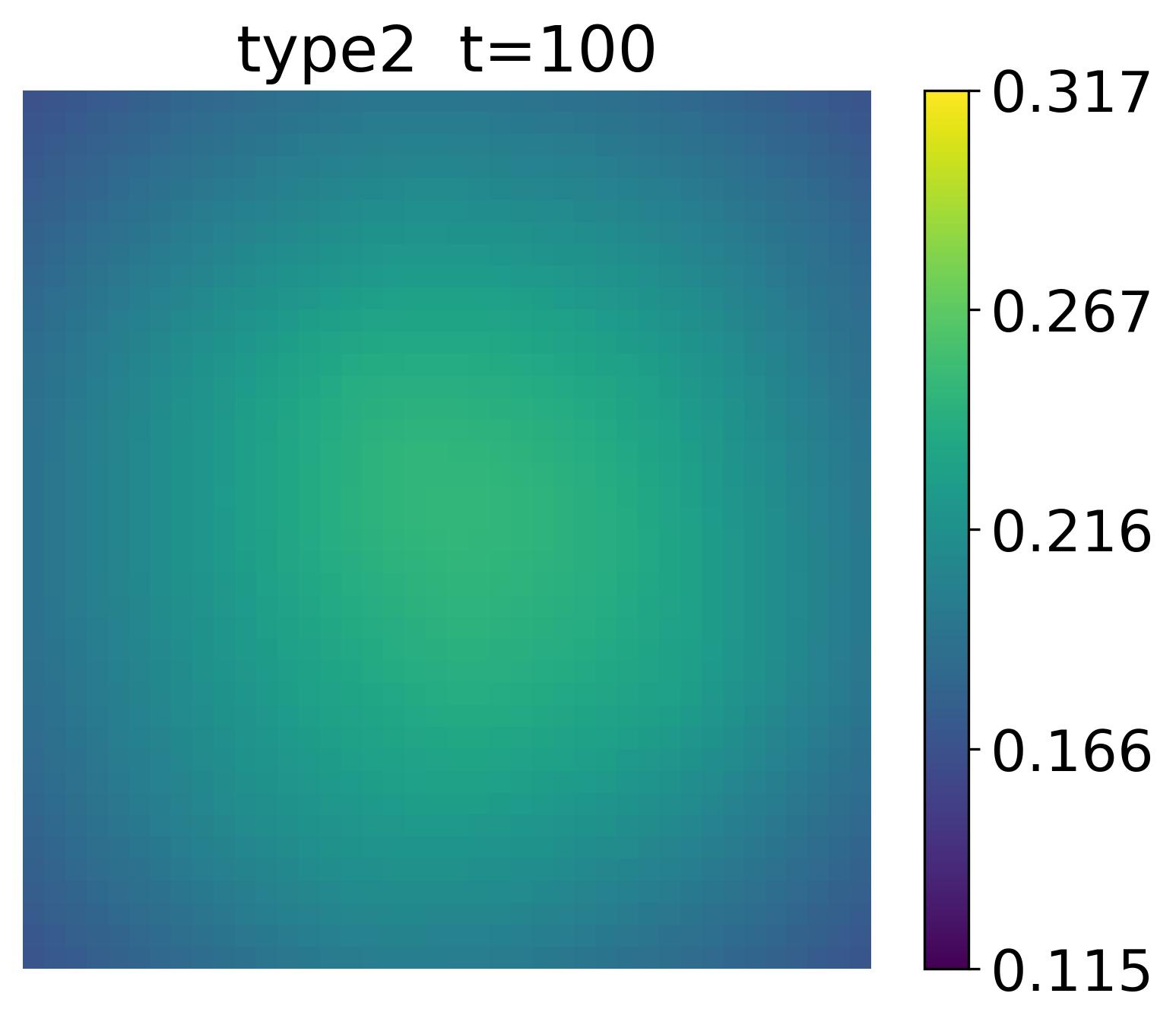}}

\caption{Cumulative time-averaged spatial intensity maps at $\tau = 100$ for the four bivariate simulation settings (Row~1–4 represents Biv~1–4 respectively). For each setting, columns 1 and 3 display the true spatial intensities, while columns 2 and 4 show the fitted counterparts for the two event types. Intensities are averaged over the entire observation window [0,100]. Cumulative maps at horizons $\tau \in $\{20,40,60,80,100$\}$ are reported in section \ref{sec:supp_mstnhp} of the supplementary material.}
\label{fig:biv1-4_spatial}
\end{figure}

\subsection{Comparison with temporal-only model}

\noindent To investigate the impact of ignoring the spatial dimension in a controlled setting, we construct a temporal-only baseline using the same spatio-temporal Hawkes datasets described in Section~\ref{sec:simulation_setting} (Biv~1–4). For each configuration, we start from the simulated space–time realizations on the unit square and time interval [0,100] and collapse them to a purely temporal bivariate point process by discarding the spatial coordinates while retaining event times and marks. The resulting temporal sequences for Biv~1–4 are then used to fit a MTNHP, trained with the same optimization scheme, train/validation/test splits, and early-stopping strategy (based on validation log-likelihood) as in the MSTNHP experiments. 

All results shown were from models trained for 2,000 epochs with 58,112 trainable parameters (64 hidden units), whereas the corresponding MSTNHP models used 32 hidden units (20,960 trainable parameters) for Biv 1–3 and 64 hidden units for Biv 4 (82,880 trainable parameters). Models trained with either significantly more or fewer parameters than  58,112 yielded markedly worse results (see section \ref{subsec:supp_biv1_study} of the supplementary materials for convergence plots).  

\begin{figure}[!hbpt]
\centering
  \subfloat[Biv1]{\includegraphics[width=0.5\columnwidth]{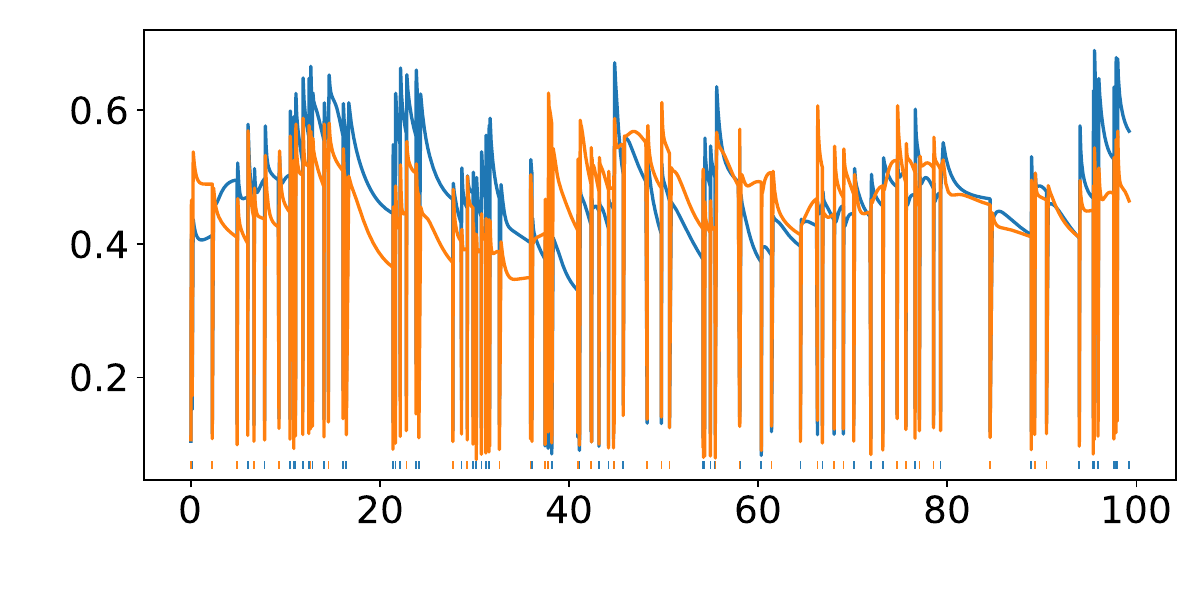}}
  \hfill
  \subfloat[Biv2]{\includegraphics[width=0.5\columnwidth]{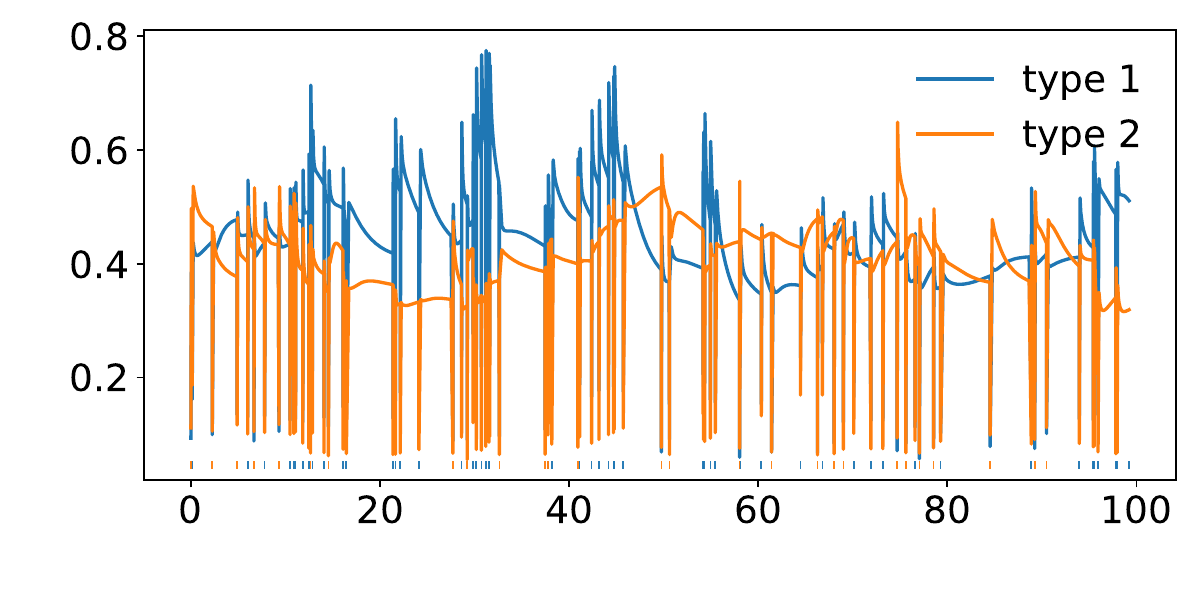}}
  
  \vspace{-8pt}

  \subfloat[Biv3]{\includegraphics[width=0.5\columnwidth]{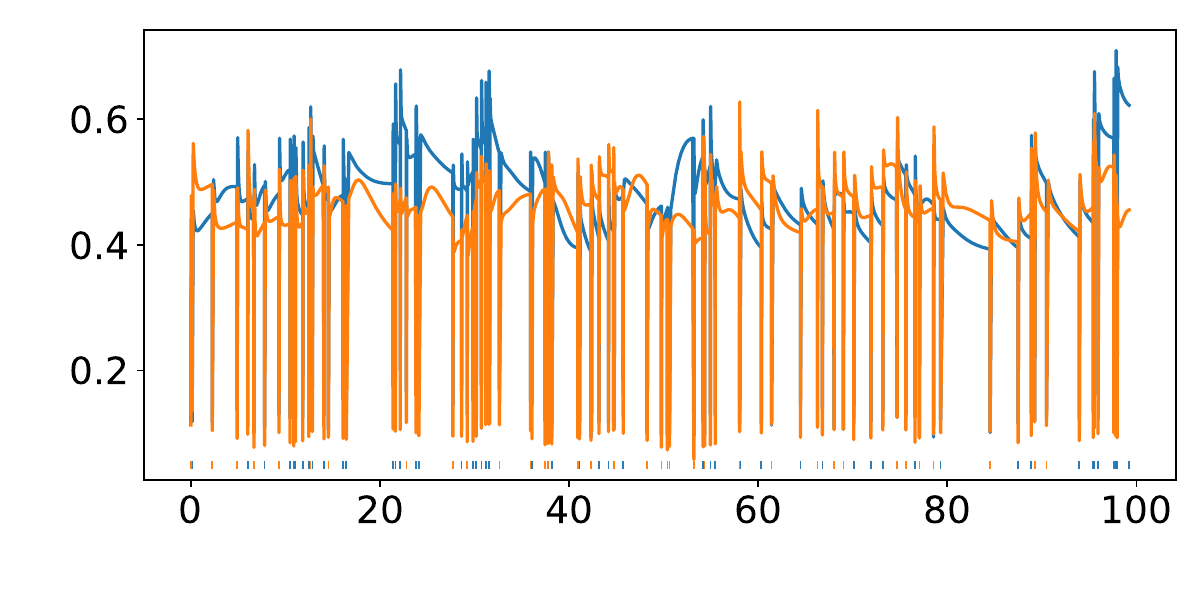}}
  \hfill
  \subfloat[Biv4]{\includegraphics[width=0.5\columnwidth]{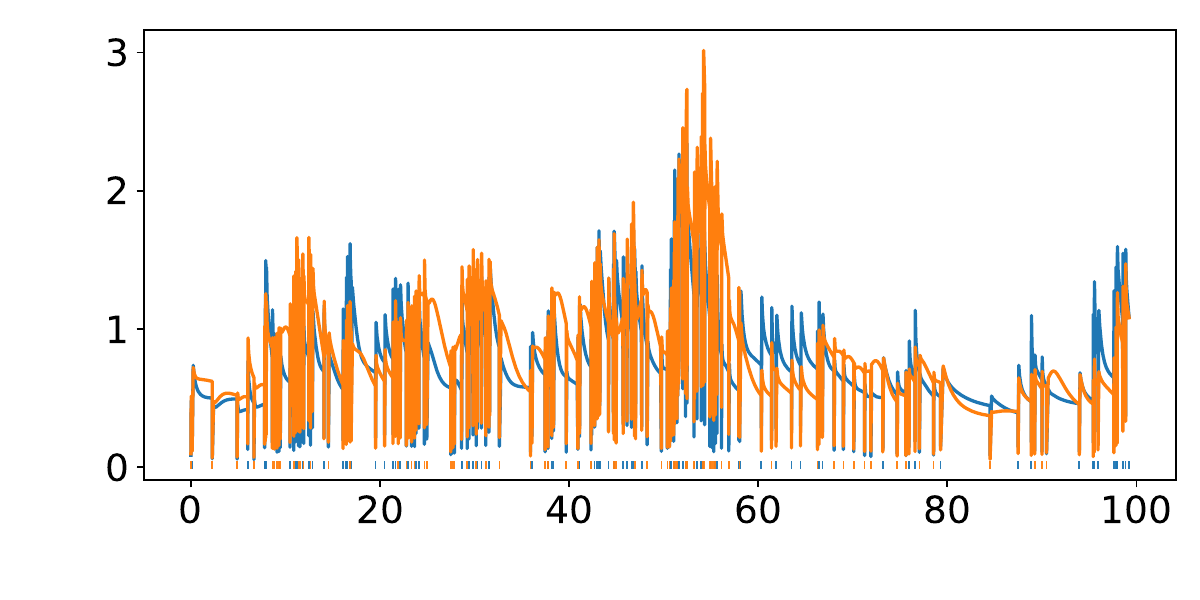}}

  \caption{Fitted temporal intensities for the simulated datasets as used in Figure~\ref{fig:biv1-4_images}, from MTNHP method. }
  
  \label{fig:biv1-4_temporal_tnhp}
\end{figure}

As a reference, Figure~\ref{fig:biv1-4_images} already shows, for each setting and event type, that MSTNHP closely matches the ground-truth temporal intensities obtained by integrating the spatio-temporal Hawkes intensity over space. Figure~\ref{fig:biv1-4_temporal_tnhp} presents the corresponding MTNHP fits on the collapsed temporal data. Despite convergence of the likelihood, the temporal-only model fails to recover the true dynamics, and the fitted curves often exhibit jagged, highly oscillatory behavior that does not resemble the smooth, interpretable Hawkes intensities. Overall, the temporal evolution produced by MTNHP is markedly inconsistent with the ground truth across all four simulation settings. This is consistent with the fact that the underlying processes are inherently spatio-temporal. By collapsing to one dimension, the model is forced to explain spatially heterogeneous clusters via a single temporal intensity, effectively averaging out spatial structure and leading to a mis-specified temporal process. In contrast, the proposed MSTNHP, which explicitly models the joint space–time dynamics, accurately recovers both the temporal (Figure~\ref{fig:biv1-4_images}) and spatial intensity patterns (Figure~\ref{fig:biv1-4_spatial}).

To probe whether the MTNHP failures are driven by spatial misspecification rather than optimization issues, we conduct an additional ablation on Biv1 dataset. We keep all the parameter values the same except for $\sigma^2$, the spatial triggering distance. We set $\sigma^2=10^{-4}$ to minimize the triggering in space and thus we are effectively transforming the system into a nearly purely temporal process. We then collapse the simulated space–time realizations to temporal sequences and train MTNHP for 500 epochs. In this regime, the fitted temporal intensities improve substantially and the spike-like “vertical bar” artifacts seen in Figure~\ref{fig:biv1-4_temporal_tnhp} vanish, while the learned curves closely track the ground truth spatially integrated intensities (Supplementary material  \ref{subsec:supp_biv1_study}).
This behavior supports our conjecture: when spatial triggering is broad, collapsing space forces a temporal-only model to explain heterogeneous, location-dependent clustering through a single temporal intensity, producing unstable and non-Hawkes-like fits; when spatial triggering is extremely localized, the collapsed process becomes closer to a well specified temporal model, and MTNHP can recover the effective temporal dynamics. Overall, the ablation reinforces the need for explicit joint space–time modeling (MSTNHP) in realistic spatio-temporal settings.
This result again reinforces the need of incorporating the spatial component of NHP model through MSTNHP, when the data are truly spatio-temporal point patterns. 



\section{Real application}

\noindent We now apply our MSTNHP framework to terrorism data from Pakistan presented in Section~\ref{sec:motivation}. 
During the period from 2008 to 2020, a total of 1,991 terrorist attacks were recorded for the four primary active groups. For our experimental setup, we partitioned the data into 13 yearly sequences: the first 10 years (1,812 events) serve as the training set, the following year (98 events) as the test set, and the final 2 years (81 events) for validation. The reported model was trained for 5,000 epochs with a batch size of 10, corresponding to the number of training sequences. For the real application, both MSTNHP and MTNHP used one V100 GPU, with MSTNHP running on 1 CPU core and MTNHP on 10 CPU cores.

Modeling these four distinct entities, Tehrik-i-Taliban Pakistan (TTP; Type 1), the Baloch Republican Army (BRA; Type 2), the Baloch Liberation Army (BLA; Type 3), and the Baloch Liberation Front (BLF; Type 4), is non-trivial and tests our model's capacity to capture a more complex, real-world conflict environment.  The model must now simultaneously learn the endogenous triggering dynamics within each of the four groups, as well as the full set of exogenous interactions between them. For example, how an attack by BLA (Type 3) might influence the subsequent event rate of TTP (Type 1) or BLF (Type 4). By successfully disentangling these interdependent dynamics, we demonstrate our model's robustness and capacity to identify unique temporal patterns of distinct actors operating within shared geographical and temporal space.

We implemented a rigorous preprocessing and normalization pipeline to ensure spatial and temporal consistency. Events are organized into independent yearly sequences, with event times recorded as days elapsed from the start of each year, resulting in a training horizon of $T \in [0, 366]$ days. To facilitate spatial learning and maintain scale invariance, we apply an affine transformation to the geographic coordinates. The raw longitude and latitude, bounded by the region were mapped onto a normalized coordinate system within the unit square $[-1, 1]^2$. We validated each normalized event location against a polygon of Pakistan, ensuring all data points fall strictly within the study's territorial boundaries. Training convergence was monitored using the validation log-likelihood. The  best model, based on the validation set, was obtained at epoch 4,165. See section \ref{sec:real_data_conv_plots} of the supplementary material for the convergence plot. 

\begin{figure}[!hpbt]
\centering
  \subfloat[MSTNHP]{\includegraphics[width=.96\columnwidth]{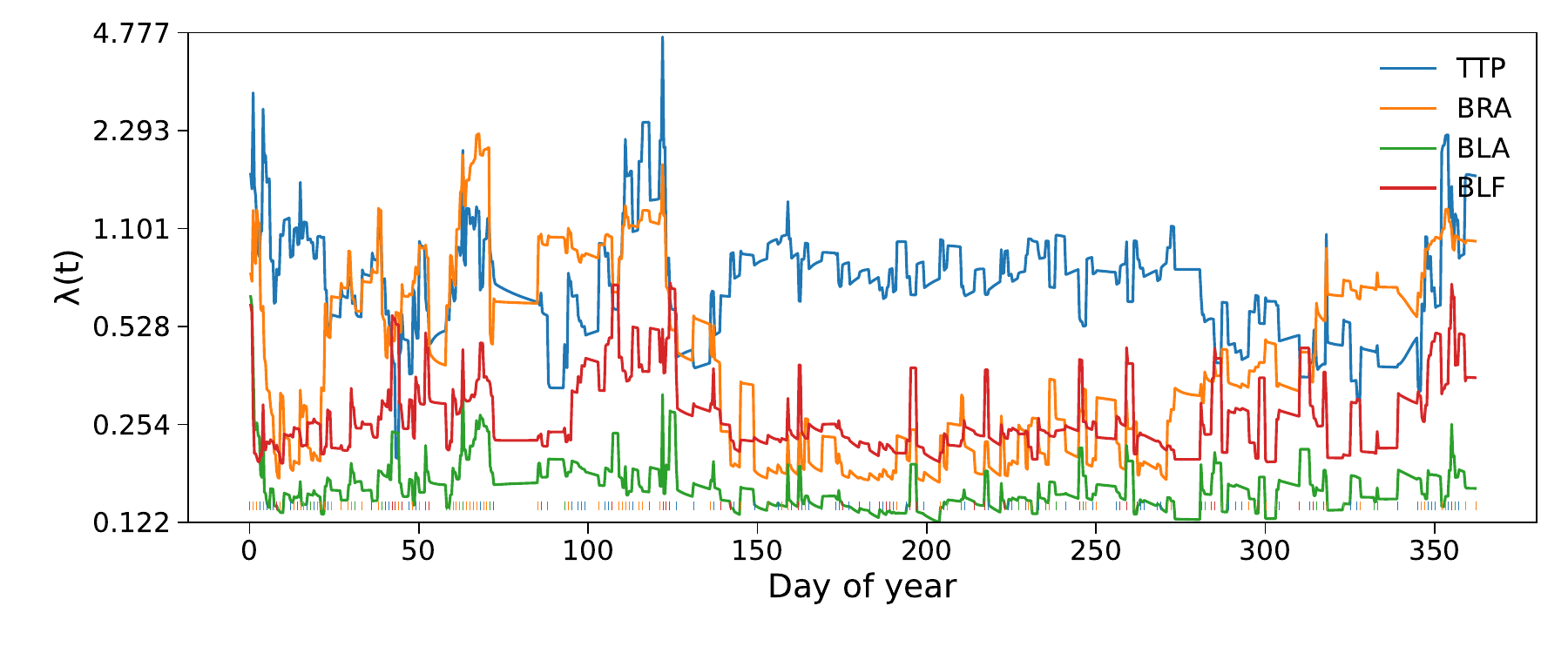}}
  \hfill
  \subfloat[MTNHP]{\includegraphics[width=.96\columnwidth]{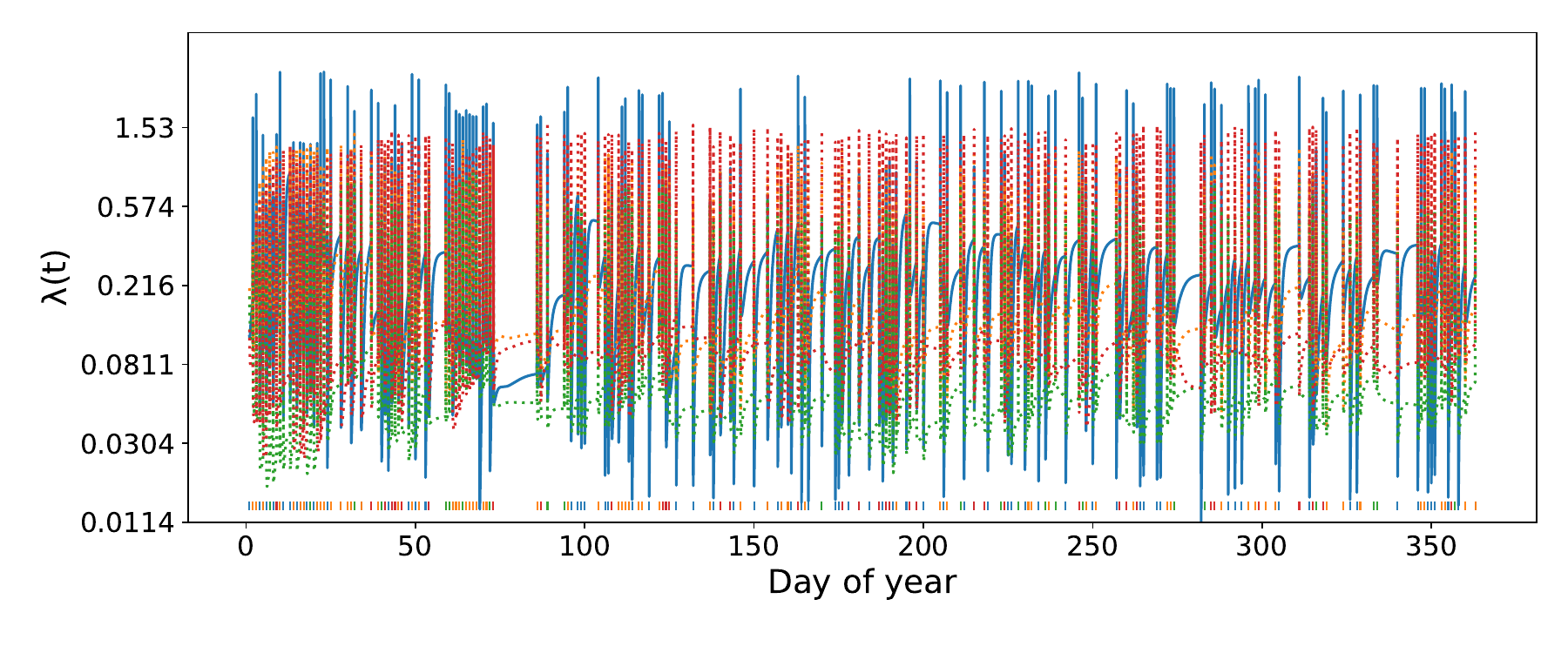}}

  \caption{Fitted temporal conditional intensity functions for Pakistan terrorism data. (a) is for MSTNHP and (b) for MTNHP. For (b), some colors are displayed with dotted lines for legibility. }
  \label{fig:gtd_pakistan_t_intensity}
\end{figure}

Figure~\ref{fig:gtd_pakistan_t_intensity} (a) shows the fitted temporal intensity function from our proposed method for the four groups. Overall, temporal triggering pattern of BRA, BLA, and BLF are similar, although there are some time period (such around $t=100$, $t=230$, or $t=340$) where there are notable difference in terms of the temporal pattern among these three groups. On the other hand, TTP shows a very different temporal pattern. Overall TTP has the largest intensity level among the four groups, and often the triggering (or modulating) pattern is the opposite to the other three groups. For each group, the triggering structure switches between triggering and modulating over time in a complex fashion.

Figure~\ref{fig:gtd_pakistan_st_intensity} shows fitted conditional spatial intensity maps for pairs of groups for a few selected dates. These two pairs (BRA \& BLA and TTP \& BLF) are chosen from six available pairs to demonstrate how the model captures interactions between two “similar” groups (BRA \& BLA) as well as two contrasting ones (TTP \& BLF). For the BRA \& BLA pair, although their spatial intensity maps look similar on most days, on some dates such as January 24 and April 8 they show opposite patterns. Furthermore, the strength of spatial triggering is not always maximized at the location of attack, which illustrates the flexibility of the model. The results for TTP and BLF further demonstrate this flexibility. Unlike the BRA \& BLA pair, for most days the spatial intensity patterns of the two groups are reversed, which may be a sign of a jointly modulating relationship between them. Overall, the fitted spatial intensity structure shows considerable versatility, which is difficult to achieve with parametric Hawkes process models.


\begin{figure}[htbp!]
\centering
\setlength{\tabcolsep}{6pt} 
\renewcommand{\arraystretch}{1.0}

\begin{tabular}{@{}c c@{}}
{\scriptsize  BRA\hspace{1.8cm}BLA} & {\scriptsize  TTP\hspace{1.8cm}BLF} \\[-2pt]
{\scriptsize  Jan 24} & {\scriptsize  Jan 1} \\[-2pt]
\includegraphics[width=0.47\columnwidth]{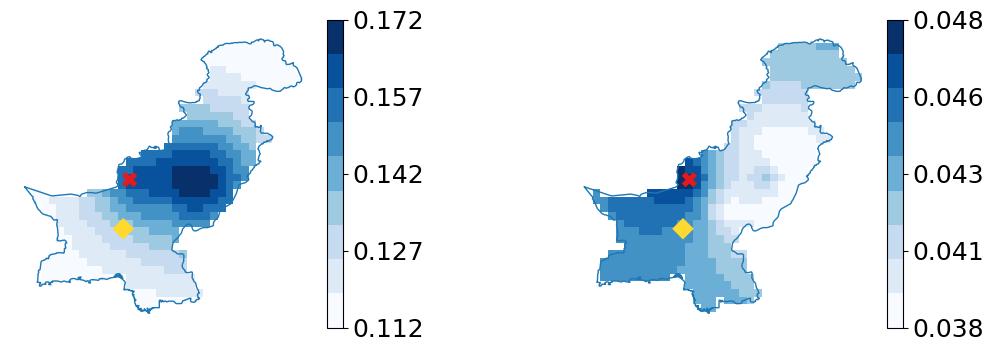} &
\includegraphics[width=0.47\columnwidth]{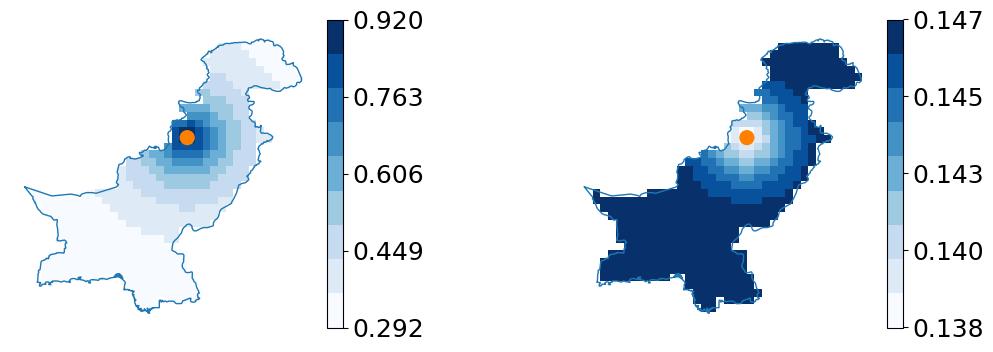} \\[6pt]

{\scriptsize  Jan 30} & {\scriptsize Mar 6}  \\[-2pt]
\includegraphics[width=0.47\columnwidth]{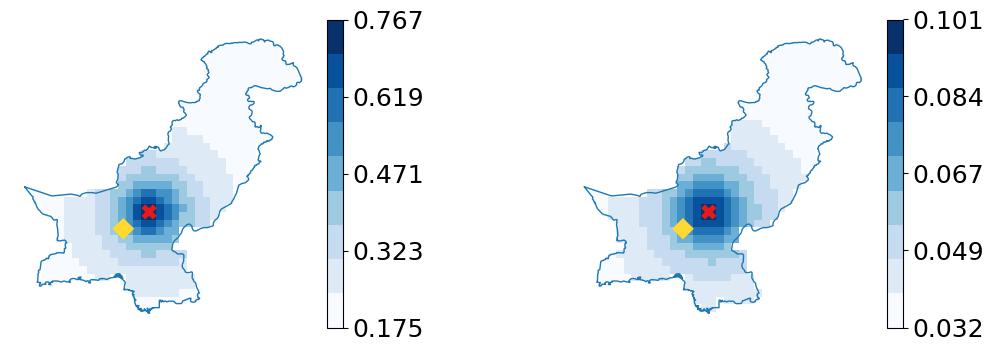} &
\includegraphics[width=0.47\columnwidth]{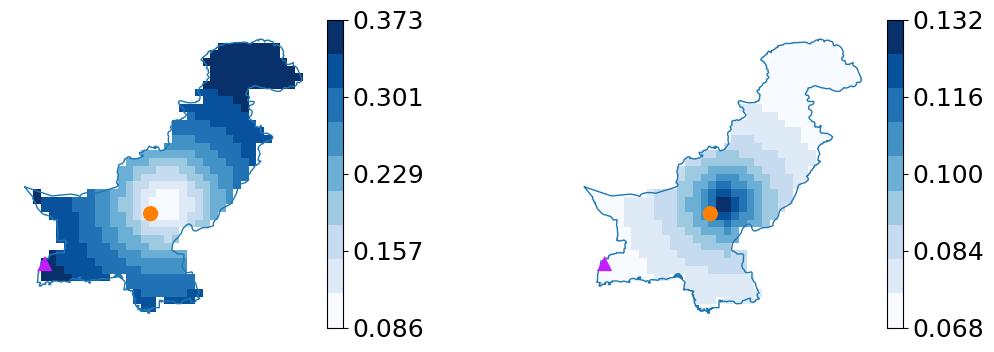} \\[6pt]

{\scriptsize Apr 8} & {\scriptsize  Apr 21} \\[-2pt]
\includegraphics[width=0.47\columnwidth]{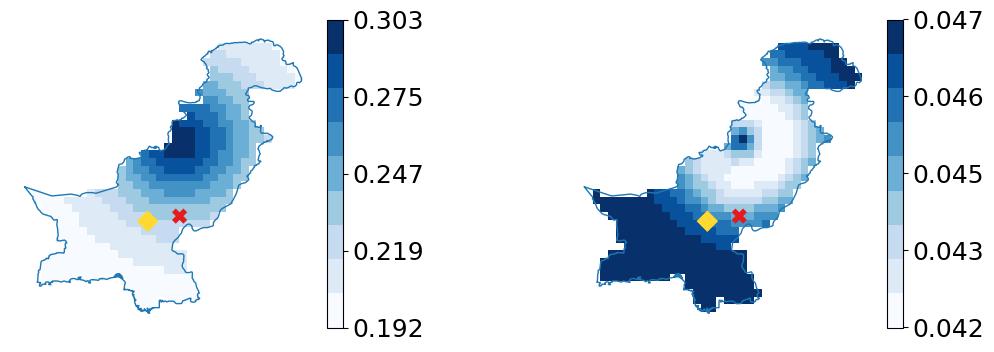} &
\includegraphics[width=0.47\columnwidth]{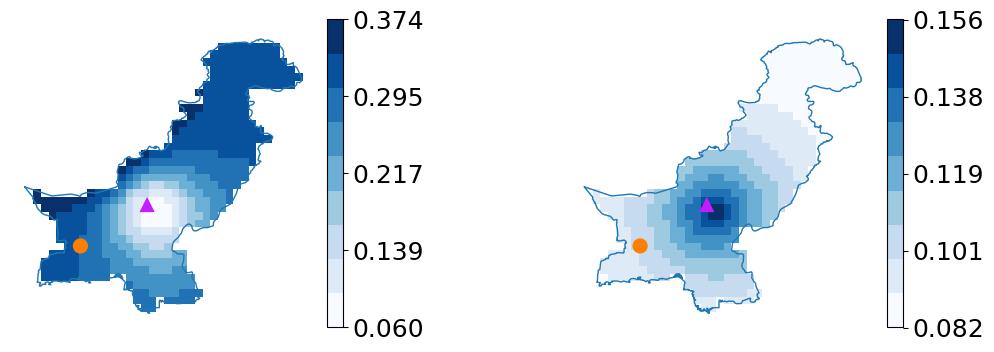} \\[6pt]

{\scriptsize  Apr 21} & {\scriptsize  Apr 23} \\[-2pt]
\includegraphics[width=0.47\columnwidth]{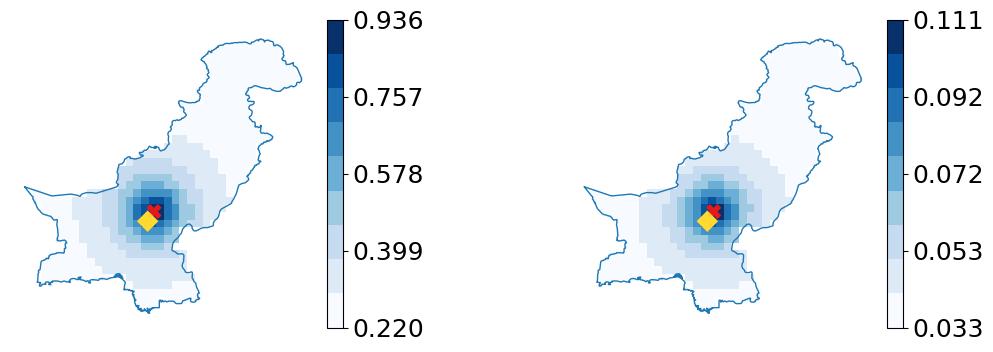} &
\includegraphics[width=0.47\columnwidth]{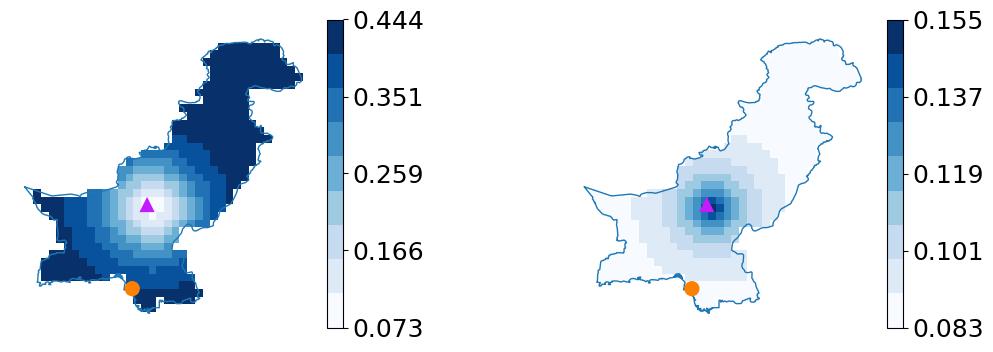} \\[6pt]

{\scriptsize  May 29} & {\scriptsize  Dec 16} \\[-2pt]
\includegraphics[width=0.47\columnwidth]{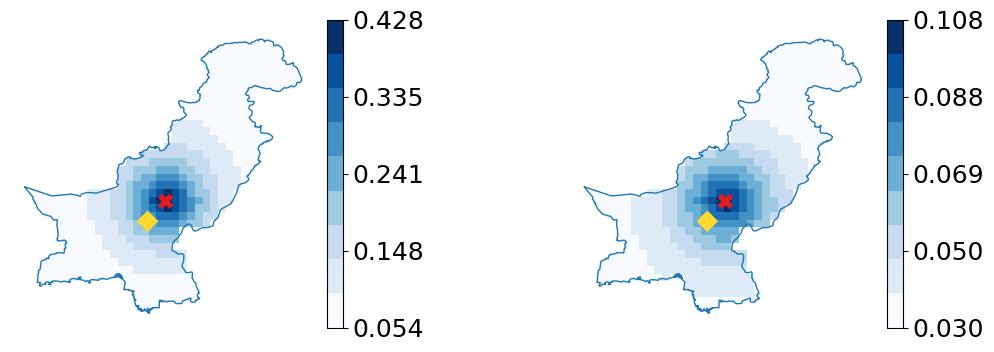} &
\includegraphics[width=0.47\columnwidth]{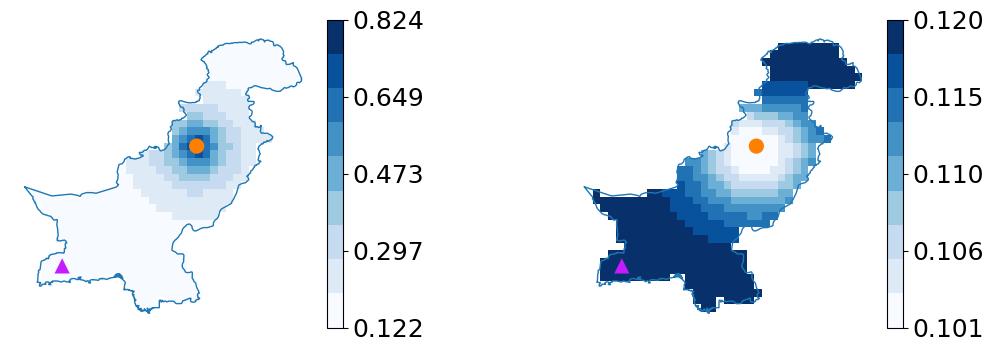} \\[6pt]

{\scriptsize  Sep 8} & {\scriptsize Dec 19} \\[-2pt]
\includegraphics[width=0.47\columnwidth]{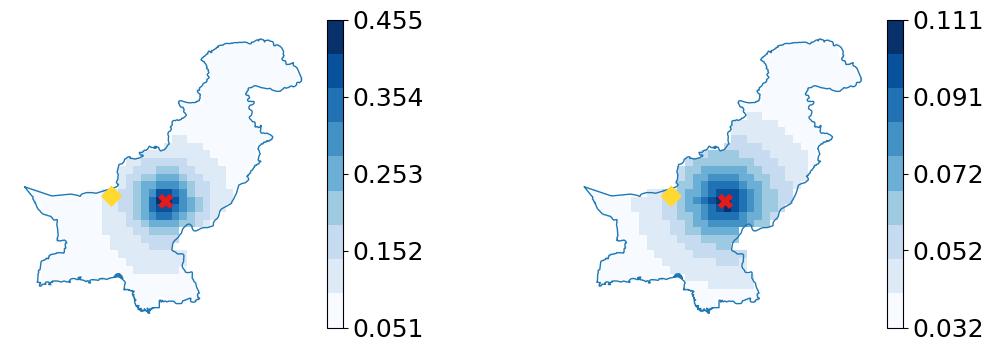} &
\includegraphics[width=0.47\columnwidth]{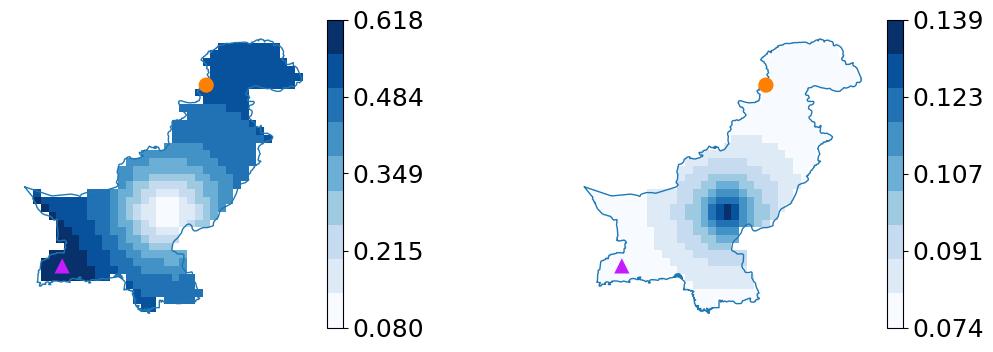} \\
\end{tabular}

\caption{Fitted spatial conditional intensity maps for selected dates in Pakistan in 2014. Left: BRA/BLA. Right: TTP/BLF. Markers indicate observed attack locations on the day: BRA ($\times$, red), BLA ($\diamond$, yellow), TTP ($\circ$, orange), BLF ($\triangle$, magenta). Per-frame color scaling is applied.}
\label{fig:gtd_pakistan_st_intensity}
\end{figure}

To further demonstrate the utility of MSTNHP we proposed, we compare the results with MTNHP model that ignores the spatial dimension.
The MTNHP utilized the same groups and observation period (2008-2020). Since now the spatial component is ignored, some of the events are ``duplicates" in the sense that more than one event are recorded at the same time. Despite the fact that they happened at different locations, the temporal only model cannot distinguish them. Most of these duplicates are indeed not the same event as they occurred in different spatial locations. We removed these duplicates to avoid coincidental points. 
As a result, for the temporal analysis, there were 1,654 total attacks over the 12 year period. The first 10 years, comprising 1,493 events were used as training data, the next 1 year, comprising 85 events were used as test data, and the last 2 years, comprising 76 events were used as validation data.  

The MTNHP model was trained for 5,000 epochs. Following the same early-stopping principle based on validation set log-likelihood, the best model was obtained at epoch 3,430. See section \ref{sec:real_data_conv_plots}
 of the supplementary materials for the convergence plot. The plot shows that the levels of maximized loglikliehood of training as well as validation data sets are comparable to those from the spatio-temporal analysis. 

In Figure~\ref{fig:gtd_pakistan_t_intensity} (b), despite extensive training of MTNHP with the convergence of the likelihood, temporal patterns of conditional intensity functions are problematic. 
This observation can be attributed to the fact that the underlying process in the Pakistan terrorism data is inherently spatio-temporal. Collapsing the data to a temporal representation removes the spatial interactions that are intrinsically coupled with temporal triggering effects. As a result, the remaining temporal process reflects an incomplete representation of the original dynamics, which MTNHP cannot capture well. 

A closer look at the MSTNHP temporal intensities in Figure~\ref{fig:gtd_pakistan_t_intensity} (a) shows that the fitted curves are well aligned with the empirical group-wise activity levels. Across most of the year, the TTP curve dominates the others, reflecting its overwhelming share of events in the full 2008–2020 sample (1,460 of 2,159 recorded attacks before preprocessing, versus 319 for BRA, 177 for BLA, and 203 for BLF). Nevertheless, the model is not simply encoding a fixed ordering by group size, in several sub-intervals (for example between roughly days 50 and 100), the BRA intensity temporarily rises above the TTP curve, indicating periods where BRA is locally more active while TTP experiences a relative lull. This behaviour matches the underlying event sequences and illustrates that MSTNHP responds to local fluctuations in group-specific activity rather than enforcing a static rank structure. By contrast, the BLA intensity remains lowest throughout the year, consistent with BLA having the fewest attacks overall, while BLF lies between BRA and TTP but exhibits occasional surges that the fitted curve captures as pronounced bumps at specific times.

The fitted curves also display occasional “down peaks,” i.e., sharp local drops in intensity following periods of heightened activity. These features appear in both the MSTNHP and the MTNHP fits. Methodologically, such downward excursions are a natural consequence of the continuous-time LSTM parameterization, after a cluster of attacks, the memory cell can encode a net inhibitory effect, causing the effective log intensity to decay not just back to its long run baseline but slightly below it, thereby creating a short “cool-down” phase during which new events are less likely. In practice, these down peaks tend to coincide with long gaps or quieter periods in the observed data rather than isolated numerical artefacts, and thus can be interpreted substantively as the model learning that intense waves of violence are often followed by temporary suppression or strategic pauses in activity.

\section{Discussion}\label{sec:discussion}

\noindent Our results reveal the disconnect between standard performance metrics and model fidelity in neural point processes. Across simulations and real data, several baseline models achieve competitive log-likelihood while producing conditional intensity functions that fail to recover the true shape, excitation patterns, and cross-type relationships. This discrepancy is further illustrated by the comparison between the purely temporal MTNHP and the proposed MSTNHP on the terrorism data. Although the two models attain similar likelihood values, MTNHP yields distorted temporal intensities with substantially lower magnitude. Collapsing inherently spatio-temporal data into a temporal-only representation removes spatial interactions that are coupled with temporal triggering, allowing a model to optimize likelihood under a misspecified representation. Consequently, likelihood alone is insufficient to assess whether a model has learned meaningful temporal or spatio-temporal dynamics, and should be complemented with structural diagnostics such as intensity recovery and spatial intensity maps.

The strong performance of NHP is not solely a consequence of architectural complexity, but is closely related to how continuous-time dynamics are integrated into the intensity. Although LSTM-based models have been reported to outperform Transformer-based models in certain domains \cite{bilokon2023transformers}, the advantage of NHP in our setting is better explained by its continuous-time formulation. NHP maintains a latent state that evolves between events and directly parameterizes the conditional intensity. This coupling enforces persistent temporal influence of past events and limits likelihood-driven collapse. In contrast, architectures that reparameterize intensity mainly at event times or through attention can decouple intensity formation from continuous-time dynamics, leading to unrealistic fitted intensities.

Proposed method can be further extended for more flexibility of the model. Although we demonstrated that our proposed spatio-temporal Neural Hawkes Process models work well with simulated and real data, the spatio-temporal memory cell in \eqref{eq:spatio_temporal_decay} is spatio-temporally separable and stationary in space and time. Extending it to spatio-temporal nonseparable or nonstationary structure may further help with real applications with complex spatio-temporal point patterns. 
Exploring potentially more effective ways of integrating advanced modeling frameworks with Hawkes processes is left for future research.

\bibliographystyle{IEEEtran}
\bibliography{main,reference}

\newpage

\section{Biography Section}

\vspace{11pt}

\begin{IEEEbiography}[{\includegraphics[width=1in,height=1.25in,clip,keepaspectratio]{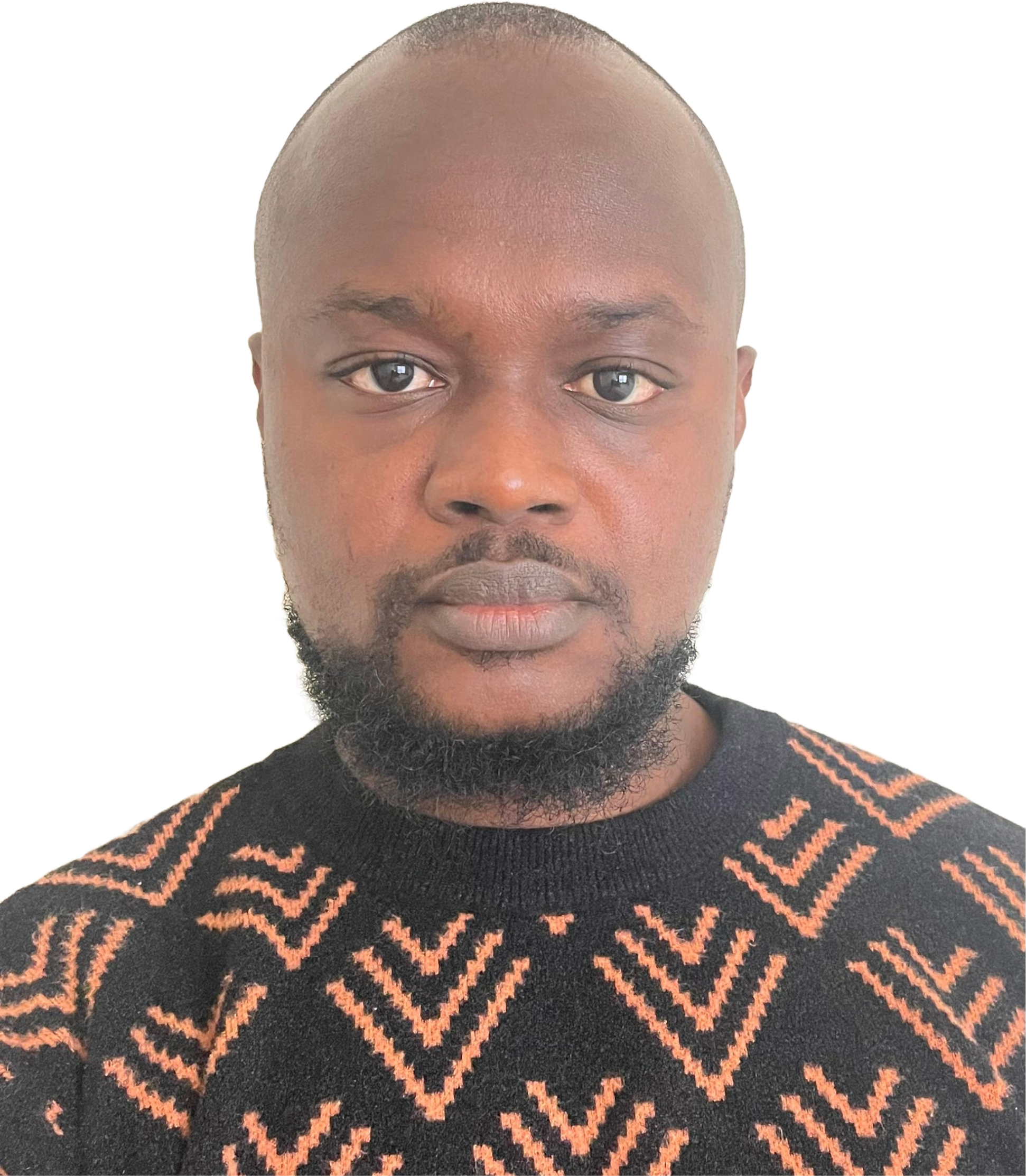}}]{Christopher Chukwuemeka}
is a Ph.D. candidate in Mathematics at the University of Houston, where he is a Graduate Teaching and Research Assistant in the Department of Mathematics. His research interests include point processes, machine learning for scientific computing, numerical optimization, and generative modeling. He is a member of the American Mathematical Society.
\end{IEEEbiography}

\vspace{11pt}

\begin{IEEEbiography}[{\includegraphics[width=1in,height=1.25in,clip,keepaspectratio]{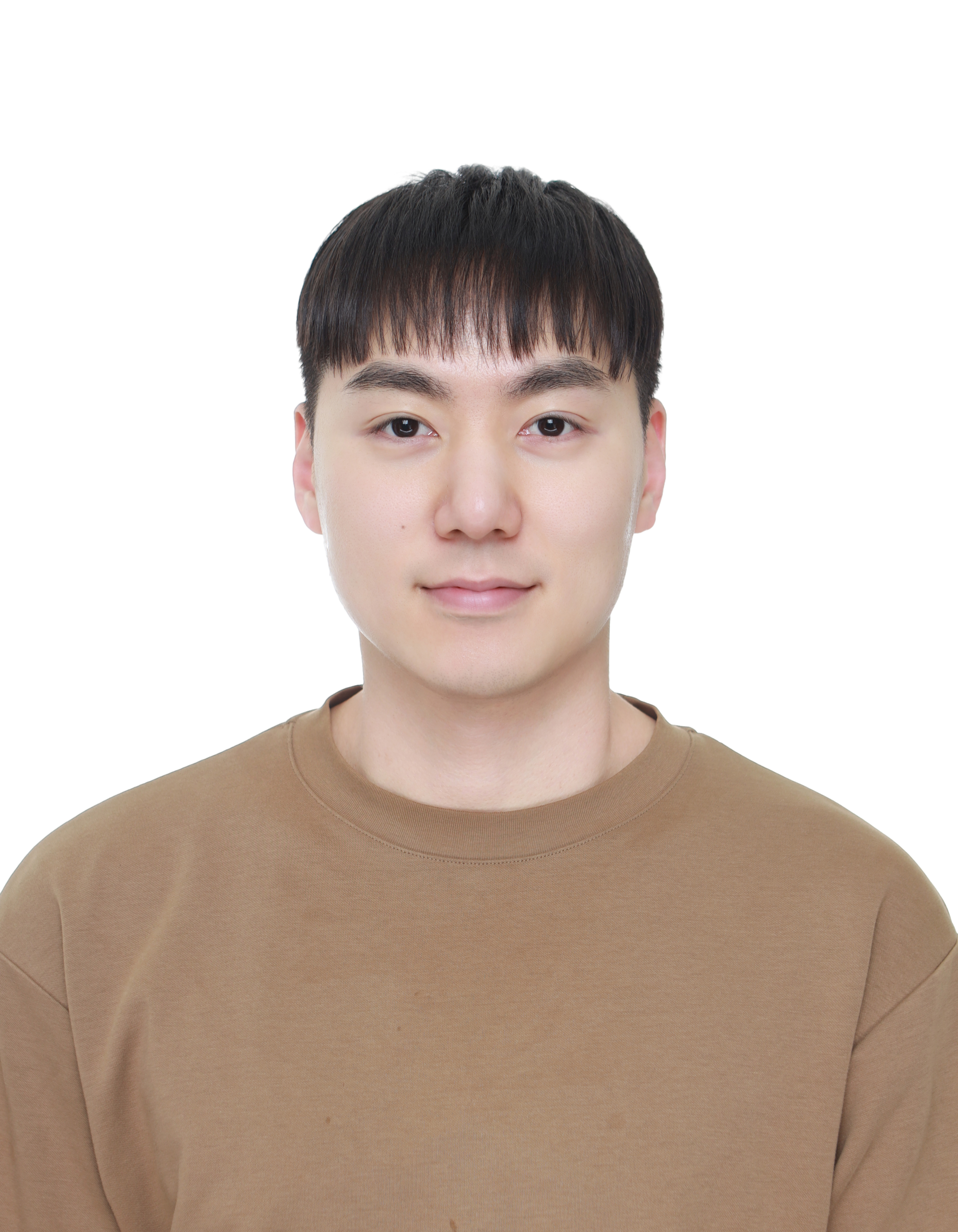}}]{Hojun You}
received the B.S. and Ph.D. degrees in Statistics from Seoul National University, South Korea. He was a Postdoctoral Fellow with the University of Houston, where he conducted research on multivariate spatio-temporal modeling and deep learning-based applications. He is currently a Machine Learning Engineer in industry. His research interests include multivariate spatio-temporal point processes and applied machine learning in statistical methodology.
\end{IEEEbiography}

\vspace{11pt}

\begin{IEEEbiography}[{\includegraphics[width=1in,height=1.25in,clip,keepaspectratio]{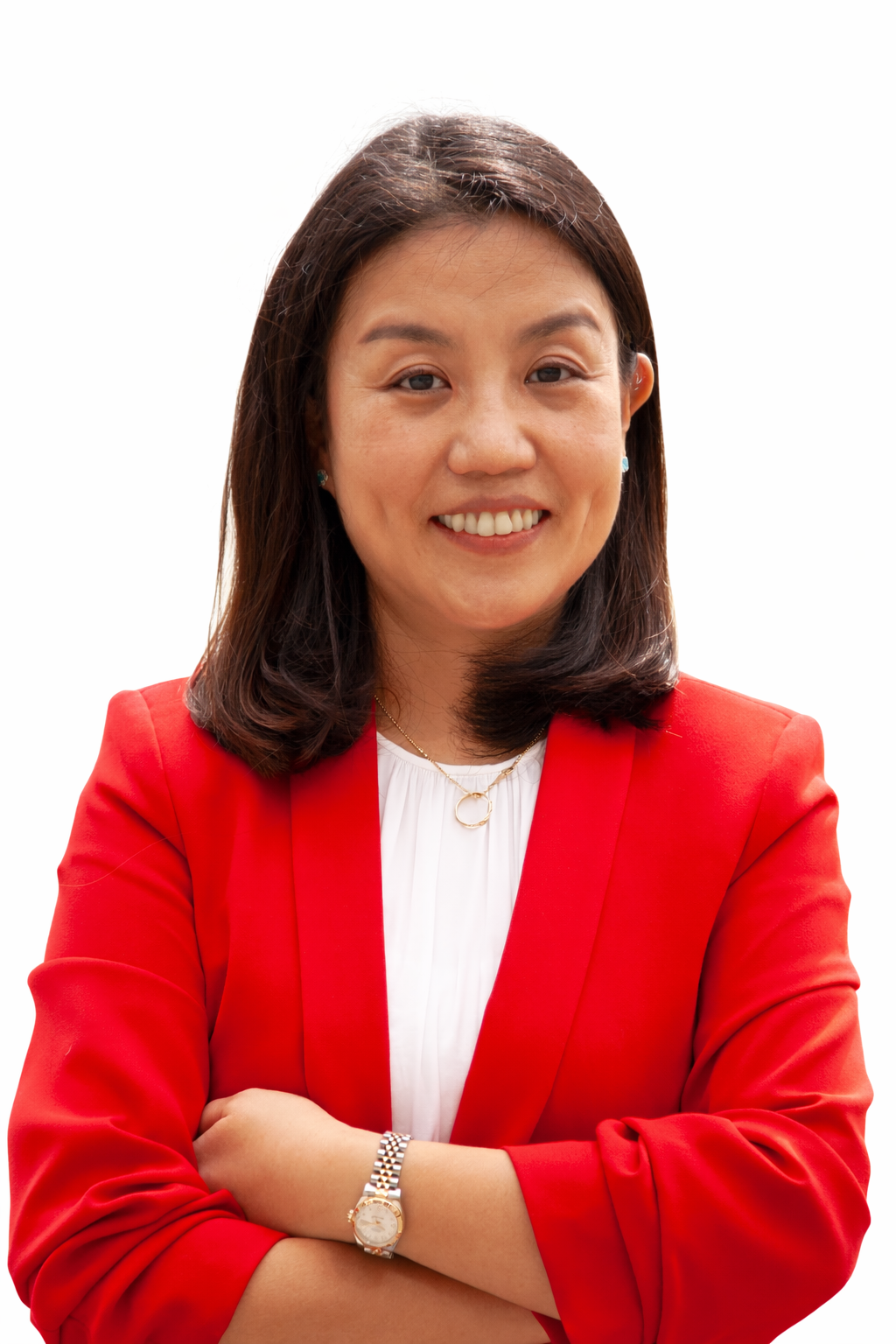}}]{Mikyoung Jun} 
is a Full Professor in the Department of Mathematics at the University of Houston. She received her Ph.D. degree in Statistics from University of Chicago in 2005. Prior to joining the University of Houston, she held an Assistant/Associate/Full Professor positions in the Department of Statistics at Texas A\&M University.
Her research interests include spatio-temporal covariance models, Gaussian processes, point processes, and their applications in environmental and social sciences. She is a Fellow of the American Statistical Association and an Elected Member of the International Statistical Institute.
\end{IEEEbiography}

\clearpage

\renewcommand{\thesection}{S.\arabic{section}}
\renewcommand{\theequation}{S.\arabic{equation}}
\renewcommand{\thefigure}{S\arabic{figure}}
\renewcommand{\thetable}{S\arabic{table}}
\setcounter{section}{0}
\setcounter{equation}{0}
\setcounter{figure}{0}
\setcounter{table}{0}

\clearpage
\twocolumn[{%
\centering
{\Large\bfseries Supplementary Material for\par}
\vspace{2pt}
{\Large\bfseries Neural Multivariate Spatio-Temporal Hawkes Processes\par}
\vspace{6pt}
{Christopher Chukwuemeka, Hojun You, and Mikyoung Jun\par}
\vspace{0.8em}
}]

\section{Comparison of six methods in EasyTPP}\label{sec:supp_easytpp}

\noindent This section contains details on the simulation experiment comparing six methods in EasyTPP in Section II. 


\begin{table}[!ht]
\centering
\resizebox{!}{0.2\textheight}{
\begin{tabular}{lcr}
\toprule
Model &  {Hyperparameters} &  {Value} \\
\midrule
\multirow{3}{*}{RMTPP} & \texttt{num\_layers} & 2 \\
& \texttt{hidden\_size} & 32 \\
& \texttt{time\_emb\_size} & 16 \\
\midrule
\multirow{3}{*}{ {NHP}} & \texttt{num\_layers} & 2 \\
& \texttt{hidden\_size} & 64 \\
& \texttt{time\_emb\_size} & 16 \\
\midrule
\multirow{3}{*}{ {ODETPP}} & \texttt{num\_layers} & 2 \\
& \texttt{hidden\_size} & 32 \\
& \texttt{time\_emb\_size} & 16 \\
\midrule
\multirow{4}{*}{ {SAHP}} & \texttt{num\_layers} & 2 \\
& \texttt{num\_heads} & 2 \\
& \texttt{hidden\_size} & 32 \\
& \texttt{time\_emb\_size} & 16 \\
\midrule
\multirow{4}{*}{ {THP}} & \texttt{num\_layers} & 2 \\
& \texttt{num\_heads} & 2 \\
& \texttt{hidden\_size} & 64 \\
& \texttt{time\_emb\_size} & 16 \\
\midrule
\multirow{4}{*}{ {AttNHP}} & \texttt{num\_layers} & 1 \\
& \texttt{num\_heads} & 2 \\
& \texttt{hidden\_size} & 32 \\
& \texttt{time\_emb\_size} & 16 \\
\bottomrule
\end{tabular}
}
\caption{Hyperparameter settings for trainining  temporal point process (TPP) models.} 
\label{tab:hyperparameters}
\end{table}


\begin{table}[!hbpt]
\centering
\begin{tabular}{lrr}
\toprule
MModel & Trainable Parameters & Runtime (min) \\ \midrule
RMTPP & 2,342 & 0.651 \\
NHP & 58,112 & 47.173 \\
ODETPP & 1,218 & 645.758 \\
SAHP & 6,514 & 0.996 \\
THP & 25,282 & 0.956 \\
AttNHP & 11,202 & 22.321 \\ \bottomrule
\end{tabular}
\caption{Comparison of model complexity and training runtime for the temporal point process models. Number of trainable parameters and the total wall clock time required to complete 2,000 training epochs using the hyperparameter configurations in Table~\ref{tab:hyperparameters} are reported.}
\label{tab:model_complexity}
\end{table}


\begin{figure}[!htbp]
\centering

\subfloat[NHP]{%
  \includegraphics[width=.48\linewidth]{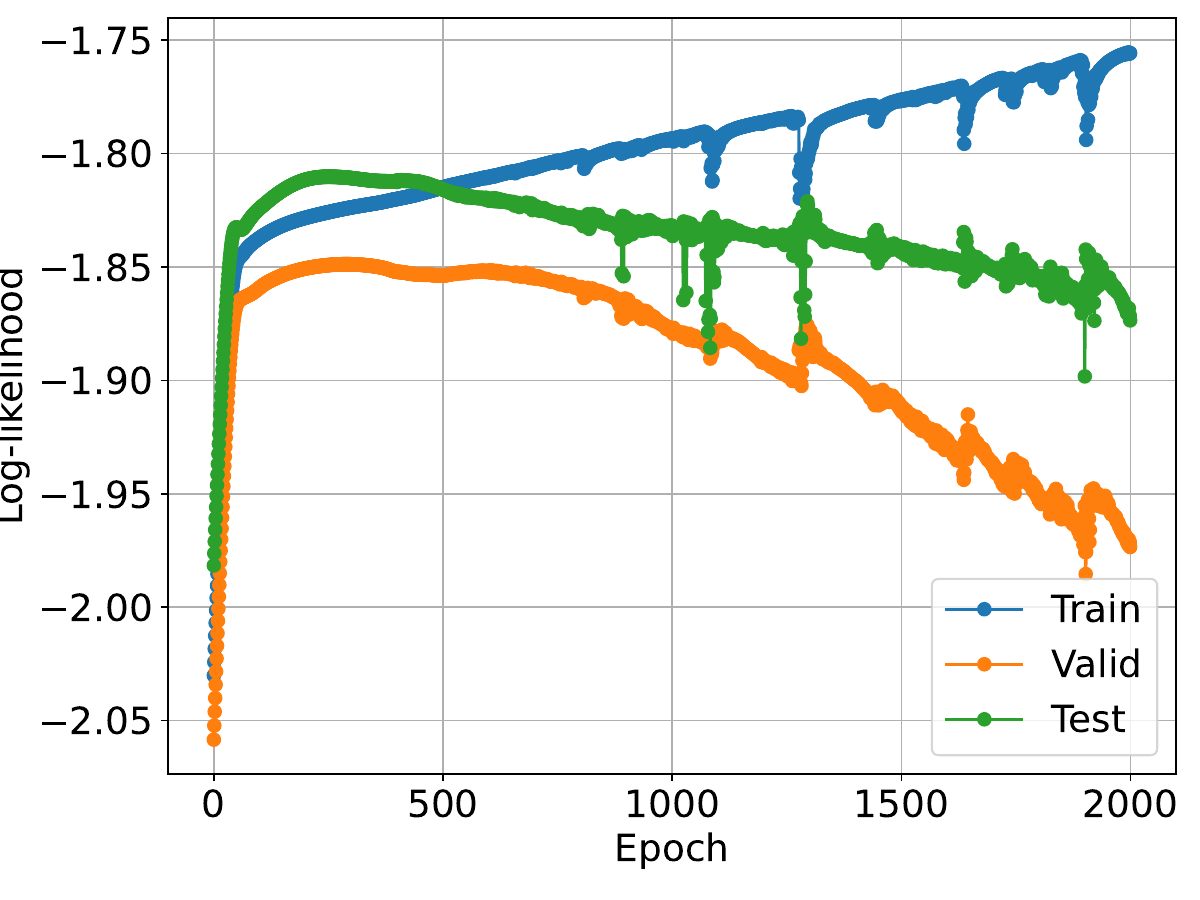}%
}\hfill
\subfloat[AttNHP]{%
  \includegraphics[width=.48\linewidth]{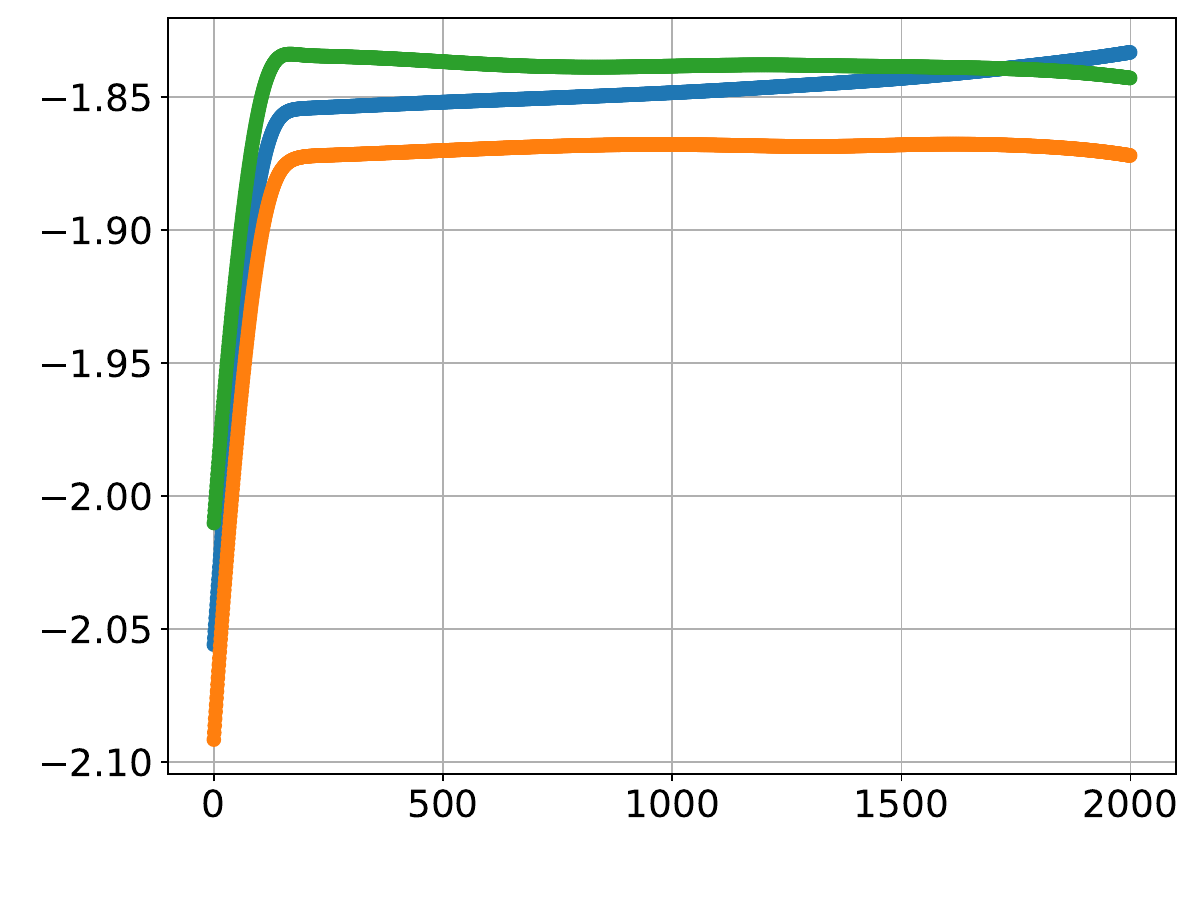}%
}\\

\subfloat[SAHP]{%
  \includegraphics[width=.48\linewidth]{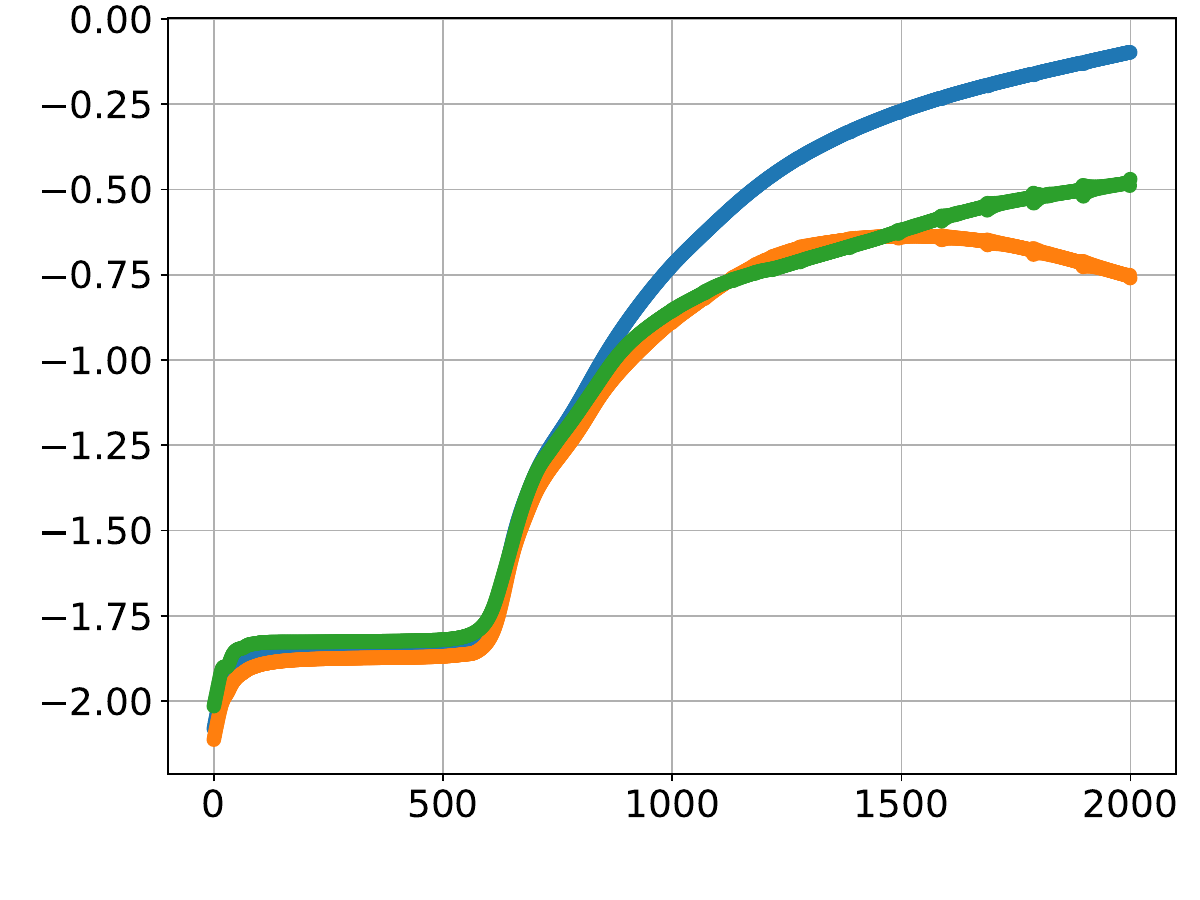}%
}\hfill
\subfloat[ODETPP]{%
  \includegraphics[width=.48\linewidth]{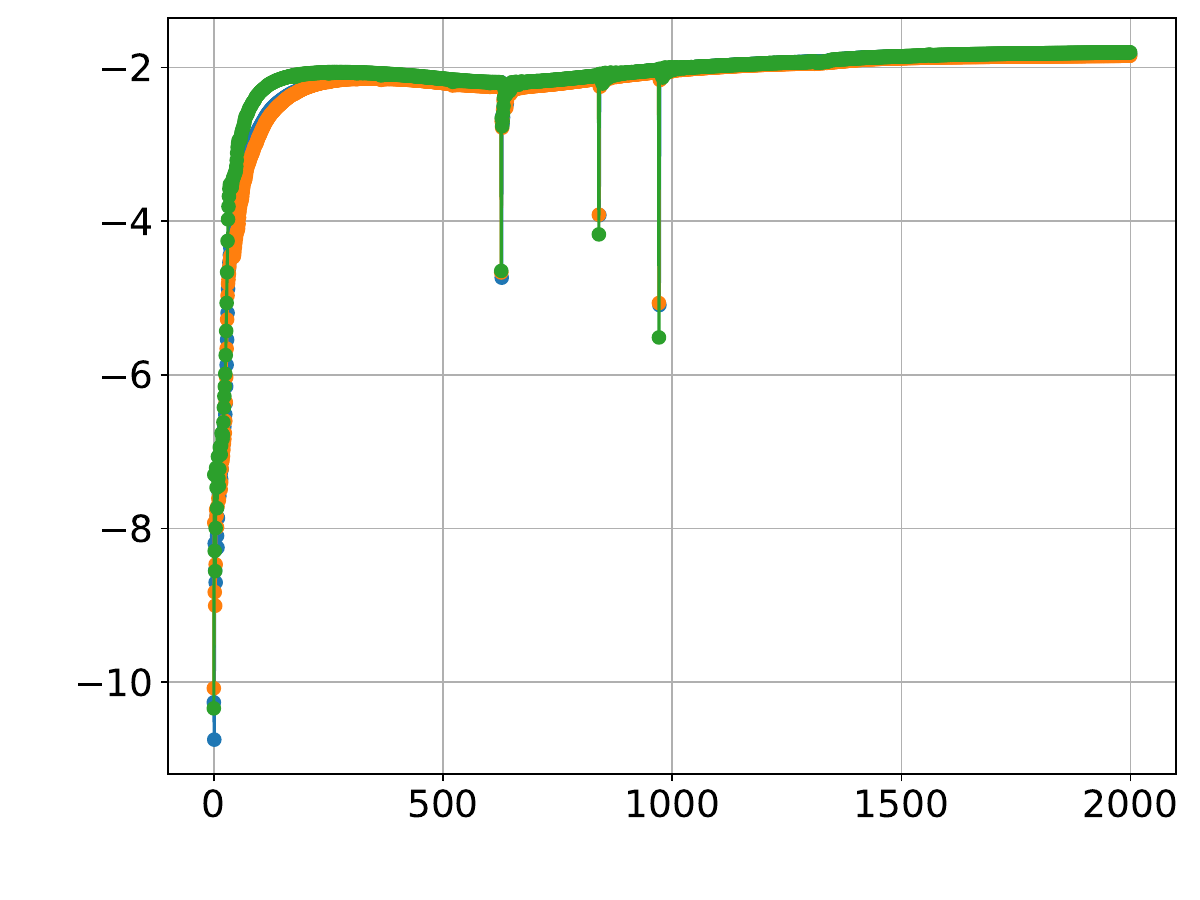}%
}\\

\subfloat[THP]{%
  \includegraphics[width=.48\linewidth]{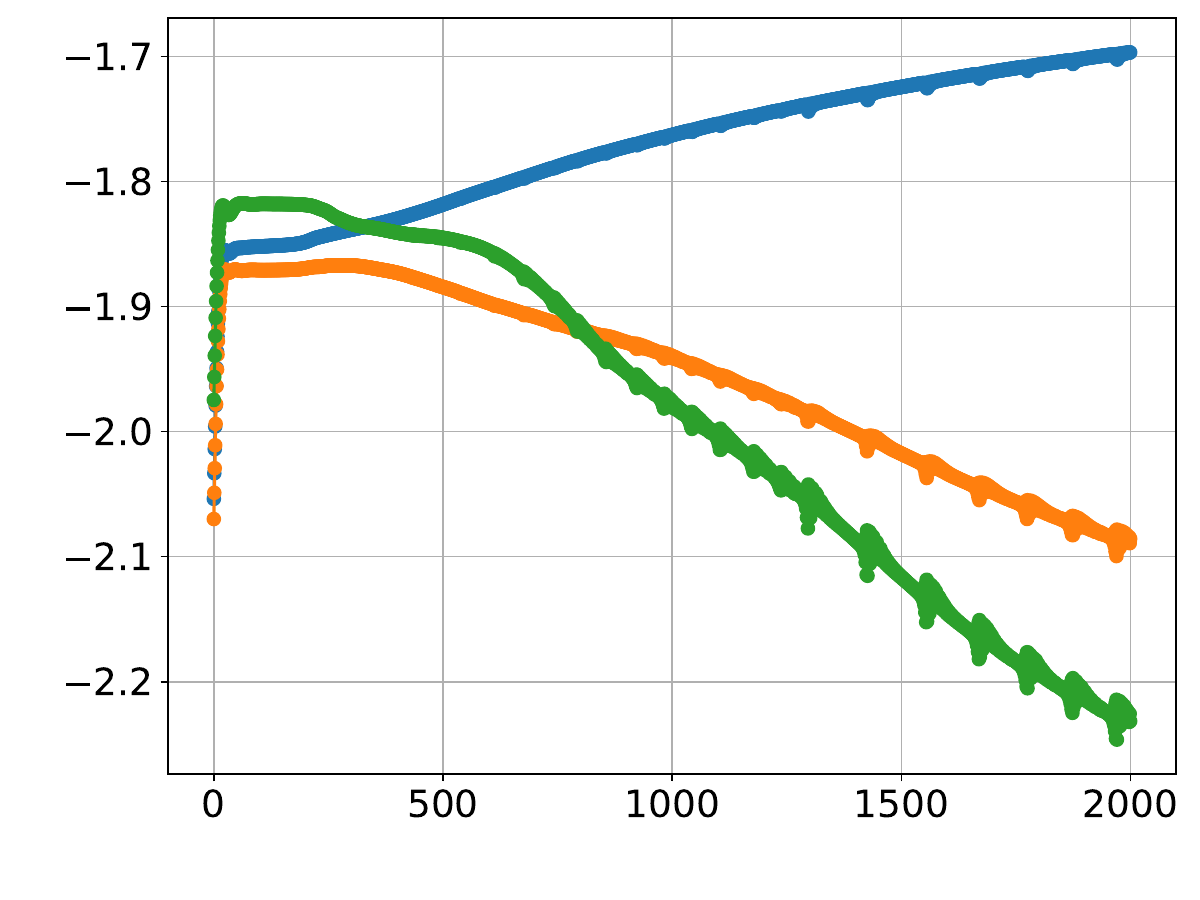}%
}\hfill
\subfloat[RMTPP]{%
  \includegraphics[width=.48\linewidth]{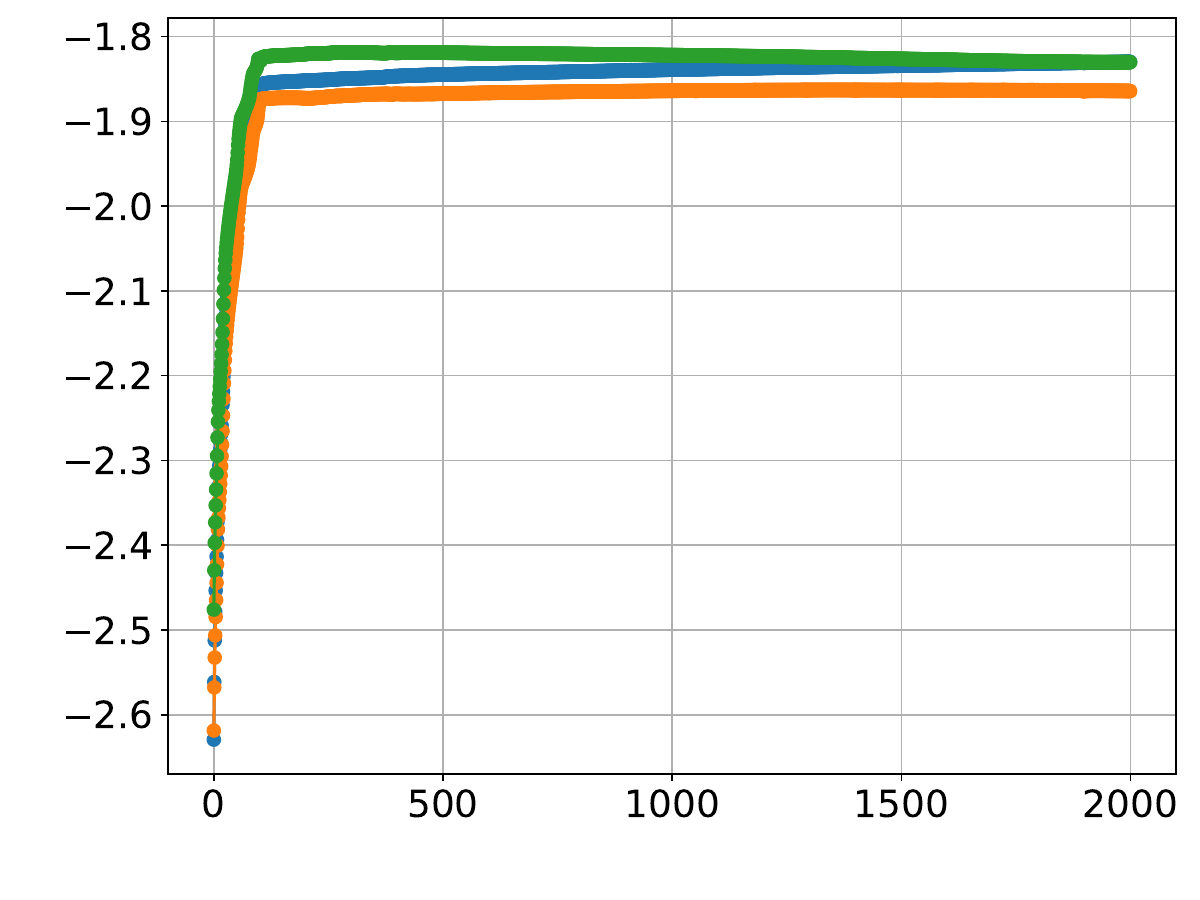}%
}
\caption{Log-likelihood values of training, testing,  validation data sets for six temporal point process models implemented in EasyTPP.}
\label{fig:convplots_2k_epochs}
\end{figure}

\clearpage

\subsection{SAHP results by epochs}\label{subsec:supp_sahp_runs}

\noindent The figure below shows the difficulty with SAHP method. 
Across each training budgets, SAHP initially learns a reasonable temporal trend but subsequently drifts toward a degenerate solution in which the intensity collapses between events, yielding highly irregular intensity profiles despite continued improvement in log-likelihood.

\begin{figure}[htpb!]
\centering

\subfloat[500 epochs]{%
  \begin{minipage}{\linewidth}
    \centering
   \includegraphics[width=.45\linewidth]{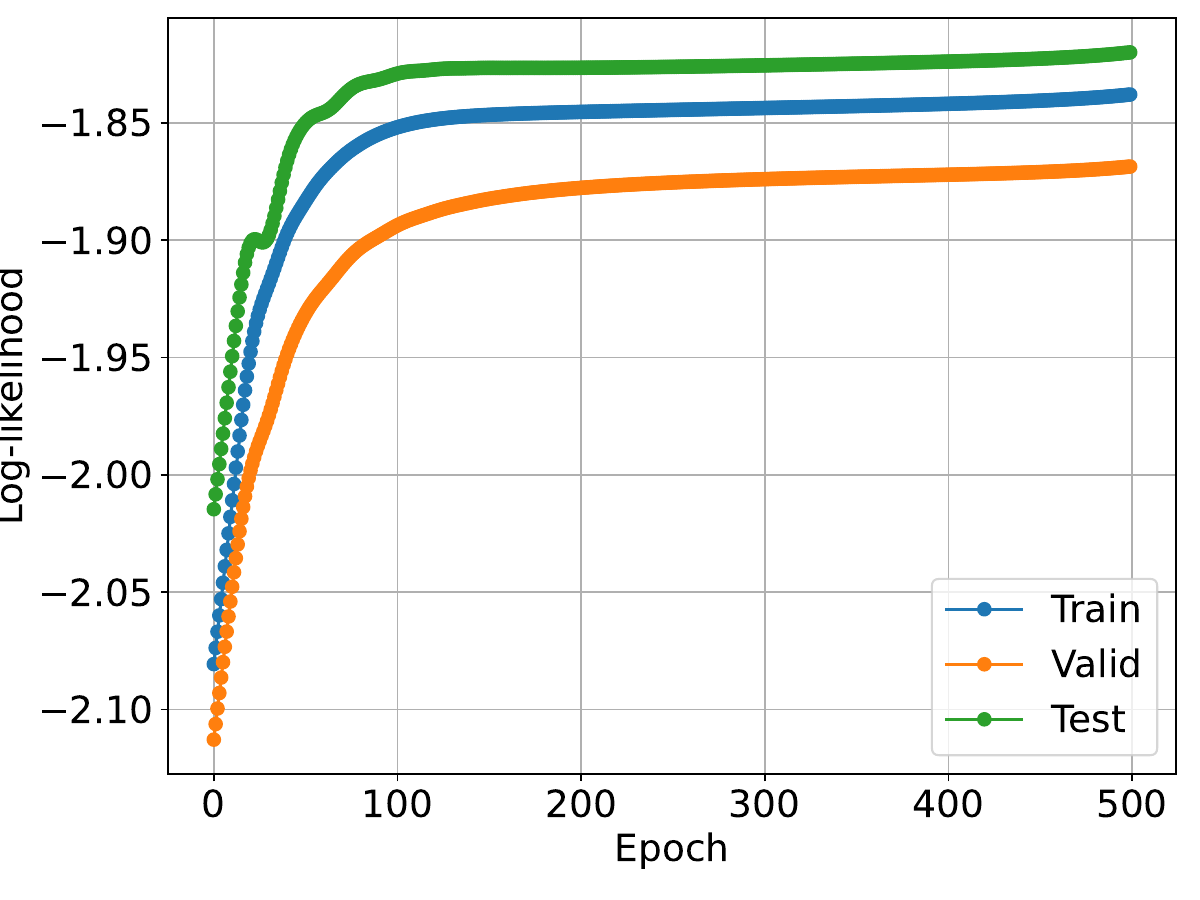}
    \includegraphics[width=.5\linewidth]{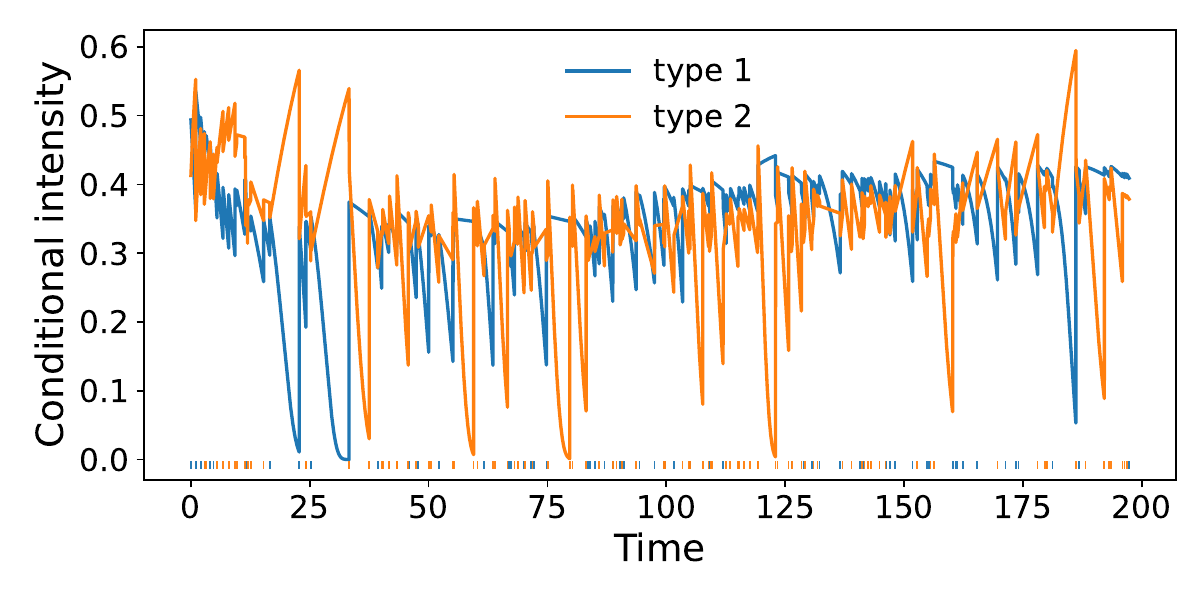}%
  \end{minipage}
}\\[8pt]

\subfloat[650 epochs]{%
  \begin{minipage}{\linewidth}
    \centering
    \includegraphics[width=.45\linewidth]{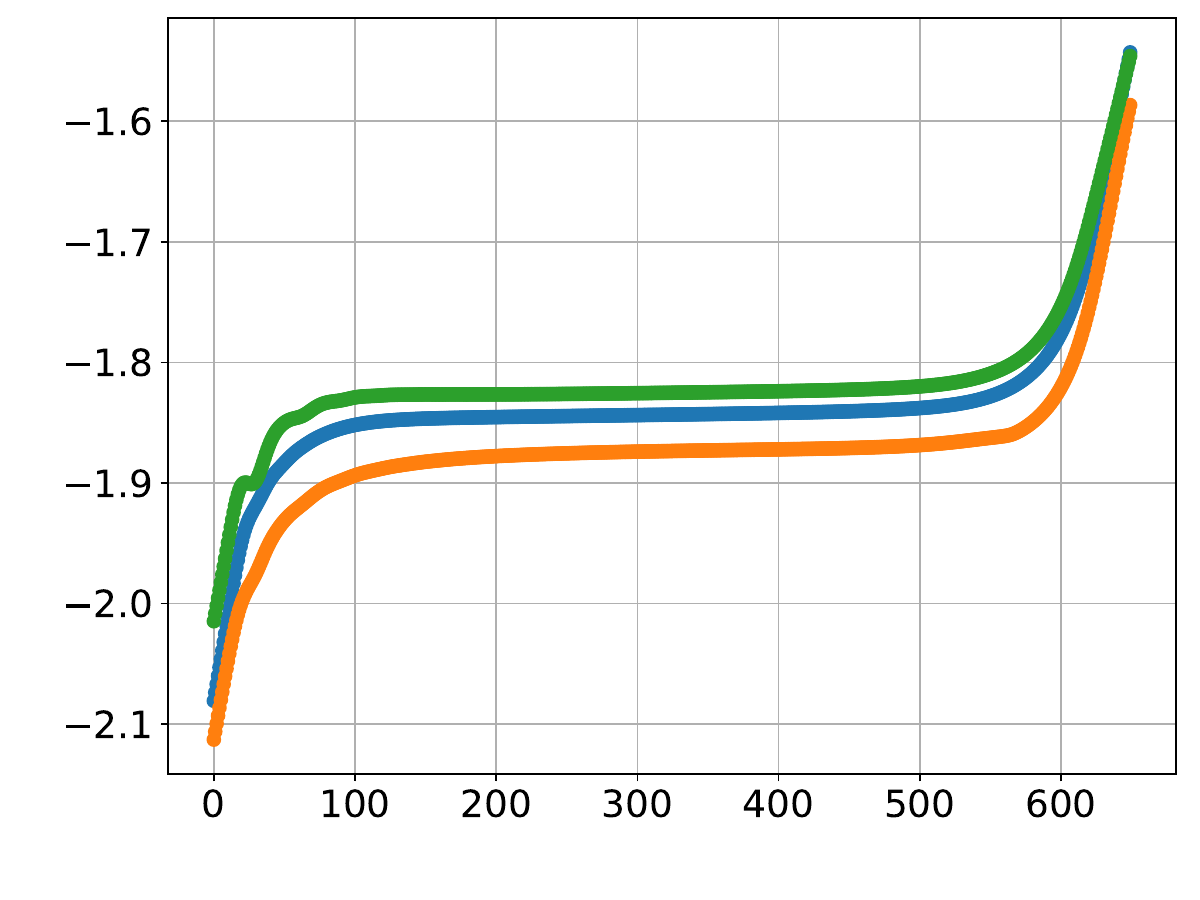}
    \includegraphics[width=.5\linewidth]{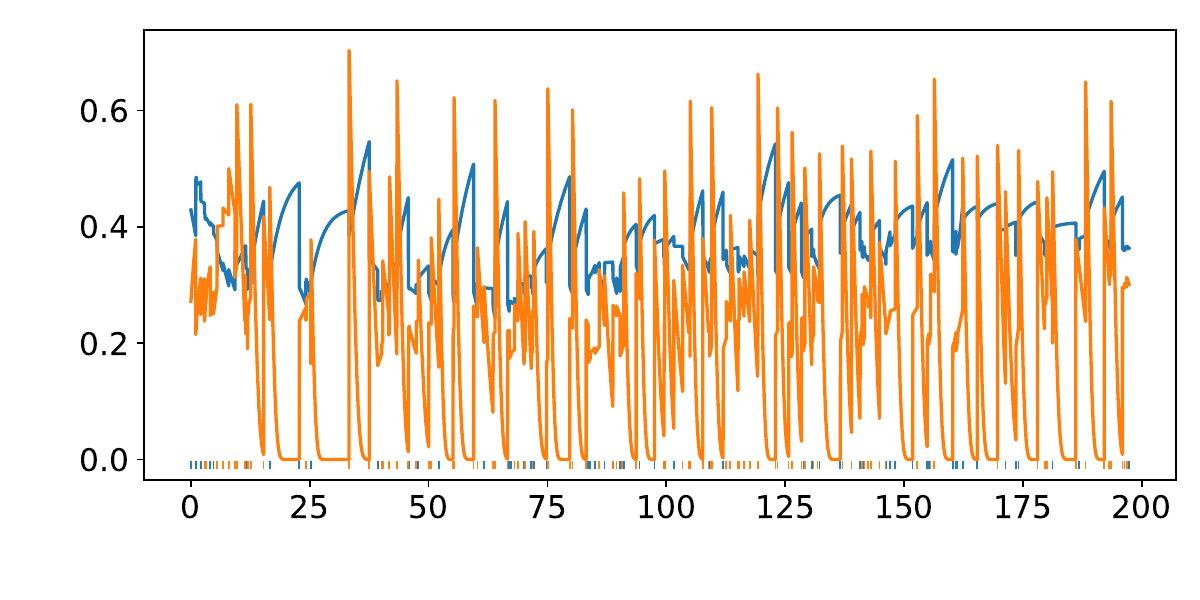}%
  \end{minipage}
}\\[8pt]

\subfloat[1k epochs]{%
  \begin{minipage}{\linewidth}
    \centering
     \includegraphics[width=.45\linewidth]{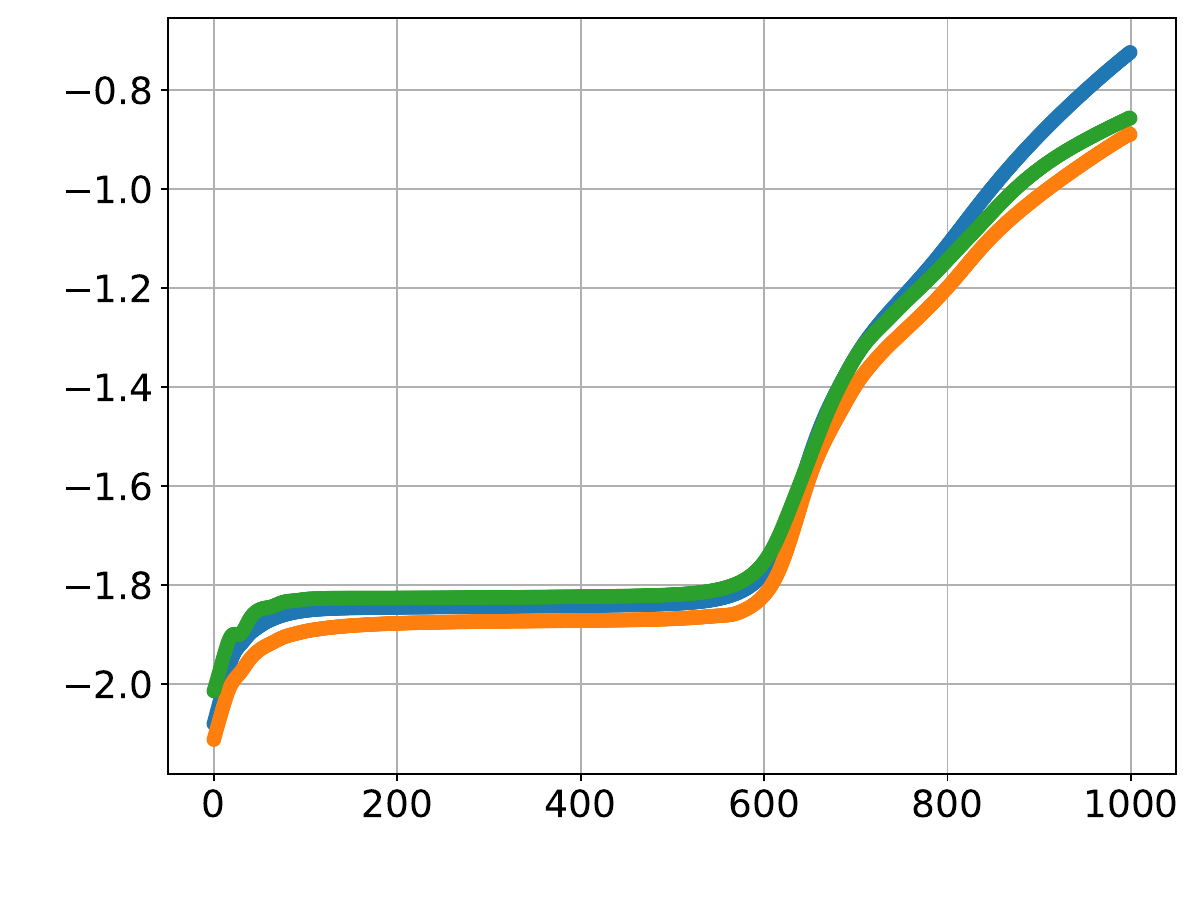}
    \includegraphics[width=.5\linewidth]{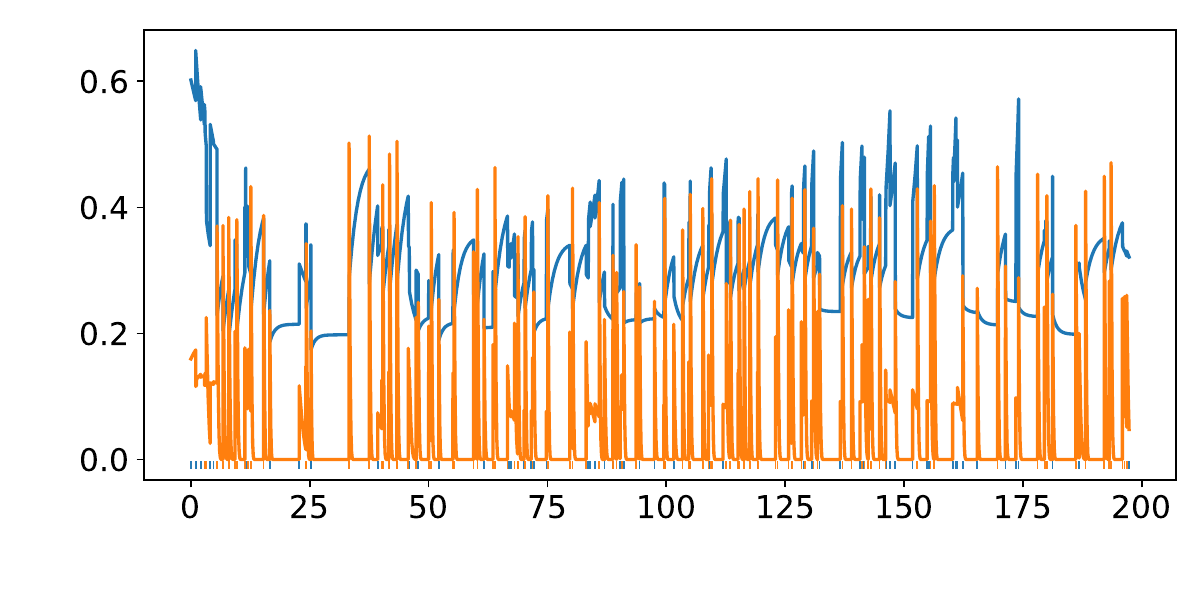}%
  \end{minipage}
}\\[8pt]

\subfloat[2k epochs]{%
  \begin{minipage}{\linewidth}
    \centering
    \includegraphics[width=.45\linewidth]{figures/biv_hawkes/sahp/sahp_2k_epoch_loss.pdf}
    \includegraphics[width=.5\linewidth]{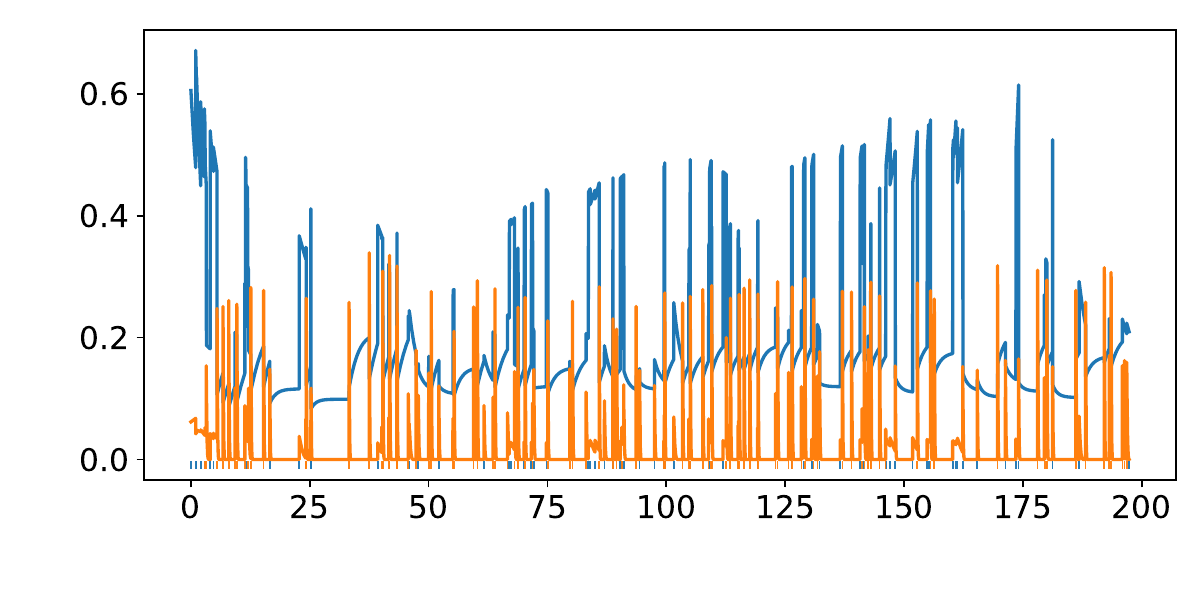}%
  \end{minipage}
}

\caption{SAHP runs under different training budgets. 
Left: Right: the evolution of the  log-likelihood as a function of epoch. Right: fitted temporal conditional intensity for a representative sequence.
}

\label{fig:sahp_runs}
\end{figure}

\section{Simulation study with bivariate spatio-temporal point patterns}\label{sec:supp_mstnhp}

This section presents details on the simulation results in Section IV. 

\begin{figure}[!htbp]
\centering

\subfloat[Biv 1\label{fig:biv1}]{%
  \includegraphics[width=.45\linewidth]{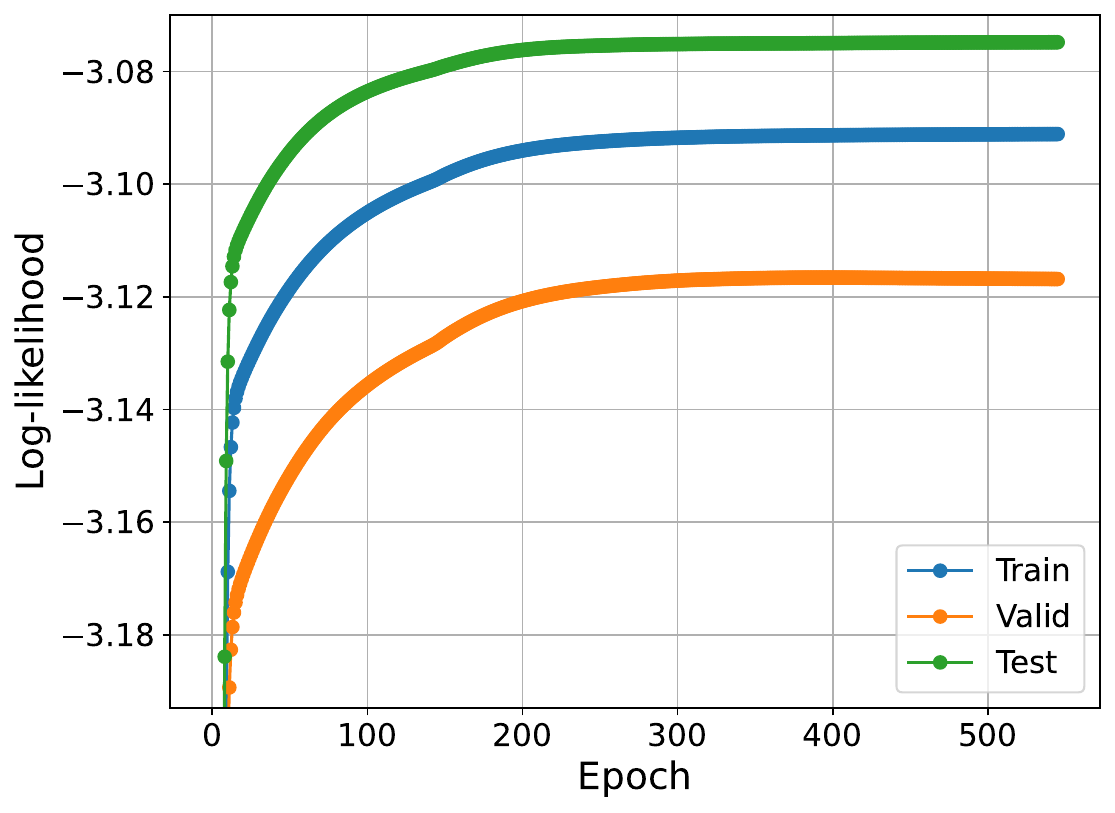}%
}
\subfloat[Biv 2\label{fig:biv2}]{%
  \includegraphics[width=.45\linewidth]{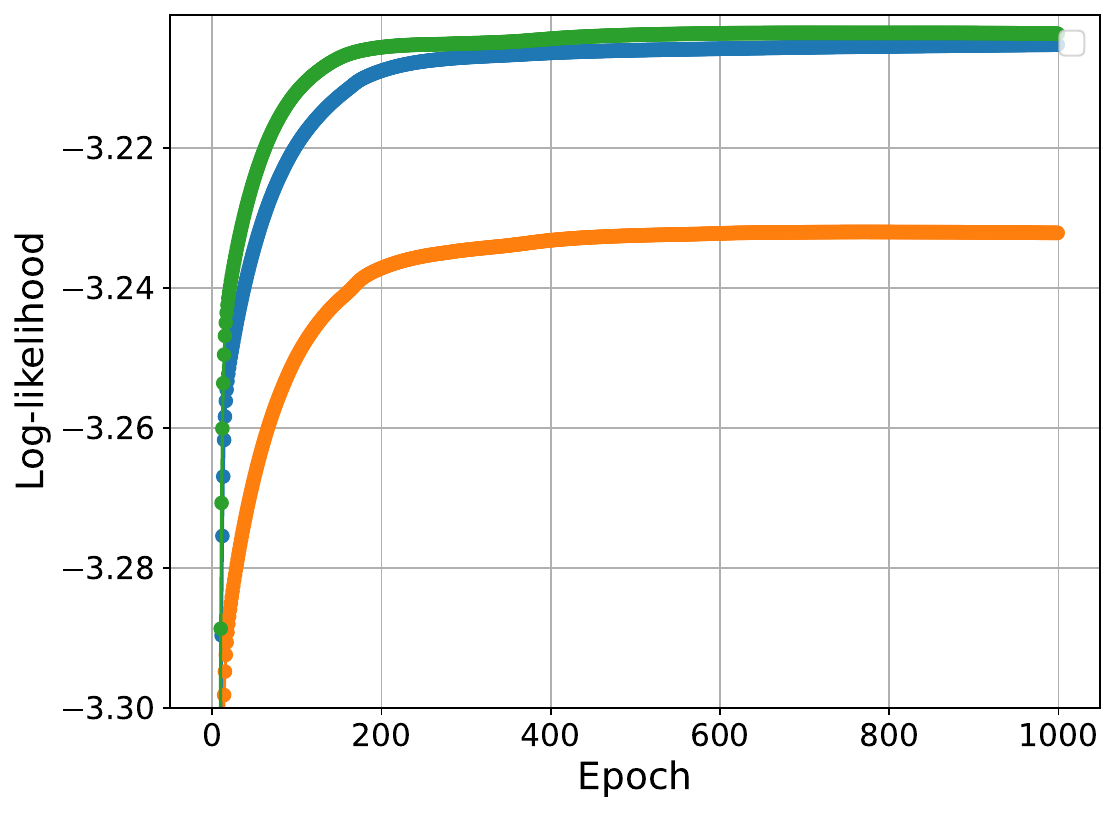}%
}\\

\subfloat[Biv 3\label{fig:biv3}]{%
  \includegraphics[width=.45\linewidth]{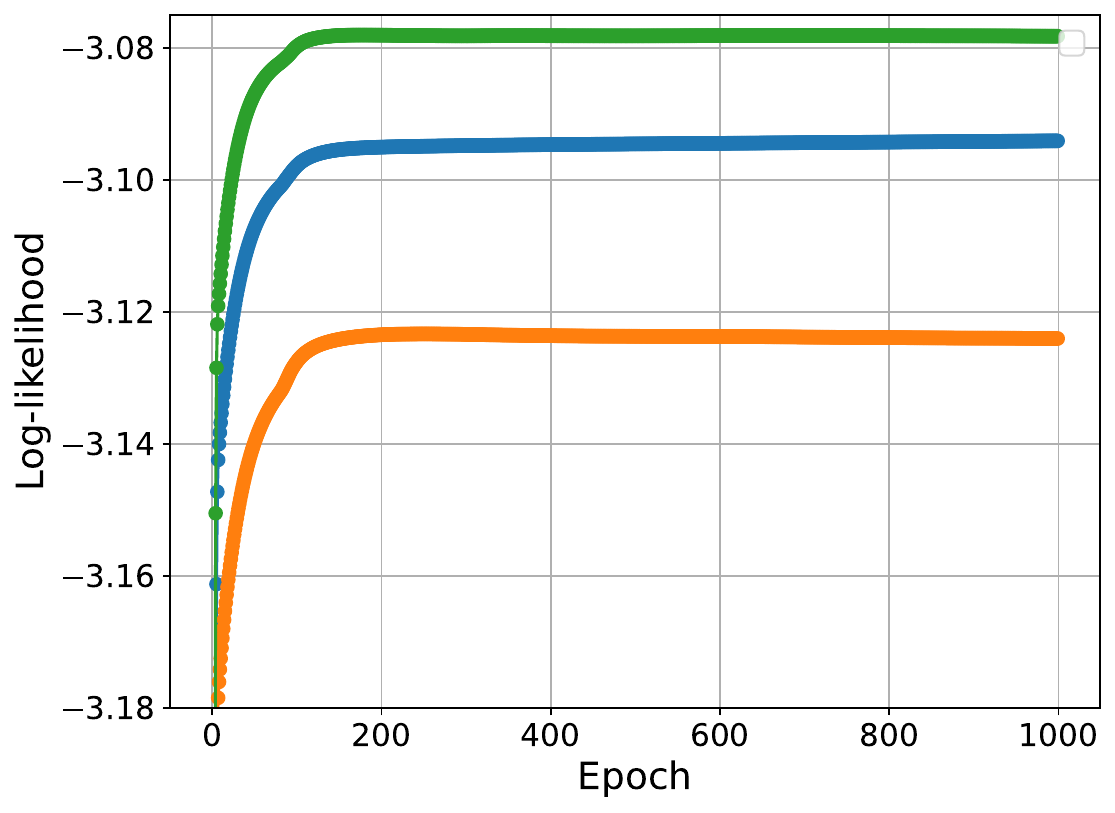}%
}
\subfloat[Biv 4\label{fig:biv4}]{%
  \includegraphics[width=.45\linewidth]{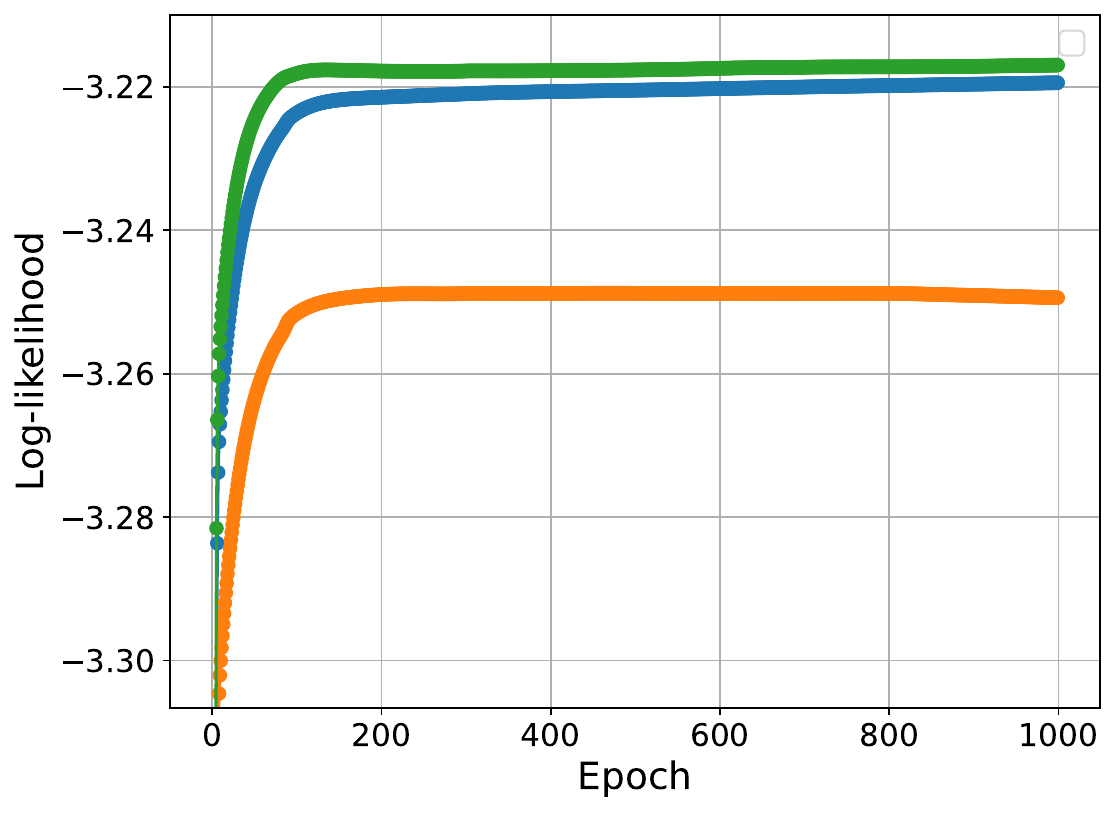}%
}
\caption{Convergence of the log-likelihood for the MSTNHP approach. }
\label{fig:biv1-4_images_conv}
\end{figure}

\begin{figure}[htbp!]
\centering
\setlength{\tabcolsep}{1pt}
\setlength{\extrarowheight}{3pt}

\begin{tabular}{cccc}
{True} & {Fitted} & {True} & {Fitted} \\[2pt]

\includegraphics[width=0.22\columnwidth]{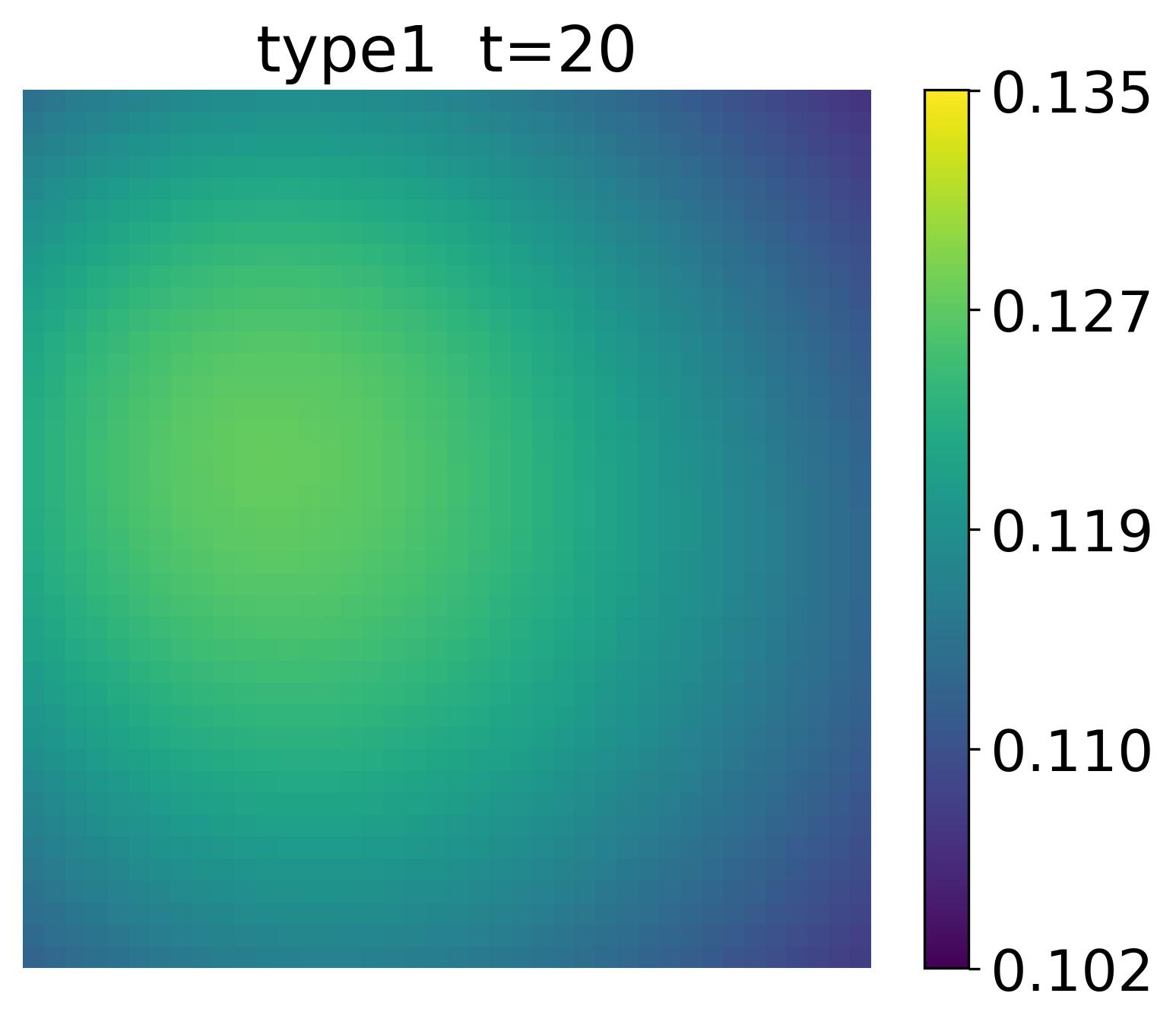} &
\includegraphics[width=0.22\columnwidth]{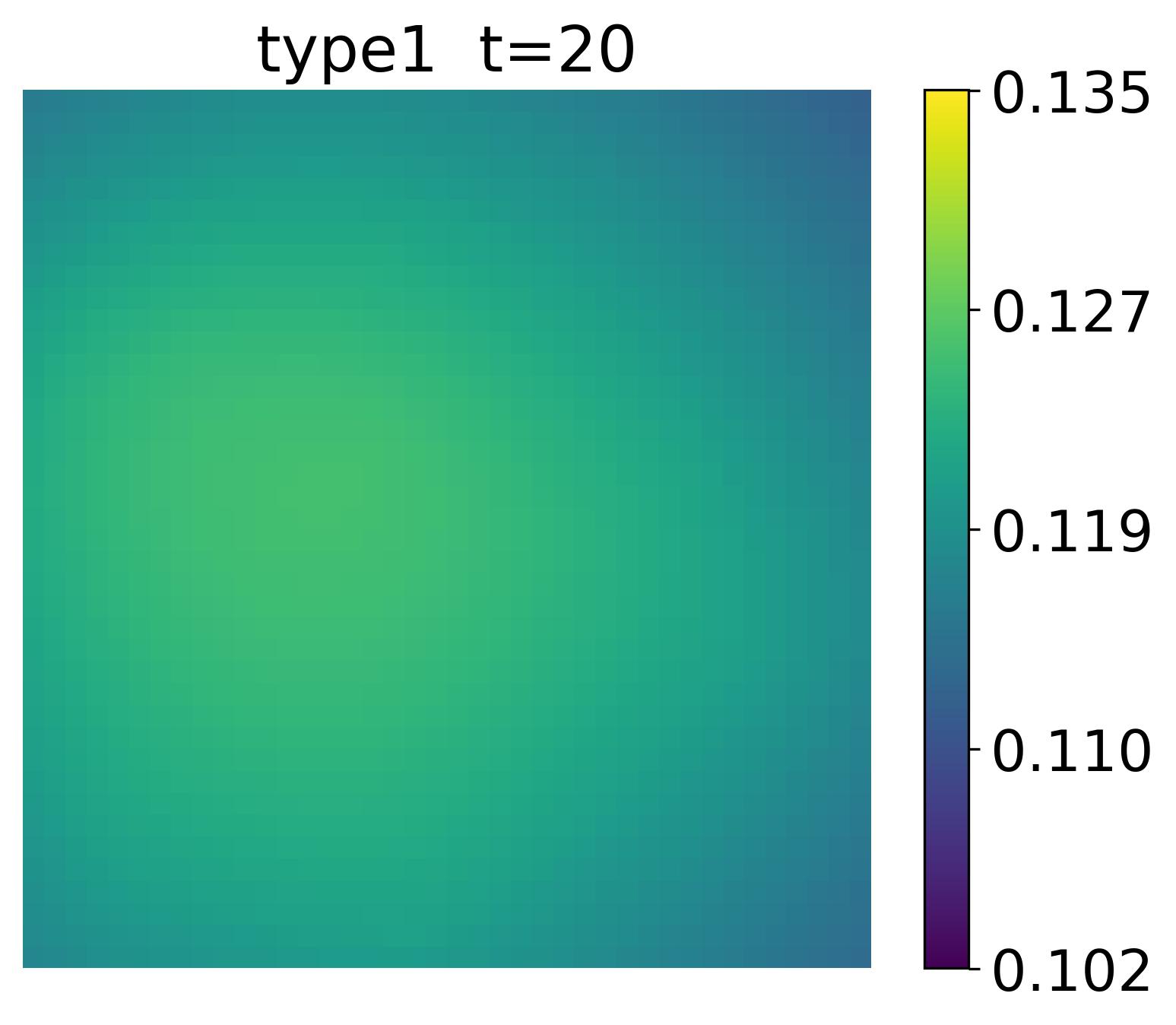} &
\includegraphics[width=0.22\columnwidth]{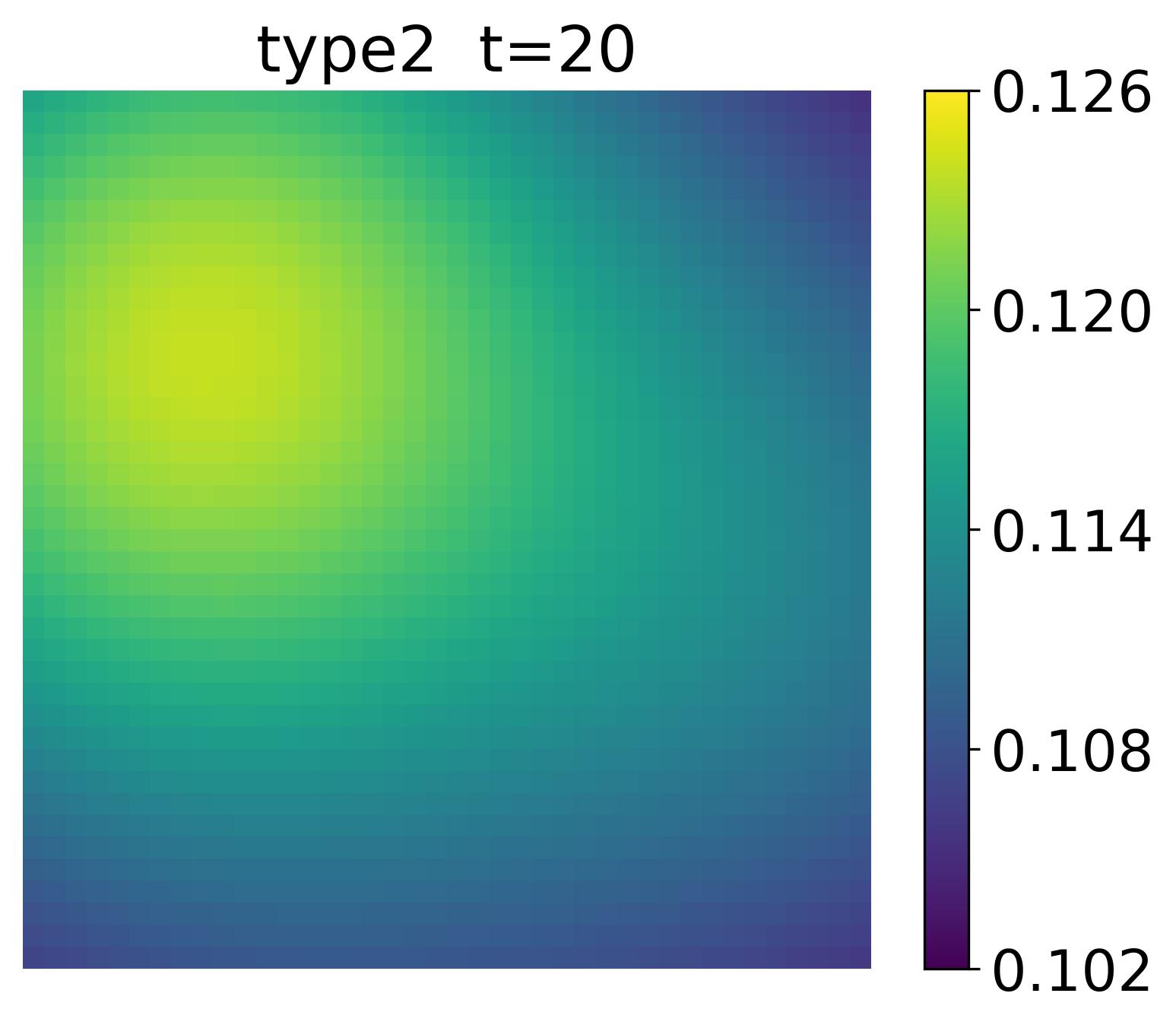} &
\includegraphics[width=0.22\columnwidth]{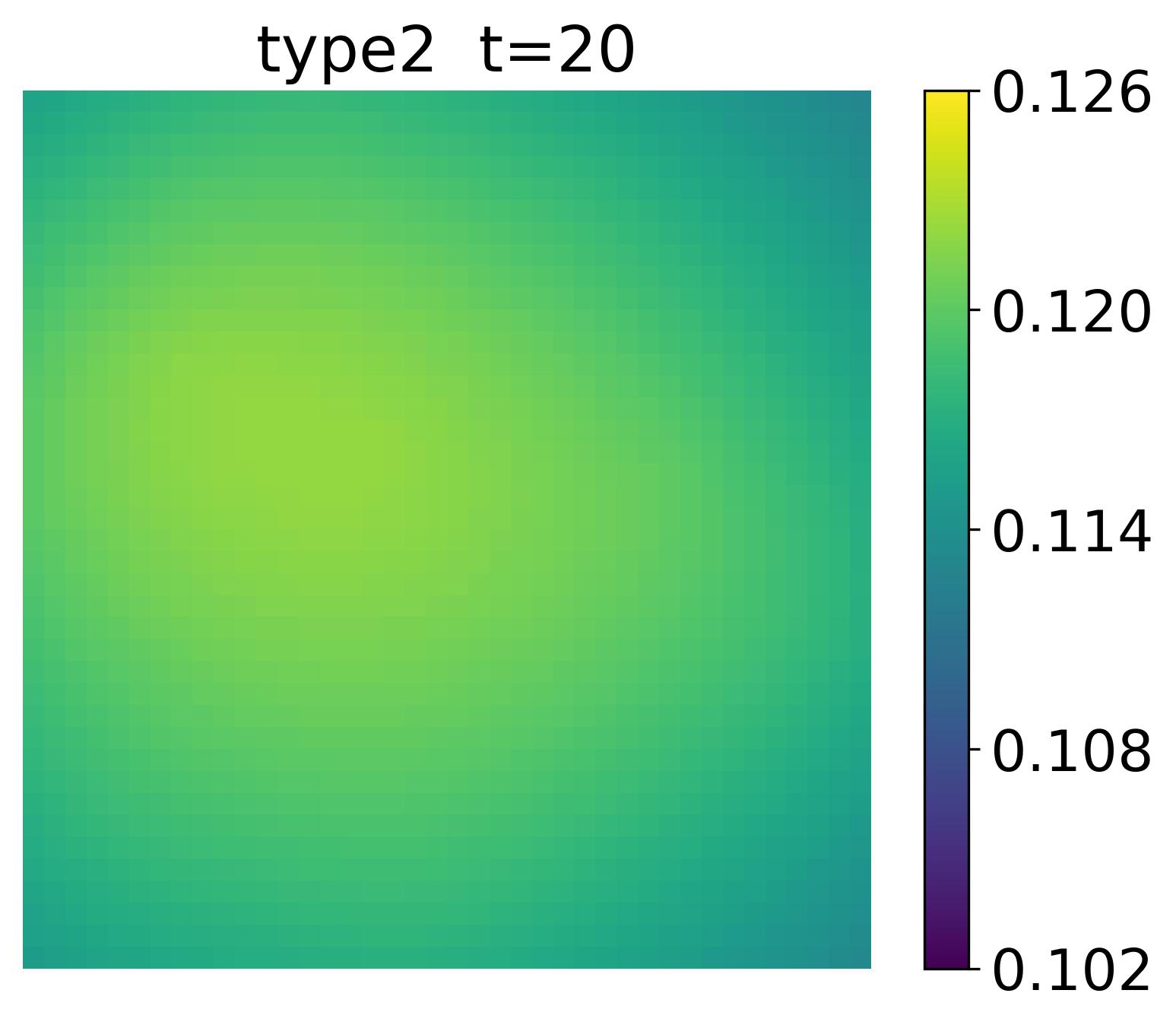} \\

\includegraphics[width=0.22\columnwidth]{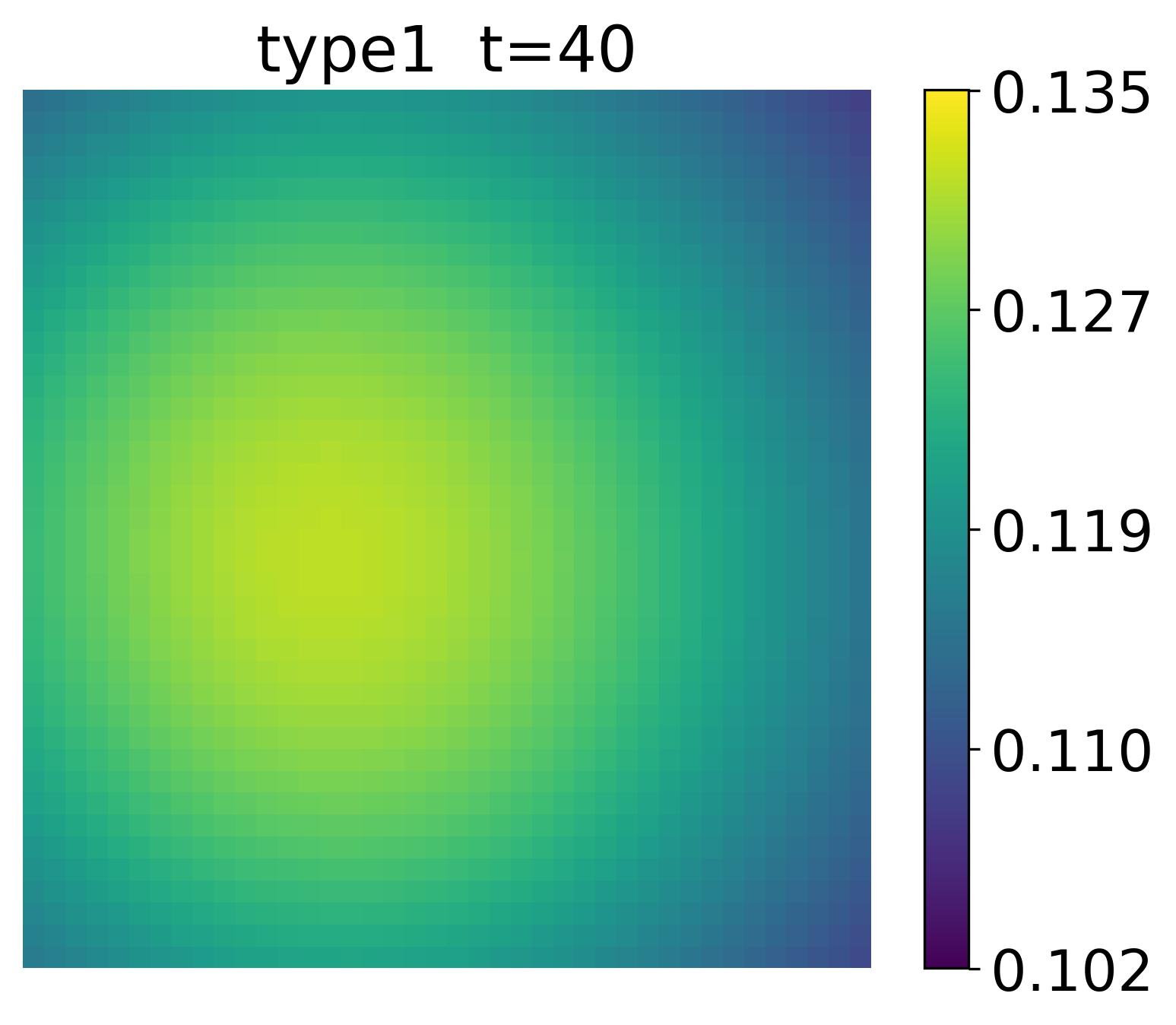} &
\includegraphics[width=0.22\columnwidth]{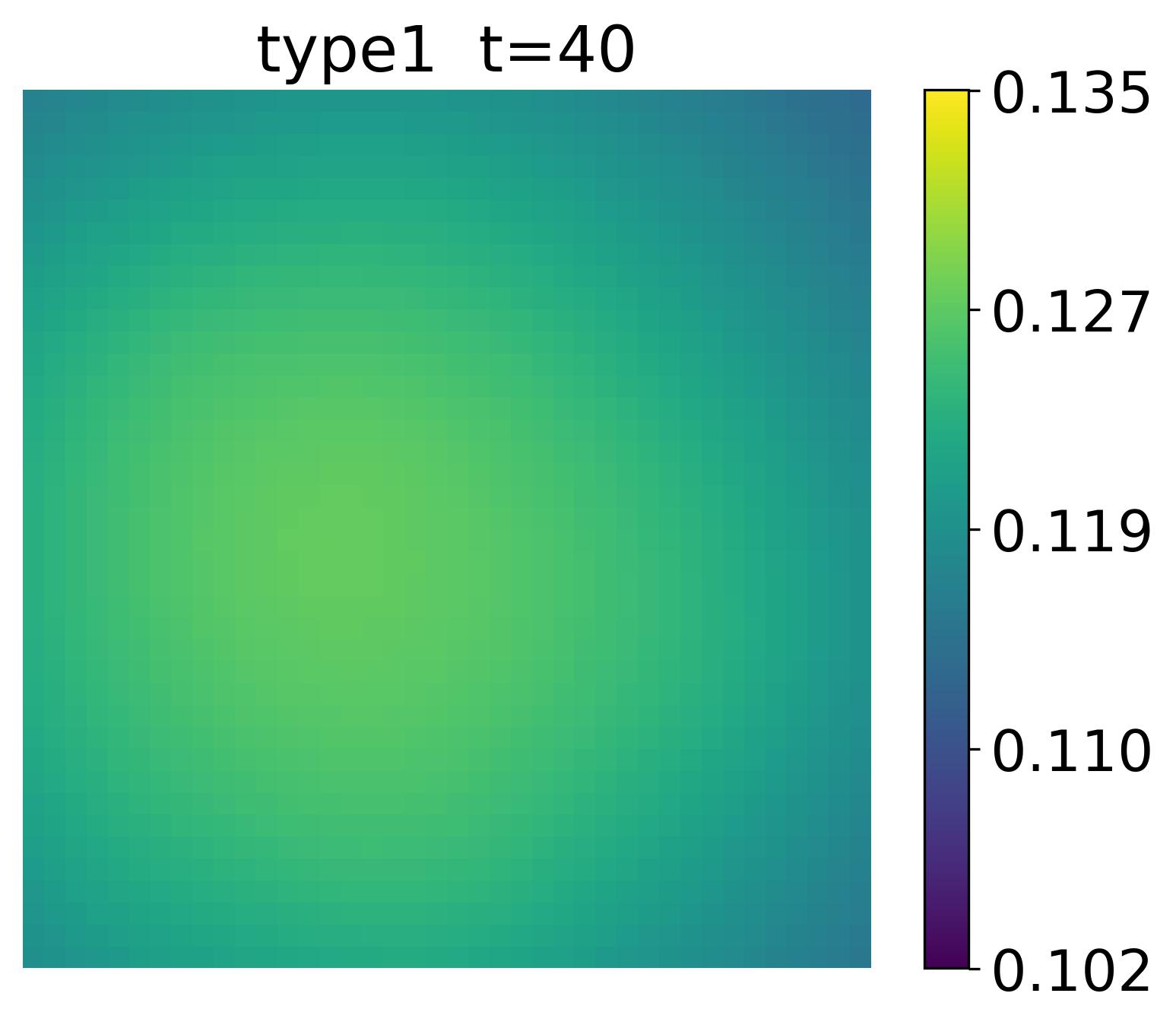} &
\includegraphics[width=0.22\columnwidth]{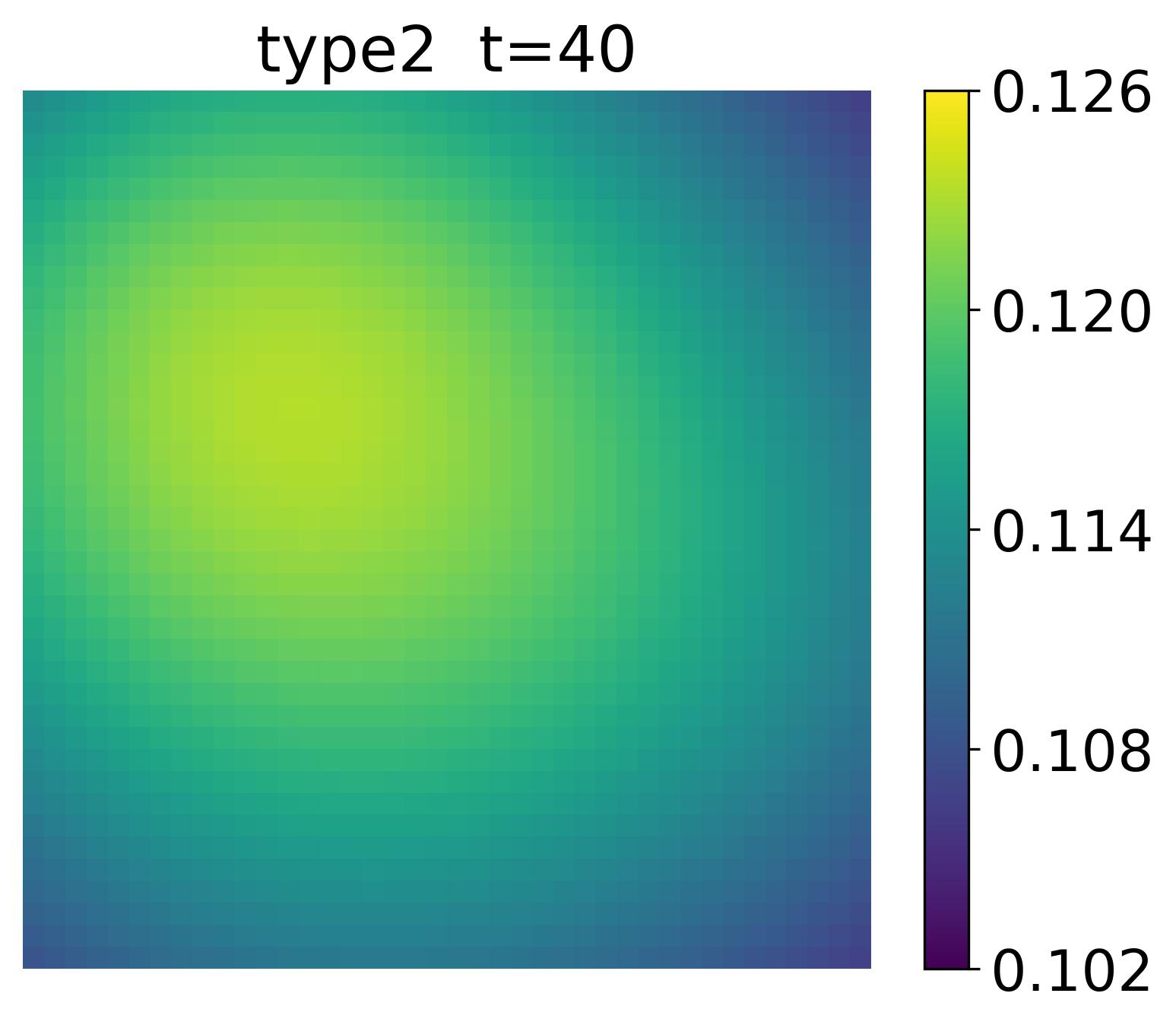} &
\includegraphics[width=0.22\columnwidth]{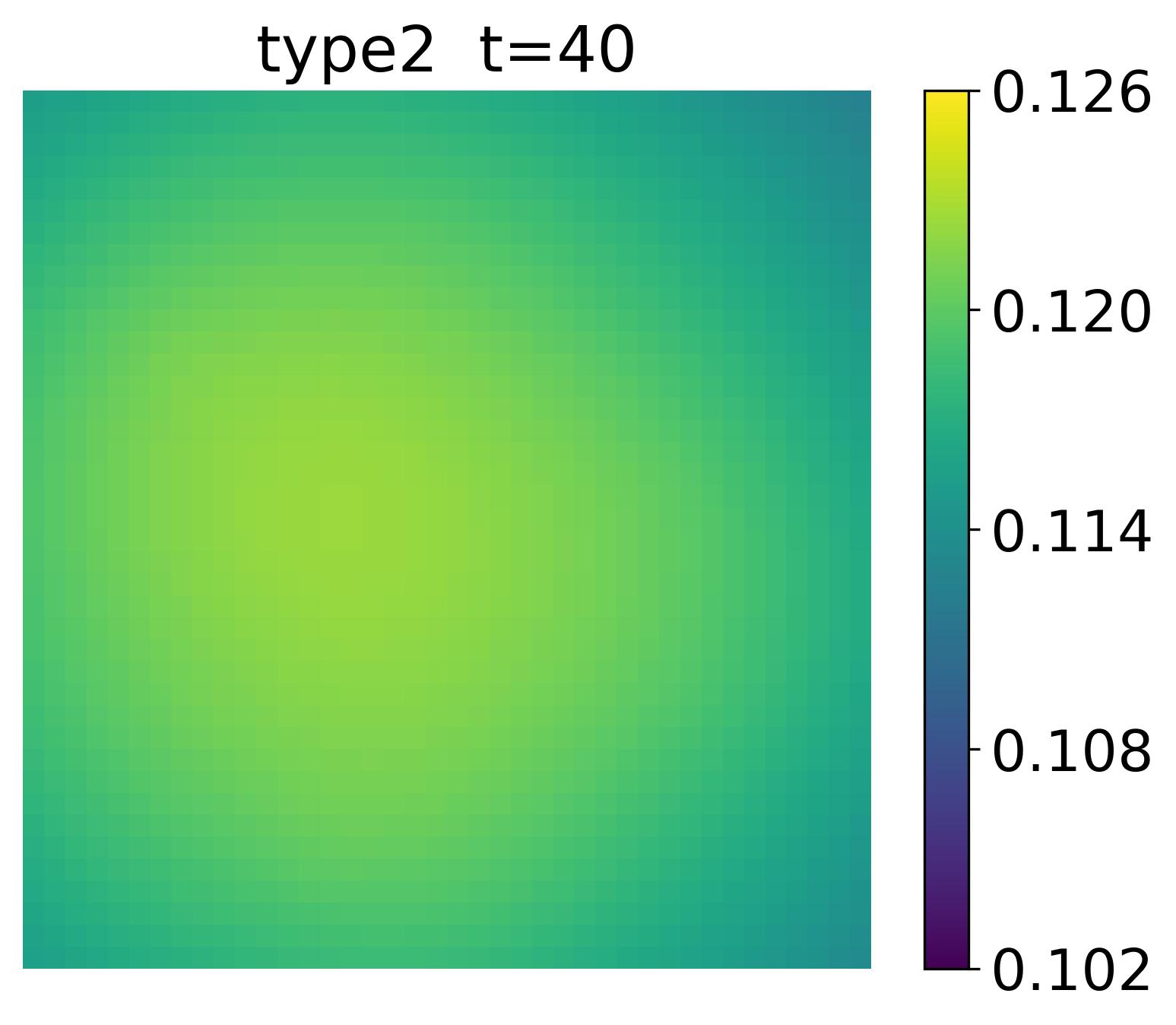} \\

\includegraphics[width=0.22\columnwidth]{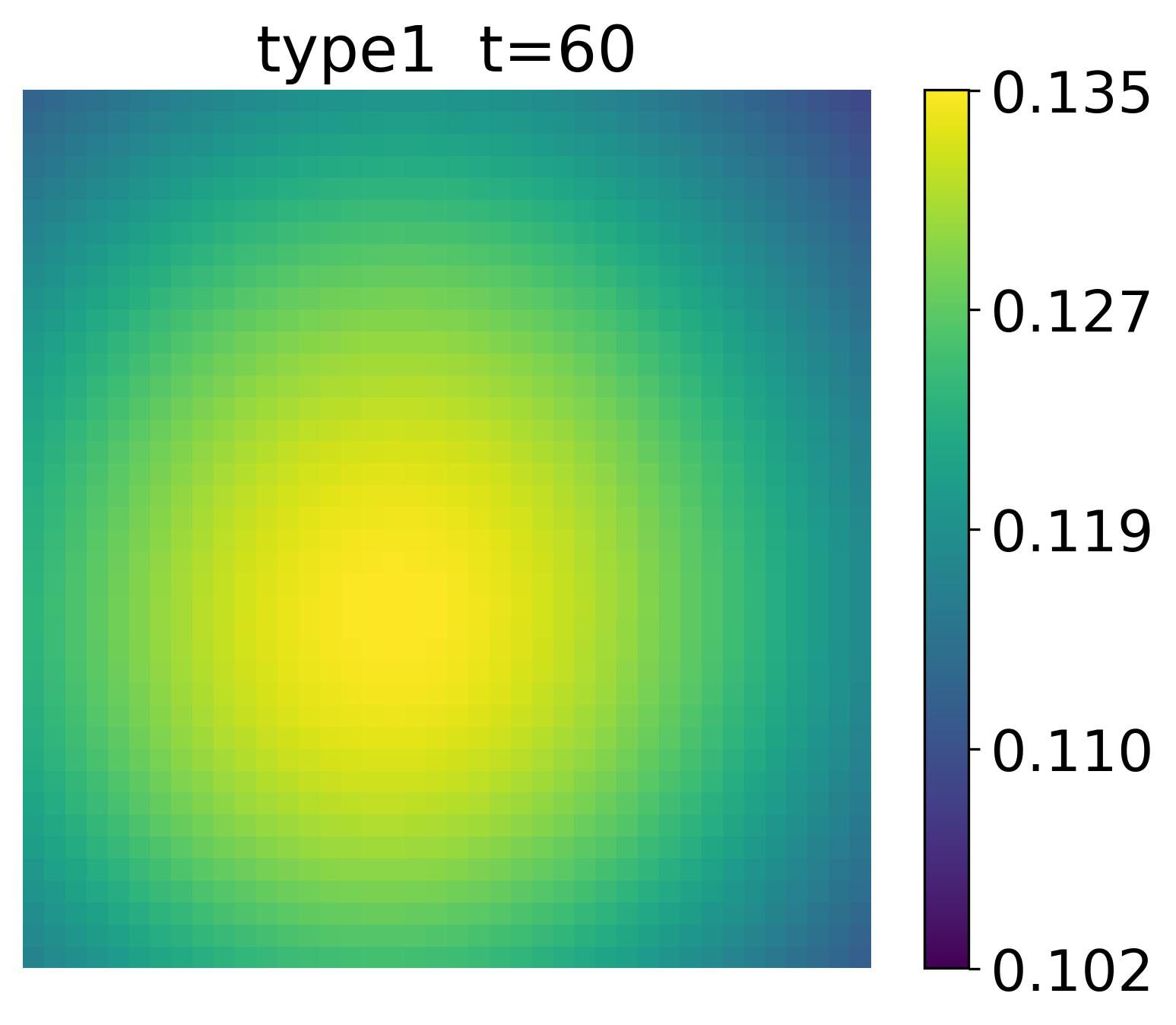} &
\includegraphics[width=0.22\columnwidth]{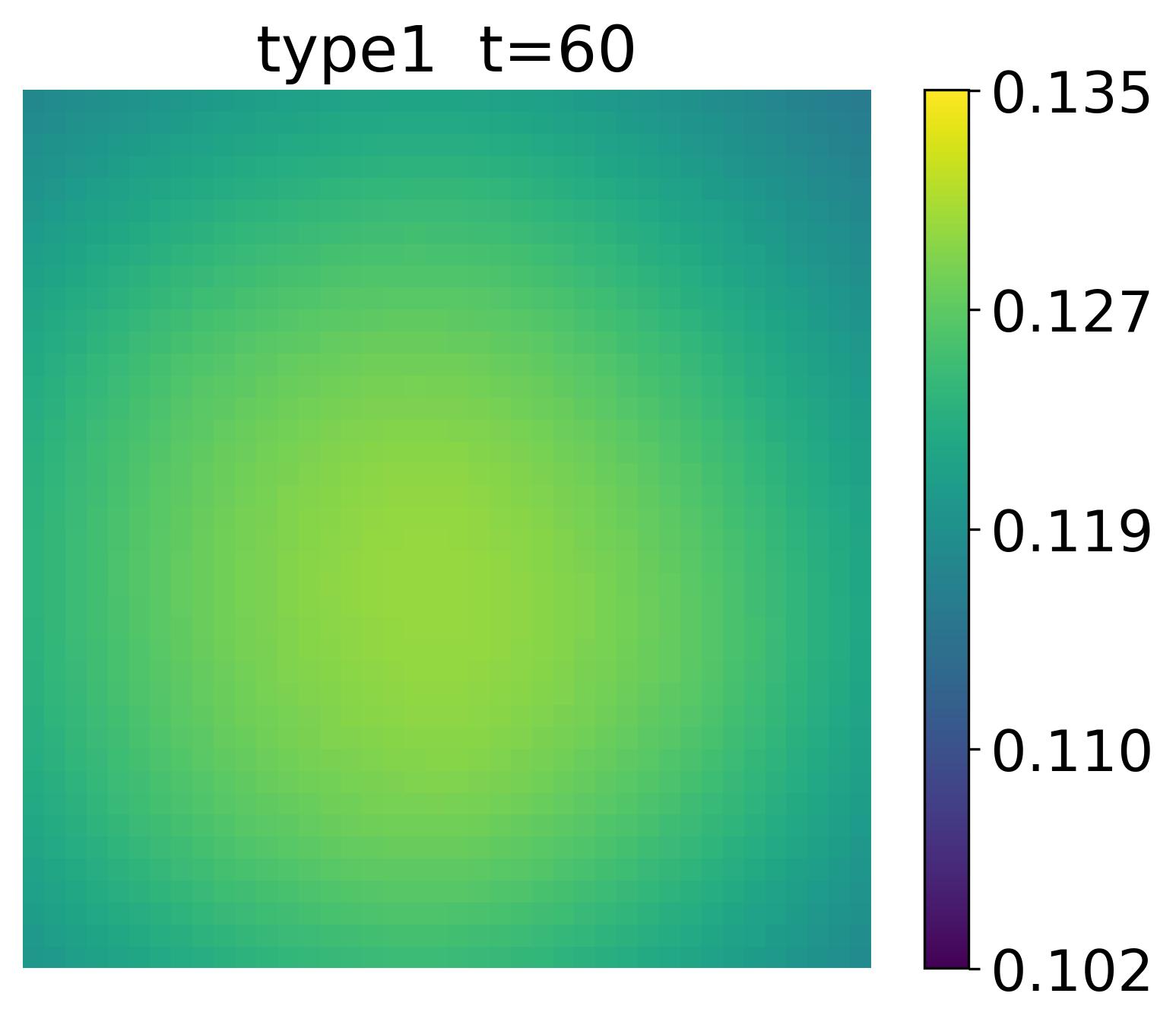} &
\includegraphics[width=0.22\columnwidth]{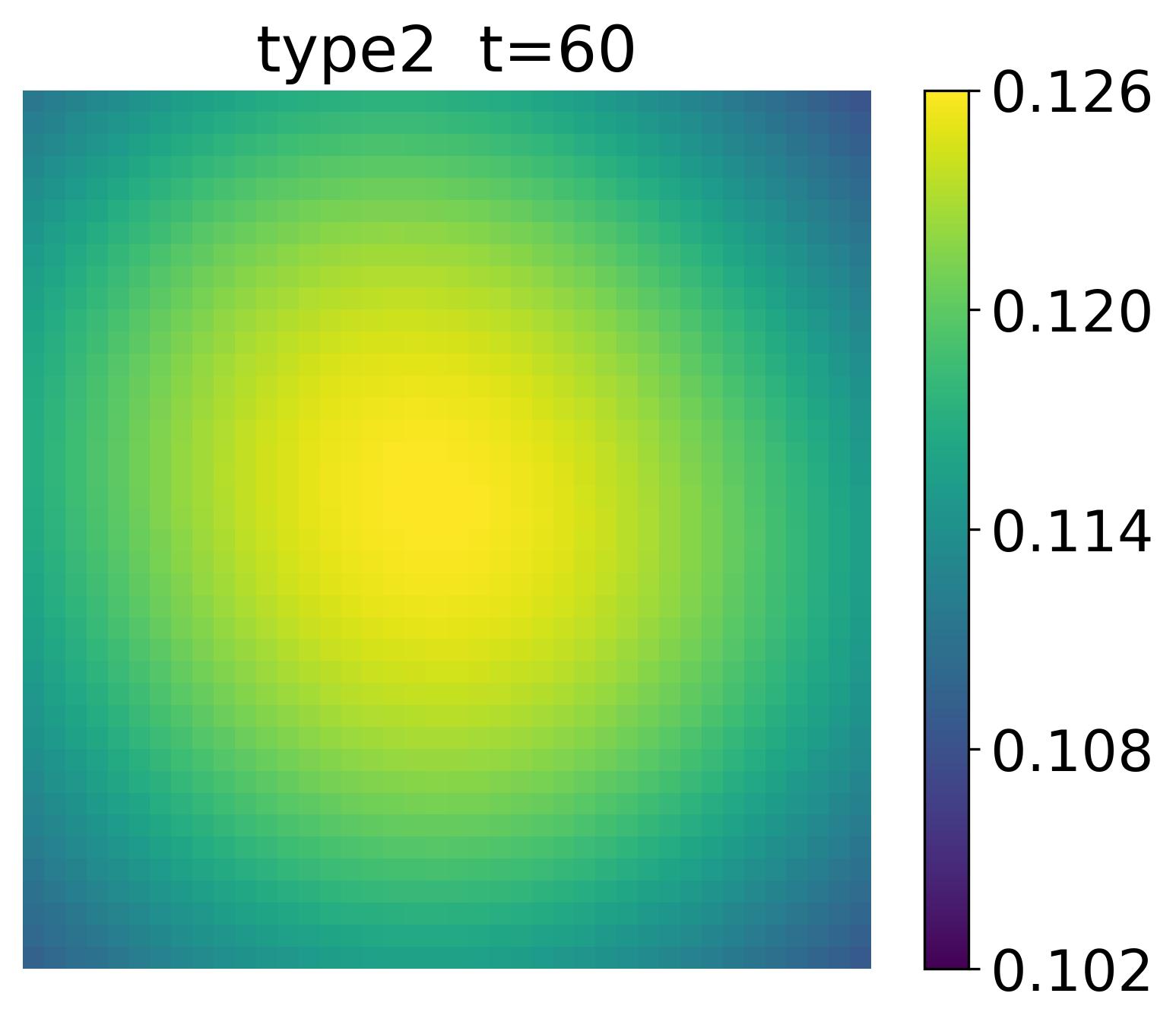} &
\includegraphics[width=0.22\columnwidth]{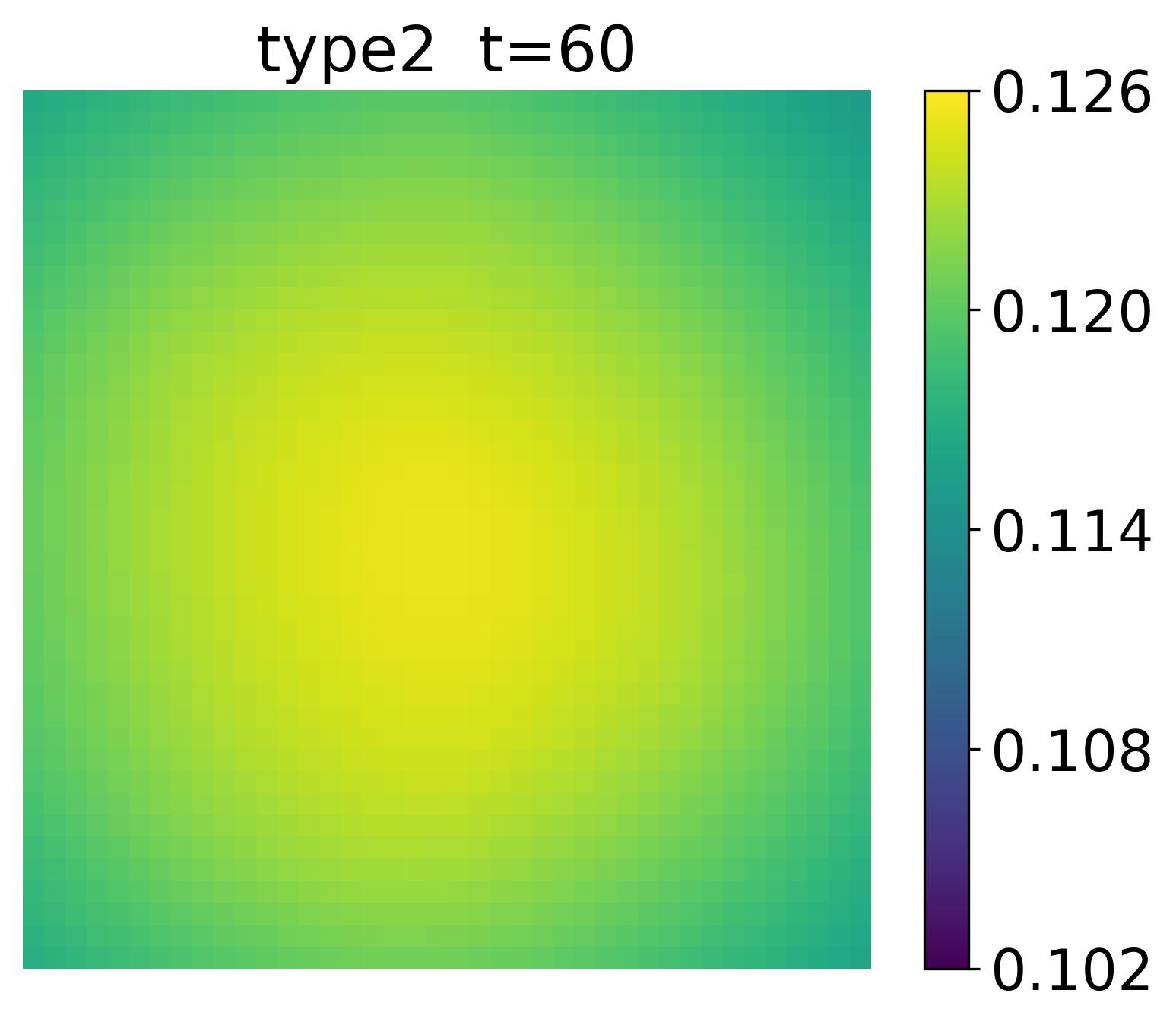} \\

\includegraphics[width=0.22\columnwidth]{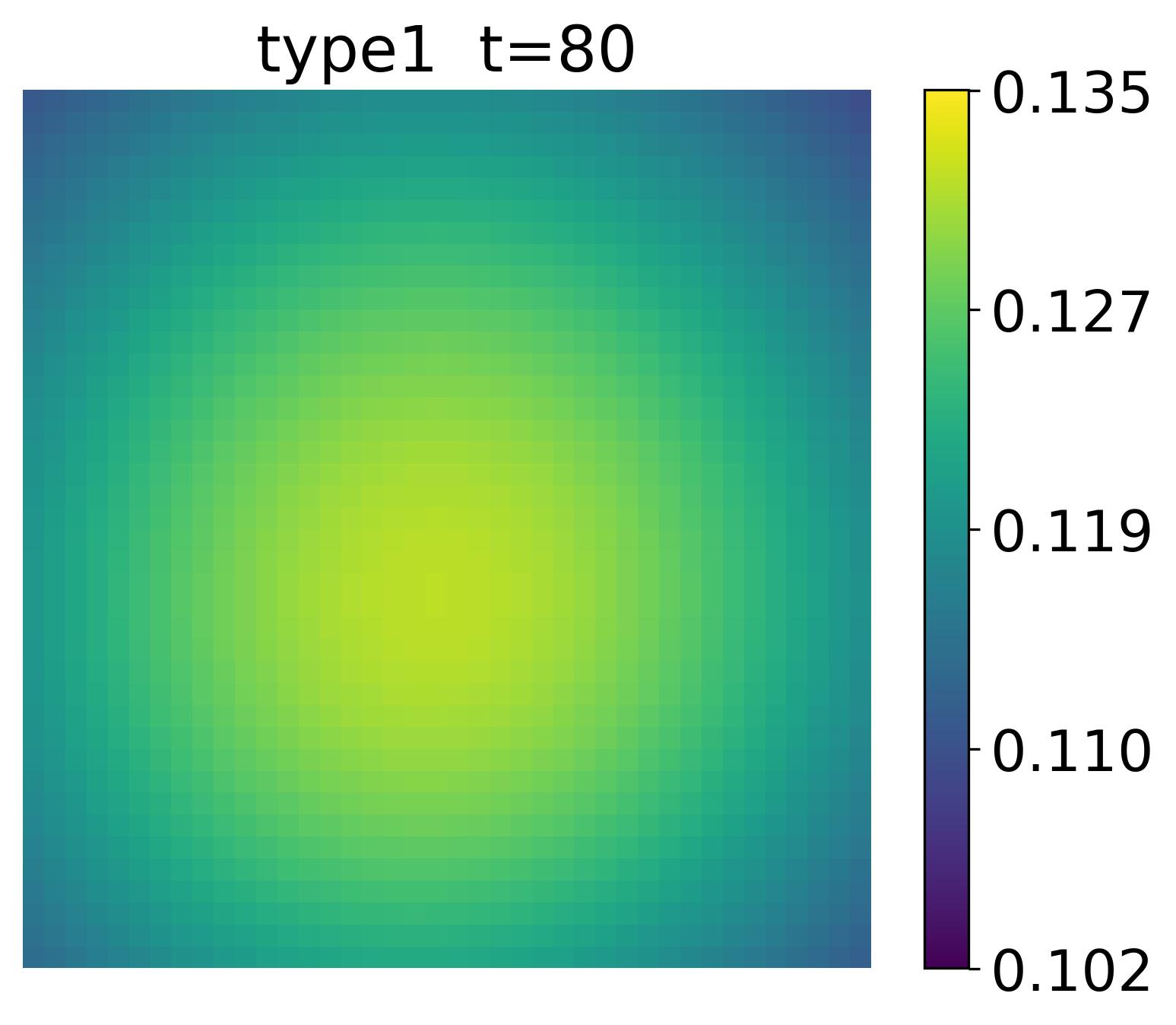} &
\includegraphics[width=0.22\columnwidth]{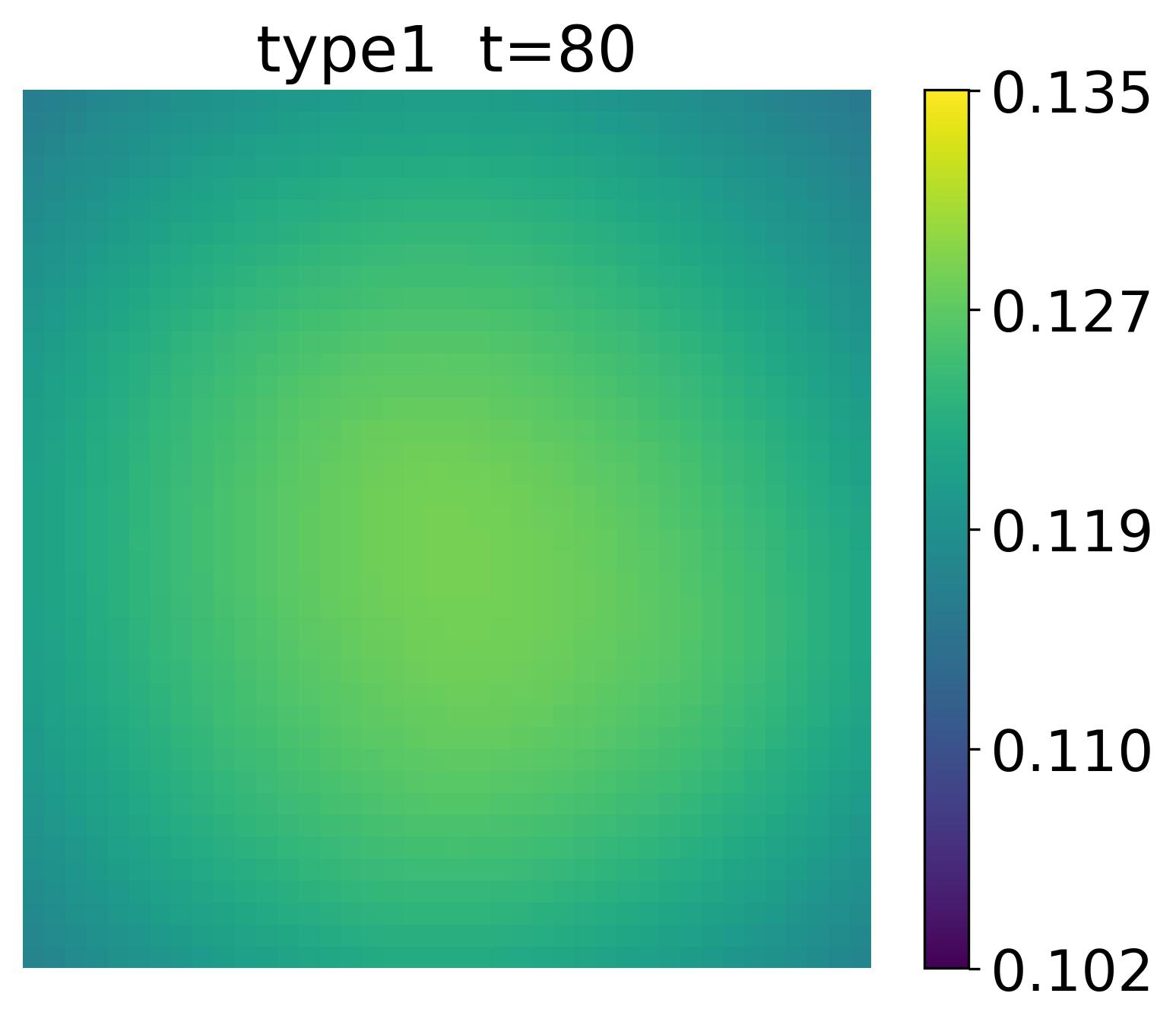} &
\includegraphics[width=0.22\columnwidth]{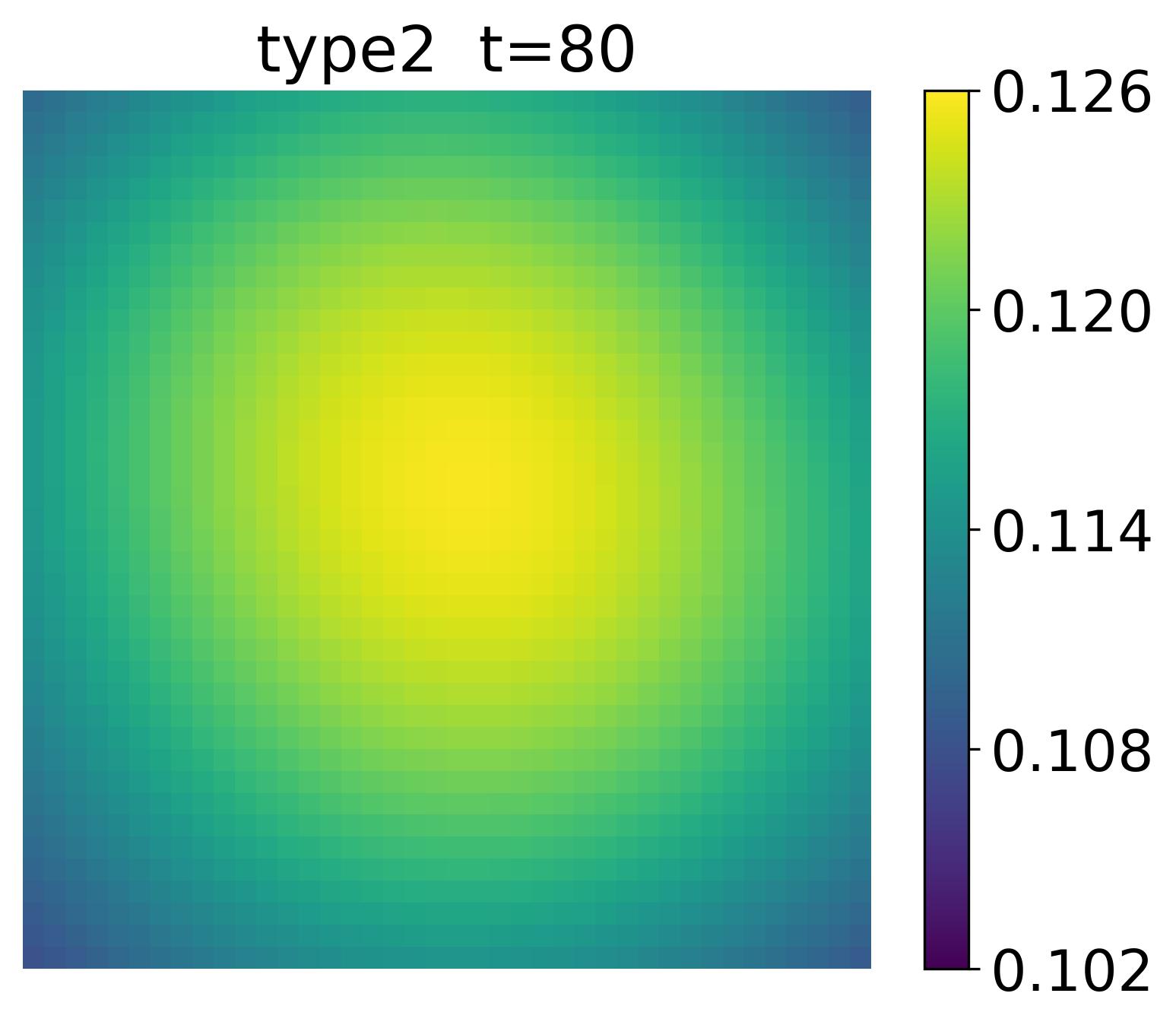} &
\includegraphics[width=0.22\columnwidth]{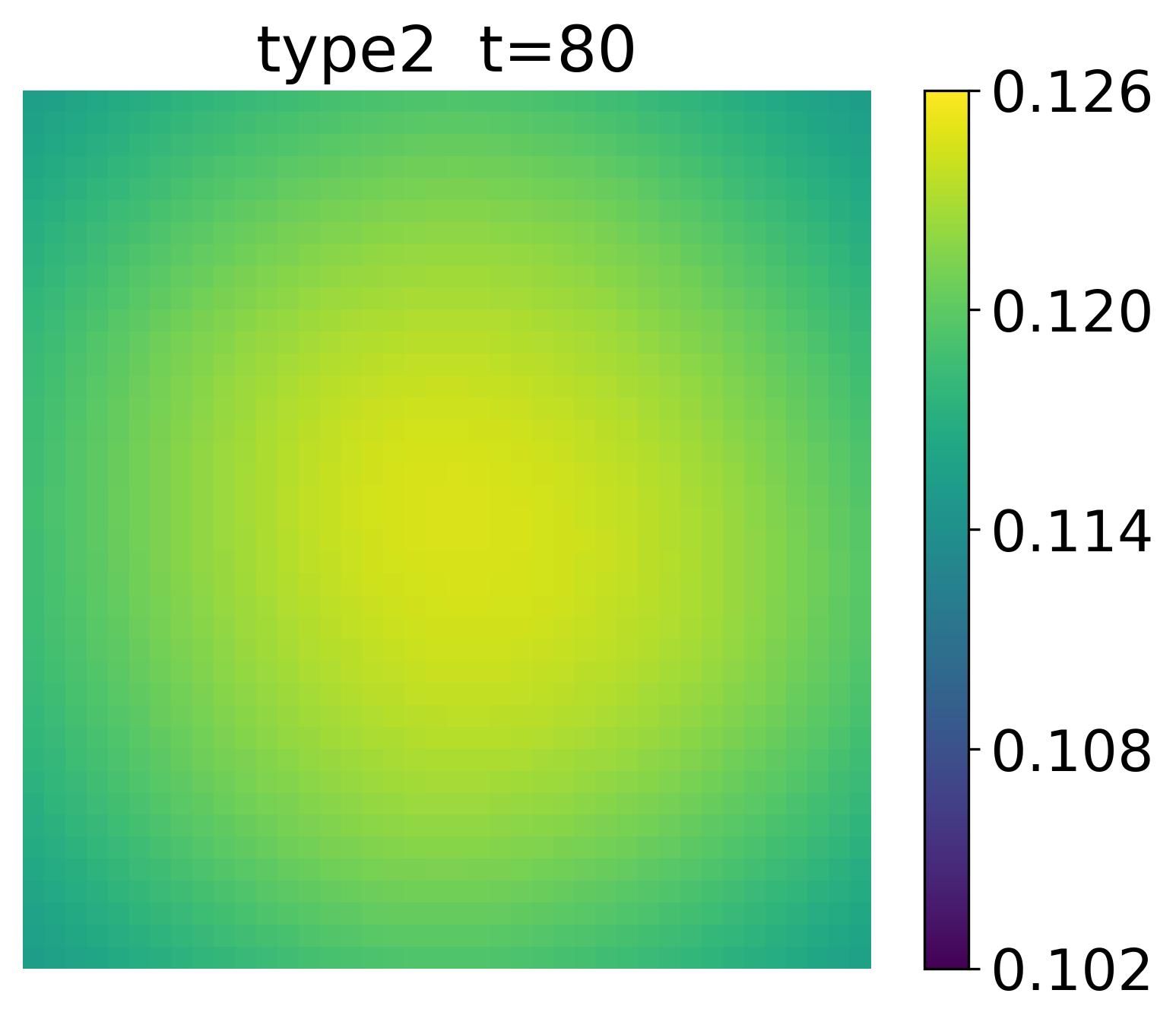} \\

\includegraphics[width=0.22\columnwidth]{figures/cum_avg/biv_1/true/type1/09.jpg} &
\includegraphics[width=0.22\columnwidth]{figures/cum_avg/biv_1/fitted/type1/09.jpg} &
\includegraphics[width=0.22\columnwidth]{figures/cum_avg/biv_1/true/type2/09.jpg} &
\includegraphics[width=0.22\columnwidth]{figures/cum_avg/biv_1/fitted/type2/09.jpg} \\
\end{tabular}

\caption{Cumulative time-averaged spatial intensity maps for Biv~1. Columns 1 and 3 give true spatial intensities, columns 2 and 4 corresponding MSTNHP-fitted intensities for each event type. Each panel displays the average spatial intensity accumulated from $t=0$ up to the indicated time horizon (i.e., a cumulative moving average over $[0,t]$).}

\label{fig:biv1_maps}
\end{figure}

\begin{figure}[htbp!]
\centering
\setlength{\tabcolsep}{1pt} 
\setlength{\extrarowheight}{3pt} 

\begin{tabular}{cccc}

{True} & {Fitted} & {True} & {Fitted} \\[2pt]

\includegraphics[width=0.22\columnwidth]{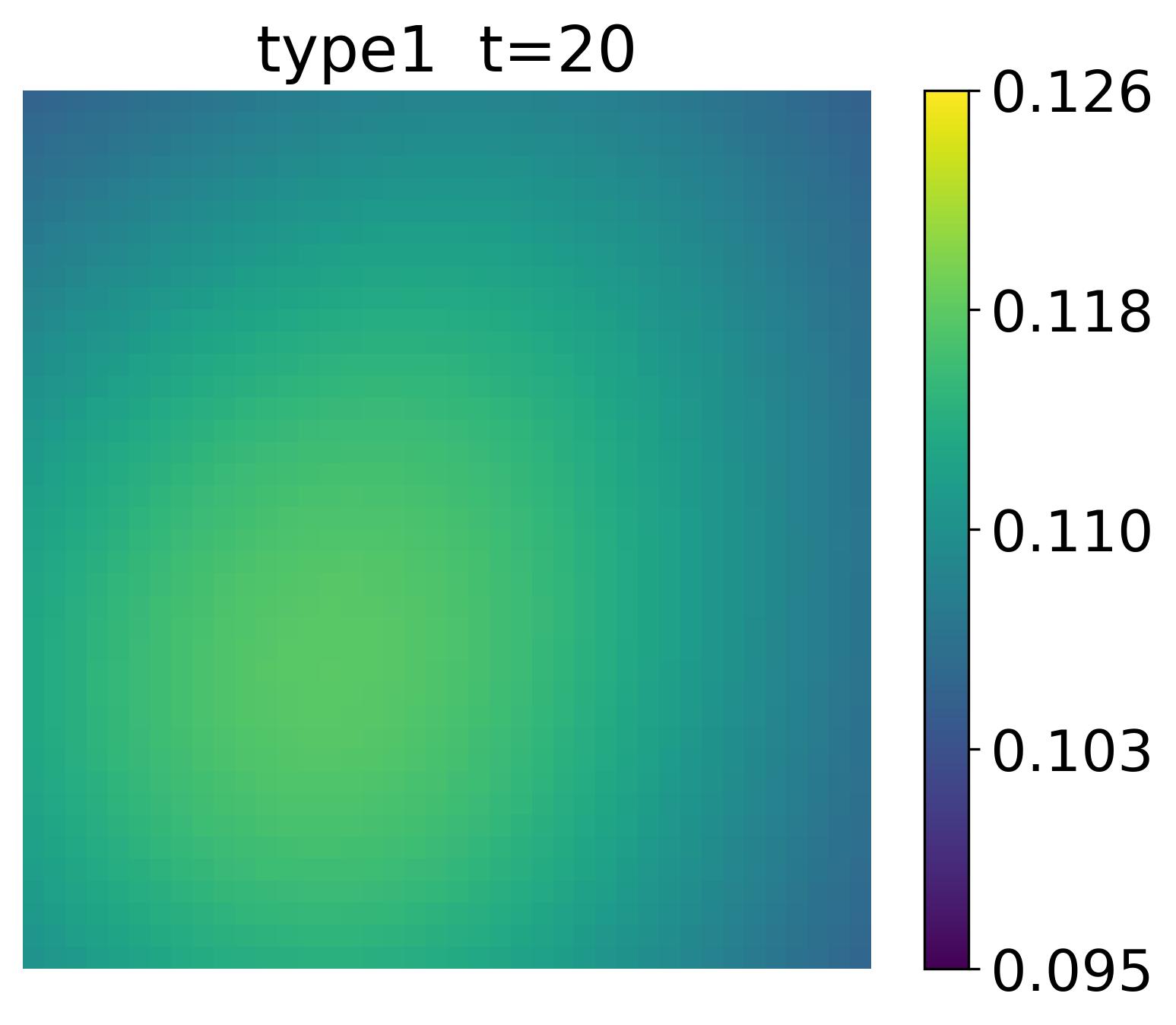} &
\includegraphics[width=0.22\columnwidth]{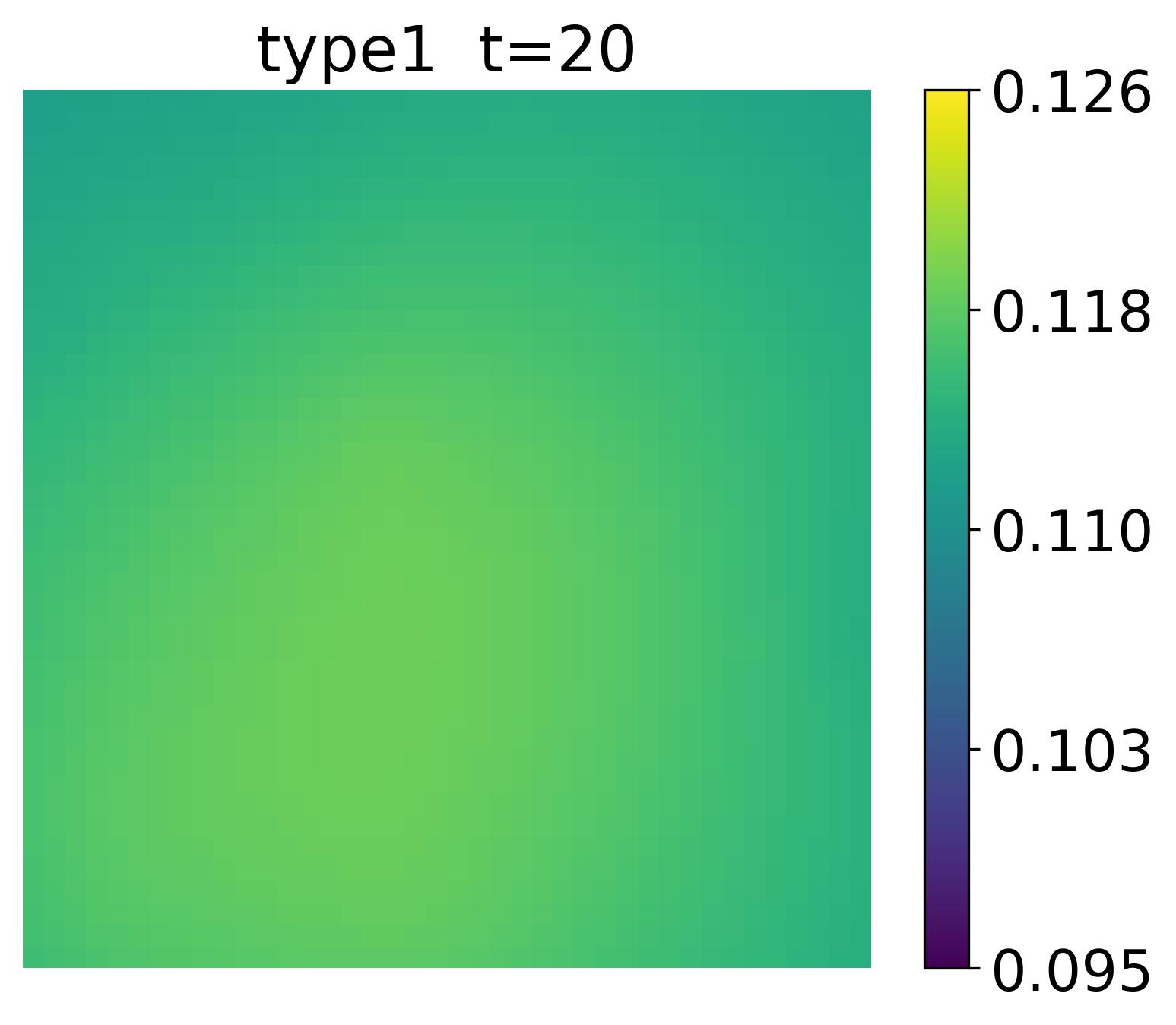} &
\includegraphics[width=0.22\columnwidth]{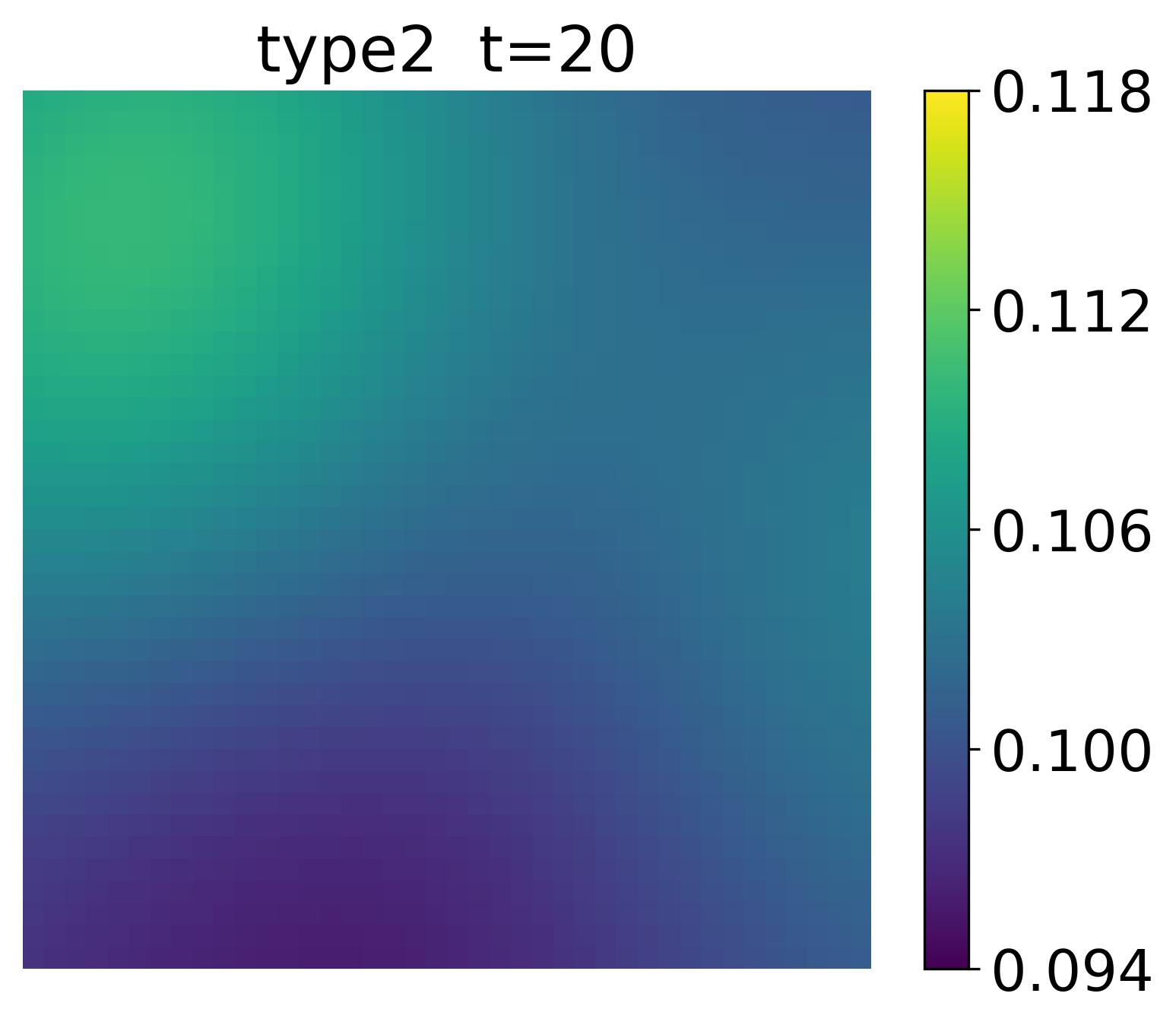} &
\includegraphics[width=0.22\columnwidth]{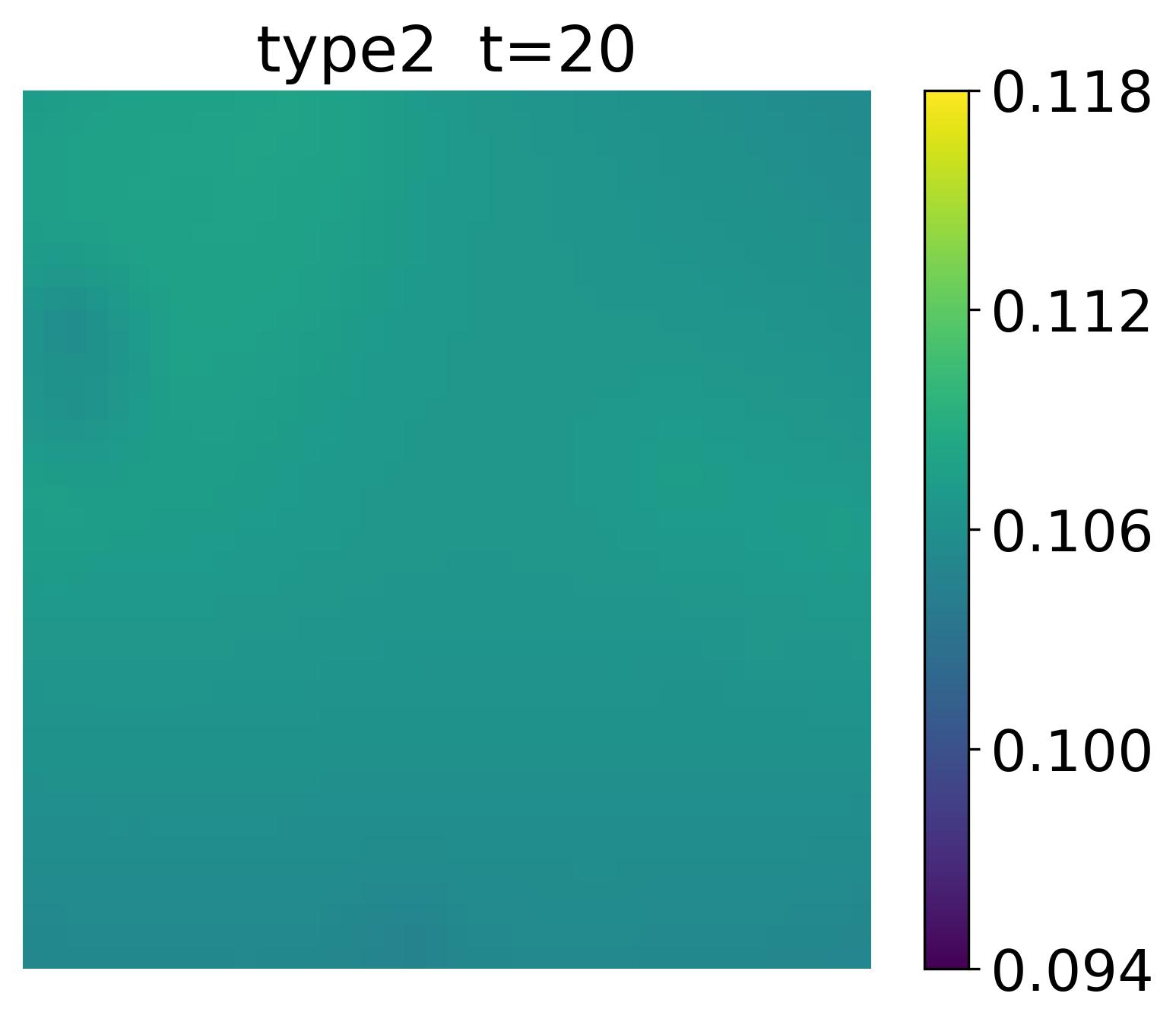} \\

\includegraphics[width=0.22\columnwidth]{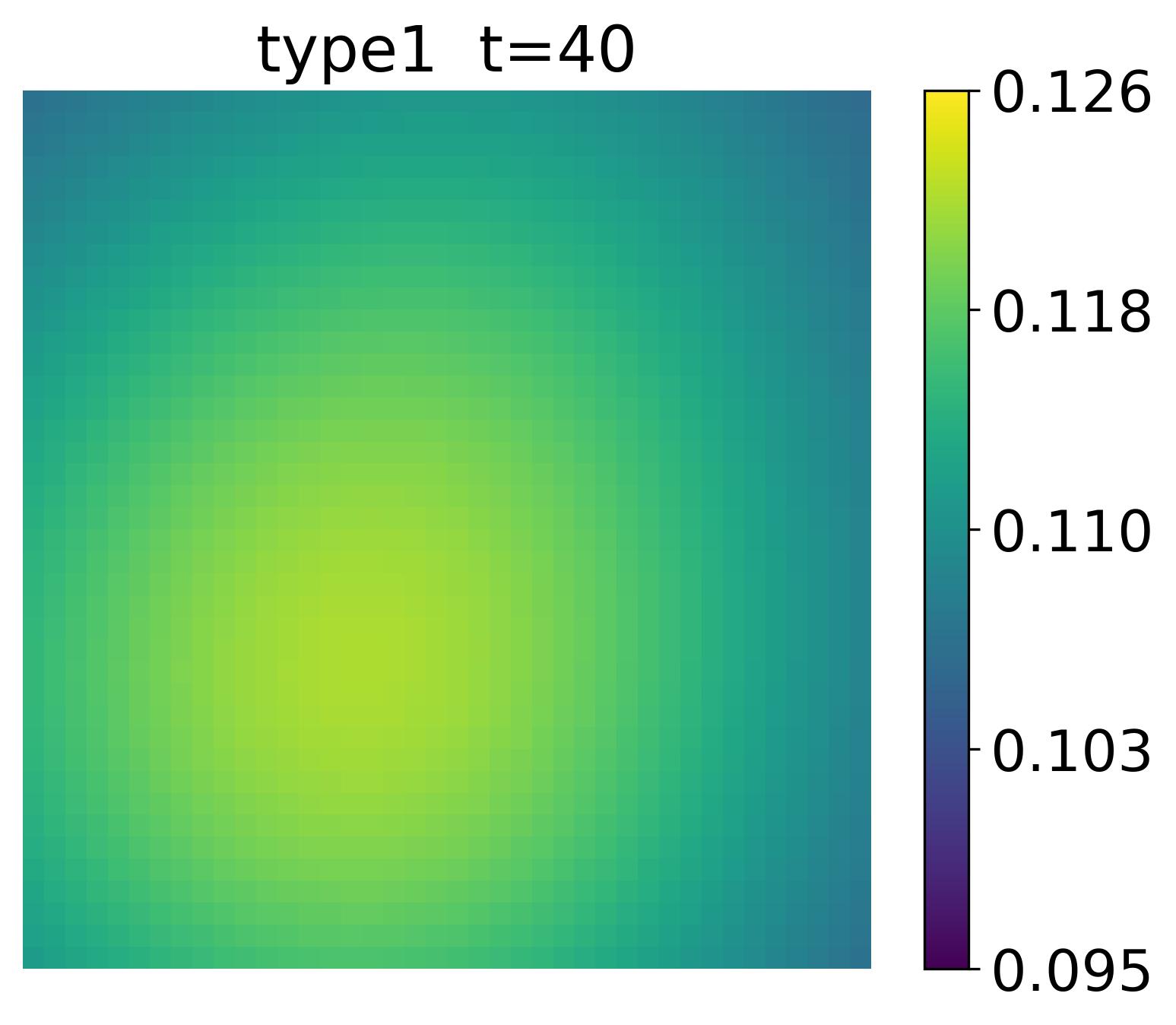} &
\includegraphics[width=0.22\columnwidth]{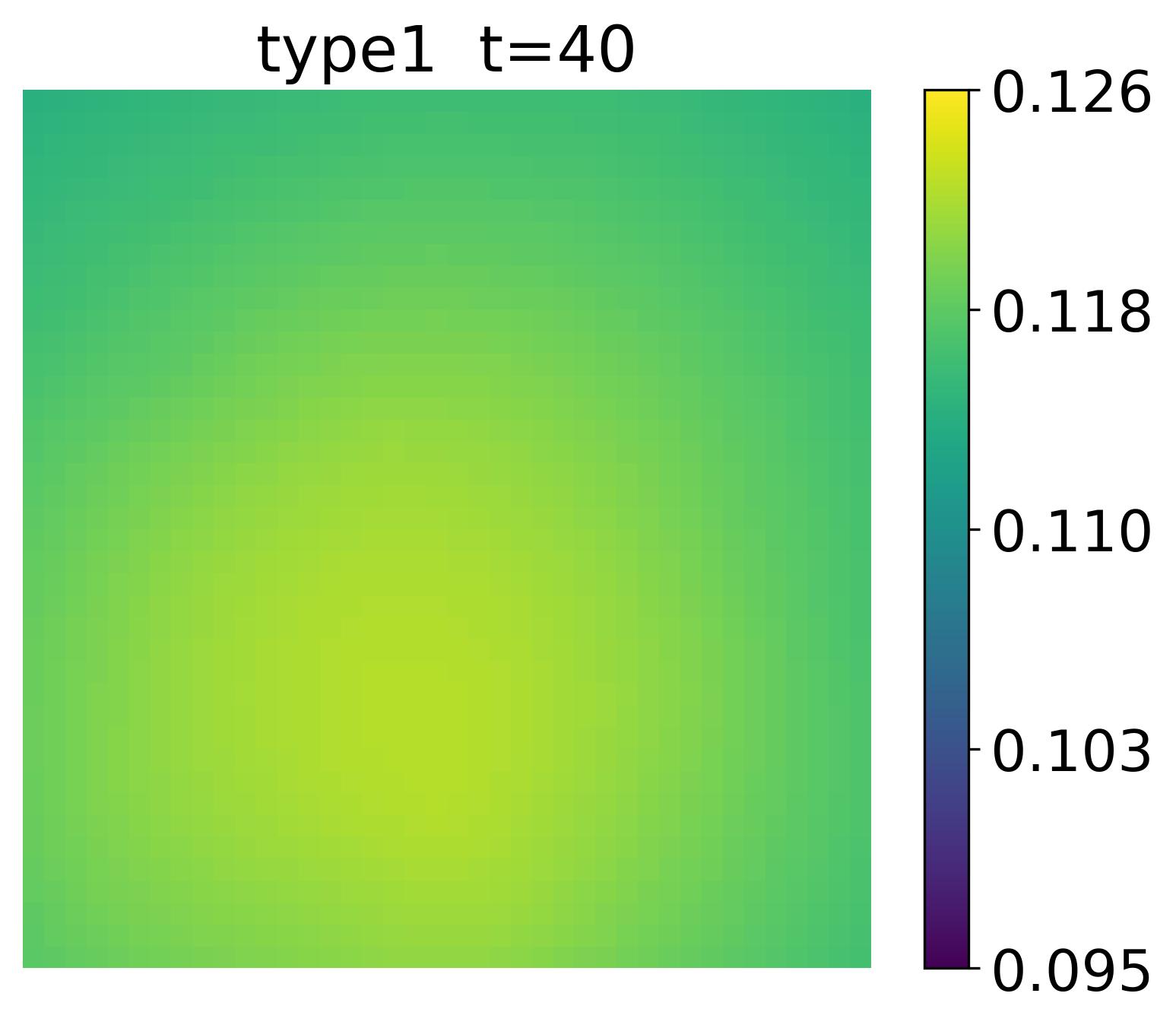} &
\includegraphics[width=0.22\columnwidth]{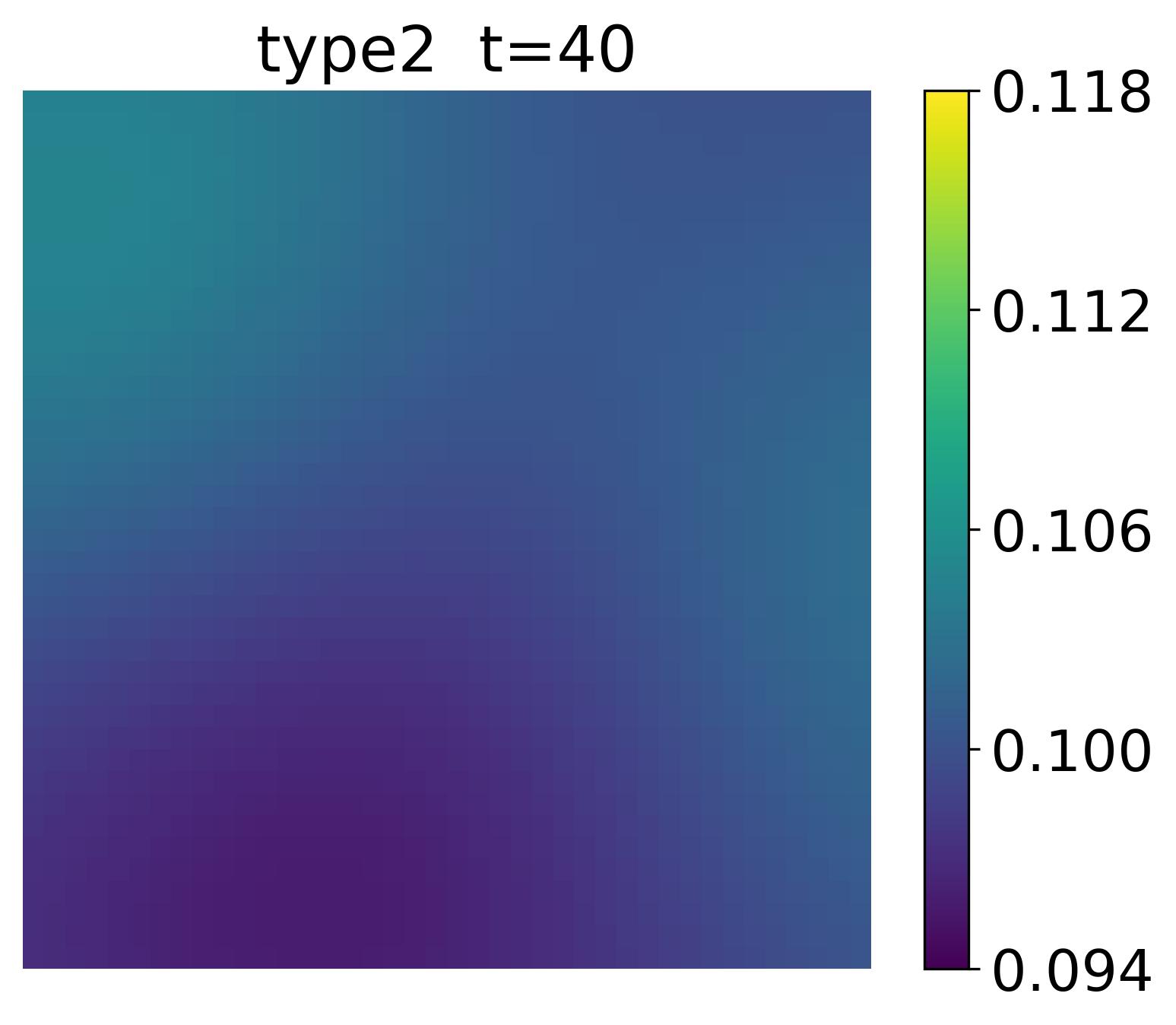} &
\includegraphics[width=0.22\columnwidth]{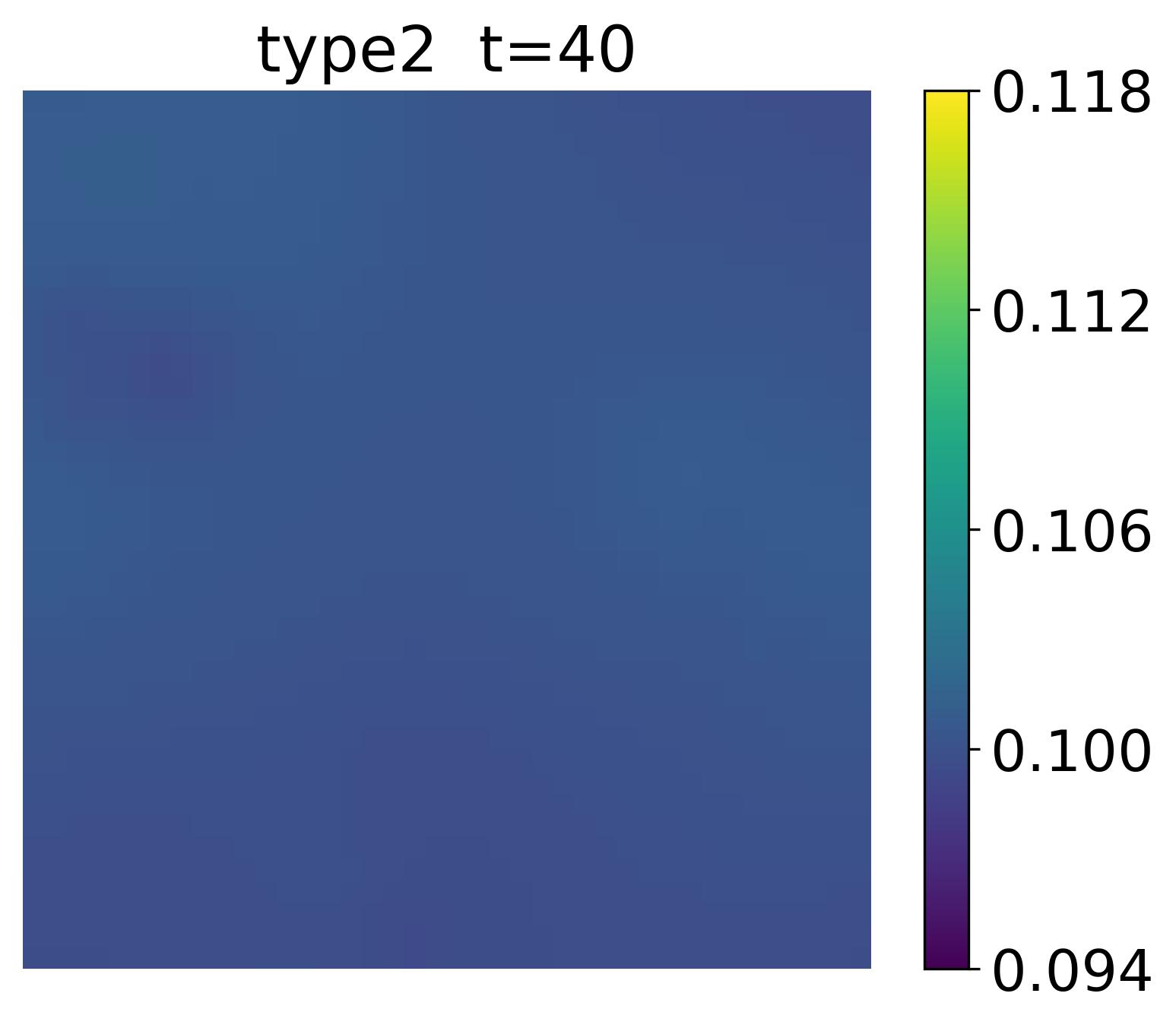} \\

\includegraphics[width=0.22\columnwidth]{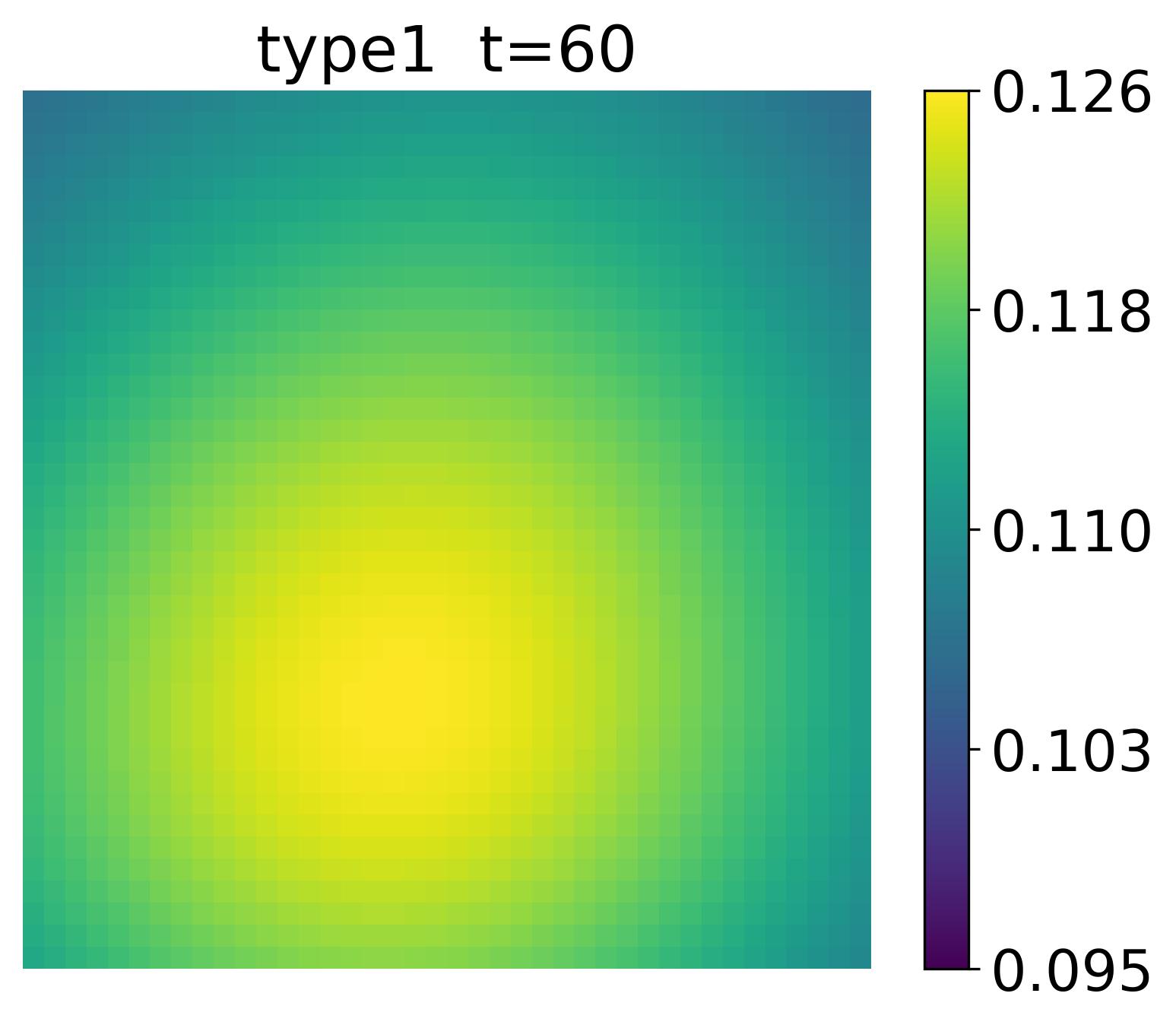} &
\includegraphics[width=0.22\columnwidth]{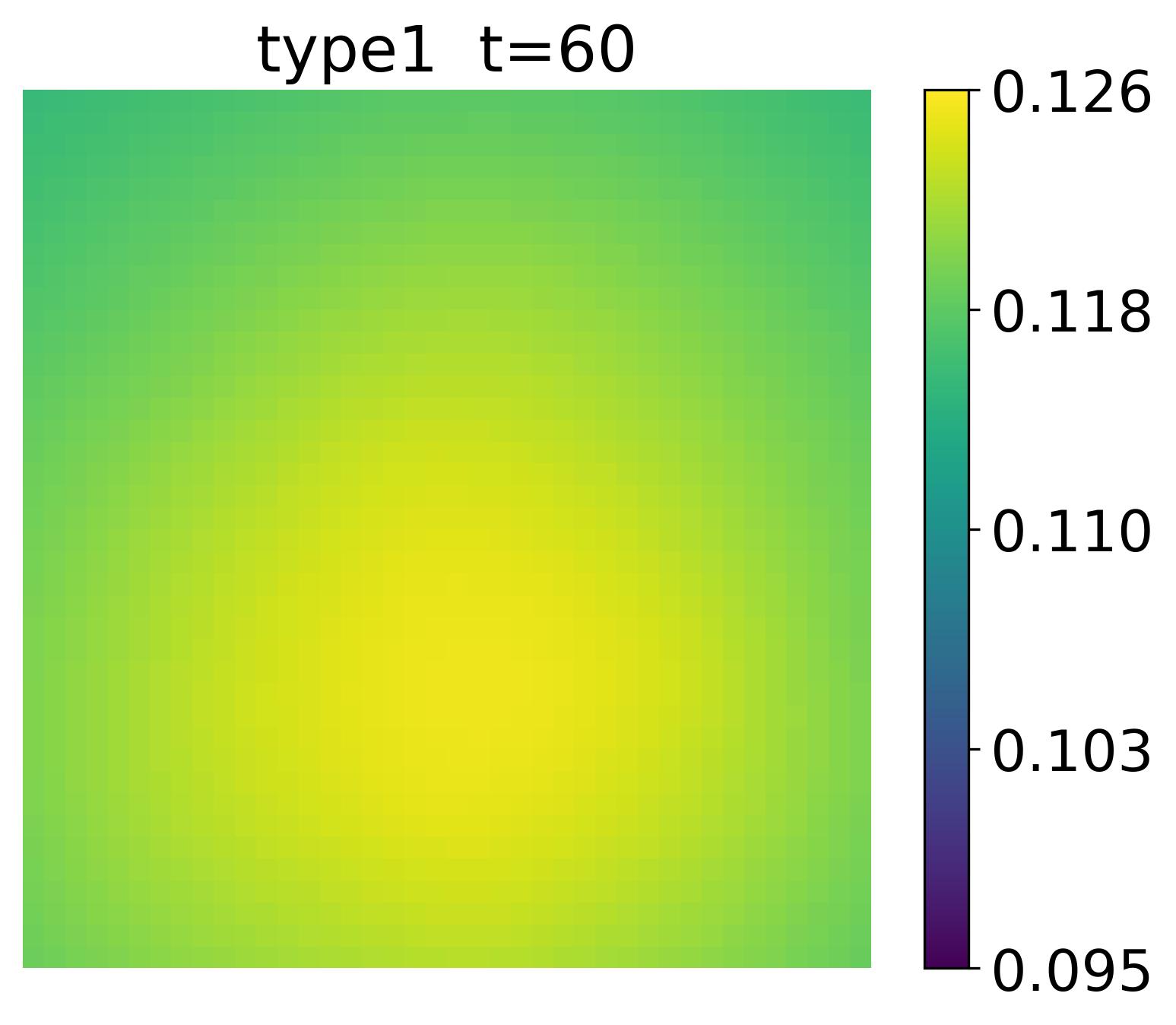} &
\includegraphics[width=0.22\columnwidth]{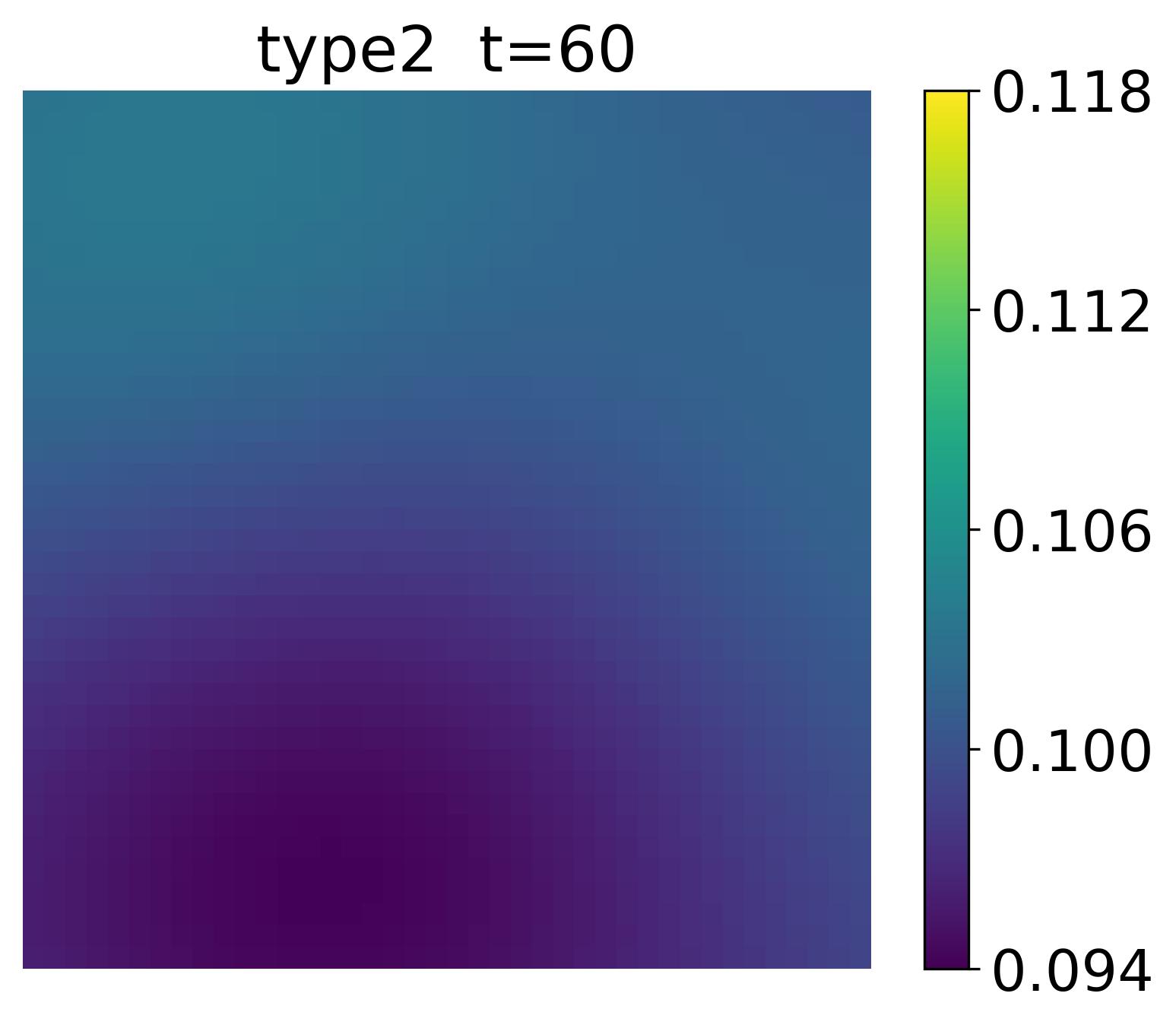} &
\includegraphics[width=0.22\columnwidth]{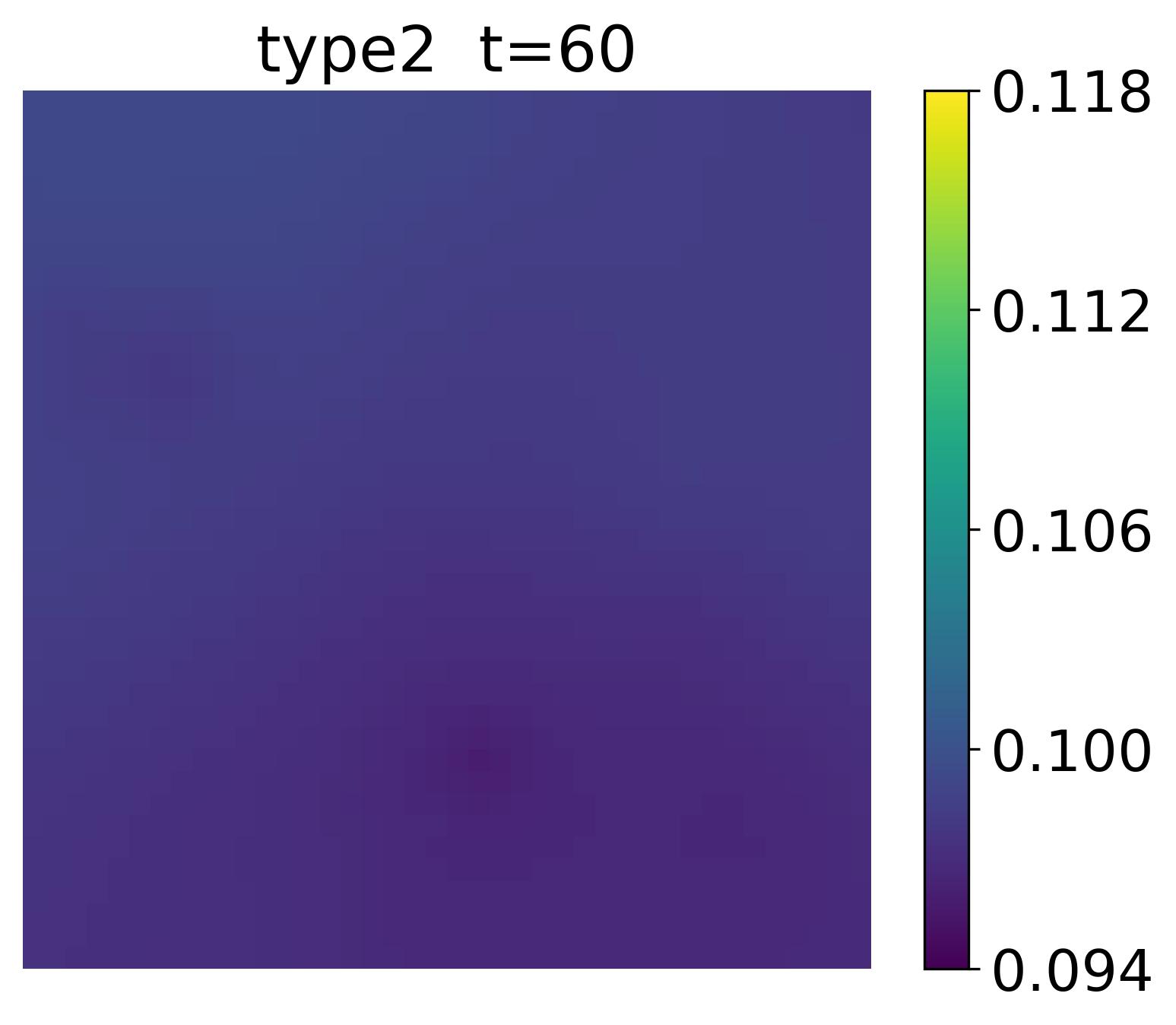} \\

\includegraphics[width=0.22\columnwidth]{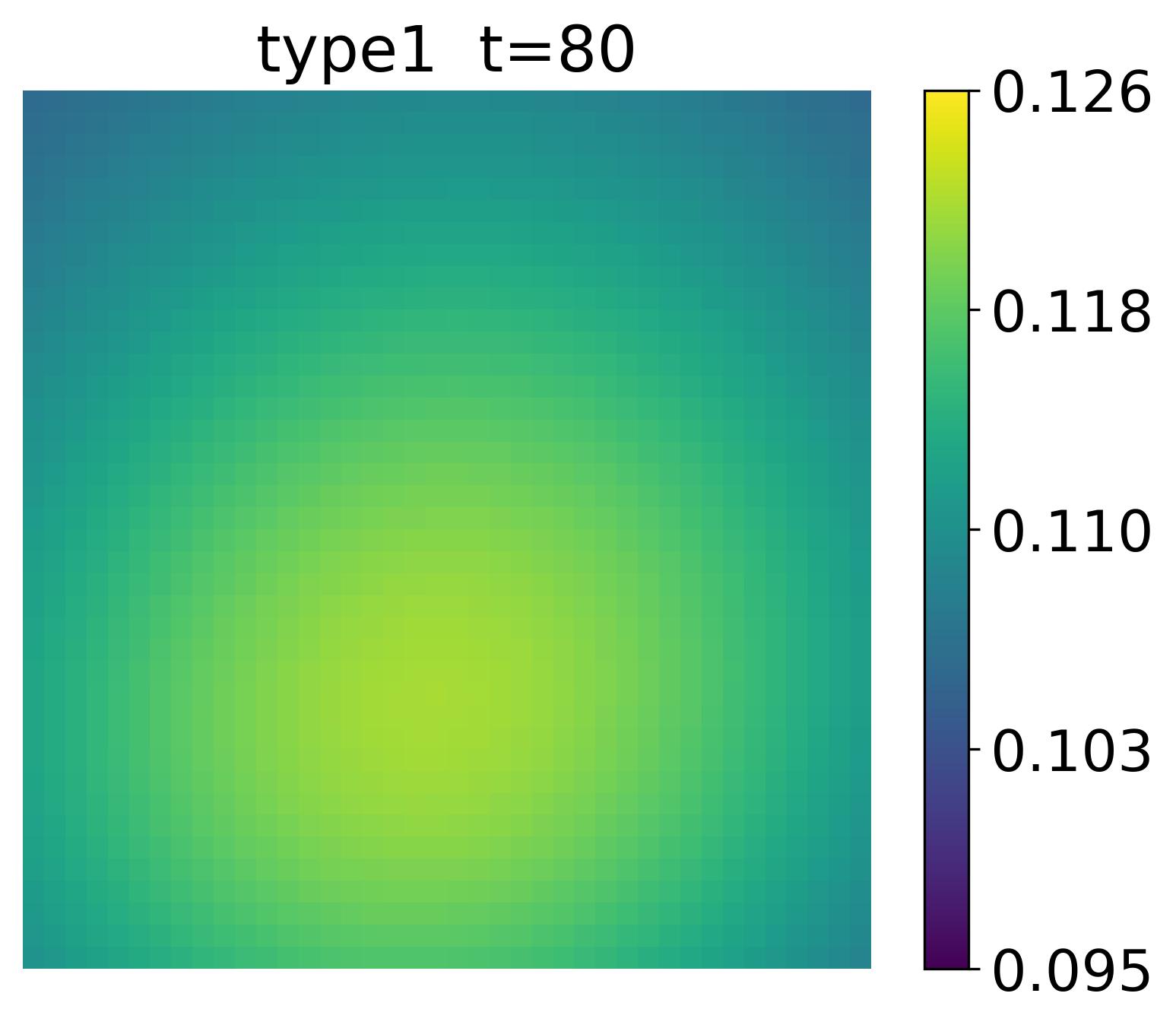} &
\includegraphics[width=0.22\columnwidth]{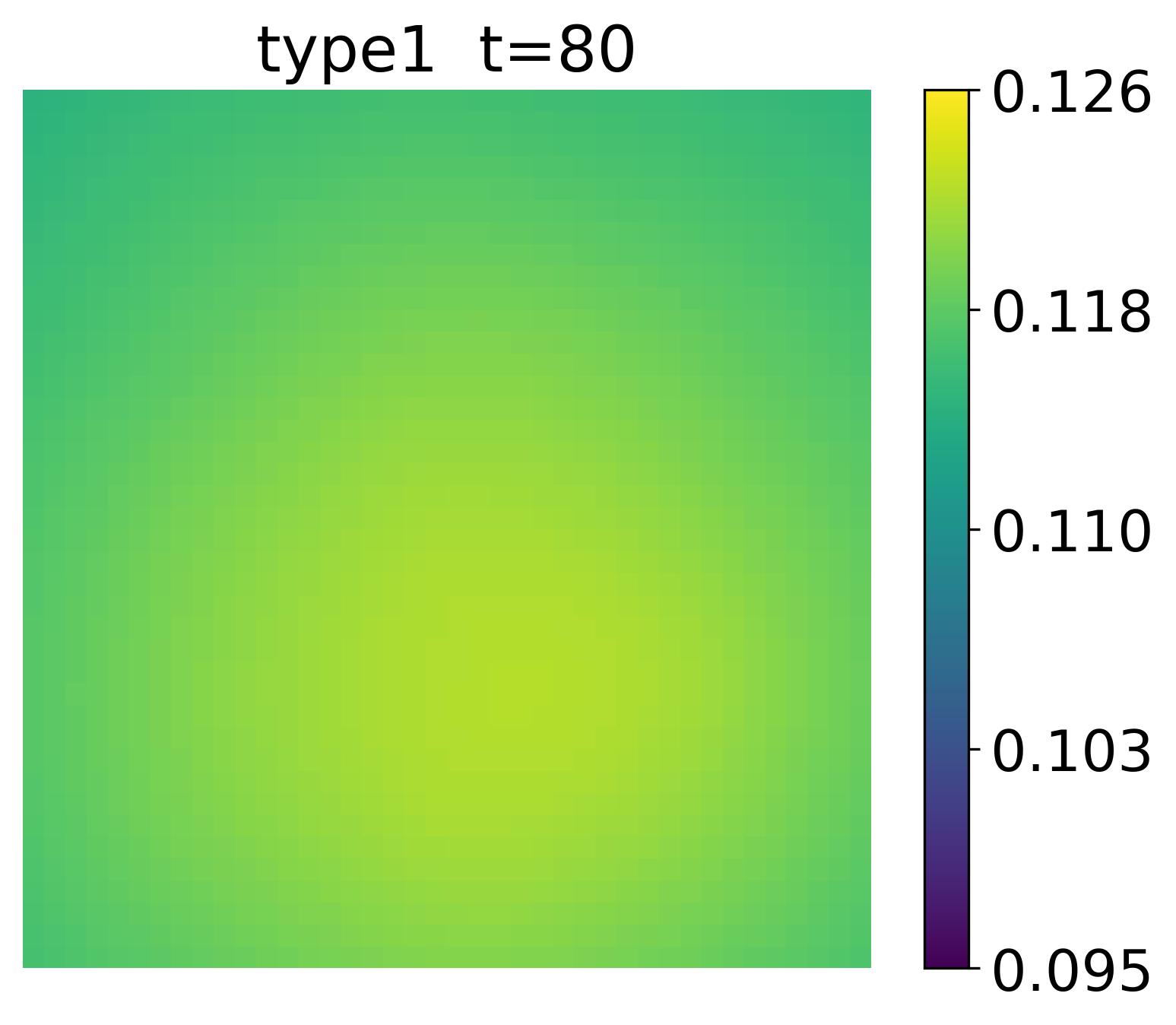} &
\includegraphics[width=0.22\columnwidth]{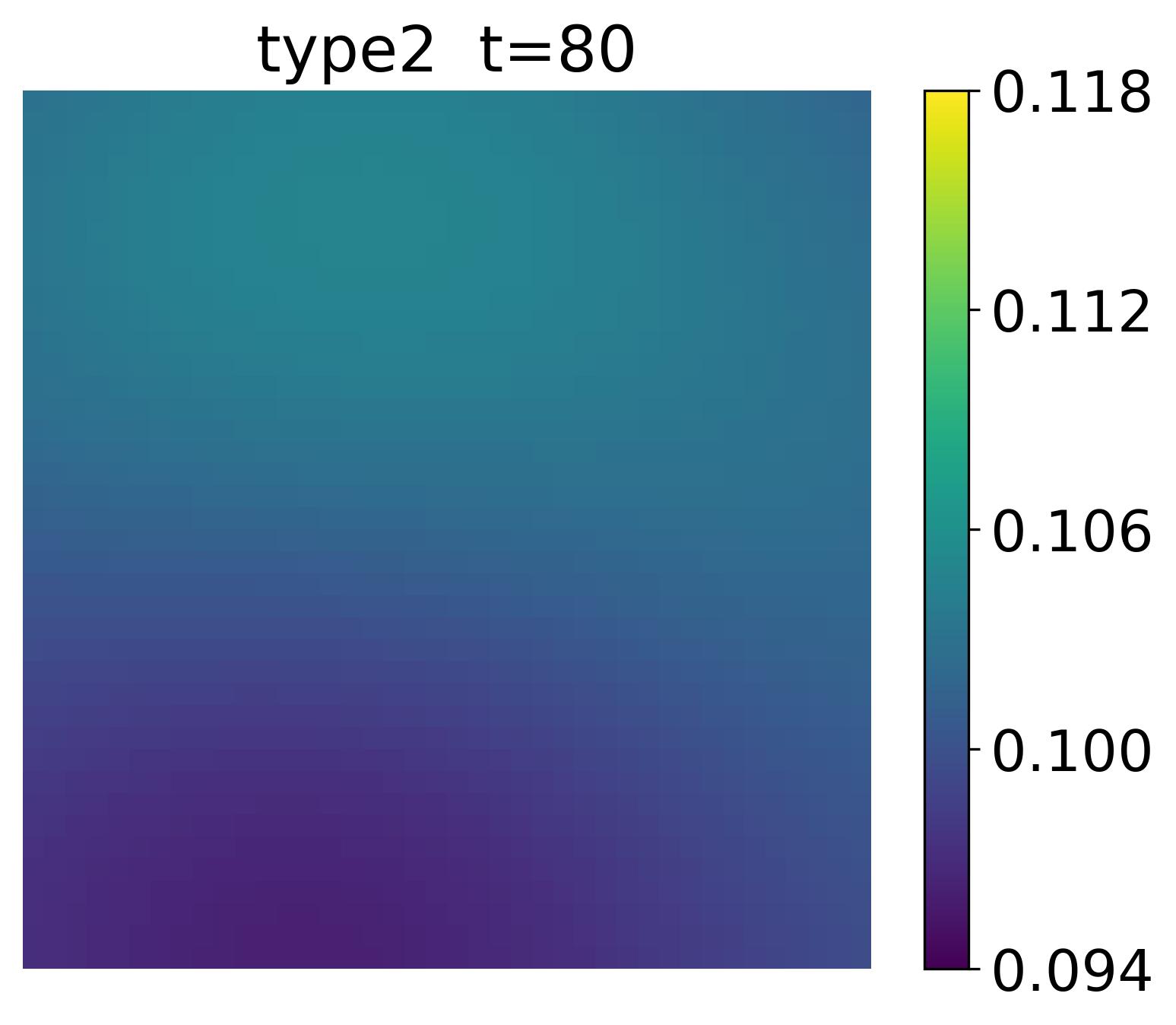} &
\includegraphics[width=0.22\columnwidth]{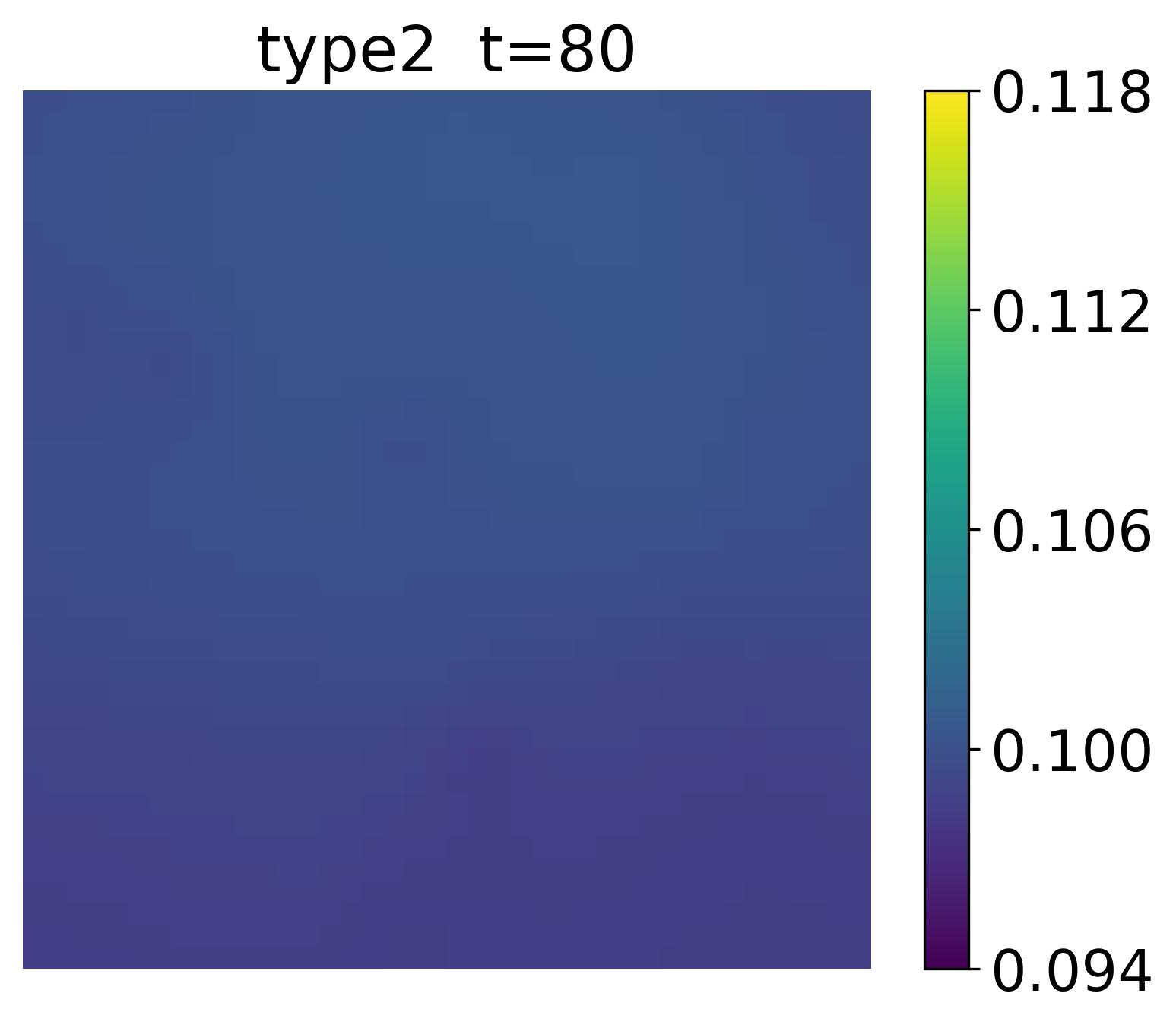} \\

\includegraphics[width=0.22\columnwidth]{figures/cum_avg/biv_2/true/type1/09.jpg} &
\includegraphics[width=0.22\columnwidth]{figures/cum_avg/biv_2/fitted/type1/09.jpg} &
\includegraphics[width=0.22\columnwidth]{figures/cum_avg/biv_2/true/type2/09.jpg} &
\includegraphics[width=0.22\columnwidth]{figures/cum_avg/biv_2/fitted/type2/09.jpg} \\
\end{tabular}

\caption{Same as Figure~\ref{fig:biv1_maps} except for Biv~2. }
\label{fig:biv2_maps}
\end{figure}

\begin{figure}[htbp!]
\centering
\setlength{\tabcolsep}{1pt}
\setlength{\extrarowheight}{3pt}

\begin{tabular}{cccc}

{True} & {Fitted} & {True} & {Fitted} \\[2pt]

\includegraphics[width=0.22\columnwidth]{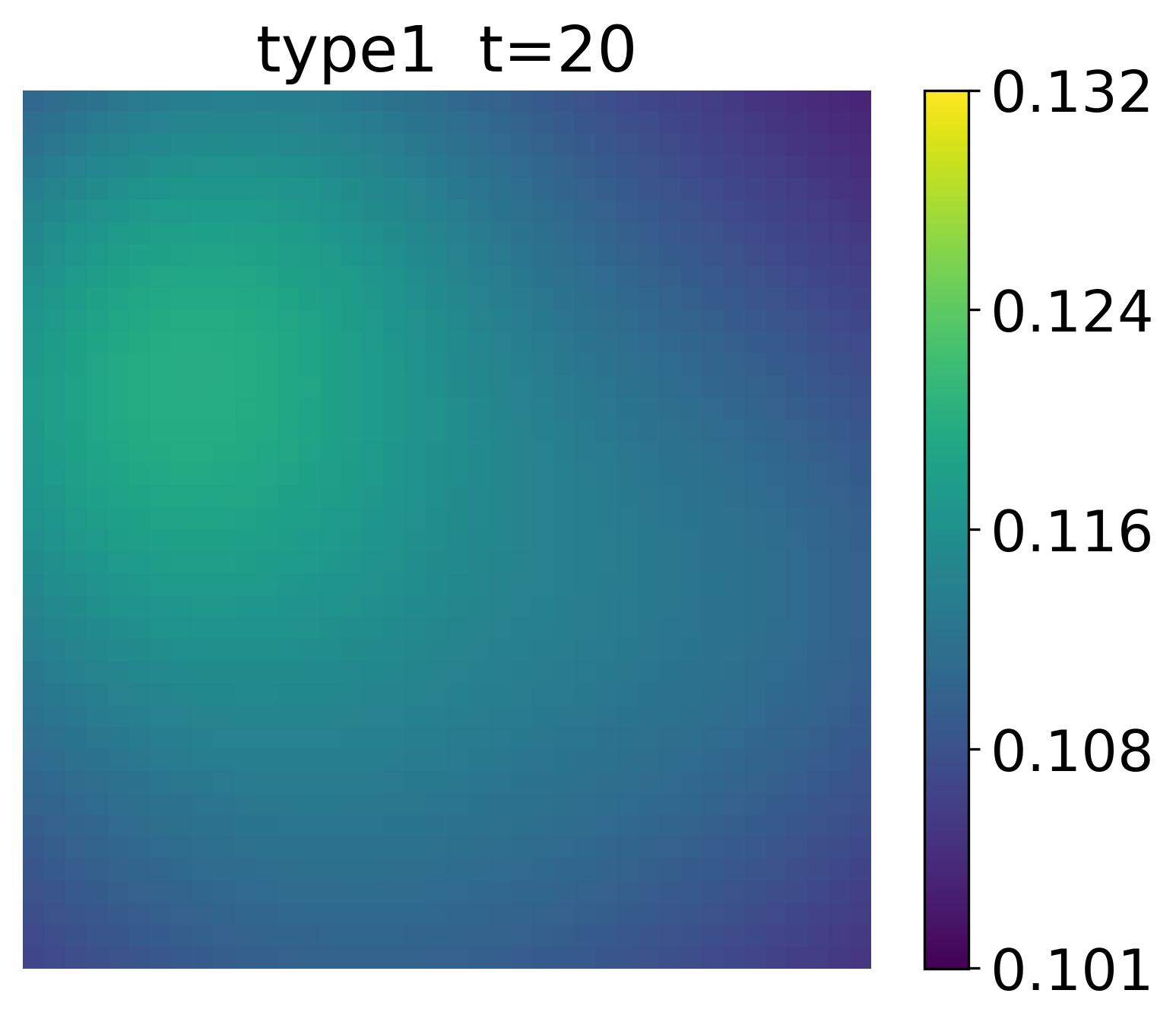} &
\includegraphics[width=0.22\columnwidth]{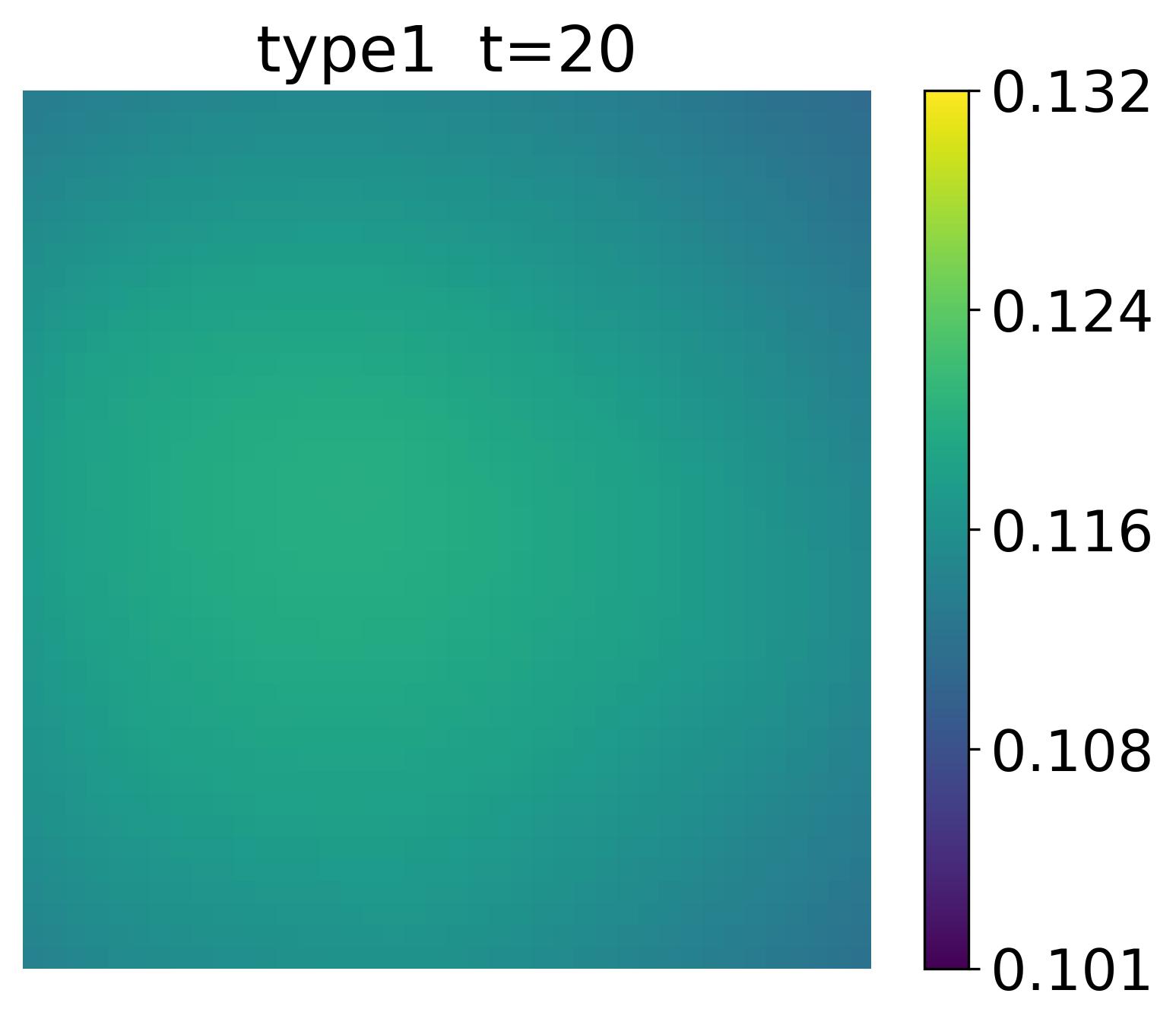} &
\includegraphics[width=0.22\columnwidth]{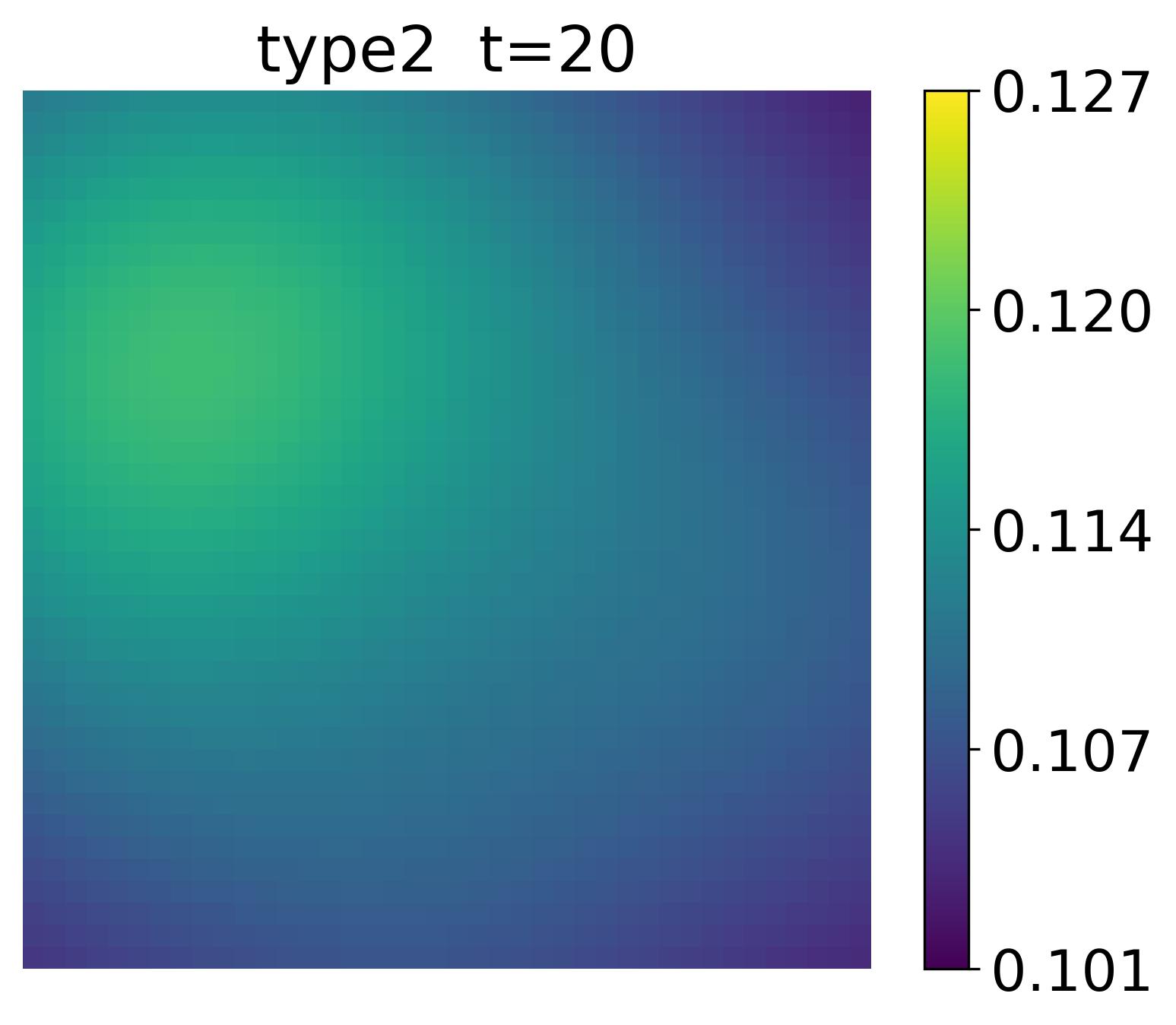} &
\includegraphics[width=0.22\columnwidth]{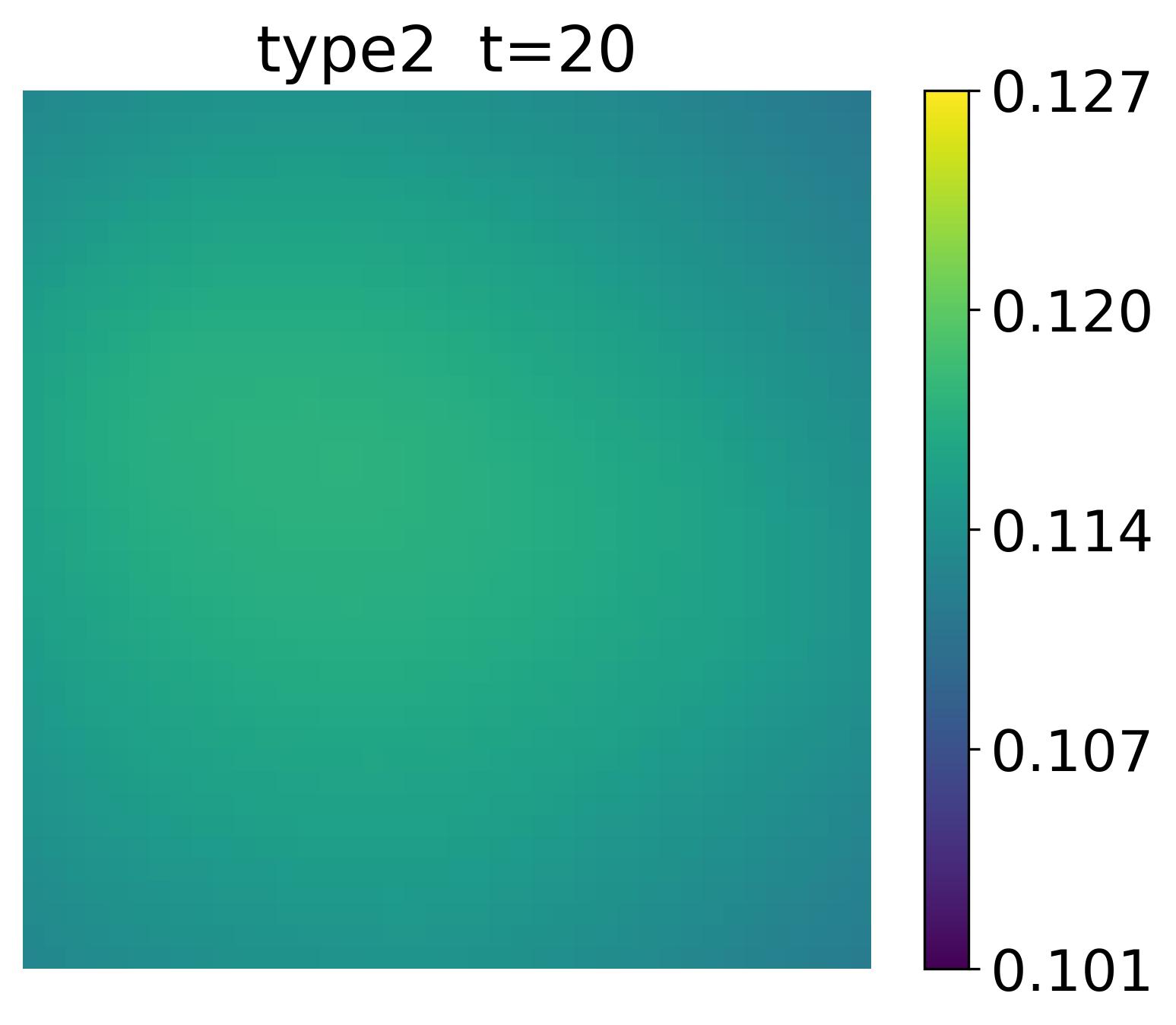} \\

\includegraphics[width=0.22\columnwidth]{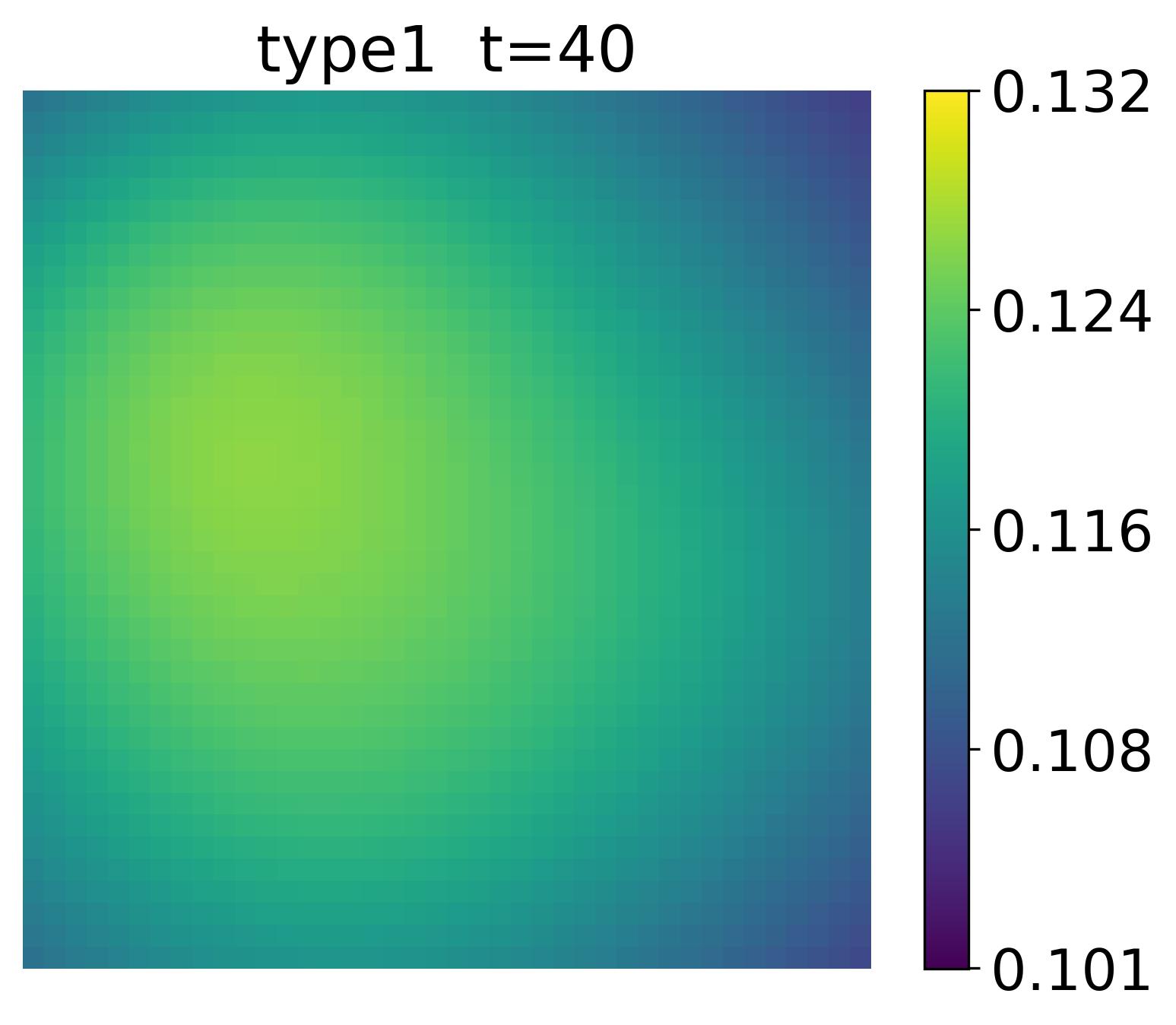} &
\includegraphics[width=0.22\columnwidth]{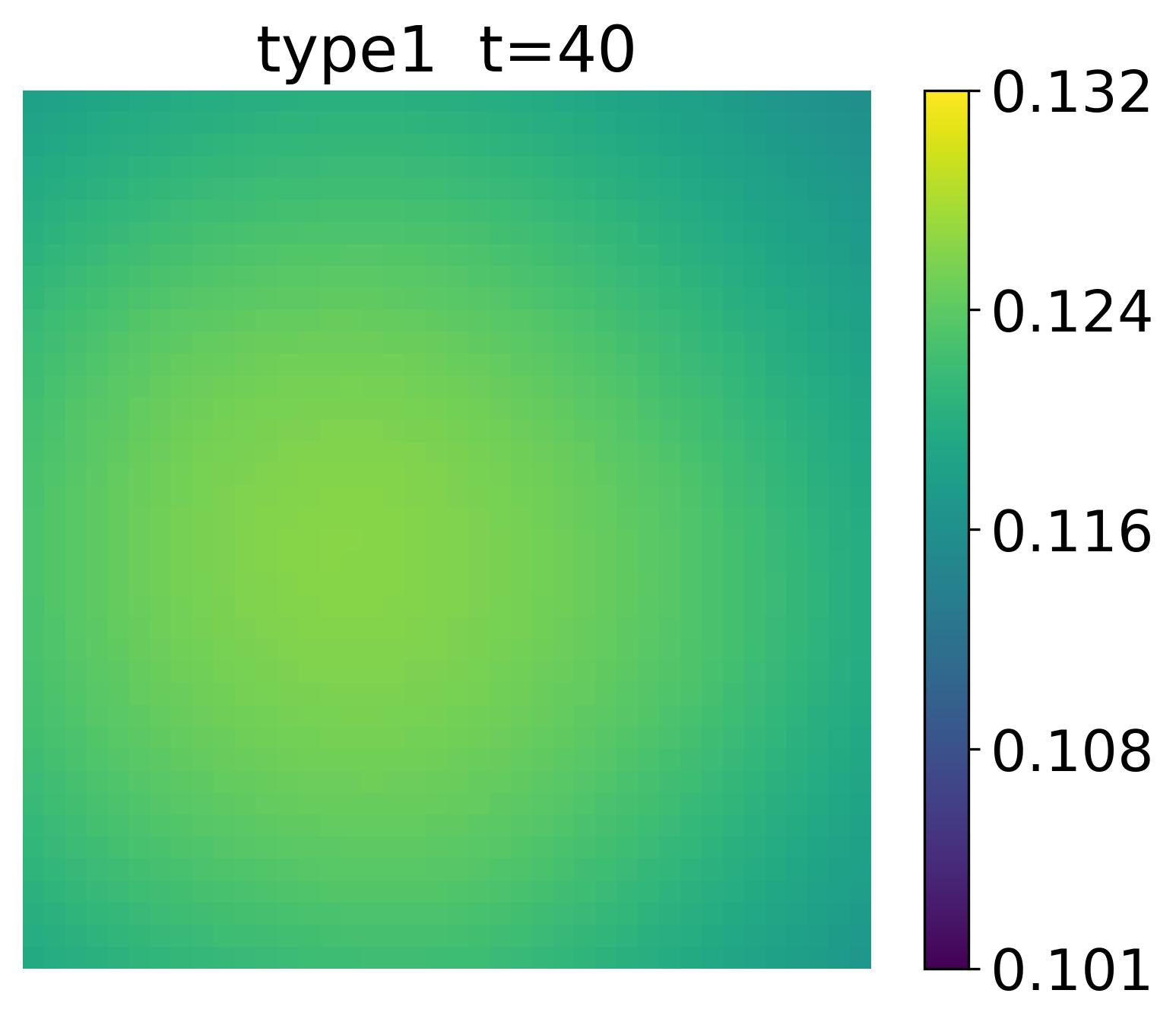} &
\includegraphics[width=0.22\columnwidth]{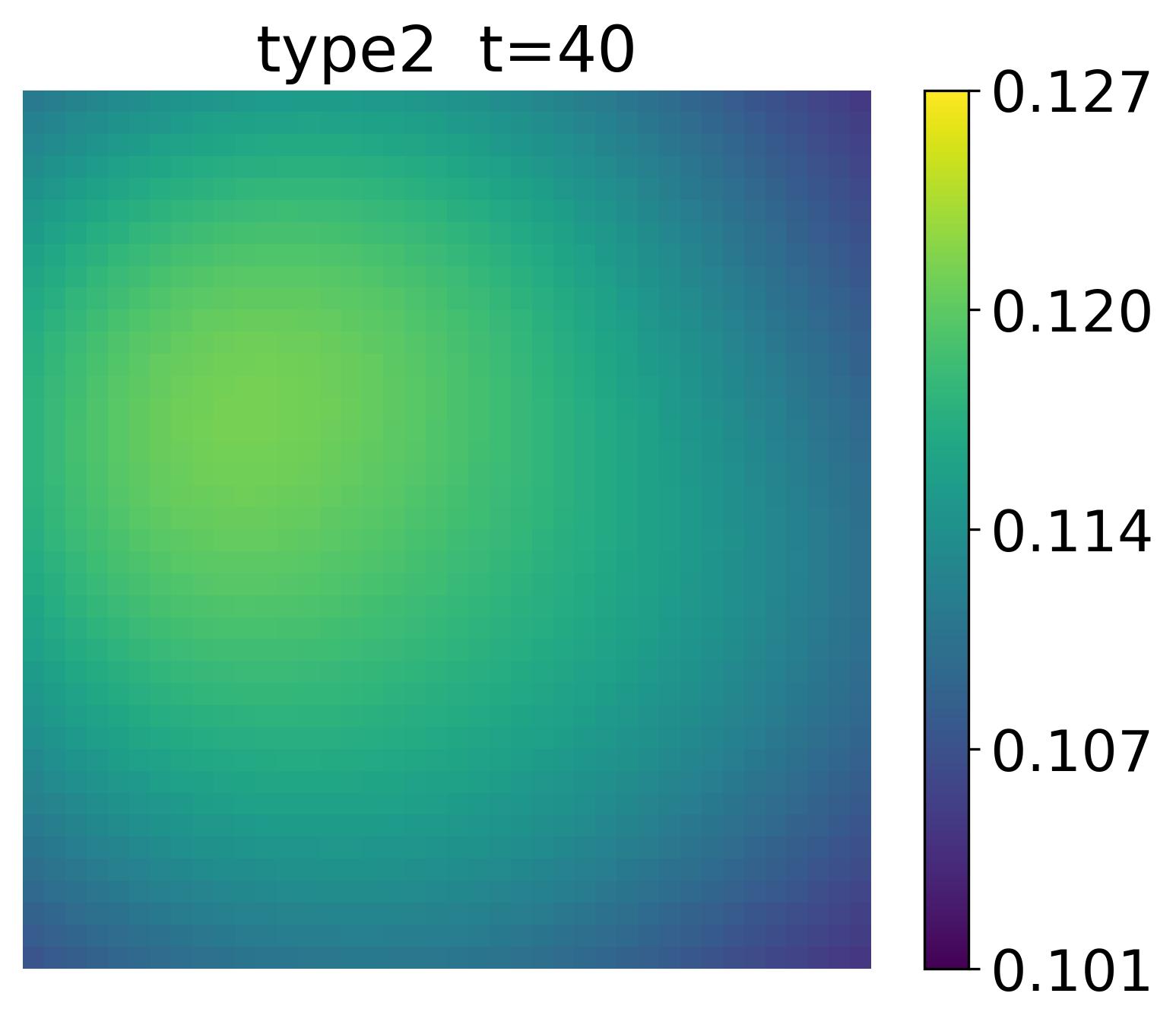} &
\includegraphics[width=0.22\columnwidth]{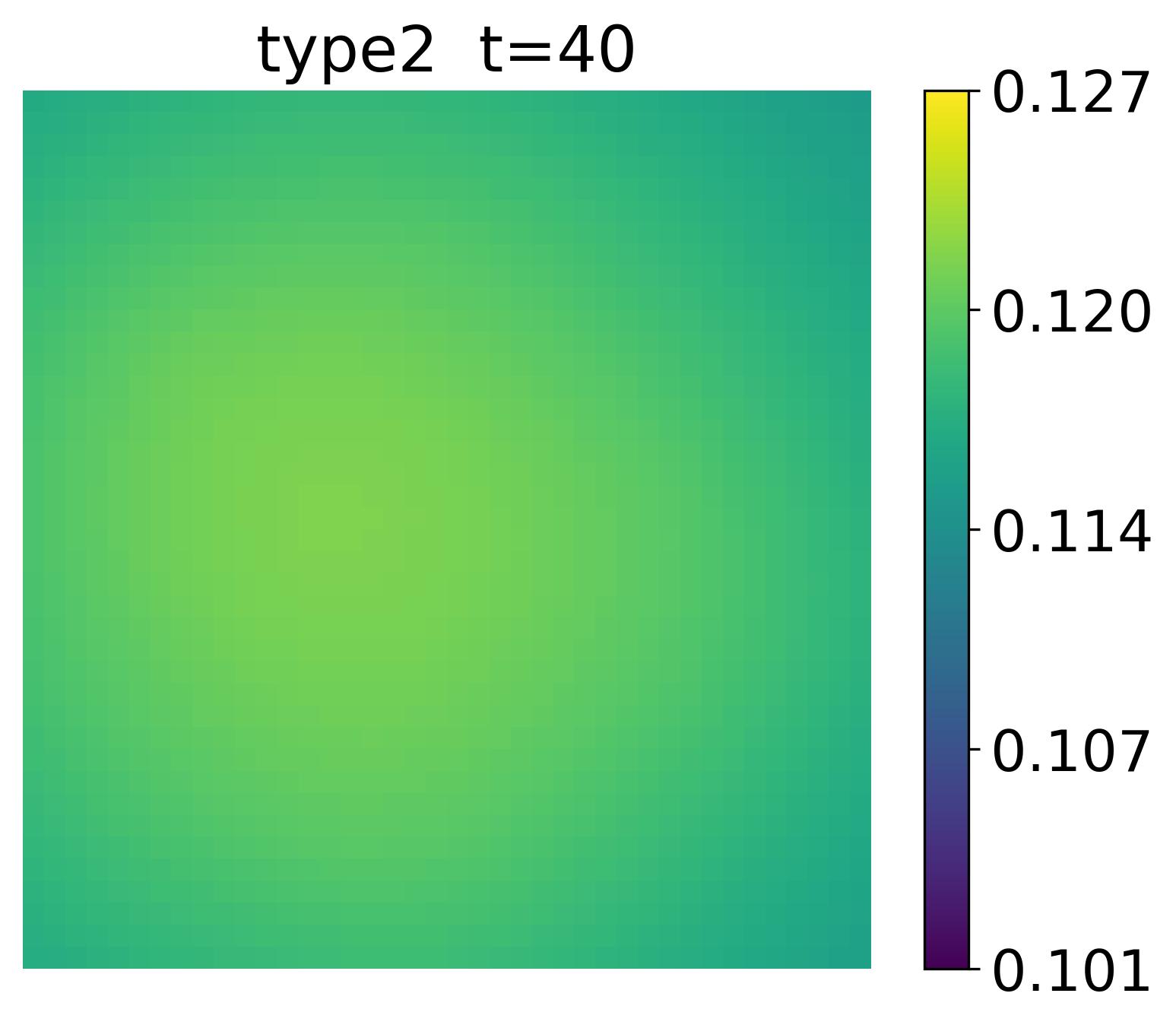} \\

\includegraphics[width=0.22\columnwidth]{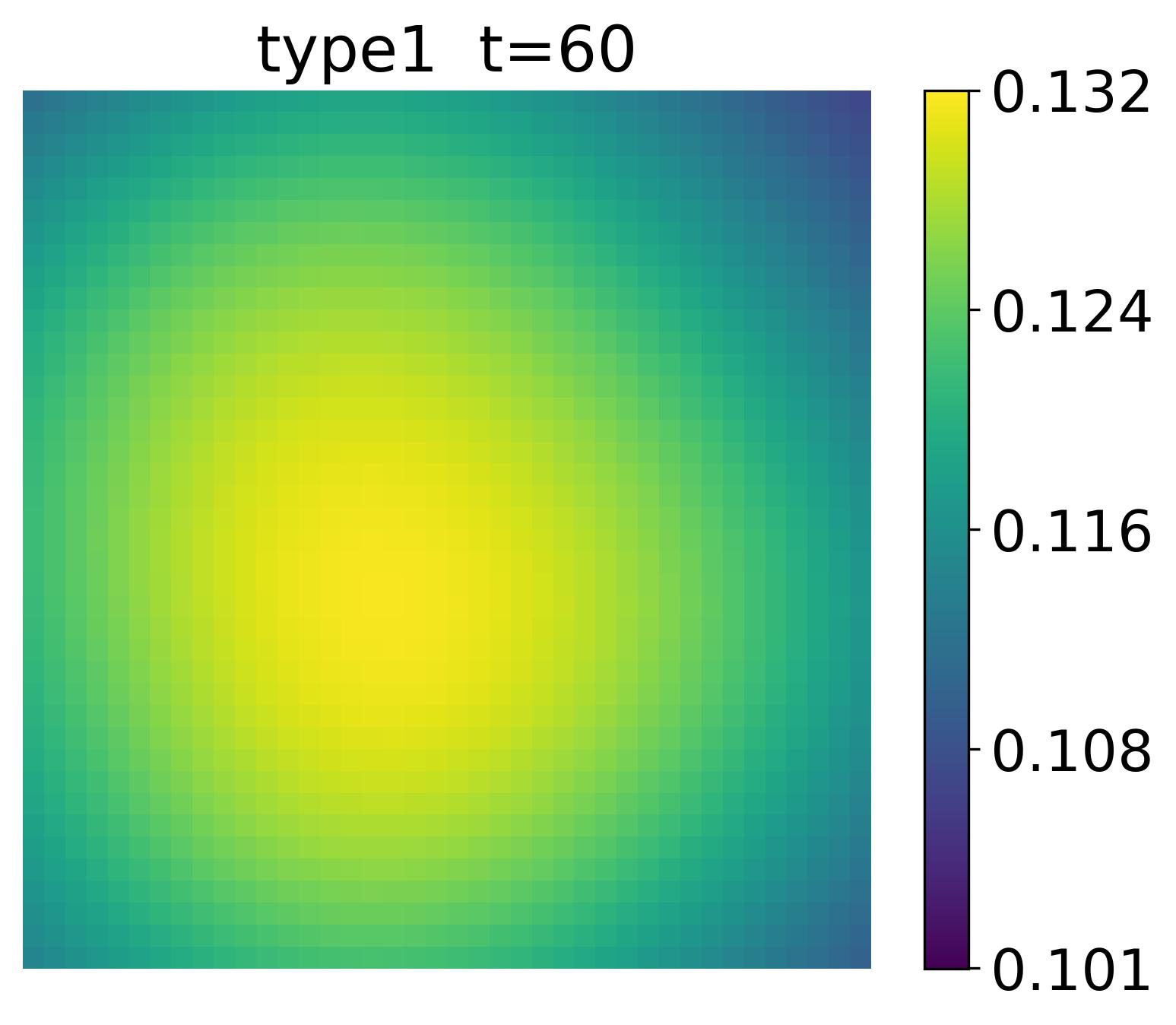} &
\includegraphics[width=0.22\columnwidth]{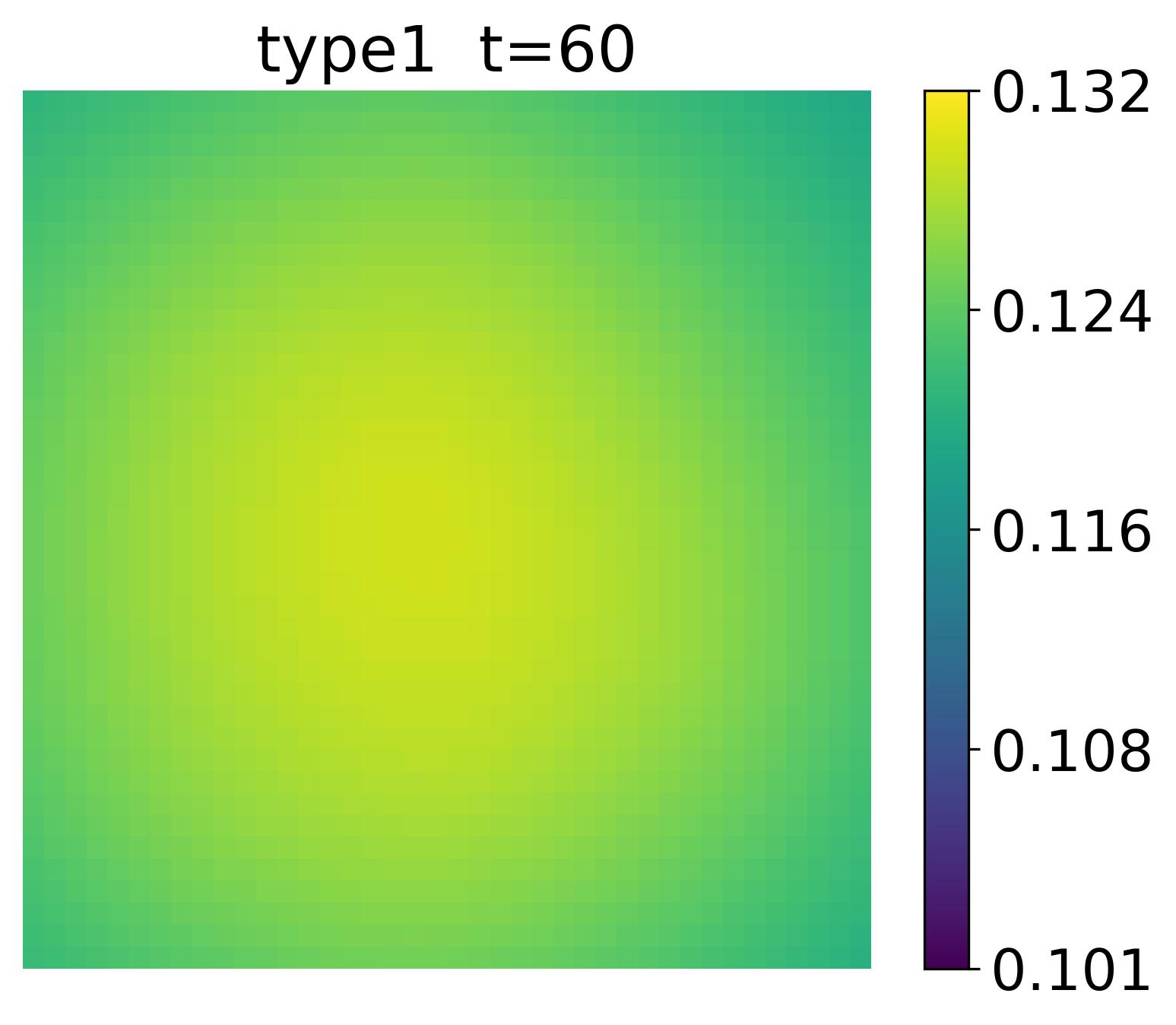} &
\includegraphics[width=0.22\columnwidth]{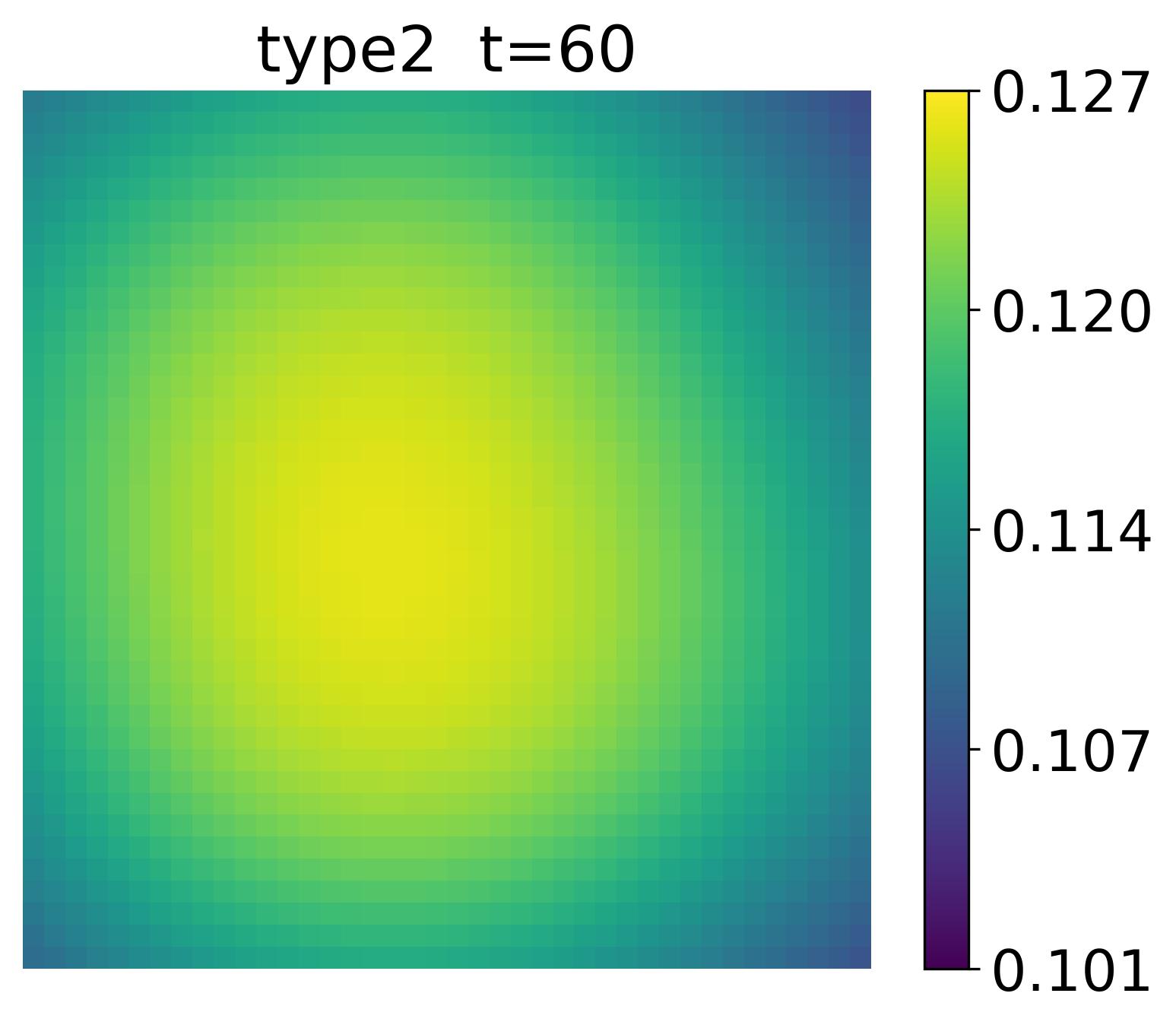} &
\includegraphics[width=0.22\columnwidth]{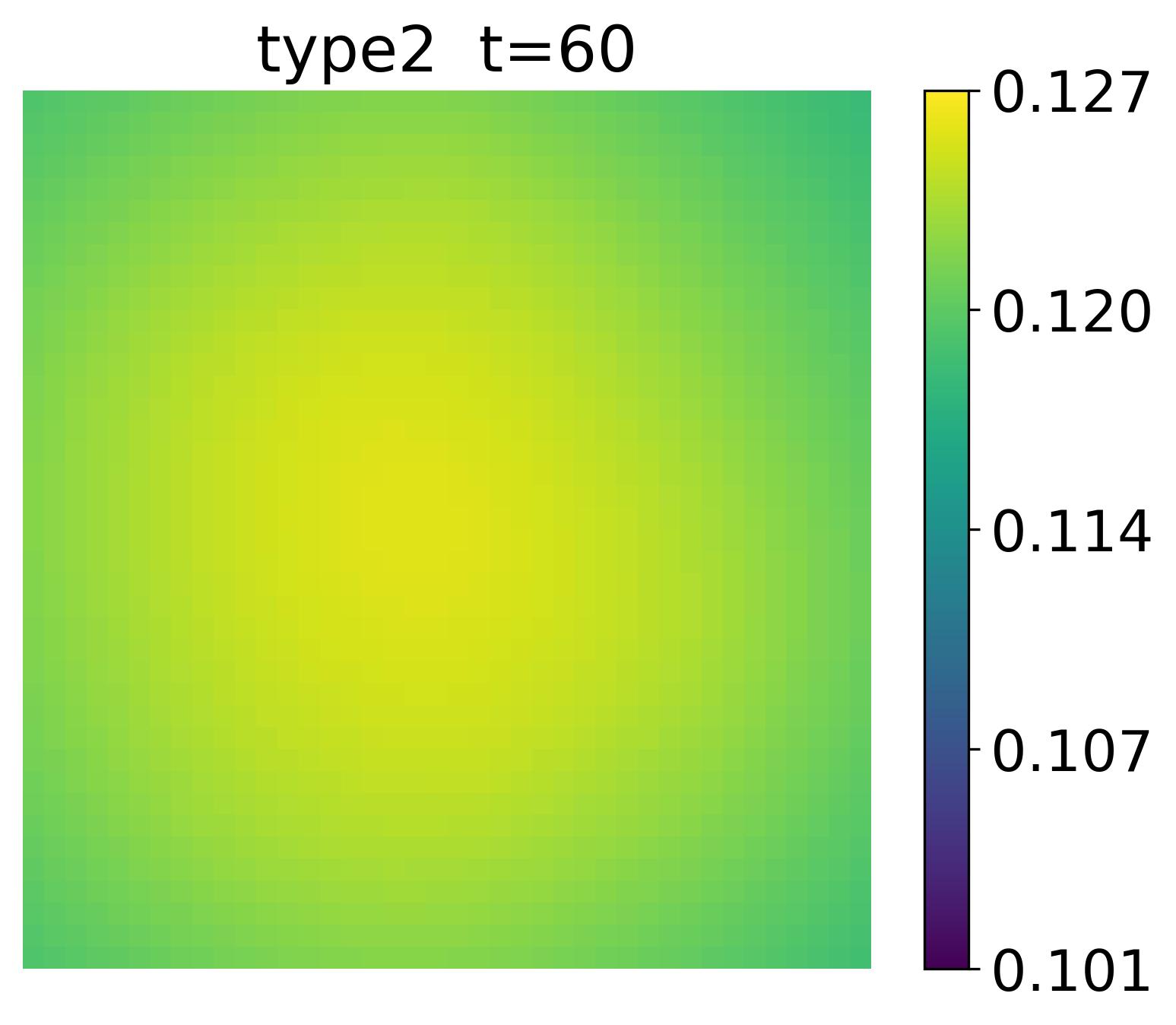} \\

\includegraphics[width=0.22\columnwidth]{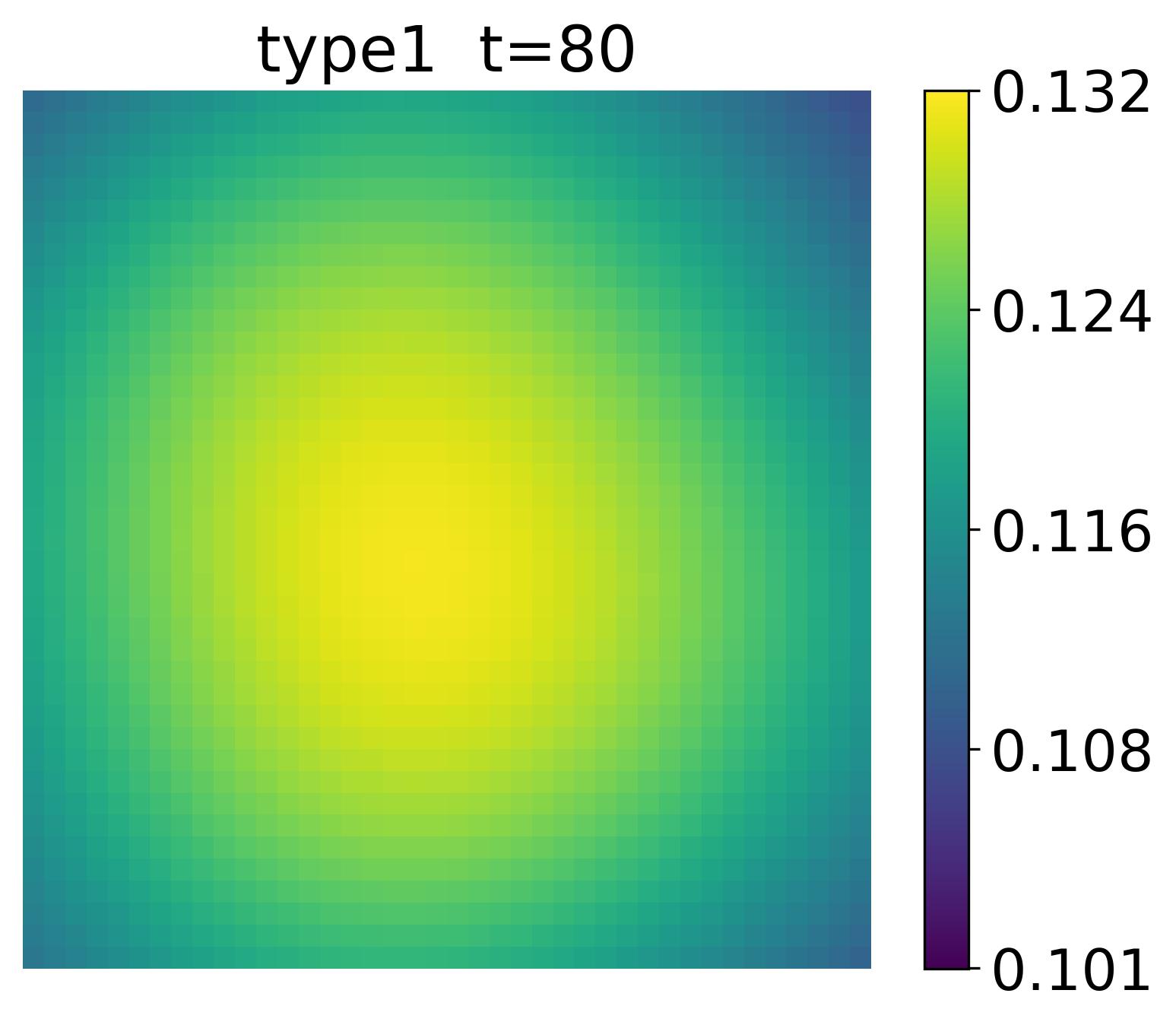} &
\includegraphics[width=0.22\columnwidth]{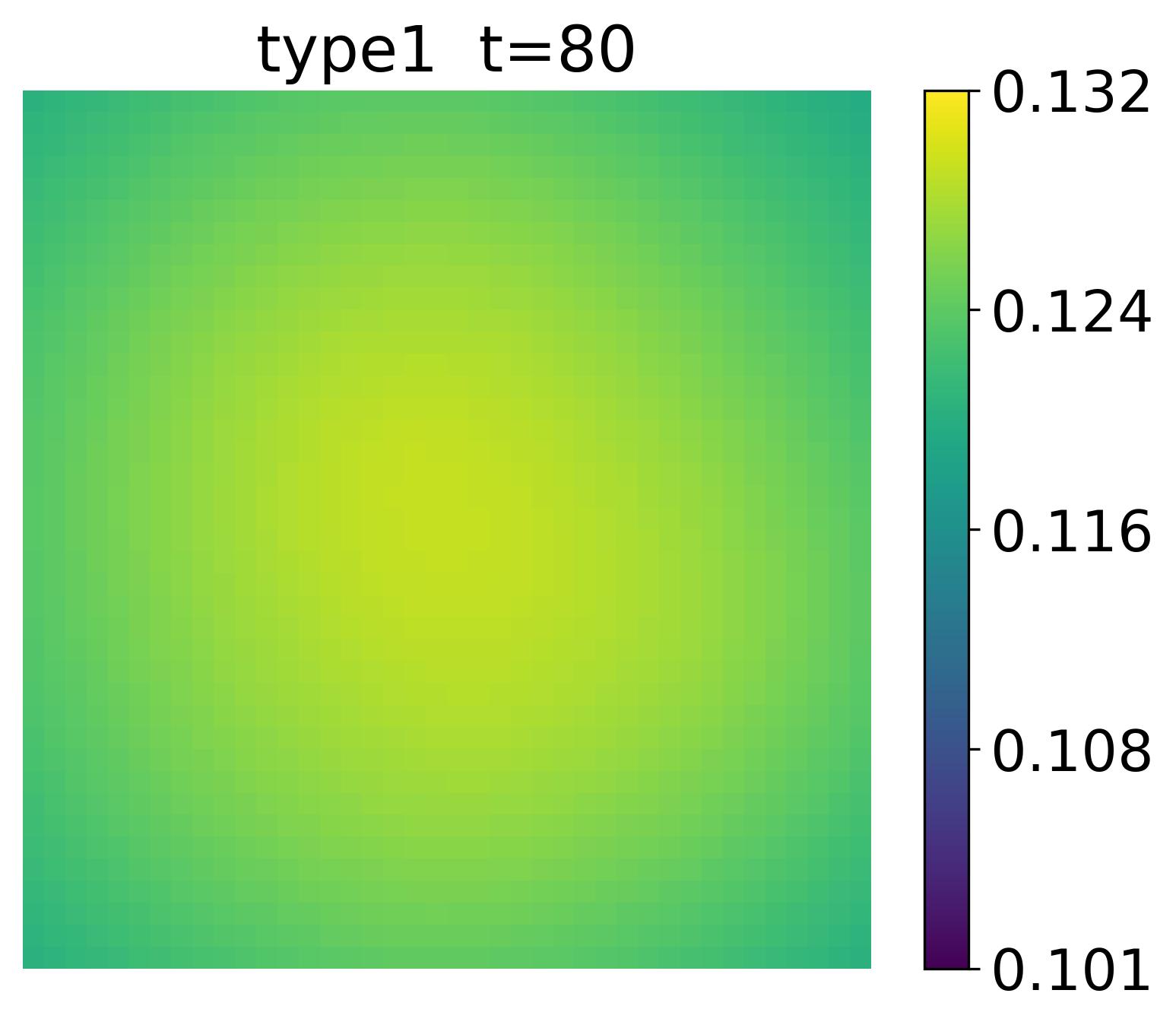} &
\includegraphics[width=0.22\columnwidth]{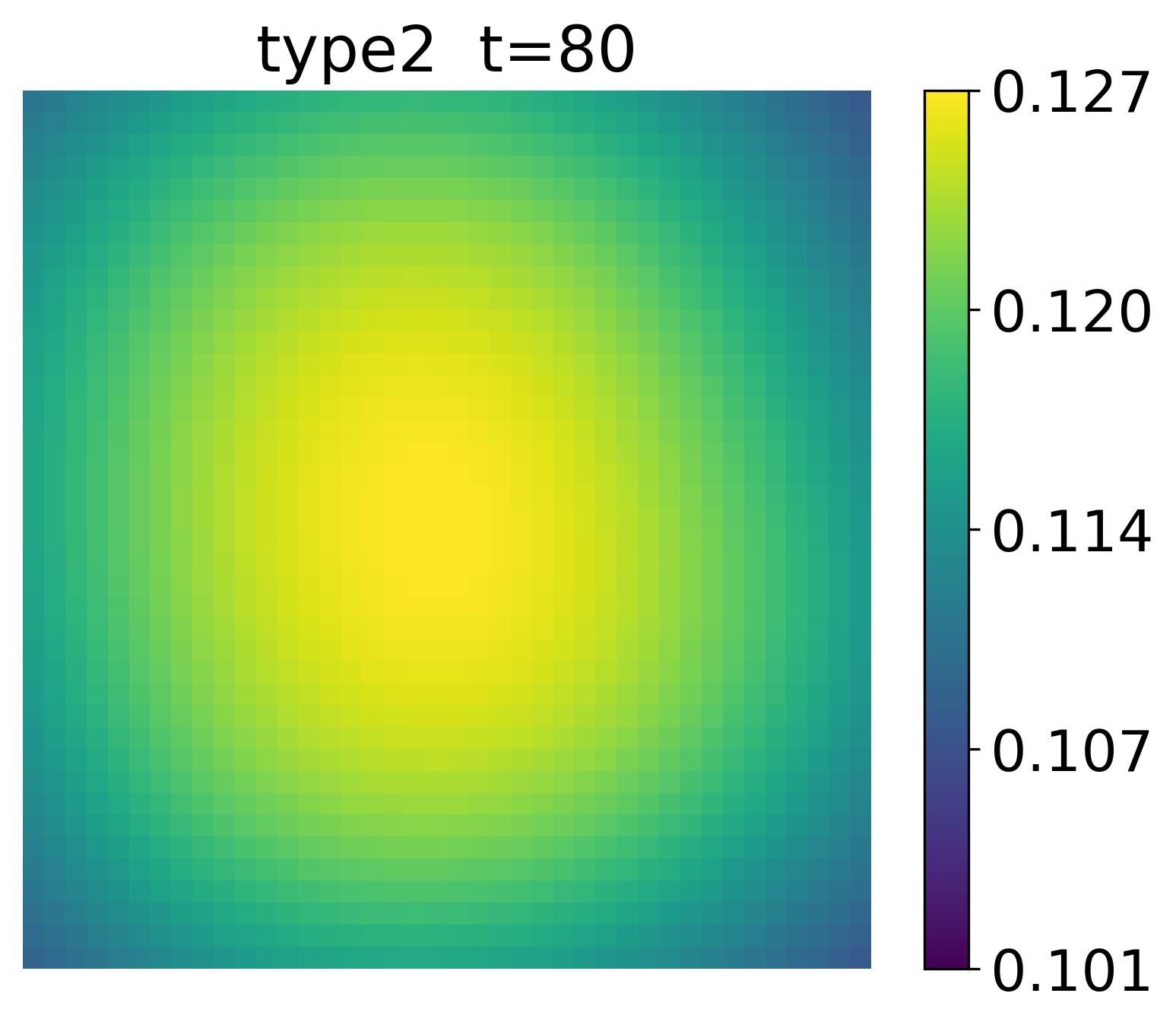} &
\includegraphics[width=0.22\columnwidth]{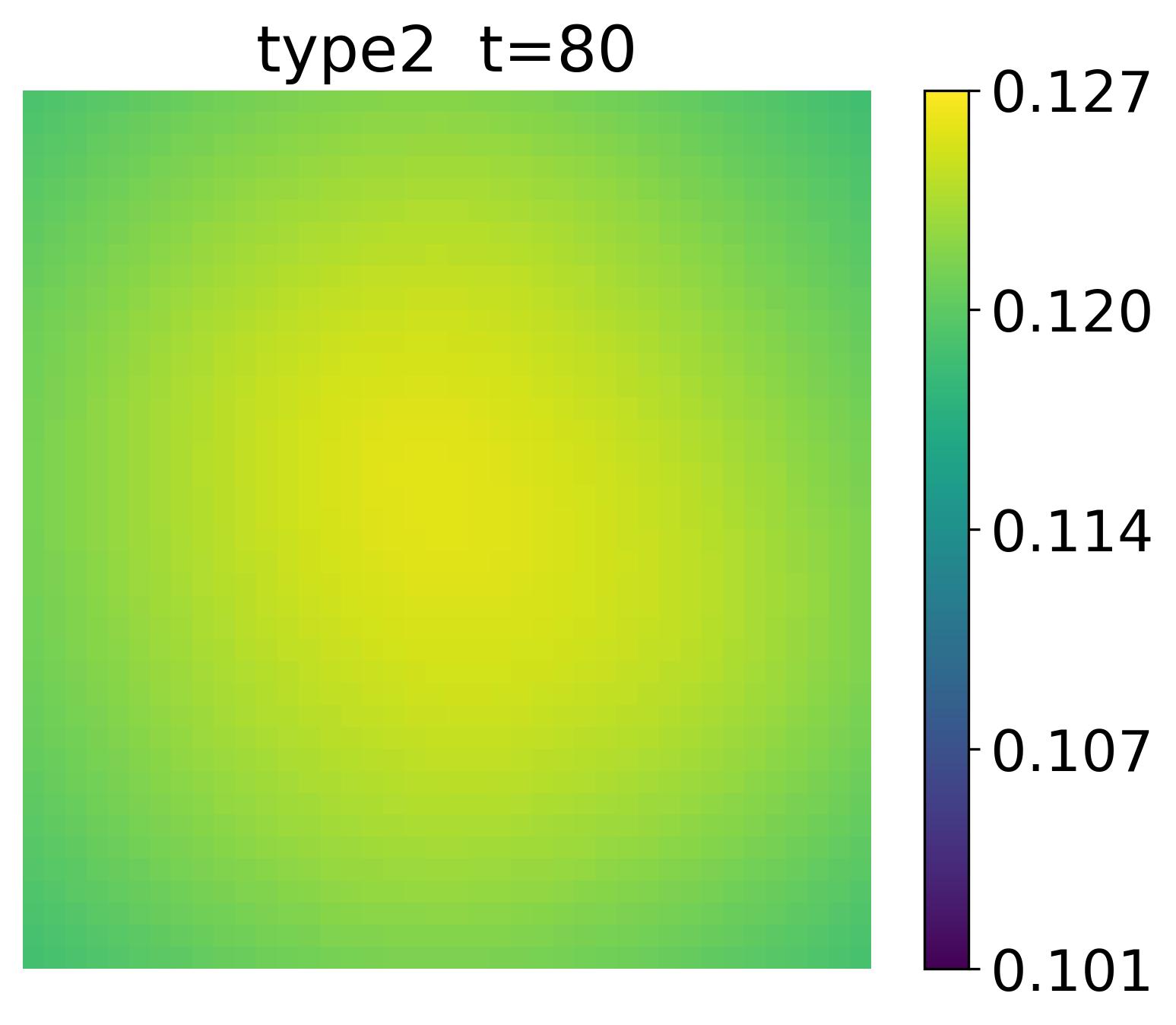} \\

\includegraphics[width=0.22\columnwidth]{figures/cum_avg/biv_3/true/type1/09.jpg} &
\includegraphics[width=0.22\columnwidth]{figures/cum_avg/biv_3/fitted/type1/09.jpg} &
\includegraphics[width=0.22\columnwidth]{figures/cum_avg/biv_3/true/type2/09.jpg} &
\includegraphics[width=0.22\columnwidth]{figures/cum_avg/biv_3/fitted/type2/09.jpg} \\
\end{tabular}

\caption{Same as Figure~\ref{fig:biv1_maps} except for Biv~3.}
\label{fig:biv3_maps}
\end{figure}

\begin{figure}[htbp!]
\centering
\setlength{\tabcolsep}{1pt}
\setlength{\extrarowheight}{3pt}

\begin{tabular}{cccc}

{True} & {Fitted} & {True} & {Fitted} \\[2pt]

\includegraphics[width=0.22\columnwidth]{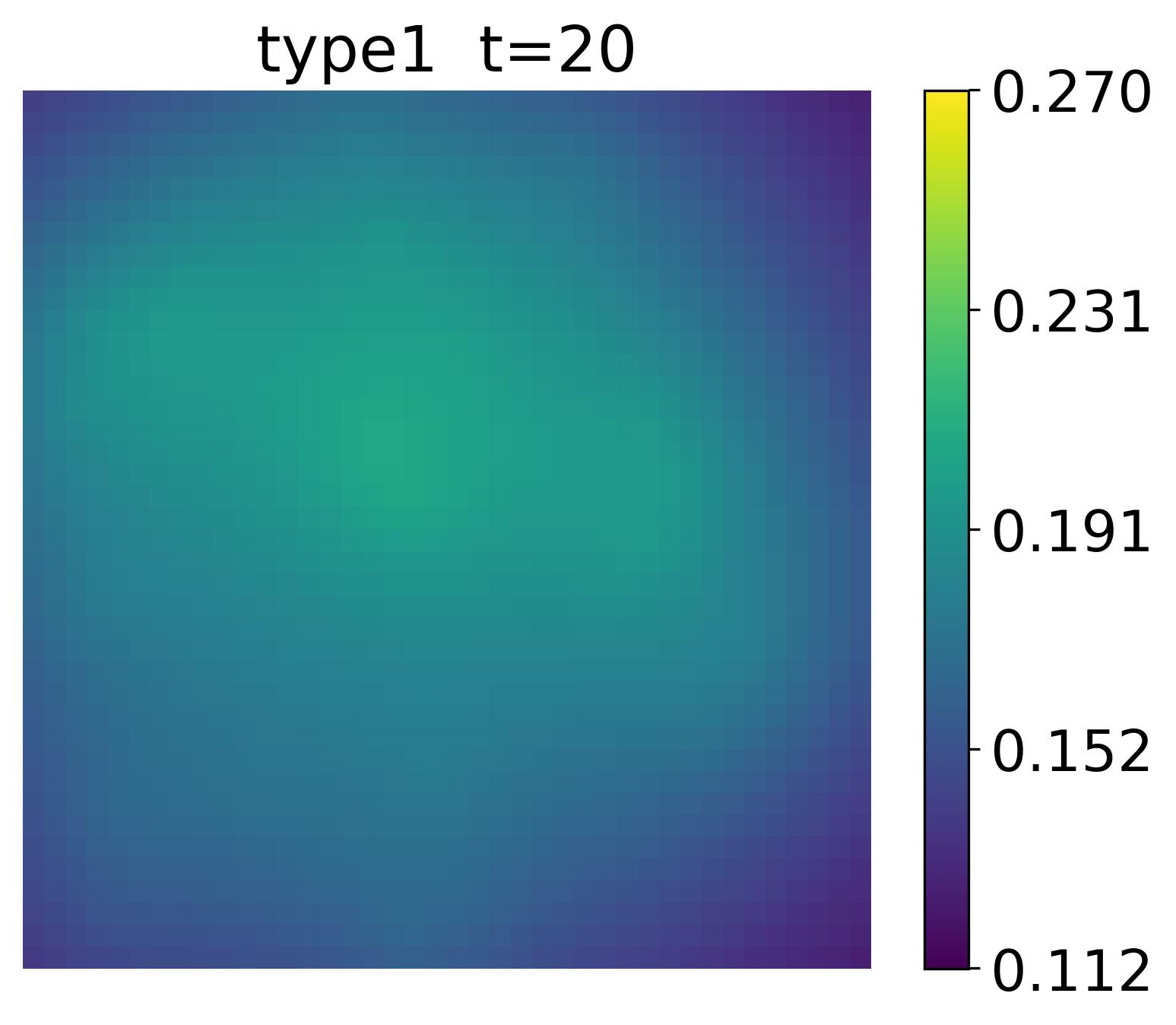} &
\includegraphics[width=0.22\columnwidth]{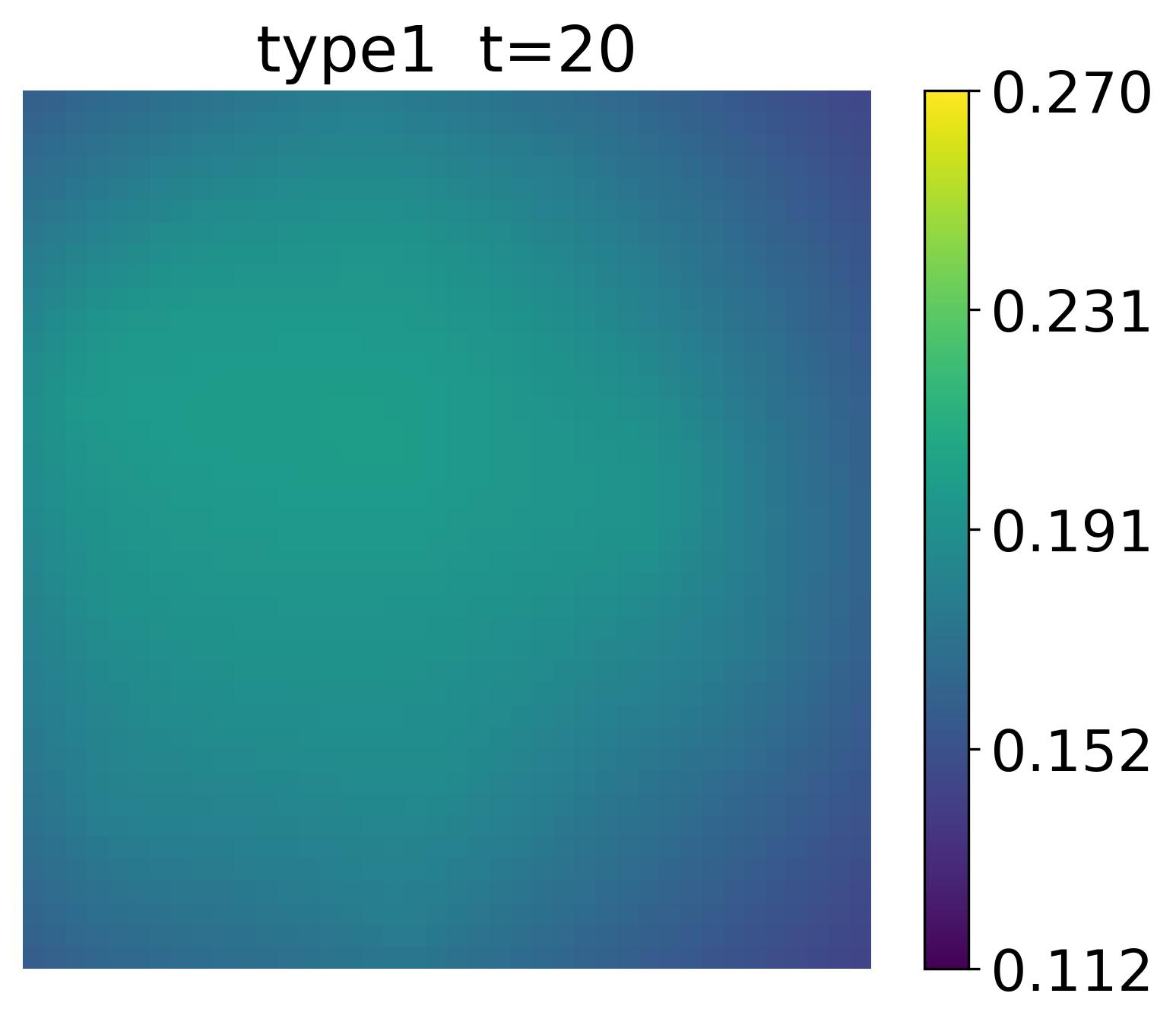} &
\includegraphics[width=0.22\columnwidth]{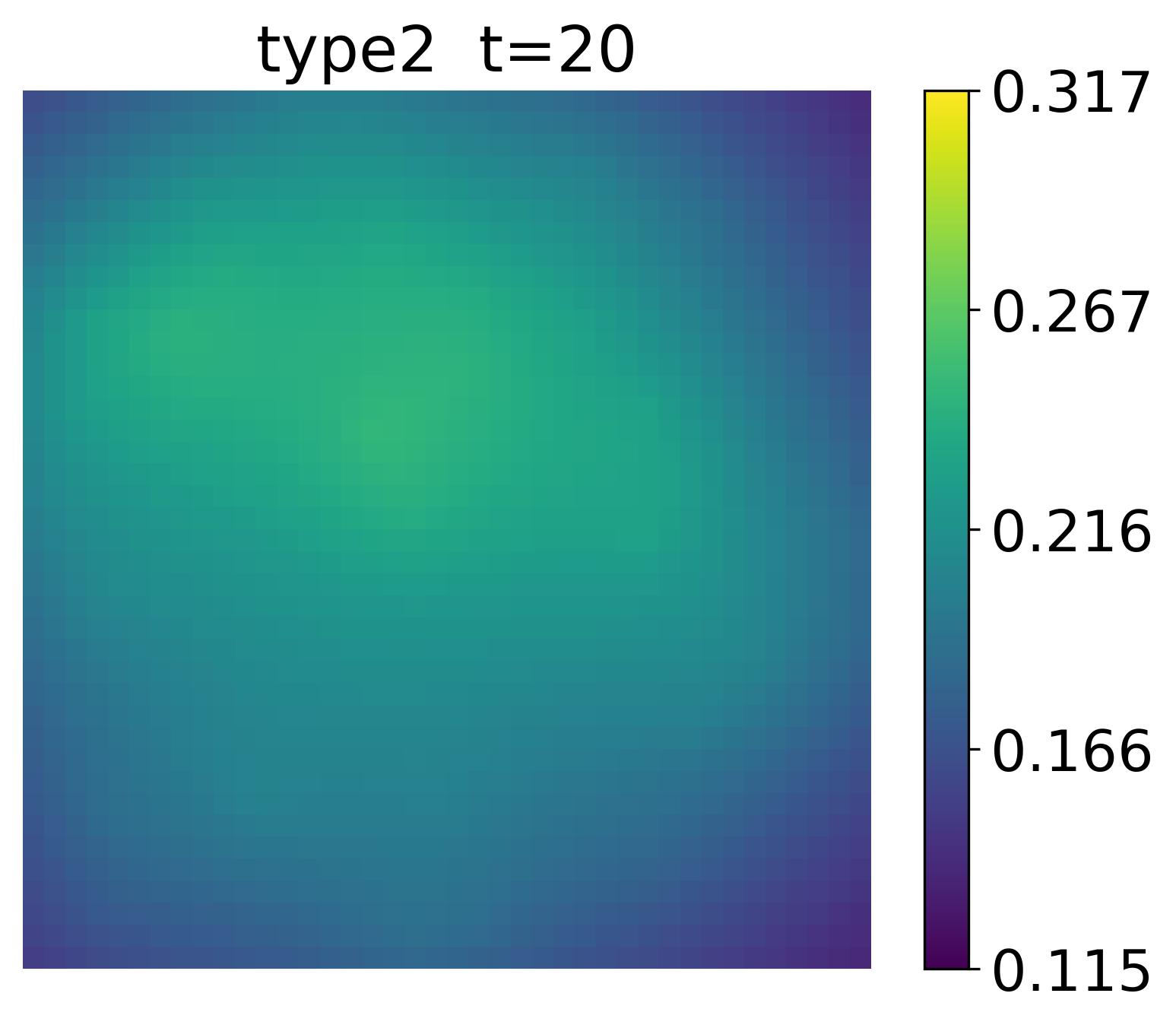} &
\includegraphics[width=0.22\columnwidth]{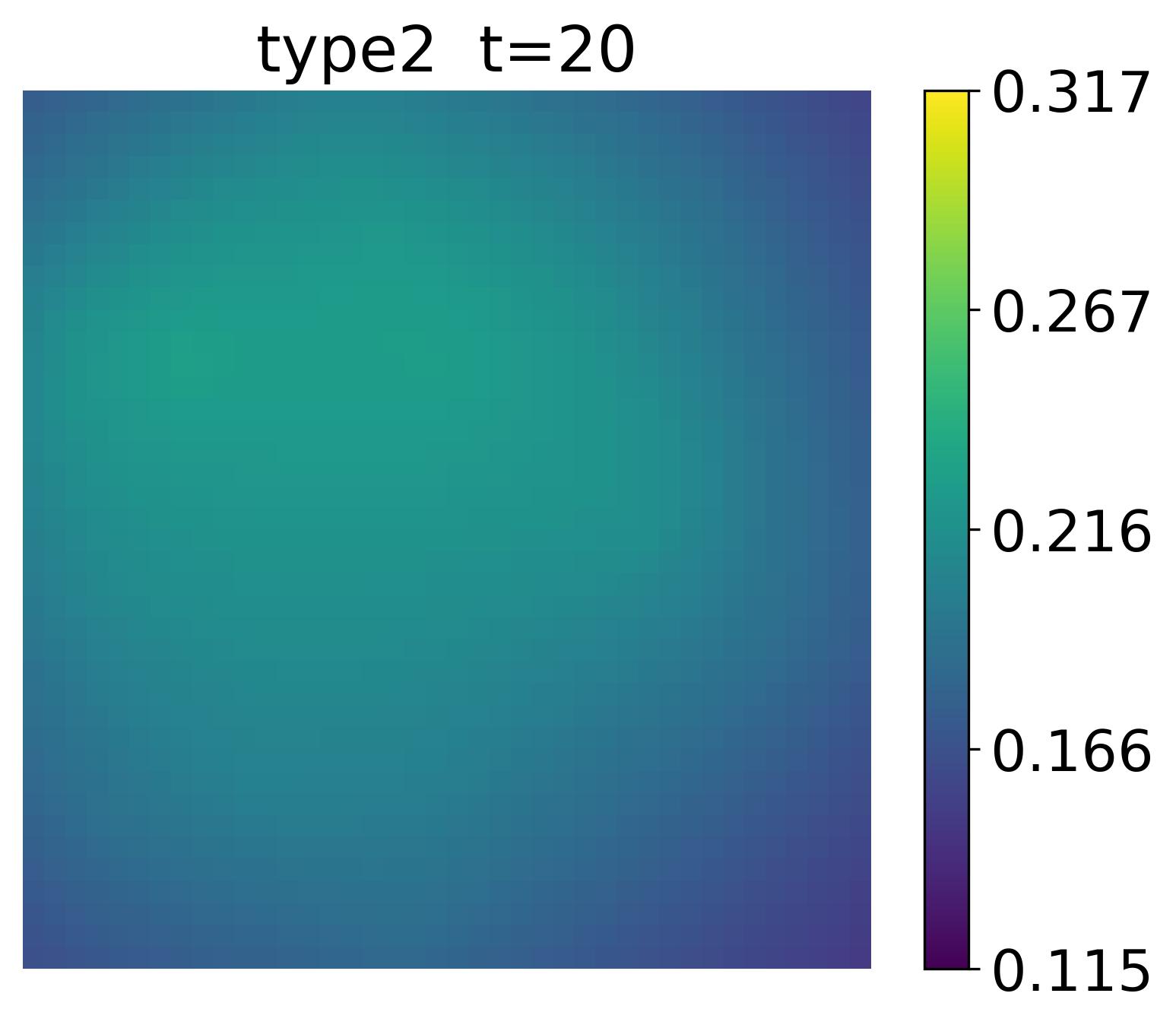} \\

\includegraphics[width=0.22\columnwidth]{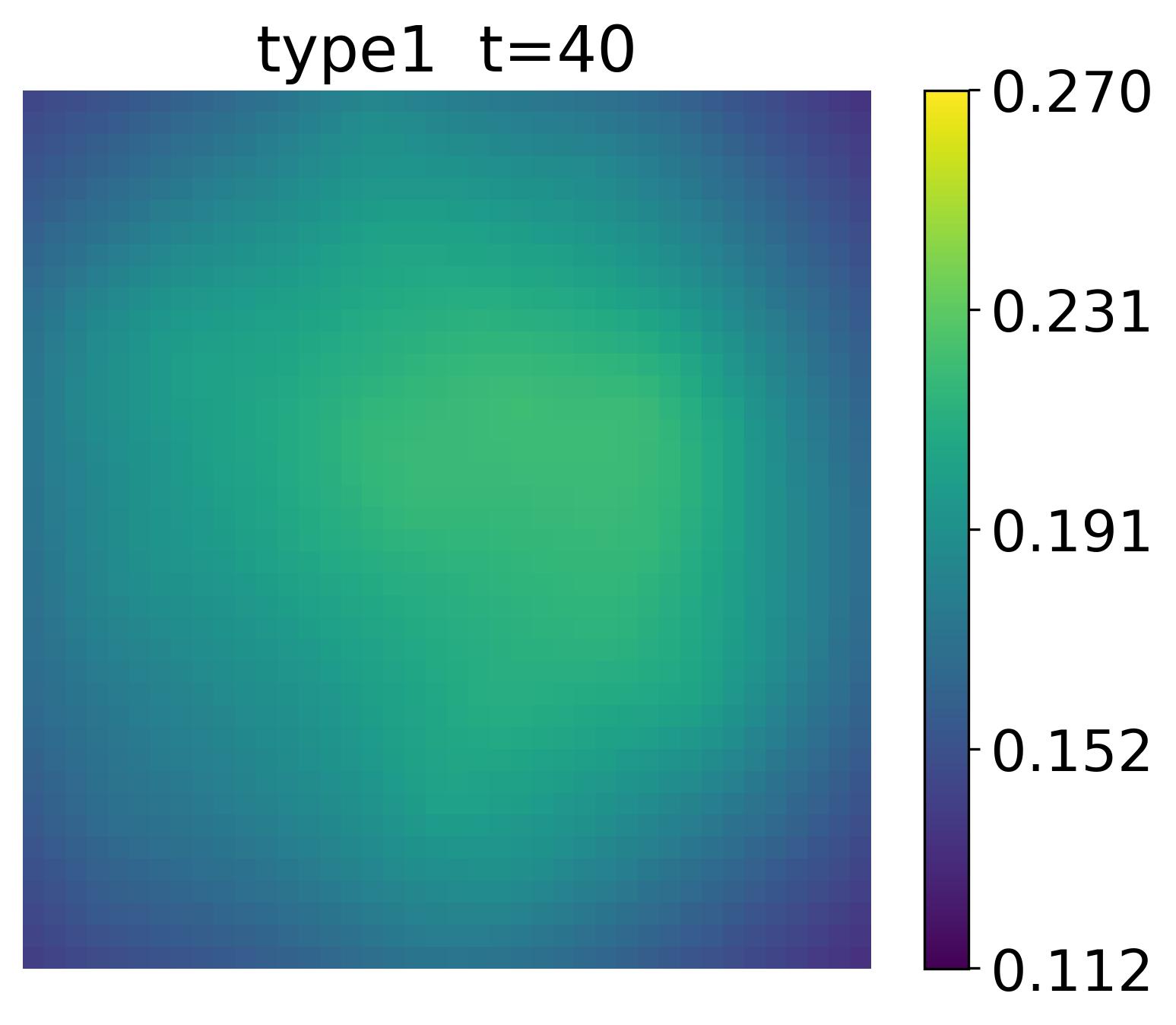} &
\includegraphics[width=0.22\columnwidth]{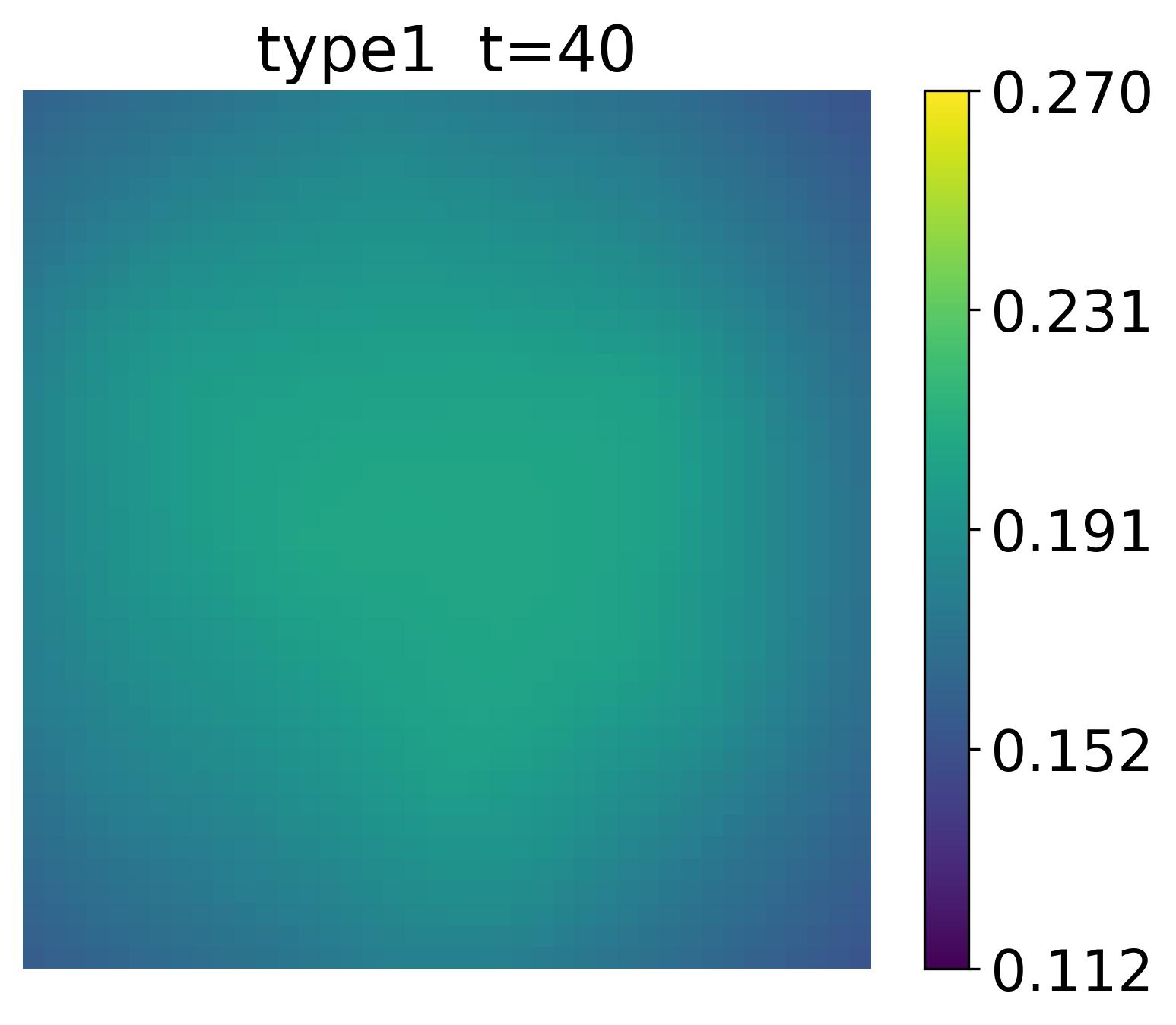} &
\includegraphics[width=0.22\columnwidth]{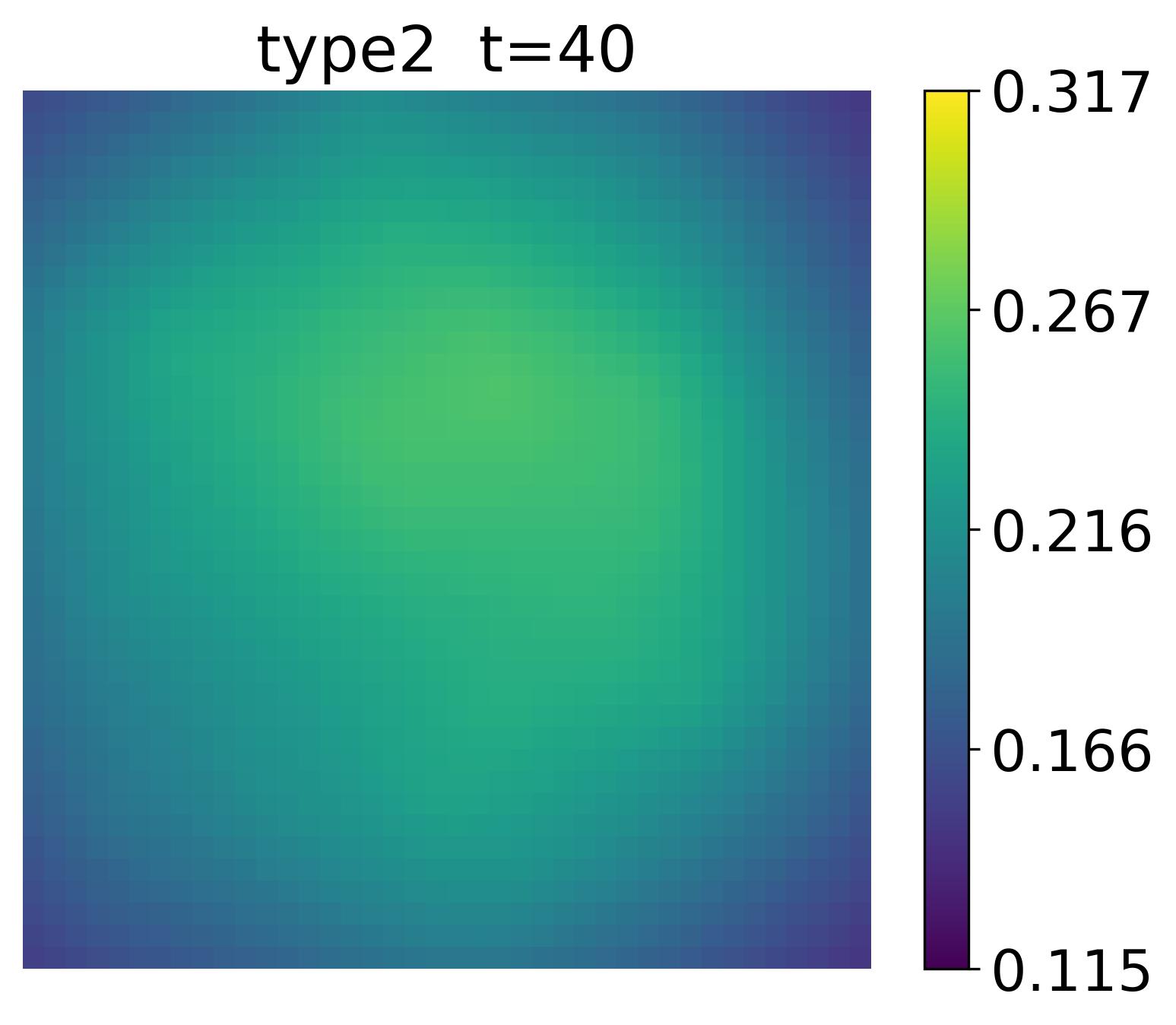} &
\includegraphics[width=0.22\columnwidth]{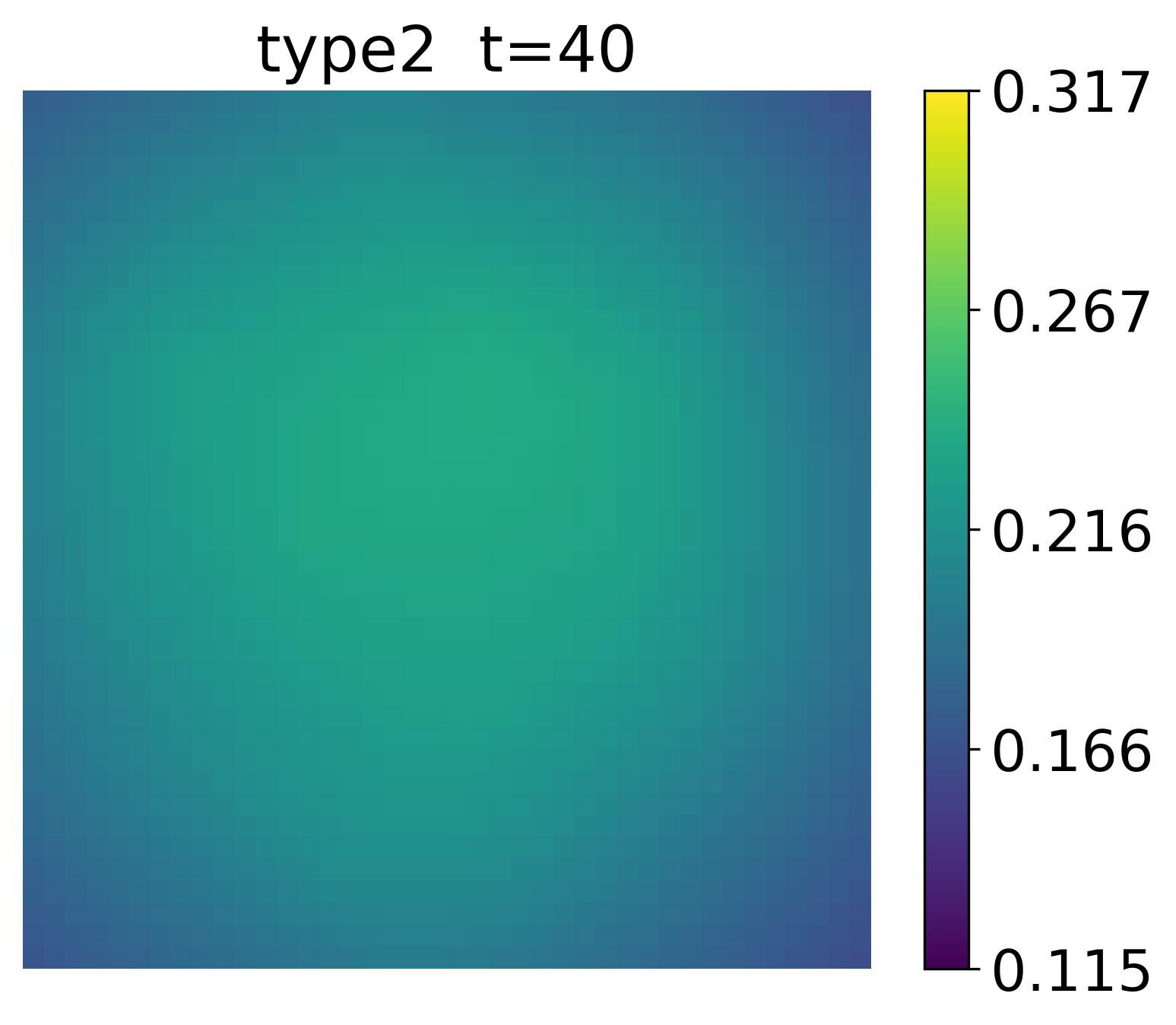} \\

\includegraphics[width=0.22\columnwidth]{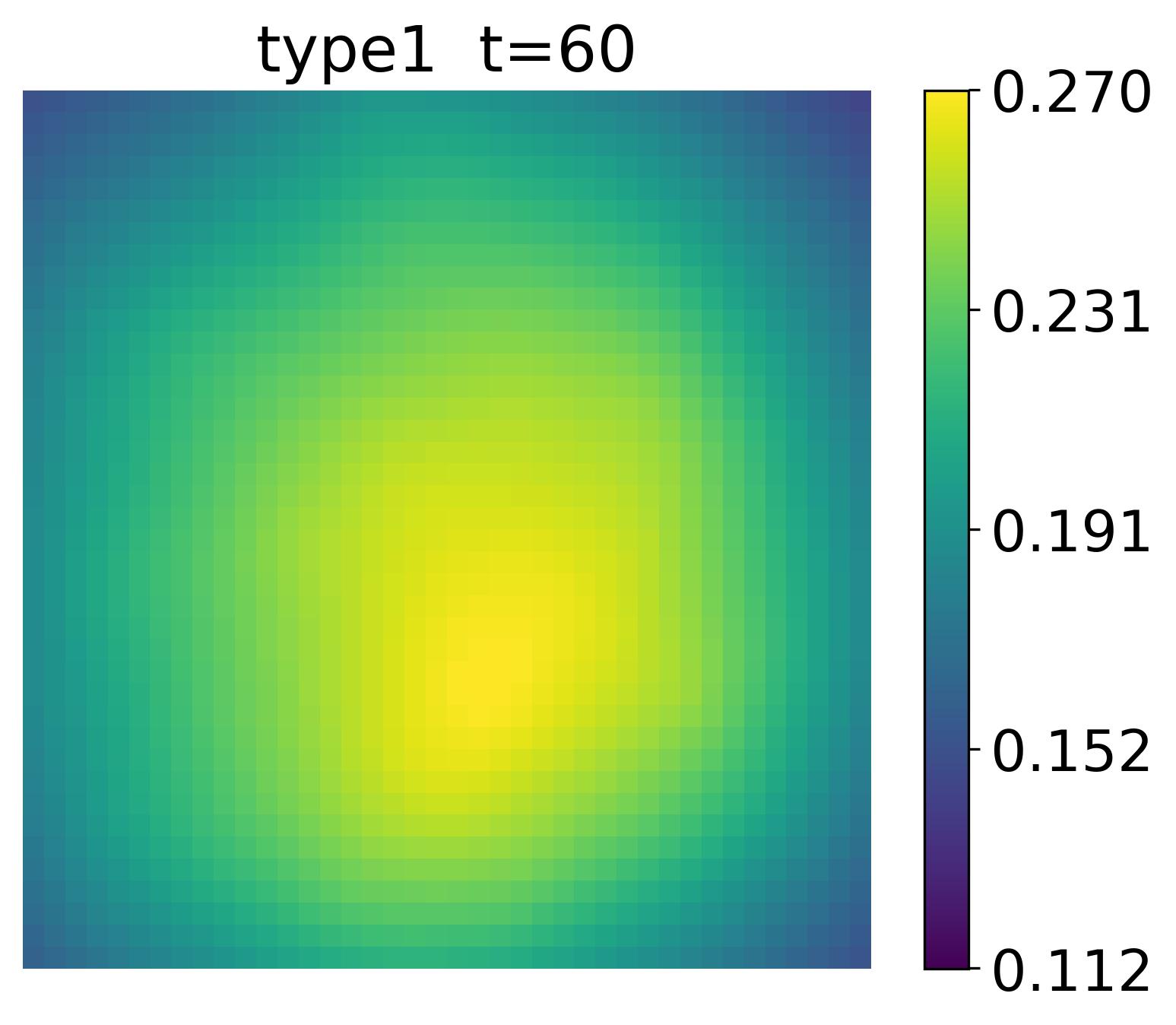} &
\includegraphics[width=0.22\columnwidth]{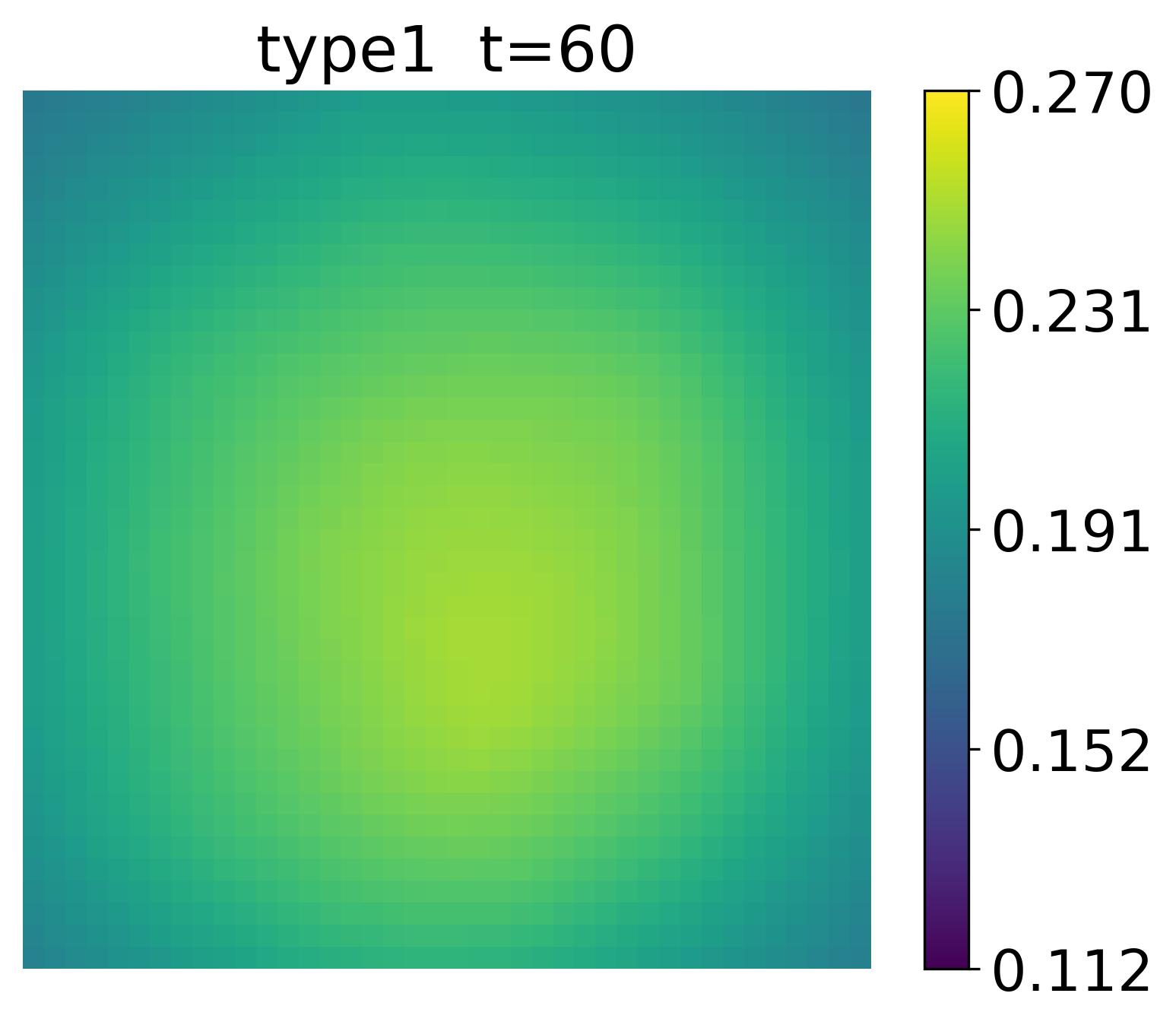} &
\includegraphics[width=0.22\columnwidth]{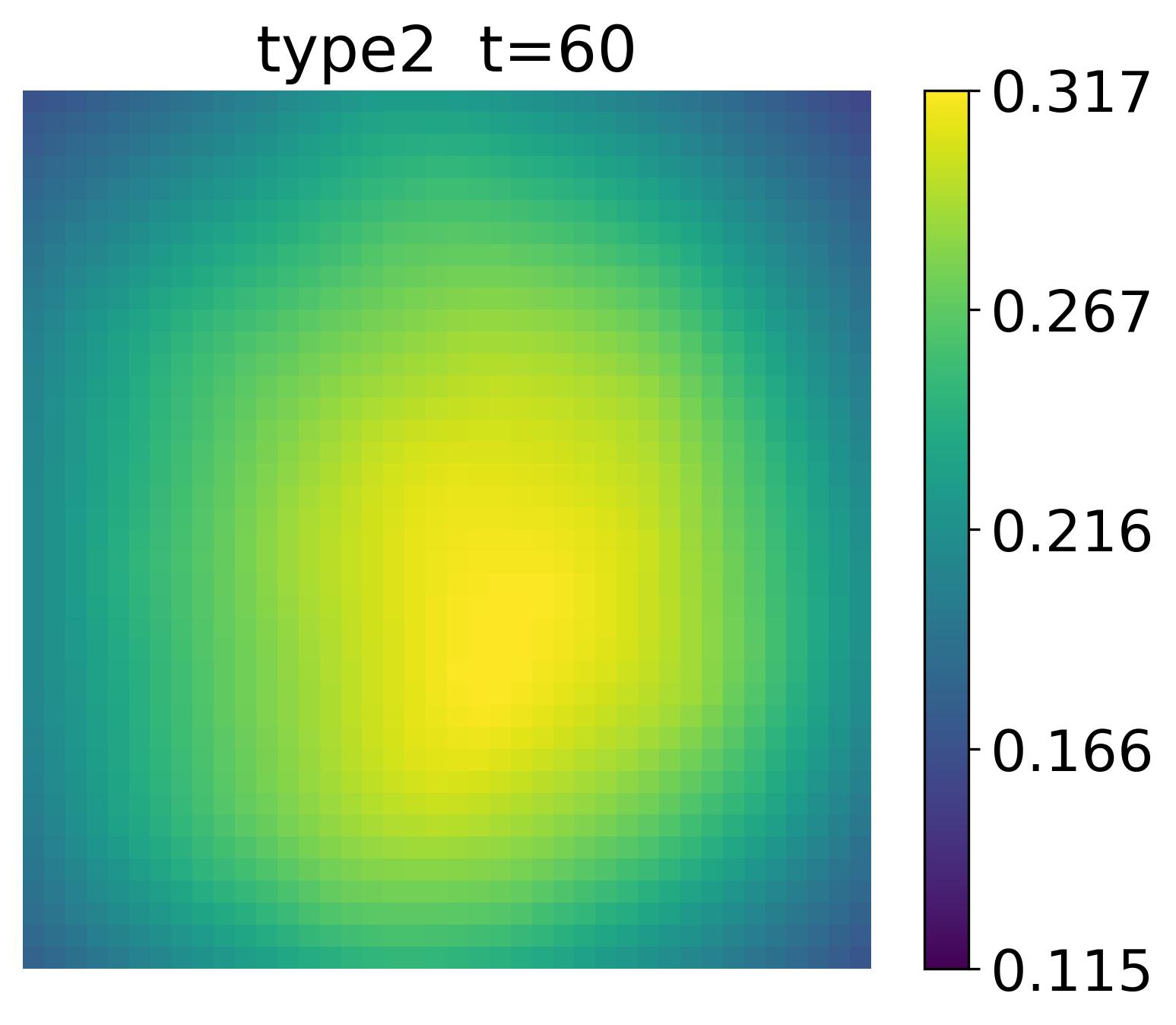} &
\includegraphics[width=0.22\columnwidth]{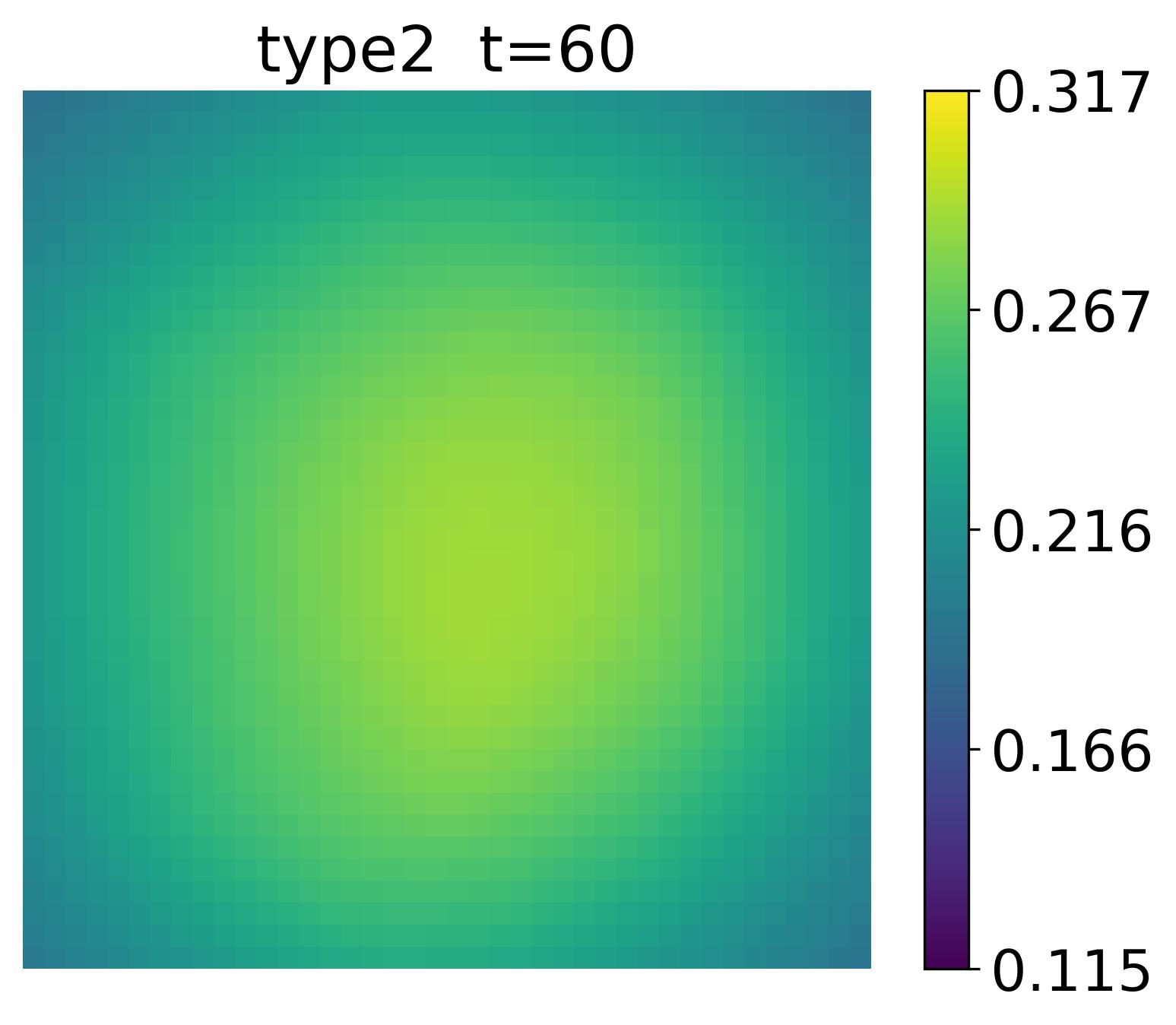} \\

\includegraphics[width=0.22\columnwidth]{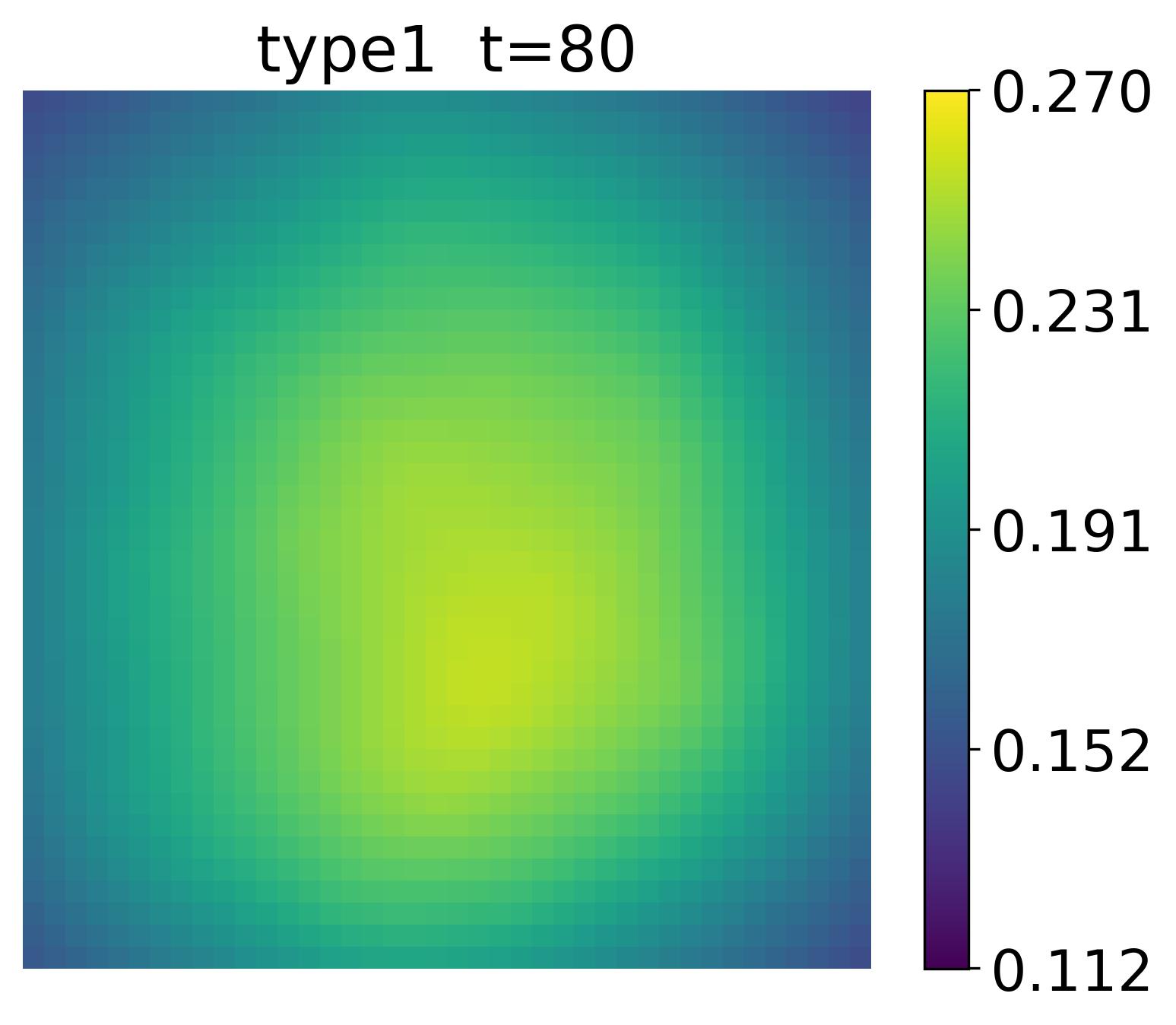} &
\includegraphics[width=0.22\columnwidth]{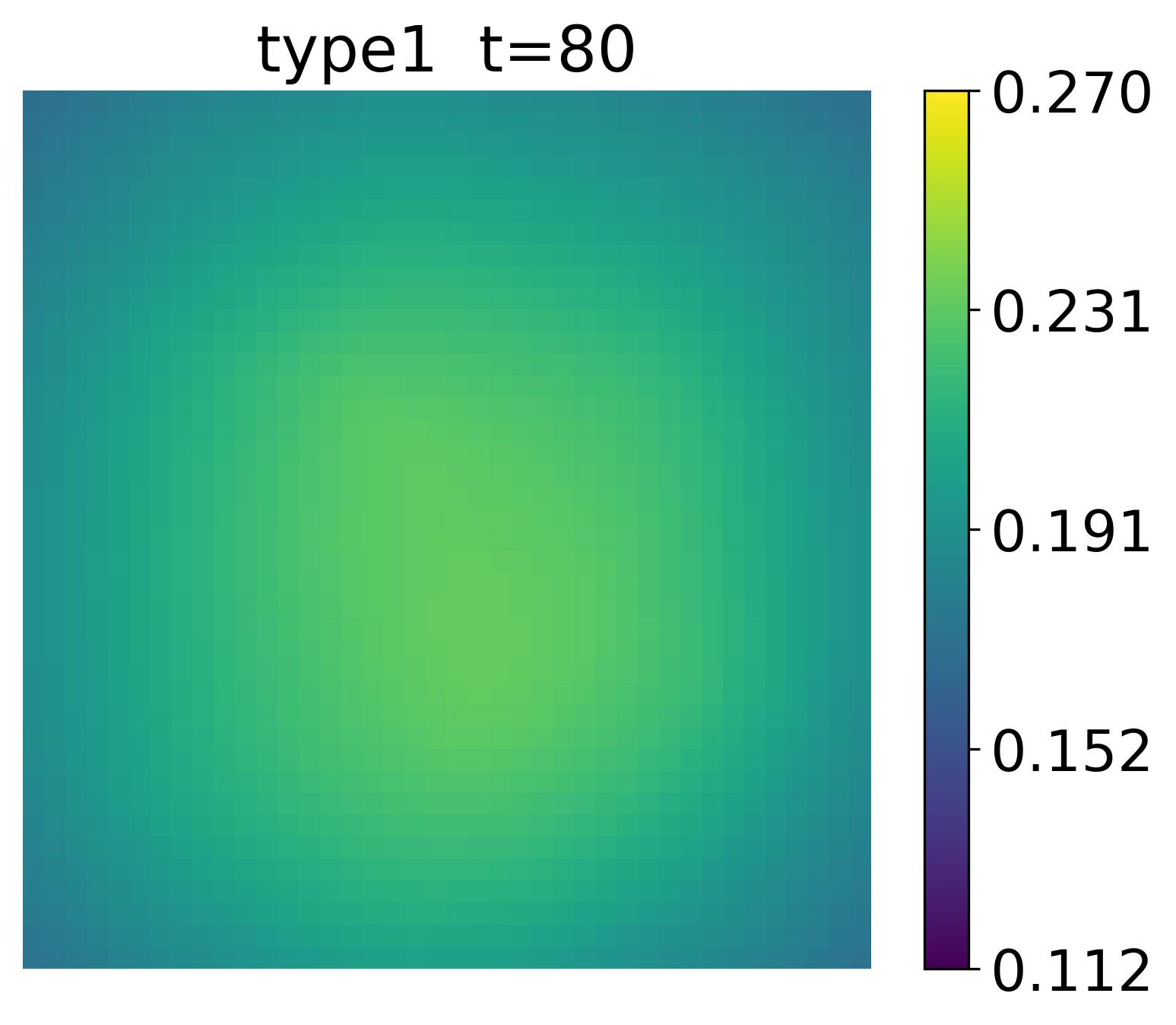} &
\includegraphics[width=0.22\columnwidth]{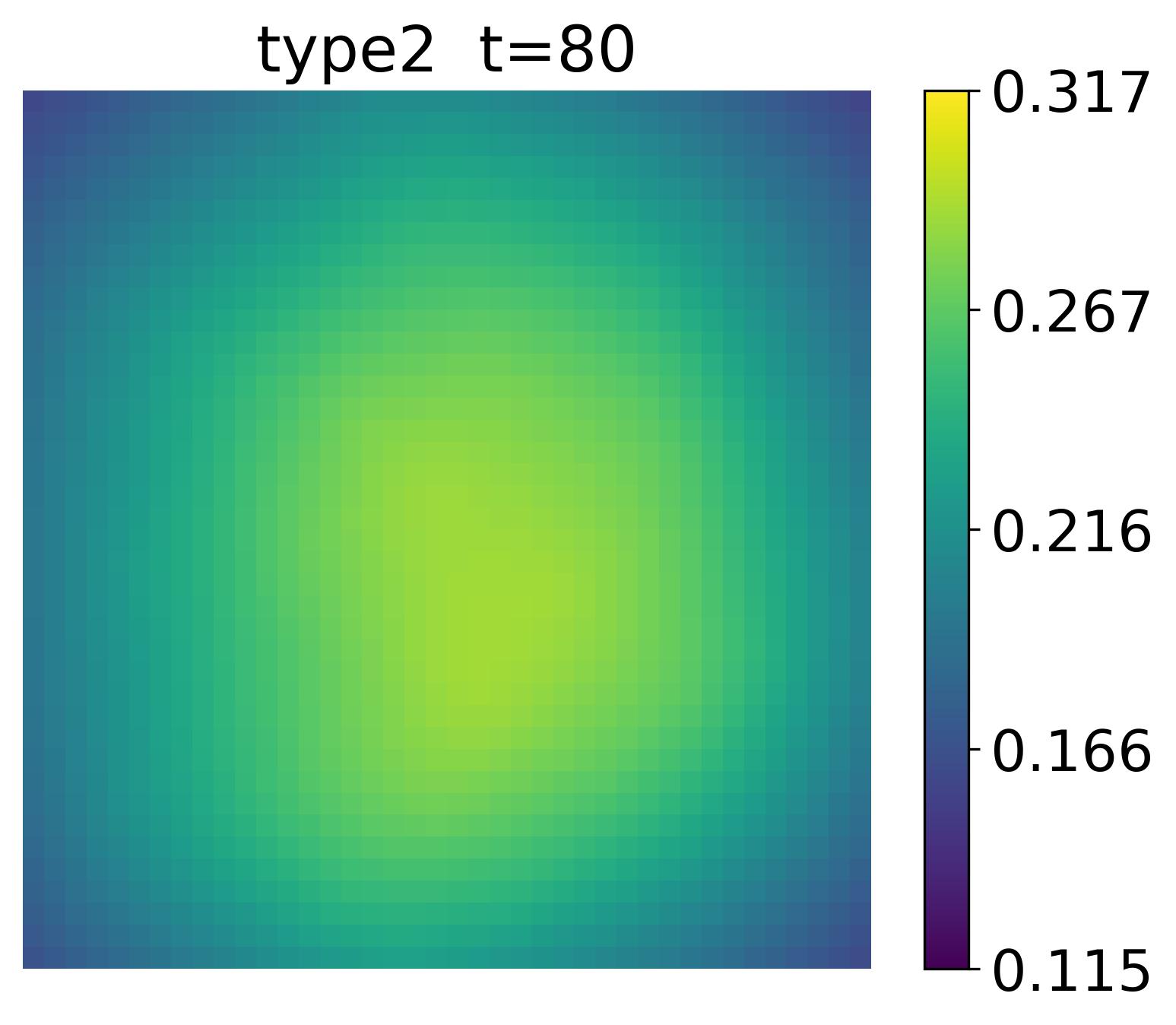} &
\includegraphics[width=0.22\columnwidth]{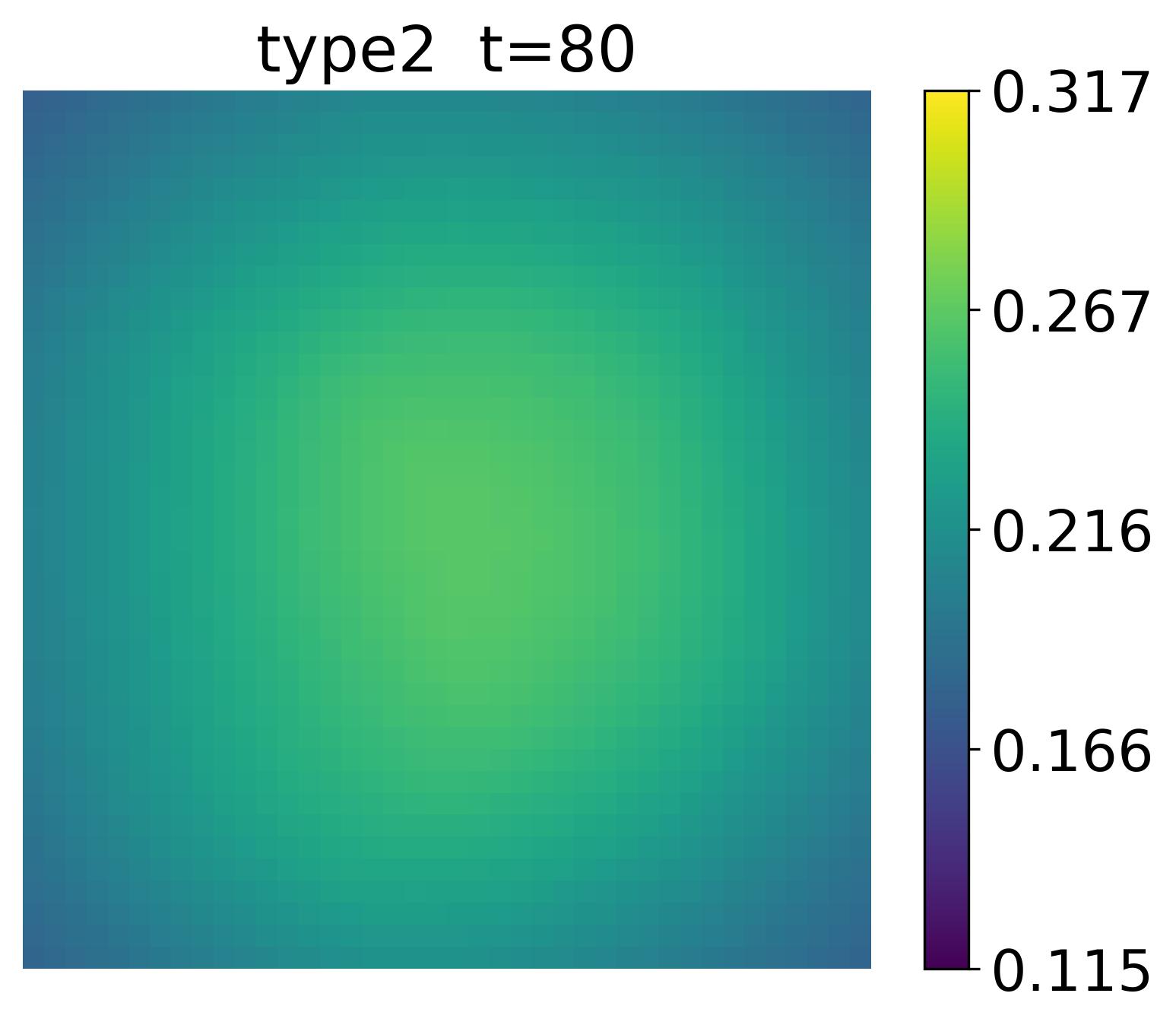} \\

\includegraphics[width=0.22\columnwidth]{figures/cum_avg/biv_4/true/type1/09.jpg} &
\includegraphics[width=0.22\columnwidth]{figures/cum_avg/biv_4/fitted/type1/09.jpg} &
\includegraphics[width=0.22\columnwidth]{figures/cum_avg/biv_4/true/type2/09.jpg} &
\includegraphics[width=0.22\columnwidth]{figures/cum_avg/biv_4/fitted/type2/09.jpg} \\
\end{tabular}

\caption{Same as Figure~\ref{fig:biv1_maps} except for Biv~4.}
\end{figure}

\begin{figure}[!htbp]
\centering

\subfloat[Biv 1\label{fig:biv1_tnhp}]{%
  \includegraphics[width=.45\linewidth]{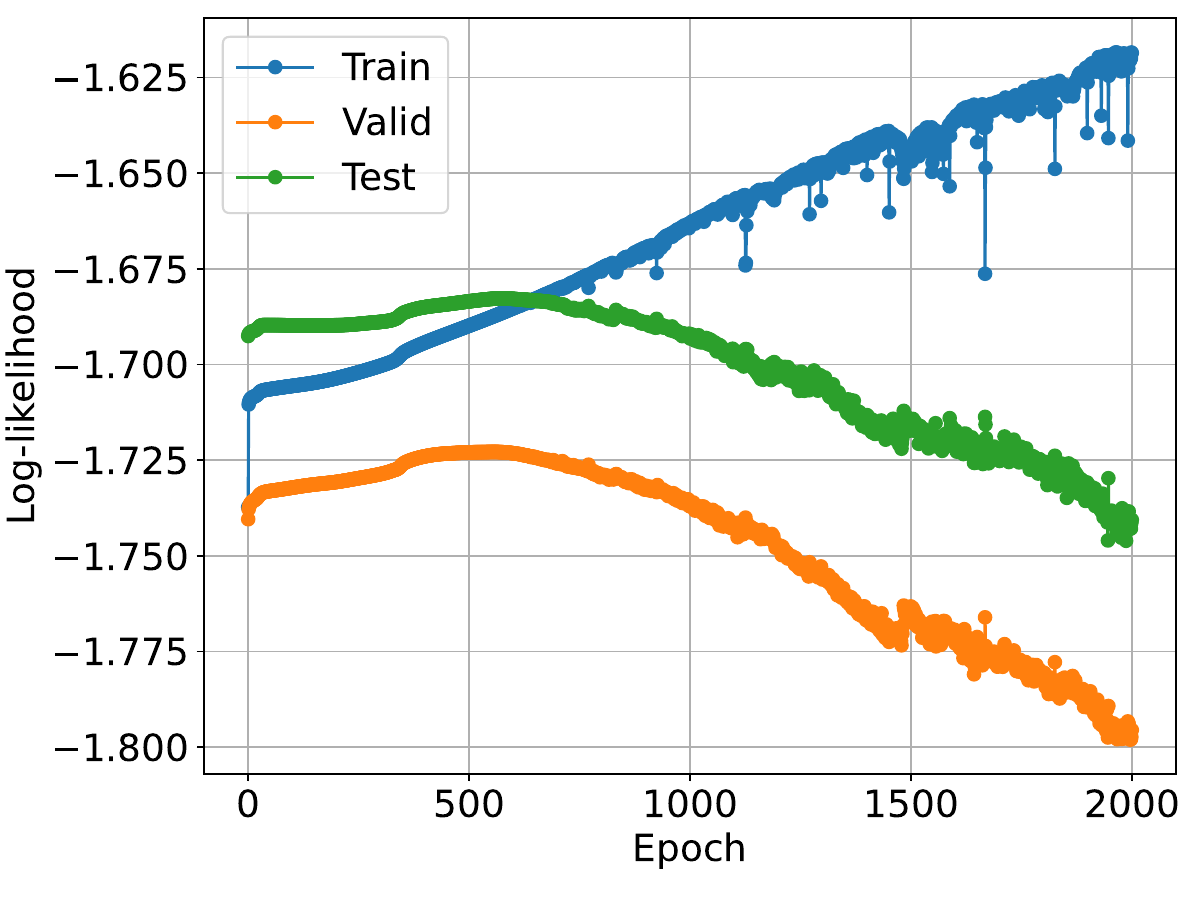}%
}
\subfloat[Biv 2\label{fig:biv2_tnhp}]{%
  \includegraphics[width=.45\linewidth]{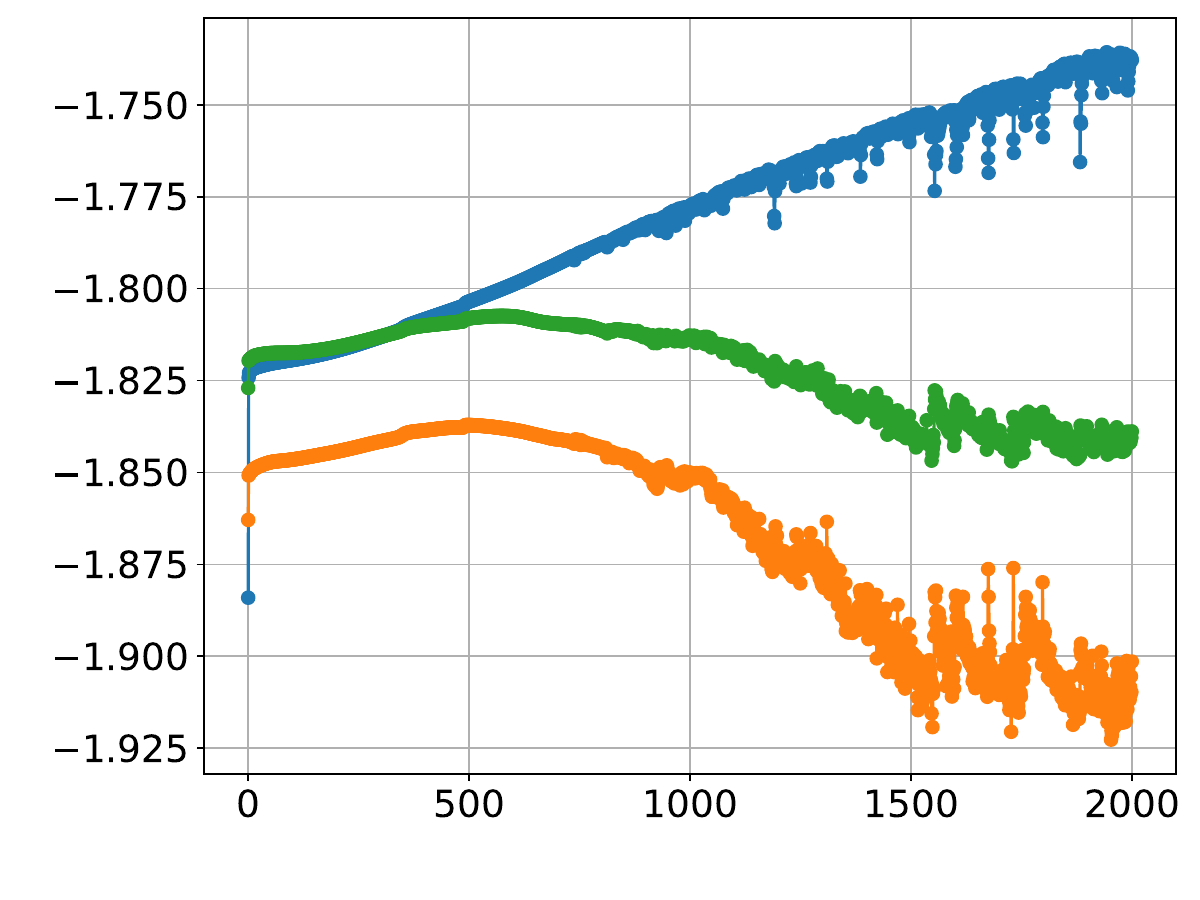}%
}\\

\subfloat[Biv 3\label{fig:biv3_tnhp}]{%
  \includegraphics[width=.45\linewidth]{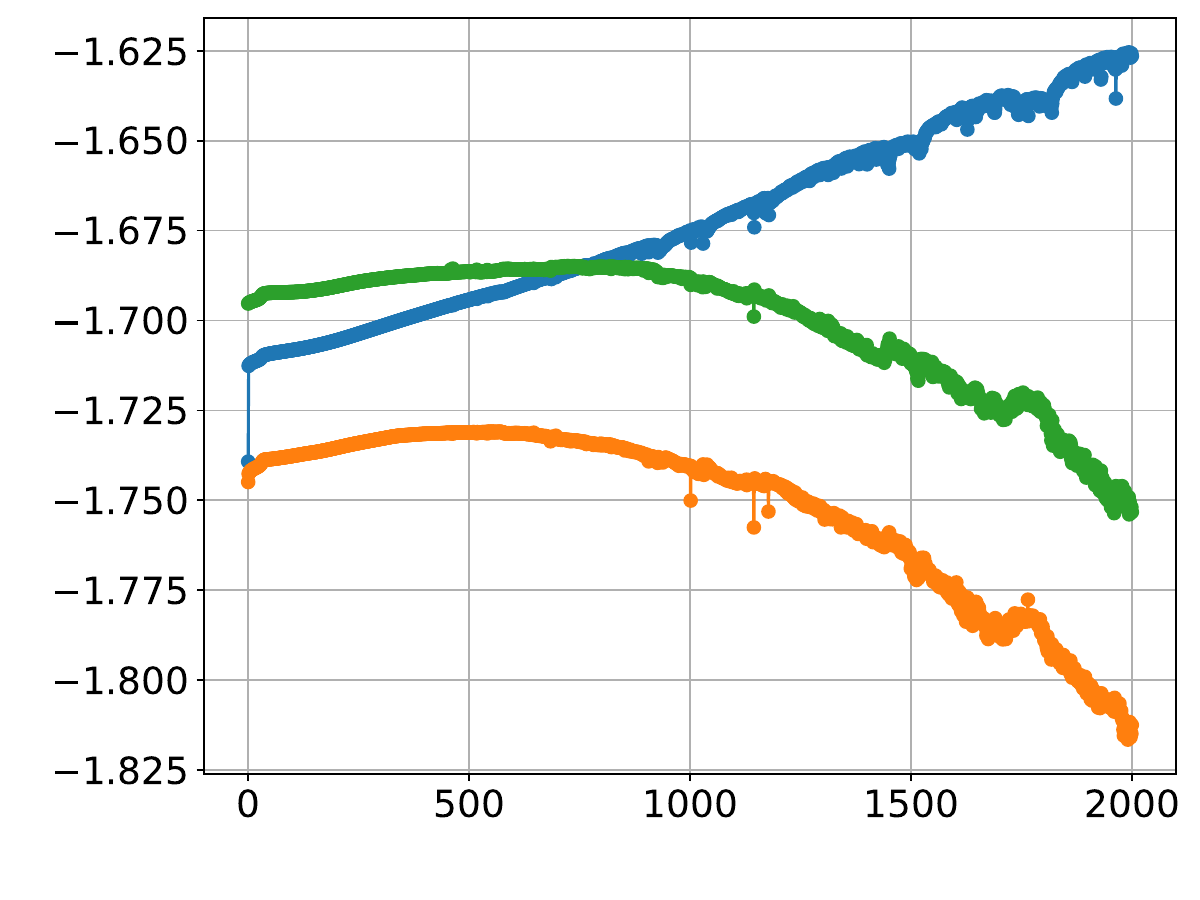}%
}
\subfloat[Biv 4\label{fig:biv4_tnhp}]{%
  \includegraphics[width=.45\linewidth]{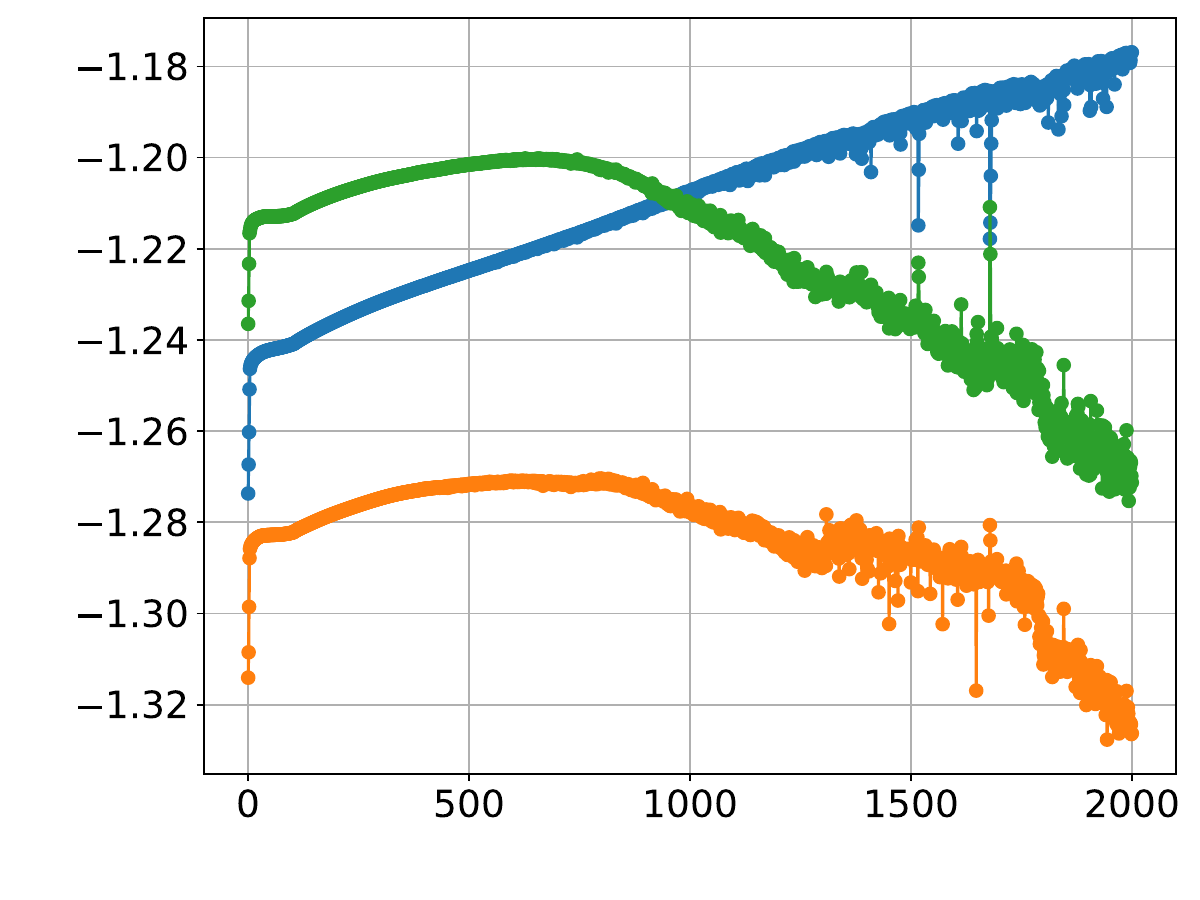}%
}
\caption{MTNHP results for Biv~1--4 datasets. All models were trained for 2k epochs with 58{,}112 trainable parameters (64 hidden units), whereas the corresponding MSTNHP models used 32 hidden units for Biv~1--3 and 64 hidden units for Biv~4.}
\label{fig:tnhp_64hsize}
\end{figure}

\subsection{Impact of Spatial Triggering Distance on MTNHP Recovery of Intensity Curves} \label{subsec:supp_biv1_study}

\begin{figure}[htbp!]
\centering
\subfloat[True\label{fig:true_intensity_biv_1}]{
  \includegraphics[width=.48\linewidth]{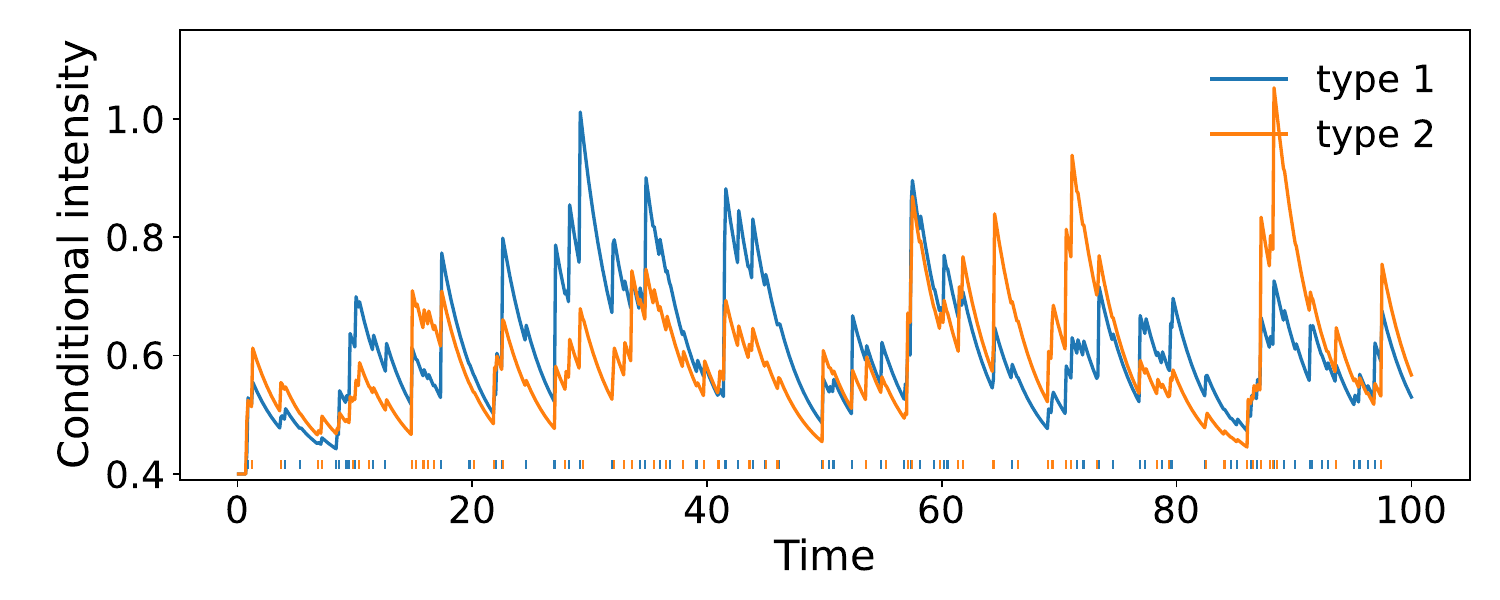}
  }
  \subfloat[Fitted\label{fig:biv1_tnhp_intensities}]{
   \includegraphics[width=.48\linewidth]{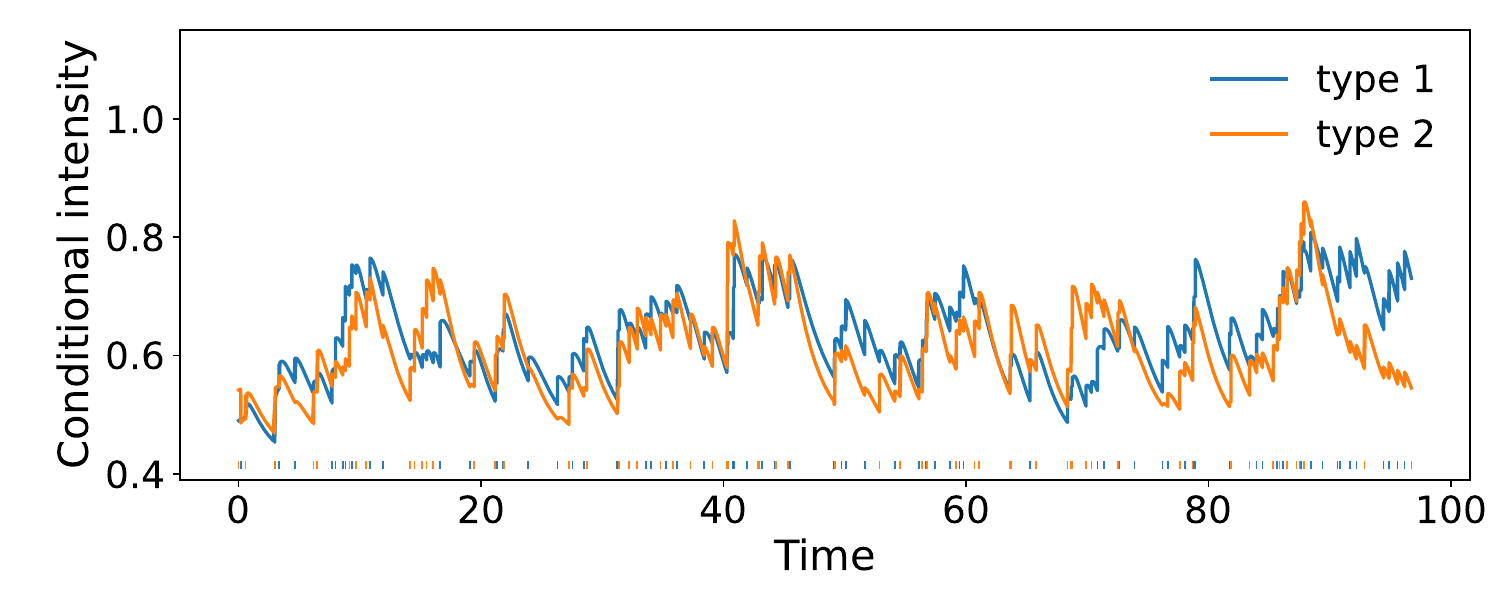}
   }%
\hfill
  \caption{True and fitted temporal intensity from MTNHP with data simulated under Biv 1 model, except for shorter spatial triggering distance (i.e., $\sigma^2=10^{-4}$).} 
  \label{fig:compare} 
\end{figure}

\begin{figure}[!htbp]
\centering
  \includegraphics[width=.45\linewidth]{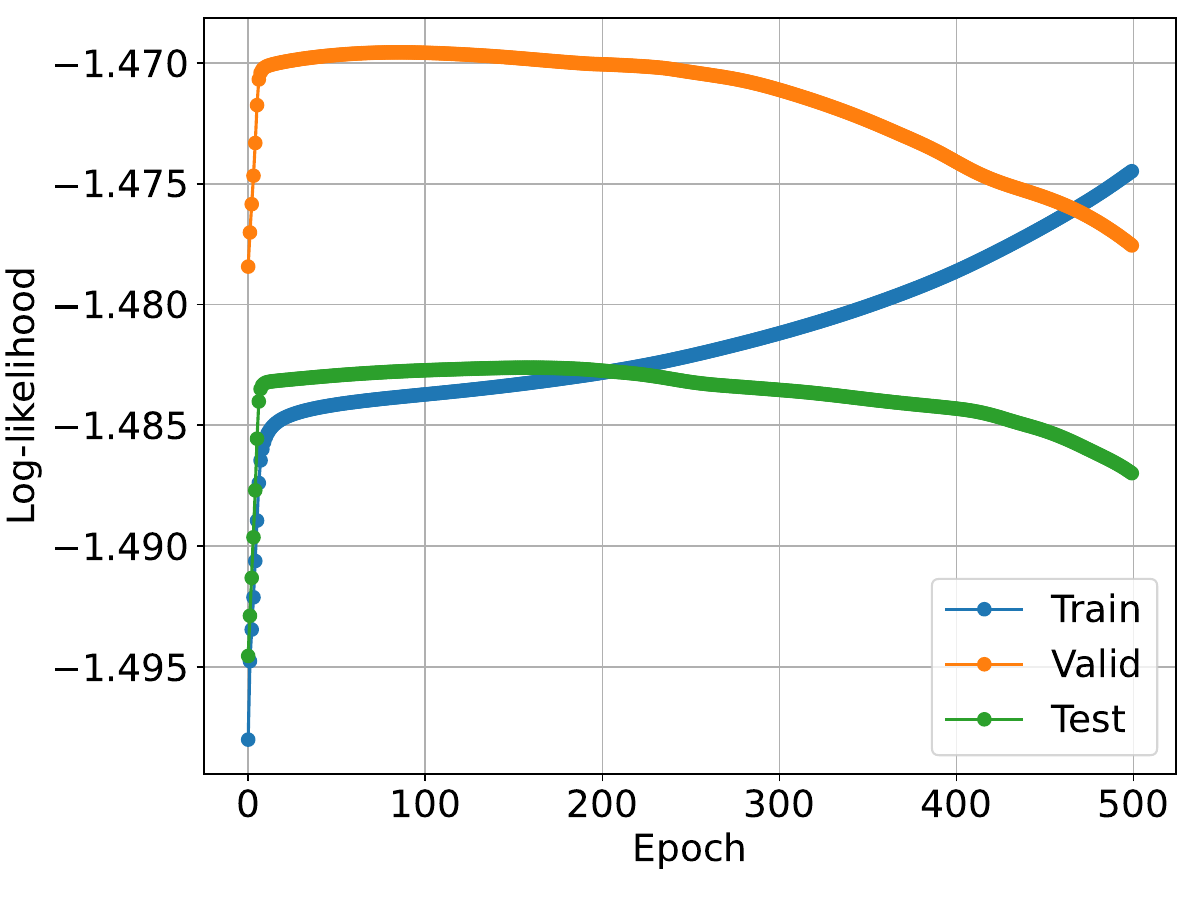}%
\caption{Log-likelihoods against epoch for the Biv 1 simulation with $\sigma^2 = 10^{-4}$. Validation likelihood has its maximum at epoch=85.}
\label{fig:biv1_tnhp_conv}
\end{figure}

\FloatBarrier
\section{Real application results}

\label{sec:real_data_conv_plots}
\begin{figure}[!htbp]
\centering

\subfloat[MSTNHP]{%
  \includegraphics[width=.45\linewidth]{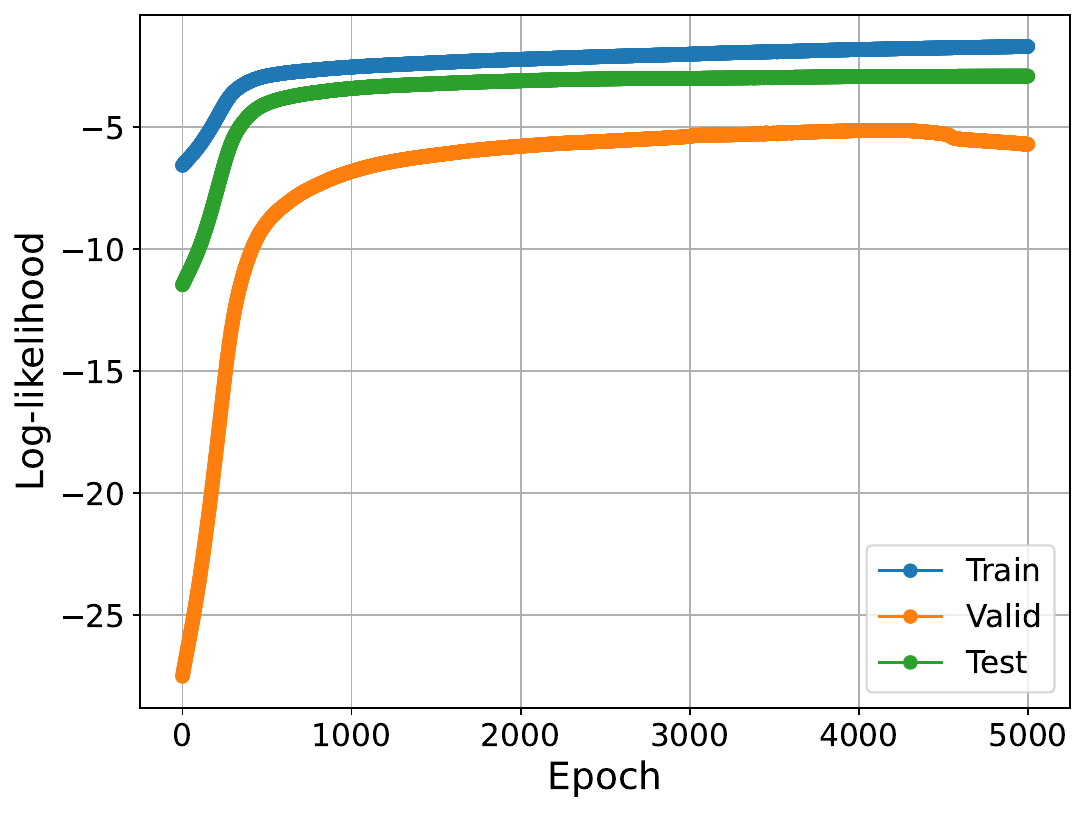}%
}
\subfloat[MTNHP]{%
  \includegraphics[width=.45\linewidth]{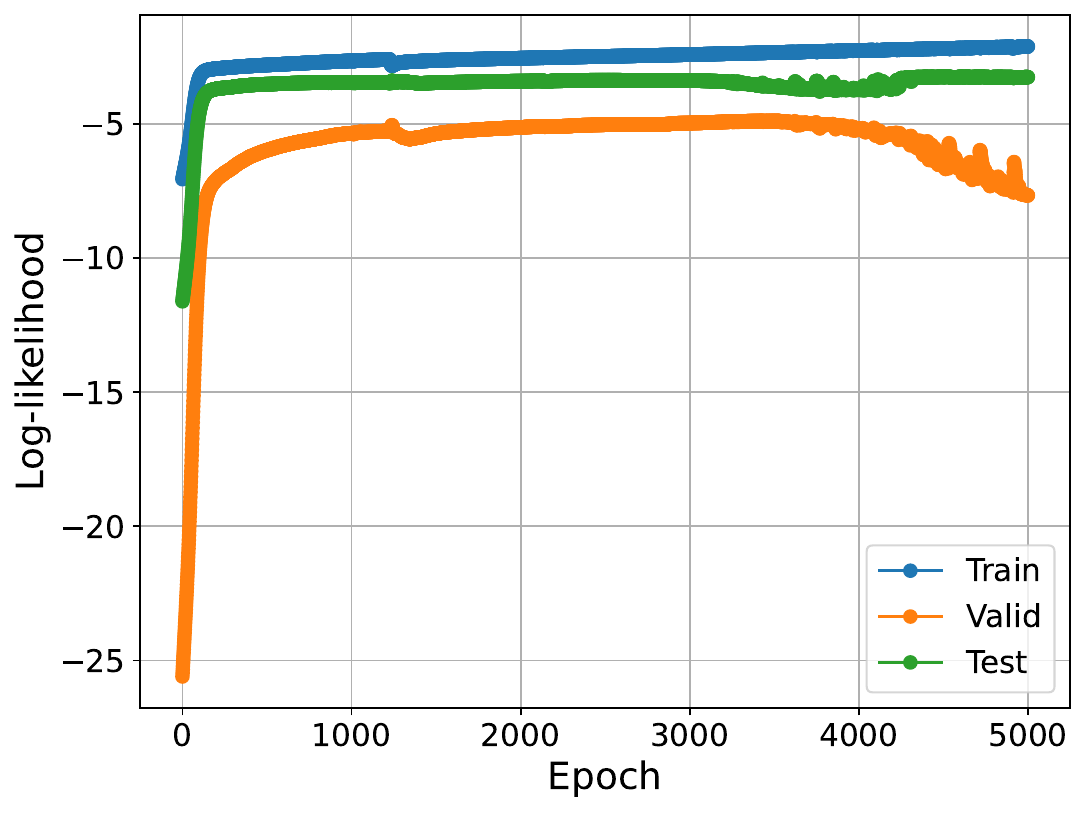}%
}
\caption{Log-likelihood values for MSTNHP and MTNHP for the Pakistan terrorism data.}

\label{fig:pakistan-convergence}
\end{figure}

\vfill

\end{document}